\documentclass[10pt,twocolumn,letterpaper]{article}

\usepackage[pagenumbers]{cvpr}

\usepackage{textcomp}
\usepackage{verbatim}
\usepackage{amsfonts}
\usepackage{threeparttable}
\usepackage{booktabs}
\usepackage{multirow}
\usepackage{xspace}
\usepackage{array}
\newcolumntype{C}[1]{>{\centering\arraybackslash}m{#1}}
\usepackage{adjustbox}
\usepackage{colortbl}
\usepackage{placeins}
\usepackage[ruled,longend,linesnumbered]{algorithm2e}
\usepackage{tcolorbox}
\usepackage{marvosym}

\definecolor{sem}{HTML}{77BEF0}
\definecolor{myblue}{RGB}{6,119,215}
\definecolor{mypink}{RGB}{245,210,210}
\definecolor{rebuttal_blue}{RGB}{0,0,0}
\DeclareMathOperator*{\argmin}{arg\,min}
\DeclareMathOperator*{\argmax}{arg\,max}

\definecolor{cvprblue}{rgb}{0.21,0.49,0.74}
\usepackage[pagebackref,breaklinks,colorlinks,allcolors=cvprblue]{hyperref}

\def\paperID{*****}
\def\confName{CVPR}
\def\confYear{2026}

\title{Follow-Your-Preference++: Rethinking Preference Alignment for Image Inpainting}

\author{
Junkun Yuan$^{1*}$ \quad
Yutao Shen$^{2*}$ \quad
Toru Aonishi$^2$ \quad
Hideki Nakayama$^2$ \quad
Yue Ma$^{3\dagger}$\\
$^1$Zhejiang University \quad
$^2$The University of Tokyo \quad
$^3$Tsinghua University\\
{\small $^*$Equal contribution. $\dagger$Corresponding author.}
}

\begin{document}
\maketitle

\begin{figure*}[h]
\centering
\includegraphics[width=\linewidth, trim=0mm 0mm 0mm 0mm, clip]{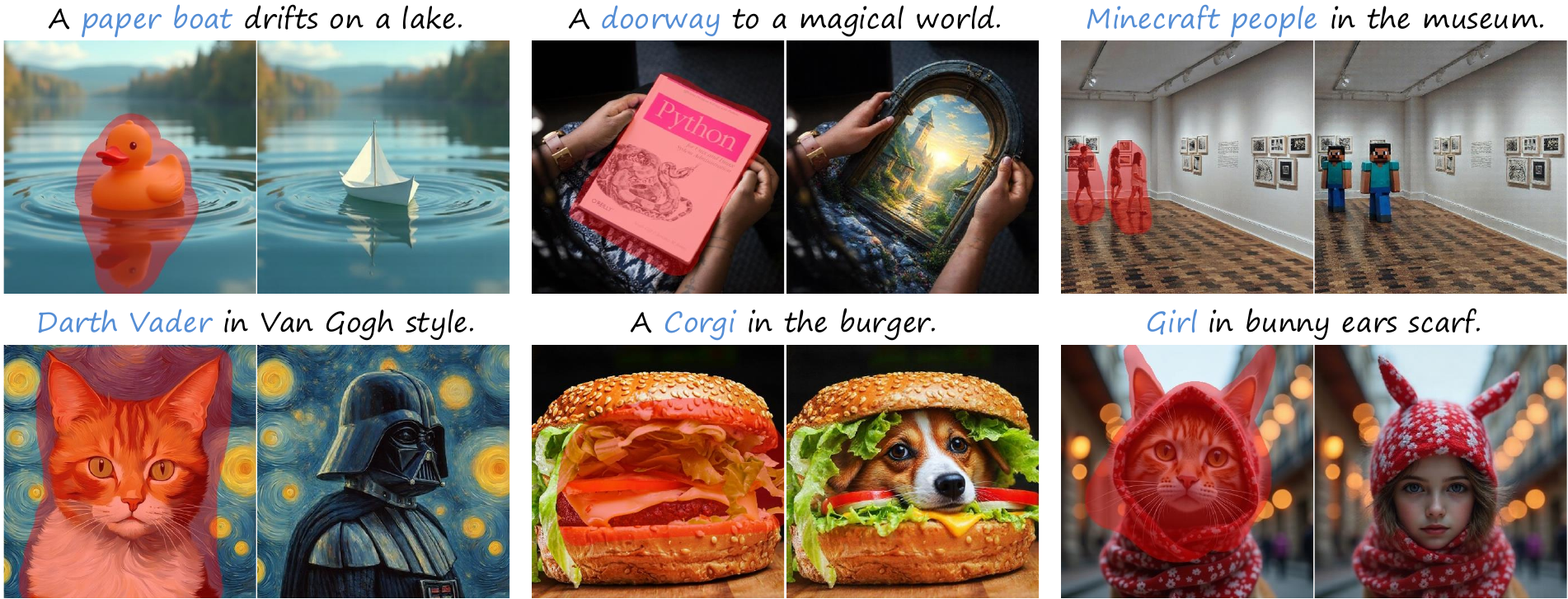}
\caption{\textbf{Results of our model.} The inpainting outputs generated by our model are visually coherent, semantically aligned with text prompts, and consistent with human aesthetic preferences.}
\end{figure*}

\begin{abstract}
We study preference alignment for image inpainting. Rather than proposing yet another method, we revisit the problem from first principles and reassess its core challenges. We adopt the widely used direct preference optimization framework and construct preference training data with publicly available reward models. Our empirical study spans nine reward models, two benchmarks, and two baseline inpainting models that differ in architecture and generative mechanism. Our main findings are: (1) Most reward models provide valid signals for preference data construction, although some are unreliable as evaluators. (2) Across models and benchmarks, preference data exhibits consistent trends under both candidate and sample scaling. (3) Reward models display pronounced biases--particularly in brightness, composition, and color scheme--that make them prone to inducing reward hacking. (4) A simple ensemble of reward models mitigates such biases and yields robust, generalizable performance. {\color{rebuttal_blue}(5) Preference alignment is transferable to the object removal task, where the goal shifts from open-ended creative generation to coherent background completion. (6) Further analysis reveals that a calibrated ensemble method further mitigates hacking and improves robustness.} Without modifying model architectures or introducing additional datasets, our models substantially outperform prior state-of-the-art models on standard metrics, large vision-language model evaluations, and human assessments. Our code is available at: \url{https://github.com/shenytzzz/Follow-Your-Preference}.
\end{abstract}

\section{Introduction}

Image inpainting~\cite{inpaint} aims to reconstruct user-specified regions of an image so that the completed result is visually coherent with the surrounding context and photorealistic. It underpins a wide range of applications, including photo restoration~\cite{restoration}, content creation~\cite{powerpaint}, and image editing~\cite{magicbrush}. Driven by the rapid progress in diffusion models~\cite{DDPM} and flow-based models~\cite{FlowMatching}, image inpainting has become a prominent research direction.

Aligning visual generation systems with human preferences has recently emerged as a central research focus~\cite{Diffusion-DPO, dancegrpo, DDPO, dpok}. Despite the substantial progress of image inpainting~\cite{PreAliSurvey, BrushNet, powerpaint, HD-Painter}, comparatively little effort has been devoted to aligning inpainting outputs with human preferences.

This paper studies preference alignment for image inpainting. Given the limited prior work on this task, our goal is not to introduce yet another method, but to revisit its fundamental questions. To this end, we adopt the widely used Direct Preference Optimization (DPO)~\cite{DPO, Diffusion-DPO, flow-dpo} as the basis of our analysis for its simplicity and efficiency. To avoid the high cost and limited scalability of human annotation, we construct preference training data using publicly available, off-the-shelf reward models. Our investigation centers on three core questions: (1) How \textit{effective} are these reward models at scoring candidates and constructing high-quality preference data? (2) How \textit{scalable} is preference data with respect to candidate quantity and sample quantity? (3) How does \textit{reward hacking}~\cite{hack} arise, and how can it be mitigated?

To address these questions, we conduct experiments across \textit{nine commonly used reward models} (e.g., HPSv2~\cite{hpsv2}, PickScore~\cite{pickscore}), \textit{two representative benchmarks} (BrushBench~\cite{BrushNet}, EditBench~\cite{EditBench}), and \textit{two baseline inpainting models} (BrushNet~\cite{BrushNet}, FLUX.1 Fill~\cite{flux1filldev}) that cover diverse architectures (U-Net~\cite{UNet}, Transformer~\cite{Transformer}) and generative paradigms (diffusion~\cite{DDPM}, flow matching~\cite{FlowMatching}). Our results show that: \textbf{(1)} Most reward models yield \textit{valid} reward signals for constructing effective preference training data, although some of them are unreliable as evaluators and share common biases. \textbf{(2)} Across baselines and benchmarks, preference data exhibits \textit{consistent trends} under both candidate scaling and sample scaling; nevertheless, biases in certain reward models (e.g., HPSv2) can induce reward hacking and weaken the effectiveness of scaling. \textbf{(3)} We uncover clear biases in reward models---particularly in \textit{brightness}, \textit{composition}, and \textit{color scheme}---that make them susceptible to reward hacking. For instance, HPSv2 favors images with bright illumination, intricate compositions, and vivid colors, whereas PickScore exhibits the opposite tendency. Consistent with this, PickScore pairs well with BrushNet, which generates vibrant outputs, while HPSv2 pairs well with FLUX.1 Fill, which produces plainer ones. {\color{rebuttal_blue}We further trace these biases to systematic differences in the training data used by different reward models.} \textbf{(4)} A simple ensemble of these reward models exhibits strong versatility across baselines and yields balanced, aesthetically pleasing inpainting results. {\color{rebuttal_blue}\textbf{(5)} Preference alignment also transfers effectively to object removal with only minor pipeline adaptations. \textbf{(6)} Going beyond the conference version~\cite{follow-your-preference}, we study several \textbf{Ensemble} variants and find that the original \textbf{Ensemble} still suffers from reward hacking; we therefore propose a calibrated ensemble that calibrates reward models, mitigates bias-induced hacking, and further improves robustness.}

{\color{rebuttal_blue}
A preliminary version of this work appeared at ICLR 2026~\cite{follow-your-preference}, with the primary focus on creative image generation in the inpainting setting. The present paper extends this line of research in three directions. \textit{First}, we generalize preference alignment to \textit{object-removal inpainting}, a more challenging setting that requires the completed background to be both structurally coherent and visually plausible. \textit{Second}, we provide a more comprehensive analysis of the Ensemble method, studying weighting strategies across reward models and the efficiency gains brought by filtering out invalid reward signals. \textit{Third}, we revisit the mathematical formulation of the conference-version Ensemble, show that it can be interpreted as a calibration across reward models, and integrate it into the IPO loss~\cite{ipo} to further mitigate reward hacking. Together, these extensions establish a new state-of-the-art over the conference version.
}

Based on this exploration, we build simple yet effective preference-aligned inpainting models via reward ensembling. Without modifying model architectures or introducing new datasets, our models substantially outperform prior state-of-the-art models on standard metrics, large vision-language model evaluations, and human assessments. Qualitative comparisons further show that our models generate more coherent and visually appealing results than competing methods. We hope this work establishes a simple yet strong baseline that helps advance this promising research direction.

\section{Related Work}
\label{sec:related-work}

\textbf{Image inpainting}~\cite{inpaint} aims to restore missing or corrupted regions of an image by generating visually plausible and semantically coherent content that is well aligned with the surrounding context. The field has progressed rapidly~\cite{HD-Painter, ASUKA, hap, collaborative, feng2025dit4edit}, largely driven by the development of diffusion~\cite{DDPM} and flow-based generative models~\cite{FlowMatching, SD3, follow-your-canvas, follow-your-emoji, ma2026group, ma2024followpose, ma2025followcreation, ma2026fastvmt, ma2025followyourmotion, ma2025controllable, wang2024taming}. These models progressively transform random noise into structured visual content through iterative denoising or rectified flow matching, thereby providing a powerful generative prior for synthesizing realistic and high-quality image regions. Building on these advances, pioneering efforts~\cite{BLD, sd, CNI} have introduced such generative paradigms to image inpainting, and subsequent methods~\cite{BrushNet, powerpaint, domain-specific} have further improved architectural design, controllability, and generation quality under more challenging inpainting settings. For example, BrushNet~\cite{BrushNet} introduces a dual-branch architecture that explicitly decouples masked-image feature extraction from the subsequent generation process, enabling the model to better preserve contextual information while producing more flexible and effective inpainting results. FLUX~\cite{flux}, in contrast, is a rectified-flow transformer that exhibits strong image generation capability, producing visually appealing results with high fidelity. Its inpainting variant, FLUX.1 Fill~\cite{flux1filldev}, serves as a strong baseline for high-quality content completion in masked regions.

{\color{rebuttal_blue}
\textbf{Object removal} can be viewed as a free-form image inpainting task, but with a stronger emphasis on seamless and plausible background completion rather than open-ended or creative generation. Its goal is to replace user-specified masked objects with realistic background content that is visually coherent with the surrounding context. Recent studies are predominantly built on diffusion and flow-based models~\cite{DDPM, DDIM, FlowMatching, SD3}, and largely fall into two paradigms: inversion-based methods~\cite{prompt2prompt, repaint} and training-based methods~\cite{instructpix2pix, rad, RORem}. For instance, RORem~\cite{RORem} alternates between training a discriminator and optimizing a diffusion model with human annotators in the loop, ultimately yielding both a well-aligned diffusion model and a high-quality dataset for object removal.
}

\textbf{Image generation with preference alignment} is an emerging field that seeks to align synthesized images with human preferences~\cite{rl-in-mm-survey, label-eff}. Earlier works~\cite{DDPO, dpok} employ reinforcement learning (RL)~\cite{rl} to fine-tune generative models with feedback from reward models. More recent works explore methods built on Direct Preference Optimization (DPO)~\cite{DPO}, which bypasses the explicit RL stage. For instance, Diffusion-DPO~\cite{Diffusion-DPO} is the first attempt to align diffusion-based generation with human preferences by optimizing a pairwise preference objective. {\color{rebuttal_blue}CaPO~\cite{capo} aligns diffusion models with multiple reward models by first calibrating the scores across reward models and matching the resulting calibrated reward differences via a regression loss.} PrefPaint~\cite{Prefpaint} refines image inpainting results using a reward model trained on human-annotated data.

\section{Preliminaries}

\begin{table*}[t]
\centering
\caption{Comparisons of reward models using \textbf{BrushNet} on BrushBench and EditBench.}
\resizebox{1.\linewidth}{!}{
\begin{threeparttable}
\renewcommand{\arraystretch}{1.8}
\setlength{\tabcolsep}{-0.5pt}
{
     \begin{tabular}{lC{1.2cm}C{1.2cm}C{1.2cm}C{1.2cm}C{1.2cm}C{1.2cm}C{1.2cm}C{1.2cm}C{1.2cm}C{1.2cm}C{1.2cm}C{1.2cm}C{1.2cm}C{1.2cm}C{1.2cm}C{1.2cm}C{1.2cm}C{1.2cm}C{1.2cm}C{1.2cm}}
          \toprule
          \multirow{2}{*}{reward model}    & \multicolumn{2}{c}{{CLIPScore}}   & \multicolumn{2}{c}{{Aesthetic}}  & \multicolumn{2}{c}{{ImageR}} & \multicolumn{2}{c}{{PickScore}}  & \multicolumn{2}{c}{{HPSv2}}   & \multicolumn{2}{c}{{VQAScore}} & \multicolumn{2}{c}{{UnifiedR}} & \multicolumn{2}{c}{{Perception}} & \multicolumn{2}{c}{{HPSv3}} & \multicolumn{2}{c}{\cellcolor{sem!15}{GPT-4}}     \\
                                             \cmidrule(lr){2-3} \cmidrule(lr){4-5} \cmidrule(lr){6-7} \cmidrule(lr){8-9} \cmidrule(lr){10-11} \cmidrule(lr){12-13} \cmidrule(lr){14-15} \cmidrule(lr){16-17} \cmidrule(lr){18-19} \cmidrule(lr){20-21}
                                        & Brush. & Edit. & Brush. & Edit. & Brush. & Edit. & Brush. & Edit. & Brush. & Edit. & Brush. & Edit. & Brush. & Edit. & Brush. & Edit. & Brush. & Edit. & \cellcolor{sem!15}Brush. & \cellcolor{sem!15}Edit. \\
          \midrule
  \color{gray}Baseline &     \color{gray}26.415 &     \color{gray}27.337 &   \color{gray}6.425 &   \color{gray}5.392 &     \color{gray}12.717 &     \color{gray}-1.296 &     \color{gray}22.133 &   \color{gray}20.616 &     \color{gray}27.509 &     \color{gray}23.076 &             \color{gray}{9.060} &                 \color{gray}6.770 &   \color{gray}3.303 &   \color{gray}2.100 &     \color{gray}26.290 &     \color{gray}26.410 &     \color{gray}5.749 &   \color{gray}0.403 &     \cellcolor{sem!15}\color{gray}79.391 &     \cellcolor{sem!15}\color{gray}57.046 \\
    \color{gray}Random &     \color{gray}26.441 &     \color{gray}27.631 &   \color{gray}6.424 &   \color{gray}5.392 &     \color{gray}12.685 &     \color{gray}-1.136 &     \color{gray}22.130 &   \color{gray}20.642 &     \color{gray}27.501 &     \color{gray}23.067 &               \color{gray}9.050 &        \color{gray}\textbf{6.917} &   \color{gray}3.302 &   \color{gray}2.110 &     \color{gray}26.292 &     \color{gray}26.422 &     \color{gray}5.738 &   \color{gray}0.425 &     \cellcolor{sem!15}\color{gray}79.177 &     \cellcolor{sem!15}\color{gray}56.753 \\
             CLIPScore &                 26.461 &        \textbf{27.710} &               6.430 &               5.393 &                 12.782 &                 -0.720 &                 22.146 &               20.680 &                 27.582 &                 23.316 &                         {9.062} &   \underline{\color{myblue}6.894} &               3.341 &               2.124 &                 26.324 &     \underline{26.548} &                 5.777 &               0.640 &                 \cellcolor{sem!15}79.661 &                 \cellcolor{sem!15}57.539 \\
             Aesthetic &                 26.465 &   \color{myblue}27.355 &   \underline{6.477} &      \textbf{5.520} &     \underline{12.994} &                 -0.877 &                 22.221 &               20.689 &                 27.594 &                 23.166 &                         {9.065} &               \color{myblue}6.828 &               3.343 &               2.140 &                 26.293 &   \color{myblue}26.342 &                 5.922 &               0.597 &                 \cellcolor{sem!15}81.603 &                 \cellcolor{sem!15}58.603 \\
                ImageR &                 26.471 &   \color{myblue}27.539 &               6.462 &               5.434 &                 12.891 &               {-0.377} &                 22.153 &               20.701 &                 27.672 &               {23.467} &             \color{myblue}9.036 &               \color{myblue}6.761 &               3.334 &               2.144 &                 26.305 &                 26.501 &                 5.913 &               0.782 &                 \cellcolor{sem!15}80.341 &                 \cellcolor{sem!15}57.806 \\
             PickScore &   \color{myblue}26.397 &   \color{myblue}27.199 &               6.454 &               5.454 &                 12.893 &   \color{myblue}-1.364 &        \textbf{22.254} &   \underline{20.732} &   \color{myblue}27.322 &   \color{myblue}22.933 &                         {9.062} &               \color{myblue}6.873 &   \underline{3.353} &      \textbf{2.178} &   \color{myblue}26.273 &                 26.427 &                 5.750 &               0.469 &        \cellcolor{sem!15}\textbf{82.726} &        \cellcolor{sem!15}\textbf{59.550} \\
                 HPSv2 &     \underline{26.481} &     \underline{27.677} &               6.476 &             {5.495} &                 12.890 &         \textbf{0.128} &                 22.137 &             {20.725} &        \textbf{27.818} &        \textbf{23.742} &                  \textbf{9.073} &               \color{myblue}6.818 &               3.332 &               2.155 &        \textbf{26.361} &        \textbf{26.678} &                 5.979 &      \textbf{1.061} &                 \cellcolor{sem!15}79.914 &                 \cellcolor{sem!15}57.658 \\
              VQAScore &                 26.442 &   \color{myblue}27.524 &               6.429 &               5.407 &   \color{myblue}12.658 &                 -0.800 &   \color{myblue}22.126 &               20.667 &                 27.527 &                 23.234 &             \color{myblue}9.038 &               \color{myblue}6.879 &               3.311 &               2.139 &                 26.326 &   \color{myblue}26.406 &   \color{myblue}5.723 &               0.555 &   \color{myblue}\cellcolor{sem!15}78.877 &   \color{myblue}\cellcolor{sem!15}56.975 \\
              UnifiedR &   \color{myblue}26.428 &   \color{myblue}27.505 &               6.433 &               5.402 &                 12.764 &                 -0.812 &                 22.157 &               20.675 &                 27.562 &                 23.204 &                         {9.061} &               \color{myblue}6.857 &               3.329 &               2.155 &                 26.320 &                 26.436 &                 5.800 &               0.540 &                 \cellcolor{sem!15}80.333 &                 \cellcolor{sem!15}57.185 \\
            Perception &                 26.448 &   \color{myblue}27.484 &               6.428 &               5.393 &                 12.789 &                 -0.973 &                 22.177 &               20.660 &                 27.519 &                 23.111 &               \underline{9.069} &   \underline{\color{myblue}6.894} &               3.327 &             {2.160} &                 26.310 &                 26.515 &                 5.764 &               0.433 &                 \cellcolor{sem!15}80.277 &                 \cellcolor{sem!15}57.254 \\
                 HPSv3 &                 26.461 &   \color{myblue}27.547 &               6.464 &               5.448 &                 12.922 &     \underline{-0.146} &                 22.176 &               20.713 &                 27.758 &                 23.491 &                         {9.065} &               \color{myblue}6.850 &               3.344 &               2.158 &                 26.317 &                 26.535 &     \underline{6.014} &               0.863 &                 \cellcolor{sem!15}80.623 &                 \cellcolor{sem!15}57.485 \\
              Ensemble &        \textbf{26.535} &   \color{myblue}27.398 &      \textbf{6.485} &   \underline{5.497} &        \textbf{13.037} &                 -0.352 &     \underline{22.229} &      \textbf{20.735} &     \underline{27.797} &     \underline{23.522} &             \color{myblue}9.053 &             {\color{myblue}6.892} &      \textbf{3.365} &   \underline{2.176} &     \underline{26.338} &     \underline{26.603} &        \textbf{6.074} &   \underline{1.015} &     \cellcolor{sem!15}\underline{82.172} &     \cellcolor{sem!15}\underline{58.986} \\
          \bottomrule
     \end{tabular}
     \begin{tablenotes}
     \item \textbf{Bold} values denote the best results. \underline{Underlined} values denote the second-best results. Values in {\color{myblue}blue} denote the results below the baseline or random chance.
     \end{tablenotes}
}
\end{threeparttable}
}
\label{tab:brushnet_metrics_8cols_twobench}
\end{table*}

\begin{table*}[t]
\centering
\caption{Comparisons of reward models using \textbf{FLUX.1 Fill} on BrushBench and EditBench.}
\resizebox{1.\linewidth}{!}{
\begin{threeparttable}
\renewcommand{\arraystretch}{1.8}
\setlength{\tabcolsep}{-0.5pt}
{
     \begin{tabular}{lC{1.2cm}C{1.2cm}C{1.2cm}C{1.2cm}C{1.2cm}C{1.2cm}C{1.2cm}C{1.2cm}C{1.2cm}C{1.2cm}C{1.2cm}C{1.2cm}C{1.2cm}C{1.2cm}C{1.2cm}C{1.2cm}C{1.2cm}C{1.2cm}C{1.2cm}C{1.2cm}}
          \toprule
          \multirow{2}{*}{reward model}    & \multicolumn{2}{c}{{CLIPScore}}   & \multicolumn{2}{c}{{Aesthetic}}  & \multicolumn{2}{c}{{ImageR}} & \multicolumn{2}{c}{{PickScore}}  & \multicolumn{2}{c}{{HPSv2}}   & \multicolumn{2}{c}{{VQAScore}} & \multicolumn{2}{c}{{UnifiedR}} & \multicolumn{2}{c}{{Perception}} & \multicolumn{2}{c}{{HPSv3}} & \multicolumn{2}{c}{\cellcolor{sem!15}{GPT-4}}     \\
                                             \cmidrule(lr){2-3} \cmidrule(lr){4-5} \cmidrule(lr){6-7} \cmidrule(lr){8-9} \cmidrule(lr){10-11} \cmidrule(lr){12-13} \cmidrule(lr){14-15} \cmidrule(lr){16-17} \cmidrule(lr){18-19} \cmidrule(lr){20-21}
                                        & Brush. & Edit. & Brush. & Edit. & Brush. & Edit. & Brush. & Edit. & Brush. & Edit. & Brush. & Edit. & Brush. & Edit. & Brush. & Edit. & Brush. & Edit. & \cellcolor{sem!15}Brush. & \cellcolor{sem!15}Edit. \\
          \midrule
  \color{gray}Baseline &     \color{gray}26.244 &     \color{gray}27.103 &   \color{gray}6.429 &   \color{gray}5.458 &     \color{gray}12.760 &     \color{gray}4.910 &   \color{gray}22.327 &     \color{gray}21.211 &   \color{gray}27.476 &   \color{gray}24.076 &                 \color{gray}9.081 &       \color{gray}8.012 &     \color{gray}3.360 &   \color{gray}2.485 &   \color{gray}25.945 &     \color{gray}26.834 &     \color{gray}6.055 &     \color{gray}2.470 &     \cellcolor{sem!15}\color{gray}83.935 &     \cellcolor{sem!15}\color{gray}66.979 \\
    \color{gray}Random &     \color{gray}26.239 &     \color{gray}27.078 &   \color{gray}6.431 &   \color{gray}5.459 &     \color{gray}12.772 &     \color{gray}4.955 &   \color{gray}22.328 &     \color{gray}21.211 &   \color{gray}27.475 &   \color{gray}24.100 &                 \color{gray}9.077 &     \color{gray}{8.030} &     \color{gray}3.356 &   \color{gray}2.491 &   \color{gray}25.944 &     \color{gray}26.838 &     \color{gray}6.056 &     \color{gray}2.490 &     \cellcolor{sem!15}\color{gray}83.517 &     \cellcolor{sem!15}\color{gray}66.942 \\
             CLIPScore &   \color{myblue}26.233 &   \color{myblue}27.072 &               6.432 &               5.477 &                 12.791 &                 4.997 &               22.329 &                 21.215 &               27.487 &               24.121 &             {\color{myblue}9.071} &          \textbf{8.071} &                 3.361 &               2.512 &               25.948 &                 26.859 &                 6.056 &                 2.499 &                 \cellcolor{sem!15}83.942 &                 \cellcolor{sem!15}66.997 \\
             Aesthetic &     \underline{26.250} &     \underline{27.200} &               6.432 &   \underline{5.478} &                 12.823 &     \underline{5.175} &               22.337 &                 21.219 &               27.520 &               24.142 &             \color{myblue}{9.075} &    \color{myblue} 8.001 &                 3.363 &               2.507 &               25.954 &                 26.878 &                 6.075 &     \underline{2.577} &                 \cellcolor{sem!15}83.950 &                 \cellcolor{sem!15}67.906 \\
                ImageR &        \textbf{26.251} &                 27.121 &               6.434 &      \textbf{5.481} &                 12.823 &                 5.001 &               22.336 &                 21.211 &               27.518 &               24.143 &             \color{myblue}{9.080} &     \color{myblue}7.977 &                 3.362 &   \underline{2.536} &               25.946 &                 26.846 &                 6.078 &                 2.550 &                 \cellcolor{sem!15}84.176 &                 \cellcolor{sem!15}67.785 \\
             PickScore &   \color{myblue}26.236 &               {27.195} &               6.436 &               5.476 &               {12.879} &                 5.134 &               22.341 &               {21.223} &               27.530 &             {24.154} &             \color{myblue}{9.076} &    \color{myblue} 8.003 &        \textbf{3.383} &               2.514 &               25.955 &     \underline{26.900} &                 6.105 &                 2.548 &                 \cellcolor{sem!15}84.188 &                 \cellcolor{sem!15}67.100 \\
                 HPSv2 &                 26.246 &                 27.160 &   \underline{6.441} &               5.475 &        \textbf{12.904} &               {5.145} &      \textbf{22.356} &        \textbf{21.232} &      \textbf{27.605} &      \textbf{24.202} &                    \textbf{9.085} &   \color{myblue}{8.028} &                 3.363 &      \textbf{2.553} &      \textbf{25.963} &               {26.895} &        \textbf{6.181} &        \textbf{2.605} &        \cellcolor{sem!15}\textbf{84.699} &        \cellcolor{sem!15}\textbf{68.186} \\
              VQAScore &   \color{myblue}26.243 &   \color{myblue}27.101 &               6.432 &               5.466 &                 12.781 &   \color{myblue}4.926 &               22.329 &                 21.215 &               27.486 &               24.104 &             {\color{myblue}9.075} &     \color{myblue}8.020 &   \color{myblue}3.353 &               2.501 &               25.950 &                 26.861 &   \color{myblue}6.046 &   \color{myblue}2.473 &   \cellcolor{sem!15}\color{myblue}83.854 &   \cellcolor{sem!15}\color{myblue}66.793 \\
              UnifiedR &   \color{myblue}26.235 &                 27.133 &               6.434 &               5.473 &   \color{myblue}12.769 &                 5.034 &               22.331 &                 21.212 &               27.485 &               24.120 &             \color{myblue}{9.076} &     \color{myblue}8.014 &               {3.366} &               2.517 &               25.954 &                 26.853 &                 6.057 &                 2.524 &                 \cellcolor{sem!15}83.950 &                 \cellcolor{sem!15}67.372 \\
            Perception &        \textbf{26.251} &   \color{myblue}27.088 &               6.433 &               5.466 &   \color{myblue}12.752 &                 4.975 &               22.331 &   \color{myblue}21.209 &               27.479 &               24.109 &                           {9.082} &     \color{myblue}8.023 &   \color{myblue}3.356 &               2.514 &               25.951 &   \color{myblue}26.818 &                 6.064 &                 2.504 &                 \cellcolor{sem!15}84.022 &                 \cellcolor{sem!15}68.024 \\
                 HPSv3 &   \color{myblue}26.238 &        \textbf{27.223} &   \underline{6.441} &               5.465 &                 12.855 &                 5.092 &               22.340 &     \underline{21.226} &               27.534 &   \underline{24.155} &                 \underline{9.083} &     \color{myblue}8.000 &     \underline{3.378} &               2.497 &   \underline{25.957} &                 26.859 &                 6.106 &                 2.568 &                 \cellcolor{sem!15}84.615 &     \cellcolor{sem!15}\underline{68.107} \\
              Ensemble &   \color{myblue}26.239 &                 27.158 &      \textbf{6.442} &               5.472 &     \underline{12.884} &        \textbf{5.239} &   \underline{22.346} &                 21.215 &   \underline{27.560} &               24.146 &                           {9.082} &       \underline{8.036} &                 3.367 &               2.507 &      \textbf{25.963} &        \textbf{26.905} &     \underline{6.151} &     \underline{2.577} &     \cellcolor{sem!15}\underline{84.628} &                 \cellcolor{sem!15}67.549 \\
          \bottomrule
     \end{tabular}
     \begin{tablenotes}
     \item \textbf{Bold} values denote the best results. \underline{Underlined} values denote the second-best results. Values in {\color{myblue}blue} denote the results below the baseline or random chance.
     \end{tablenotes}
}
\end{threeparttable}
}
\label{tab:flux_metrics_8col_twobench}
\end{table*}

\subsection{Diffusion Models and Flow-Based Models}
Diffusion models~\cite{thermodynamics, DDIM}, such as \textbf{DDPM}~\cite{DDPM}, are a class of generative models that learn to reverse a gradual noise corruption process. DDPM assumes a forward process that gradually applies noise to real data. At timestep $t$, the real data $x_0$ is destroyed to $x_t$: $q(x_t|x_0)=\mathcal{N}(x_t;\sqrt{\bar{\alpha}_t}x_0,(1-\bar{\alpha}_t)\mathbf{I})$, where $\bar{\alpha}_t$ is a noise scheduling hyper-parameter. It has a reparameterization formula: $x_t=\sqrt{\bar{\alpha}_t}x_0 + \sqrt{1-\bar{\alpha}_t}\epsilon$, where noise $\epsilon\sim\mathcal{N}(0, \mathbf{I})$. DDPM learns a reverse process using a denoising model $\epsilon_{\theta}$ with parameters $\theta$, inverting the forward process: $p_{\theta}(x_{t-1}|x_t)=\mathcal{N}(\mu_{\theta}(x_t),\Sigma_{\theta}(x_t))$. The denoising model $\epsilon_{\theta}$ can be trained by minimizing:
\begin{equation}
    \mathcal{L}_{\mathrm{DDPM}}=\mathbb{E}_{t, x_0,\epsilon}\left[|| \epsilon - \epsilon_{\theta}(\sqrt{\bar{\alpha}_t}x_0 + \sqrt{1-\bar{\alpha}_t}\epsilon, t)||^2\right].
\label{eq:ddpm}
\end{equation}
Flow-based models~\cite{SD3} are generative models that learn to model data distributions using invertible transformations. Recently, \textbf{Flow Matching}~\cite{FlowMatching} has emerged as a prominent approach for visual generation~\cite{Sit, Seedream}. It usually learns a continuous-time flow that transforms a simple prior distribution into the data distribution by solving an ODE. The process, with an optimal-transport path, employs a linear interpolation scheme: $x_t=(1-t)x_0+t\epsilon$. A denoising model $v_{\theta}$ is trained to predict the velocity field by minimizing:
\begin{equation}
\mathcal{L}_{\mathrm{FM}}=\mathbb{E}_{t,x_0,\epsilon}\left[|| v_{\theta}((1-t)x_0+t\epsilon, t)-(\epsilon - x_0) ||^2\right].
\label{eq:fm}
\end{equation}
\textbf{U-Net}~\cite{UNet} is used as the basic model structure by many previous denoising models~\cite{DDPM, DDIM}. U-Net is a symmetric encoder-decoder architecture that captures multi-scale features through progressive downsampling and upsampling. \textbf{Transformers}~\cite{Transformer}, employed in recent works~\cite{Seedance, Seedream, Wan, HunyuanVideo}, process all data elements in parallel using attention, facilitating training scalability.

To improve the reliability and generalization of conclusions drawn in our studies, we conduct investigations using two different \textbf{baseline models}---\textbf{BrushNet}~\cite{BrushNet} and \textbf{FLUX.1 Fill}~\cite{flux1filldev}, introduced in~\autoref{sec:related-work}. BrushNet is built on a U-Net-like architecture and trained with the DDPM loss, while FLUX leverages transformers and learns via Flow Matching.

\subsection{Preference Alignment}
The standard pipeline for training large-scale models typically involves pre-training, supervised fine-tuning, and preference alignment. Preference alignment refines model outputs to better match human values. Reinforcement Learning from Human Feedback (RLHF)~\cite{RLHF} is a popular alignment approach. It utilizes human preferences on model outputs to train a \textit{separate reward model}, which subsequently provides rewards for alignment via reinforcement learning algorithms such as PPO~\cite{PPO} and GRPO~\cite{GRPO}. In comparison, \textbf{Direct Preference Optimization (DPO)}~\cite{DPO}, which performs direct supervised learning, offers higher training efficiency. It constructs a preference dataset that comprises \textit{preferred samples} and \textit{dispreferred samples}. DPO learns human preferences implicitly contained within the data by maximizing:
\begin{equation}
    \mathbb{E}_{x,y^w,y^l}[\log\sigma(\beta\log\frac{\pi_{\theta}(y^w|x)}{\pi_{\mathrm{ref}}(y^w|x)}-\beta\log\frac{\pi_{\theta}(y^l|x)}{\pi_{\mathrm{ref}}(y^l|x)})],
\label{eq:dpo}
\end{equation}
where $\sigma$ is the sigmoid function; $\pi_{\theta}$ and $\pi_{\mathrm{ref}}$ are the \textit{policy} and the \textit{reference policy}, respectively. In image generation, given a text prompt $x$, $y^w$ and $y^l$ denote the generated preferred image and dispreferred image, respectively. The hyper-parameter $\beta$ controls the strength of regularization: a large value of $\beta$ increases regularization pressure, dampening preference learning. In visual generation, \autoref{eq:dpo} can be derived to yield a simplified loss~\cite{Diffusion-DPO, flow-dpo}:
\begin{equation}
    \mathcal{L}_{\mathrm{DPO}}=-\mathbb{E}[\log\sigma(-\beta((\mathcal{L}_{\theta}^{w}-\mathcal{L}_{\mathrm{ref}}^{w})-(\mathcal{L}_{\theta}^{l}-\mathcal{L}_{\mathrm{ref}}^{l})))],
\label{eq:dpo-loss}
\end{equation}
where $\mathcal{L}_{\theta}^{w}$ and $\mathcal{L}_{\theta}^{l}$ denote the loss (\autoref{eq:ddpm} or \autoref{eq:fm}) applied to the policy on preferred samples and dispreferred samples, respectively; similarly, $\mathcal{L}_{\mathrm{ref}}^{w}$ and $\mathcal{L}_{\mathrm{ref}}^{l}$ denote the losses applied to the reference policy. This loss function aligns the distribution of generated samples with the preferred data distribution and encourages it to diverge from the dispreferred distribution. Due to the simplicity, efficiency, and stability of DPO, \textit{this paper will explore preference alignment for image inpainting by optimizing~\autoref{eq:dpo-loss} on different preference datasets that are constructed for investigation.}

\subsection{Reward Models}
Reward models play an important role in preference alignment: they provide real-time rewards in RLHF~\cite{PPO}, and offer scores for constructing offline preference data in DPO~\cite{unifiedreward,capo}. However, prior works~\cite{SimpleAR, dancegrpo} directly employ off-the-shelf reward models for visual preference alignment without systematic evaluation. In this paper, we evaluate the effectiveness of these reward models in constructing preference data via extensive studies. Specifically, we examine the following \textbf{public reward models}:
(1) \textbf{CLIPScore}~\cite{clipscore} measures semantic alignment between images and text prompts by calculating cosine similarities of their CLIP embeddings~\cite{CLIP}.
(2) \textbf{Aesthetic}~\cite{aesthetic} predicts human aesthetic preferences on top of the CLIP embeddings.
(3) \textbf{ImageReward (ImageR)}~\cite{imagereward} is trained by fine-tuning BLIP~\cite{BLIP} on 137K preference samples. 
(4) \textbf{PickScore}~\cite{pickscore} is a CLIP-based image scoring model, trained on over 500K synthesized image samples with users' preference choices.
(5) \textbf{HPSv2}~\cite{hpsv2} is also a CLIP-based model that evaluates both image quality and text-image alignment by learning from 798K human preferences on 433K sample pairs.
(6) \textbf{VQAScore}~\cite{vqascore} provides a semantic alignment score by computing the probability of a VQA model answering ``yes'' to each question: ``Does this figure show \{text\}?''.
(7) \textbf{UnifiedReward (UnifiedR)}~\cite{unifiedreward} is a unified model that assesses both visual generation and understanding.
(8) \textbf{Perception Encoder (Perception)}~\cite{perception} is trained by contrastive visual-language pre-training, producing semantically aligned multimodal embeddings.
(9) \textbf{HPSv3}~\cite{hpsv3} is trained on 1.5M annotated sample pairs using Qwen2VL-7B~\cite{qwen2-vl}.

To assess their efficacy, these models are employed to assign reward scores to candidate samples that are generated by the baseline models with different random seeds. The resulting highest- and lowest-scoring samples for each text prompt are subsequently utilized as the preferred and dispreferred samples for DPO training. Based on the evaluation of training results, the top-performing ones are designated as the most effective reward models to provide accurate rewards, whereas the lowest-performing models are designated as the least effective.

\section{How Effective are Reward Models?}
\label{sec:effective}

\begin{figure*}[!h]
\begin{subfigure}[t]{\linewidth}
    \centering
    \includegraphics[width=0.172\linewidth, trim=5mm 5mm 5mm 5mm, clip]{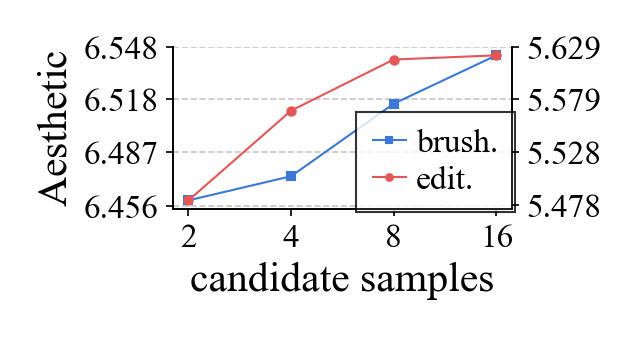}
    \includegraphics[width=0.172\linewidth, trim=5mm 5mm 5mm 5mm, clip]{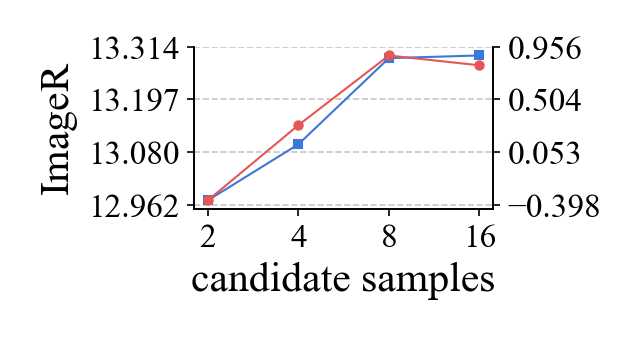}
    \includegraphics[width=0.172\linewidth, trim=5mm 5mm 5mm 5mm, clip]{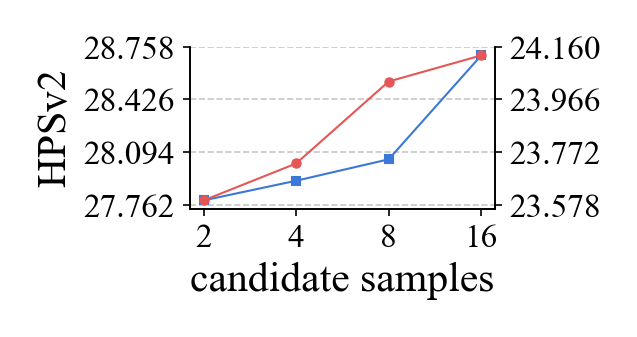}
    \includegraphics[width=0.172\linewidth, trim=5mm 5mm 5mm 5mm, clip]{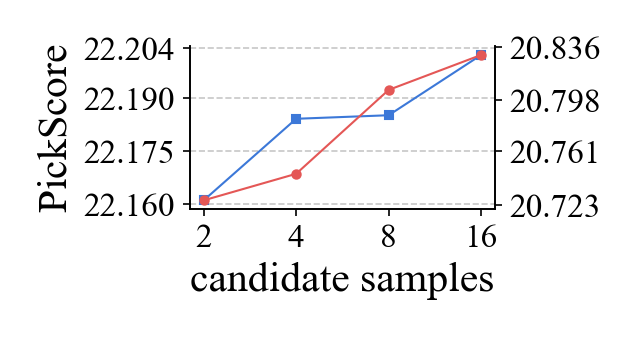}
    \includegraphics[width=0.172\linewidth, trim=5mm 5mm 5mm 5mm, clip]{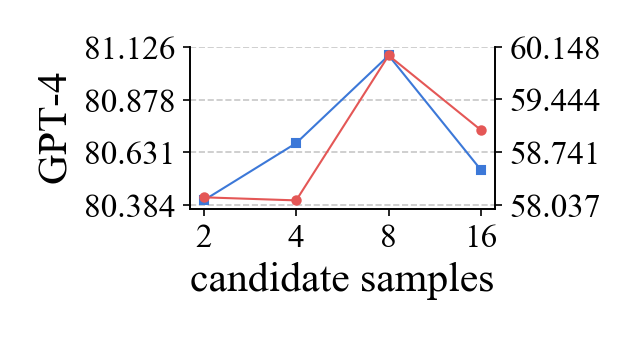}
    \includegraphics[width=0.172\linewidth, trim=5mm 5mm 5mm 5mm, clip]{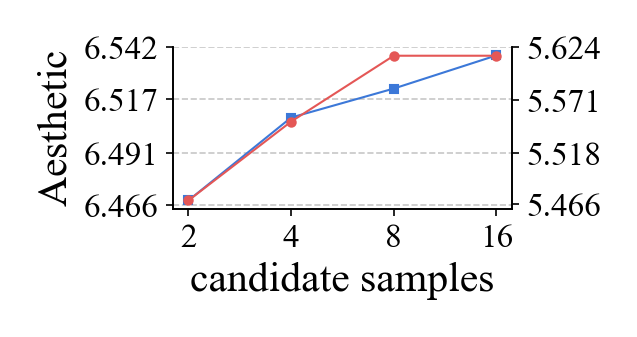}
    \includegraphics[width=0.172\linewidth, trim=5mm 5mm 5mm 5mm, clip]{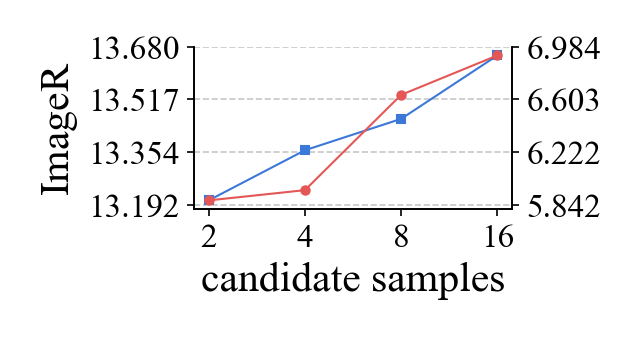}
    \includegraphics[width=0.172\linewidth, trim=5mm 5mm 5mm 5mm, clip]{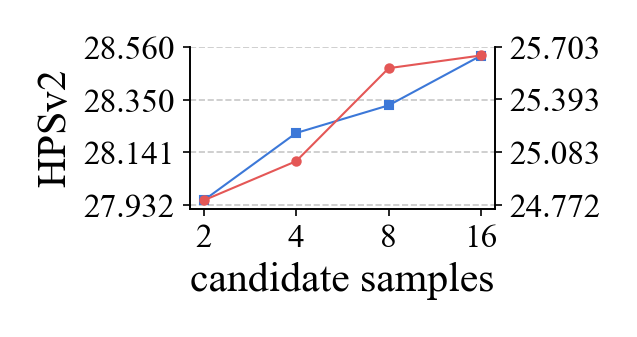}
    \includegraphics[width=0.172\linewidth, trim=5mm 5mm 5mm 5mm, clip]{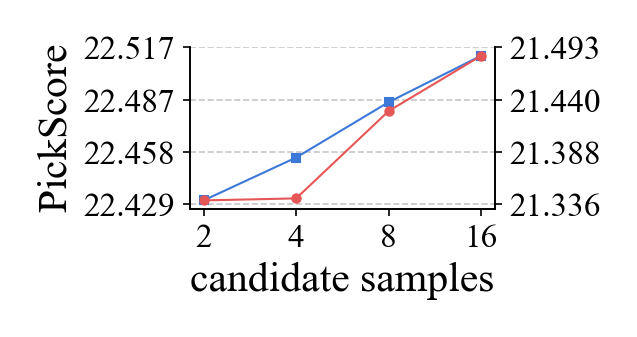}
    \includegraphics[width=0.172\linewidth, trim=5mm 5mm 5mm 5mm, clip]{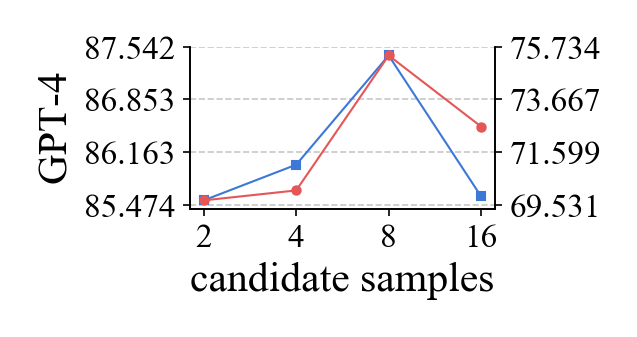}
    \caption{Candidate scaling using HPSv2.}
  \end{subfigure}
  \begin{subfigure}[t]{\linewidth}
    \centering
    \includegraphics[width=0.172\linewidth, trim=5mm 5mm 5mm 5mm, clip]{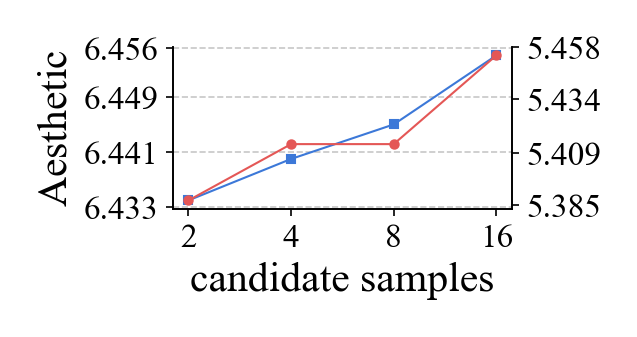}
    \includegraphics[width=0.172\linewidth, trim=5mm 5mm 5mm 5mm, clip]{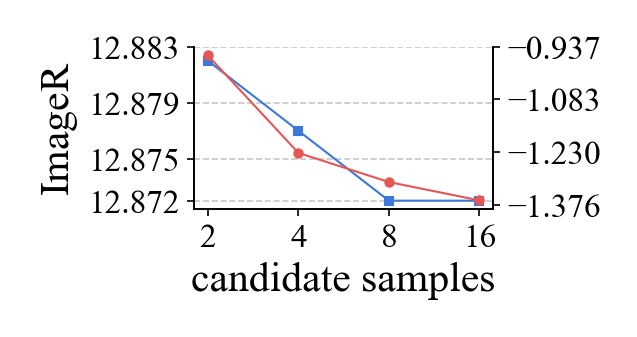}
    \includegraphics[width=0.172\linewidth, trim=5mm 5mm 5mm 5mm, clip]{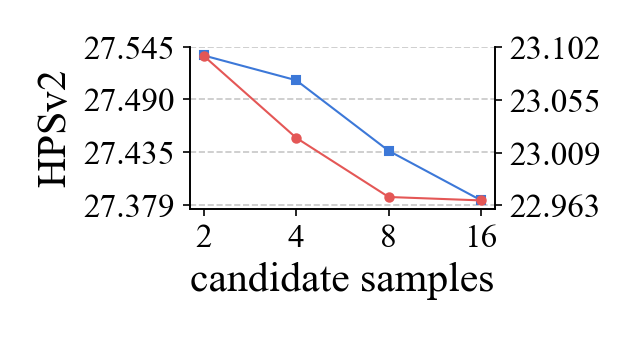}
    \includegraphics[width=0.172\linewidth, trim=5mm 5mm 5mm 5mm, clip]{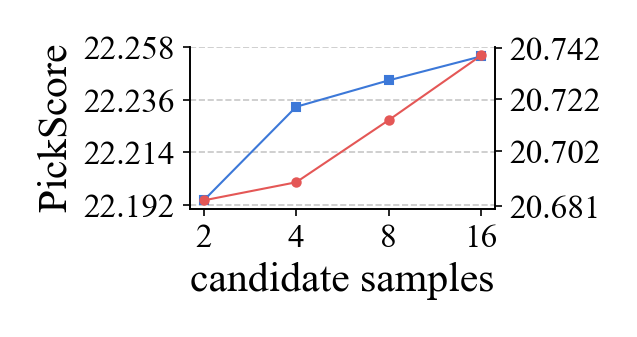}
    \includegraphics[width=0.172\linewidth, trim=5mm 5mm 5mm 5mm, clip]{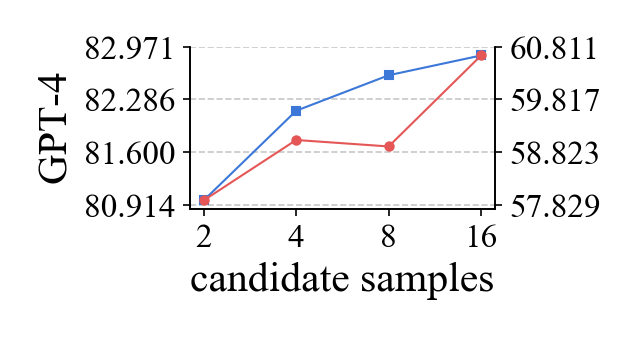}
    \includegraphics[width=0.172\linewidth, trim=5mm 5mm 5mm 5mm, clip]{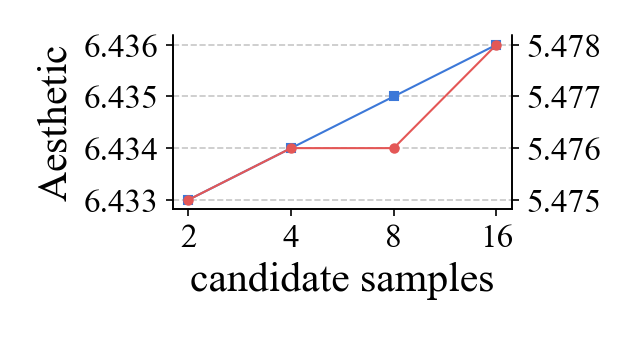}
    \includegraphics[width=0.172\linewidth, trim=5mm 5mm 5mm 5mm, clip]{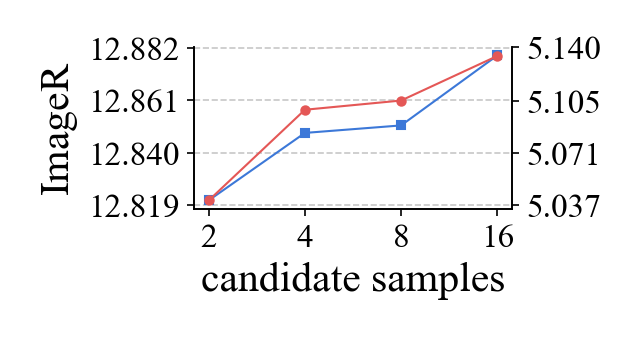}
    \includegraphics[width=0.172\linewidth, trim=5mm 5mm 5mm 5mm, clip]{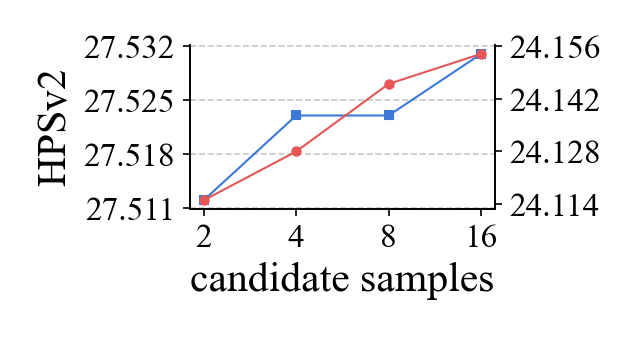}
    \includegraphics[width=0.172\linewidth, trim=5mm 5mm 5mm 5mm, clip]{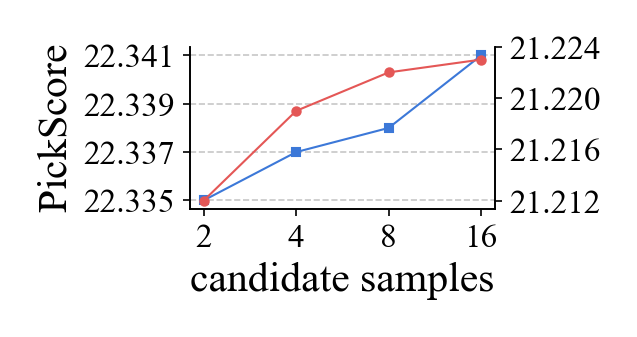}
    \includegraphics[width=0.172\linewidth, trim=5mm 5mm 5mm 5mm, clip]{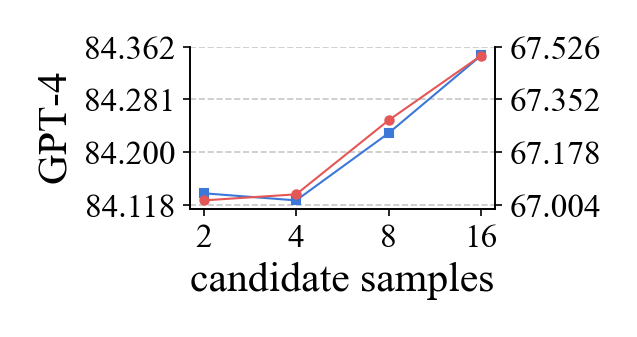}
    \caption{Candidate scaling using PickScore.}
\end{subfigure}
\begin{subfigure}[t]{\linewidth}
    \centering
    \includegraphics[width=0.172\linewidth, trim=5mm 5mm 5mm 5mm, clip]{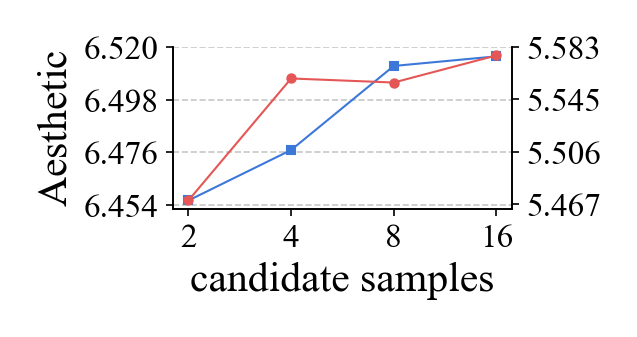}
    \includegraphics[width=0.172\linewidth, trim=5mm 5mm 5mm 5mm, clip]{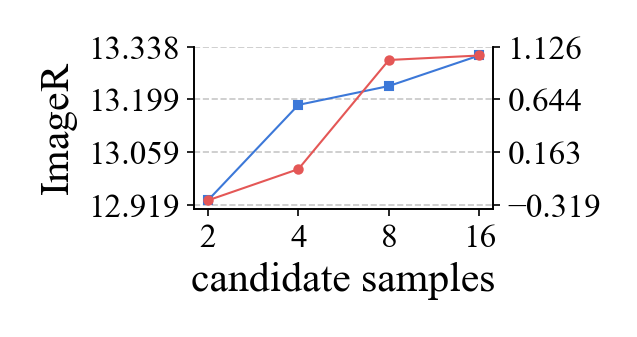}
    \includegraphics[width=0.172\linewidth, trim=5mm 5mm 5mm 5mm, clip]{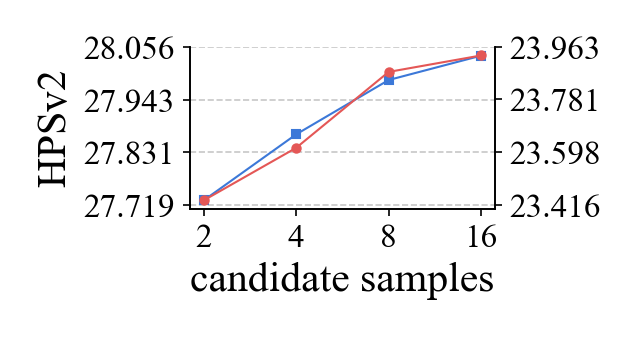}
    \includegraphics[width=0.172\linewidth, trim=5mm 5mm 5mm 5mm, clip]{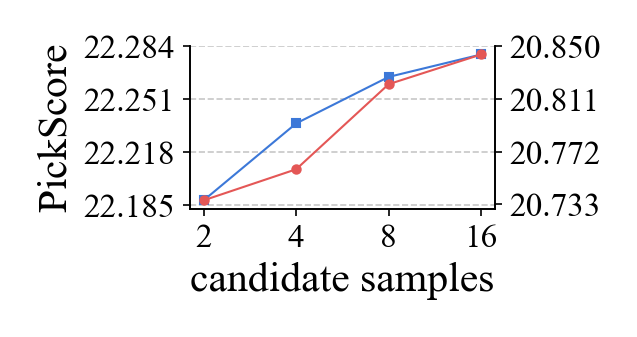}
    \includegraphics[width=0.172\linewidth, trim=5mm 5mm 5mm 5mm, clip]{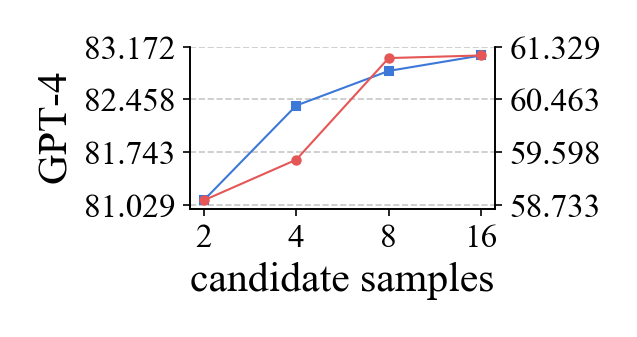}
    \includegraphics[width=0.172\linewidth, trim=5mm 5mm 5mm 5mm, clip]{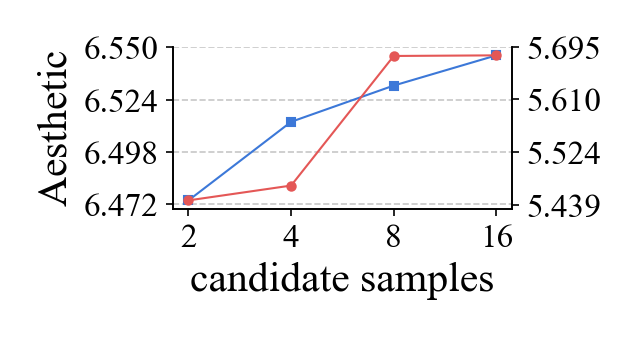}
    \includegraphics[width=0.172\linewidth, trim=5mm 5mm 5mm 5mm, clip]{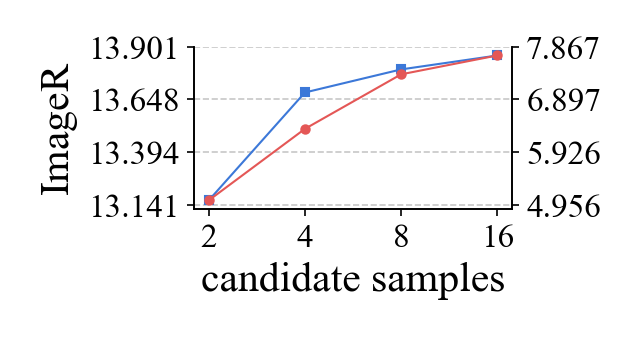}
    \includegraphics[width=0.172\linewidth, trim=5mm 5mm 5mm 5mm, clip]{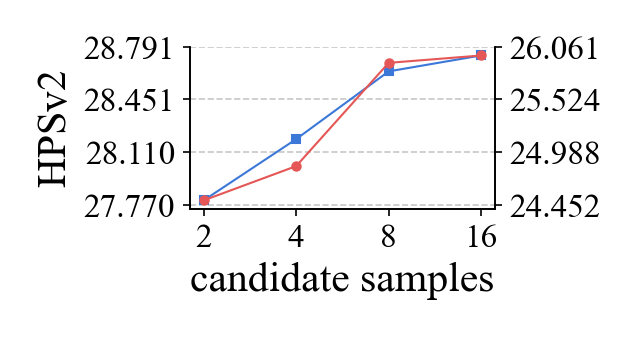}
    \includegraphics[width=0.172\linewidth, trim=5mm 5mm 5mm 5mm, clip]{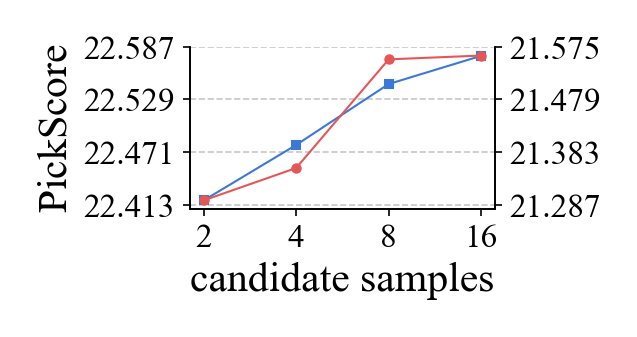}
    \includegraphics[width=0.172\linewidth, trim=5mm 5mm 5mm 5mm, clip]{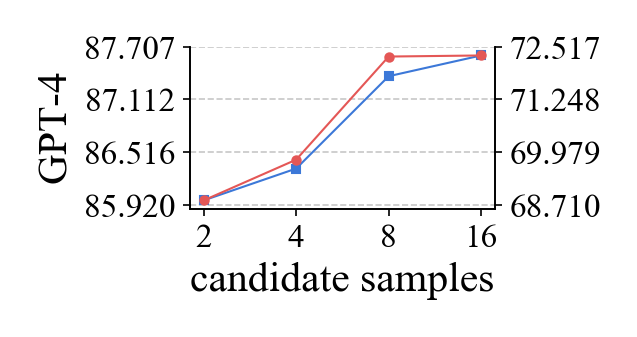}
    \caption{Candidate scaling using Ensemble.}
\end{subfigure}
  \begin{subfigure}[t]{\linewidth}
    \centering
    \includegraphics[width=0.172\linewidth, trim=5mm 5mm 5mm 5mm, clip]{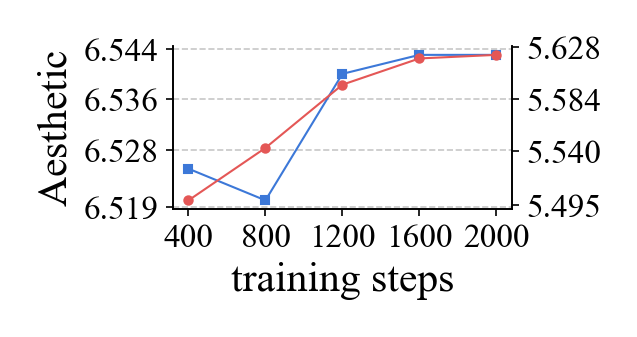}
    \includegraphics[width=0.172\linewidth, trim=5mm 5mm 5mm 5mm, clip]{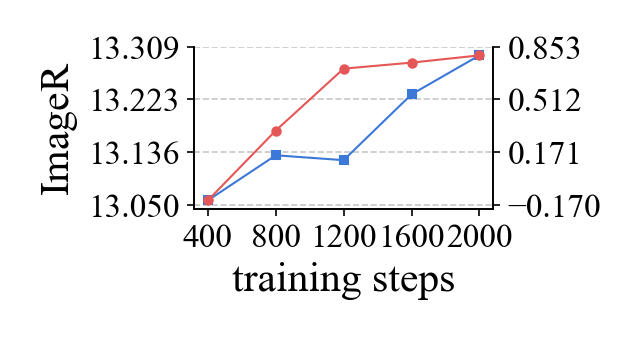}
    \includegraphics[width=0.172\linewidth, trim=5mm 5mm 5mm 5mm, clip]{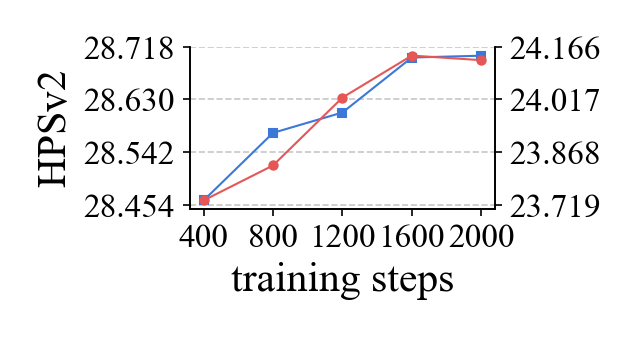}
    \includegraphics[width=0.172\linewidth, trim=5mm 5mm 5mm 5mm, clip]{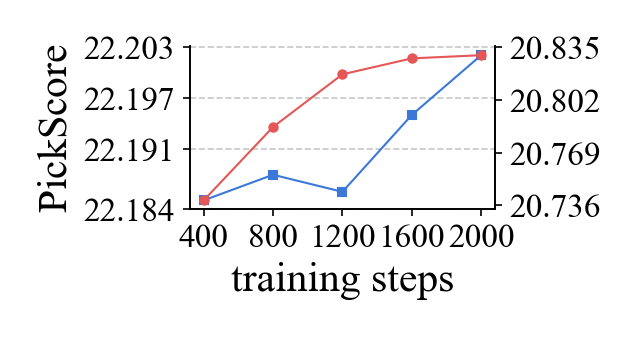}
    \includegraphics[width=0.172\linewidth, trim=5mm 5mm 5mm 5mm, clip]{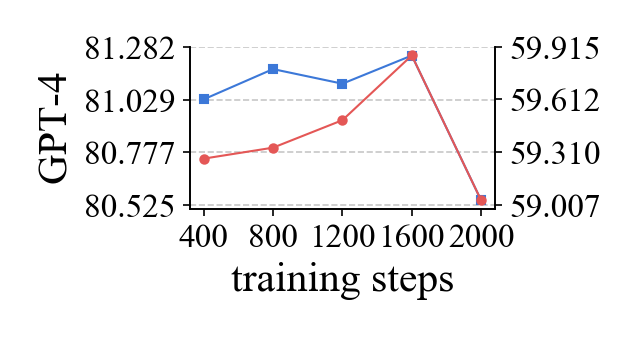}
    \includegraphics[width=0.172\linewidth, trim=5mm 5mm 5mm 5mm, clip]{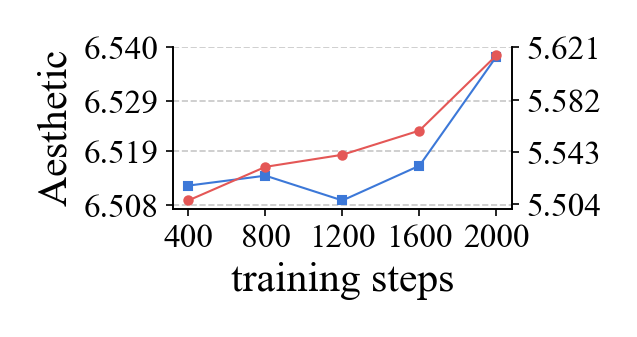}
    \includegraphics[width=0.172\linewidth, trim=5mm 5mm 5mm 5mm, clip]{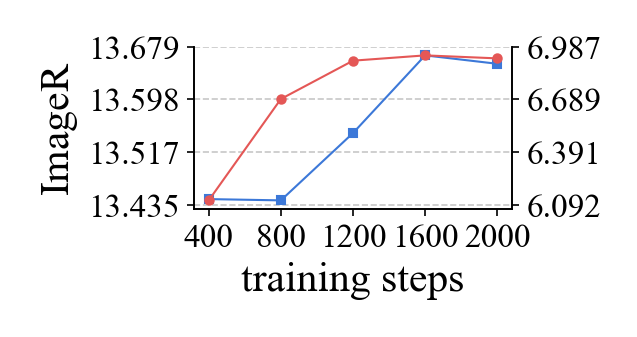}
    \includegraphics[width=0.172\linewidth, trim=5mm 5mm 5mm 5mm, clip]{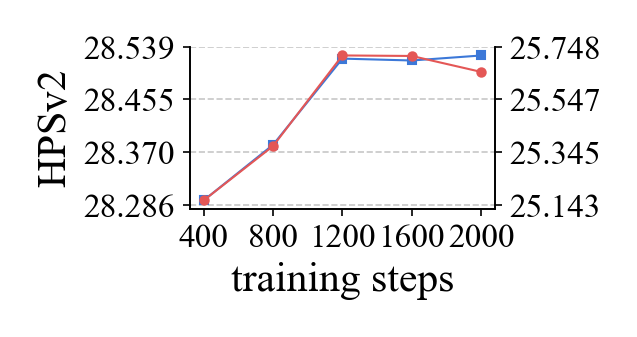}
    \includegraphics[width=0.172\linewidth, trim=5mm 5mm 5mm 5mm, clip]{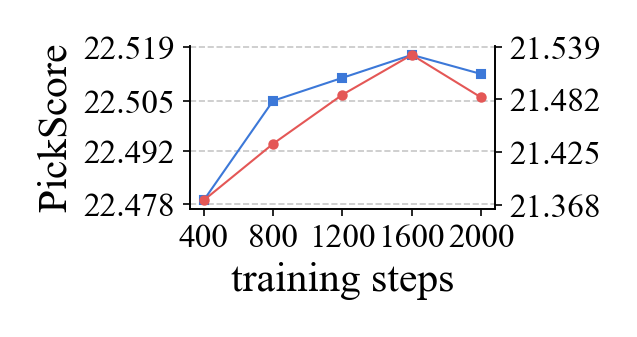}
    \includegraphics[width=0.172\linewidth, trim=5mm 5mm 5mm 5mm, clip]{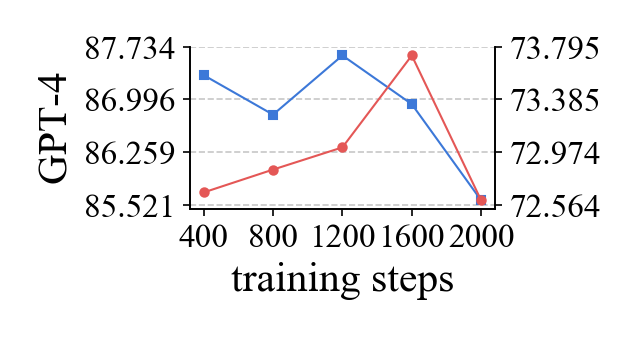}
    \caption{Sample scaling using HPSv2.}
  \end{subfigure}
  \begin{subfigure}[t]{\linewidth}
    \centering
    \includegraphics[width=0.172\linewidth, trim=5mm 5mm 5mm 5mm, clip]{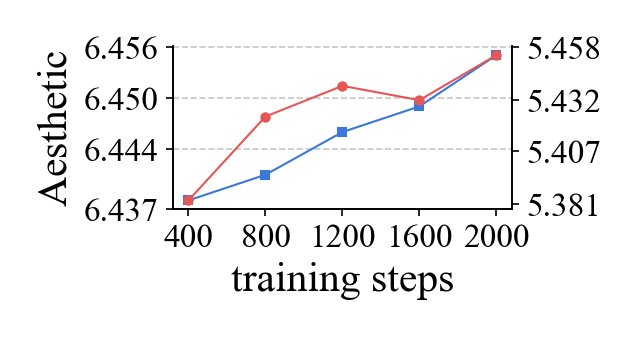}
    \includegraphics[width=0.172\linewidth, trim=5mm 5mm 5mm 5mm, clip]{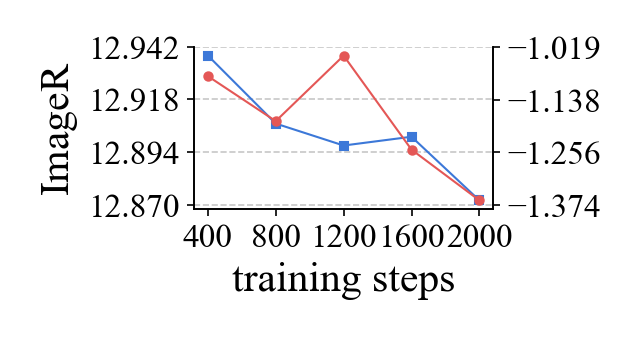}
    \includegraphics[width=0.172\linewidth, trim=5mm 5mm 5mm 5mm, clip]{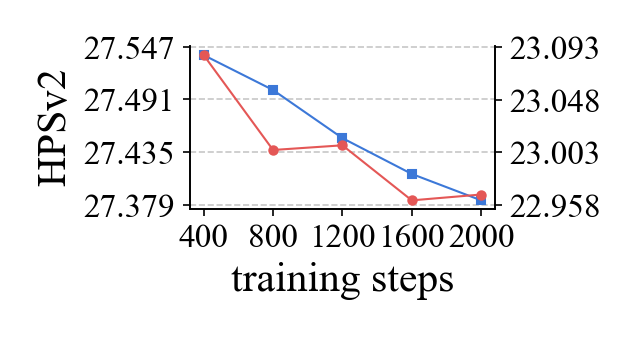}
    \includegraphics[width=0.172\linewidth, trim=5mm 5mm 5mm 5mm, clip]{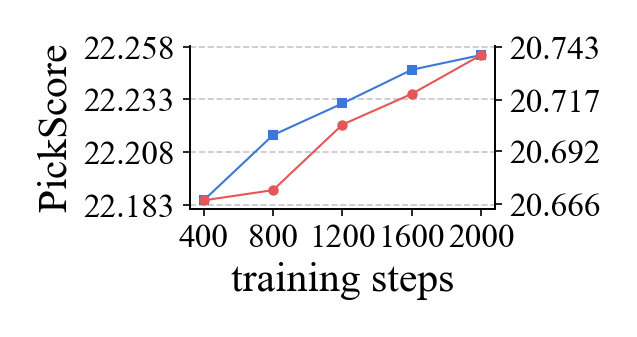}
    \includegraphics[width=0.172\linewidth, trim=5mm 5mm 5mm 5mm, clip]{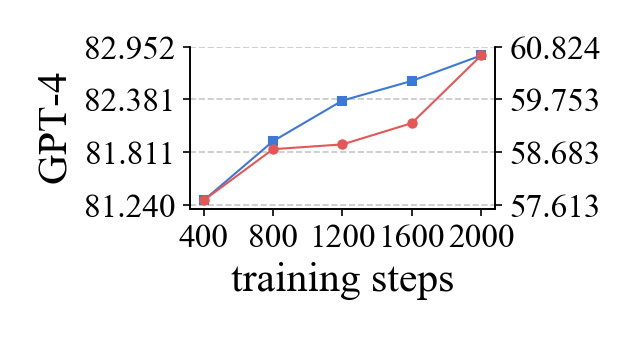}
    \includegraphics[width=0.172\linewidth, trim=5mm 5mm 5mm 5mm, clip]{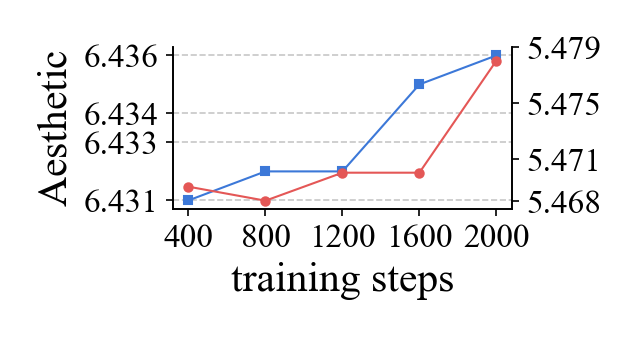}
    \includegraphics[width=0.172\linewidth, trim=5mm 5mm 5mm 5mm, clip]{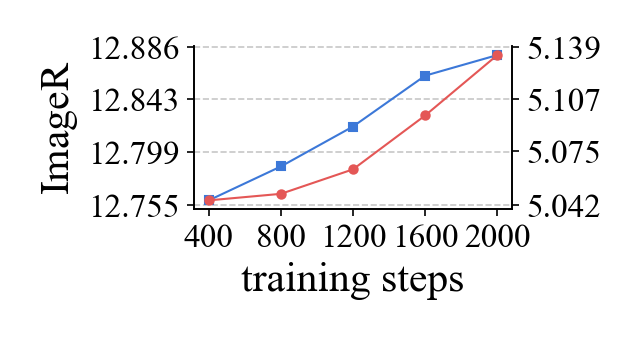}
    \includegraphics[width=0.172\linewidth, trim=5mm 5mm 5mm 5mm, clip]{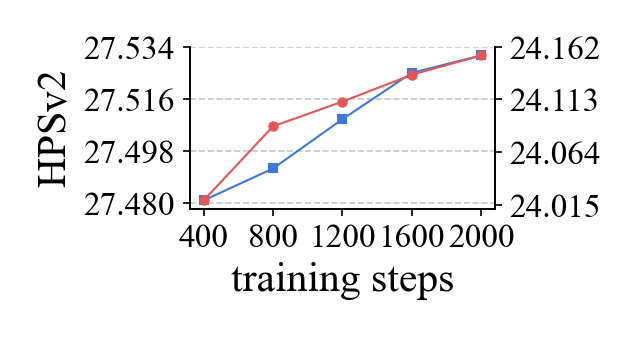}
    \includegraphics[width=0.172\linewidth, trim=5mm 5mm 5mm 5mm, clip]{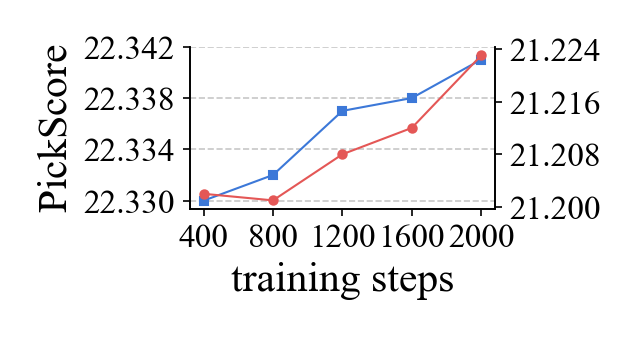}
    \includegraphics[width=0.172\linewidth, trim=5mm 5mm 5mm 5mm, clip]{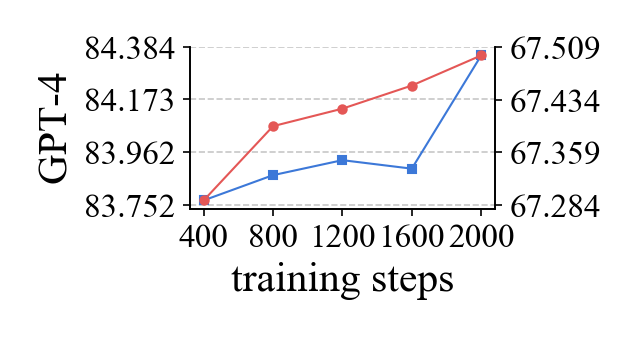}
    \caption{Sample scaling using PickScore.}
\end{subfigure}
\begin{subfigure}[t]{\linewidth}
    \centering
    \includegraphics[width=0.172\linewidth, trim=5mm 5mm 5mm 5mm, clip]{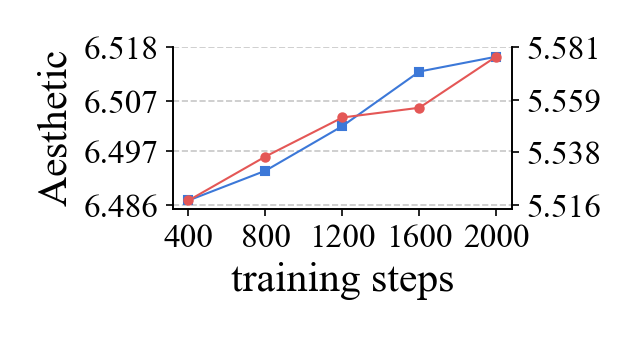}
    \includegraphics[width=0.172\linewidth, trim=5mm 5mm 5mm 5mm, clip]{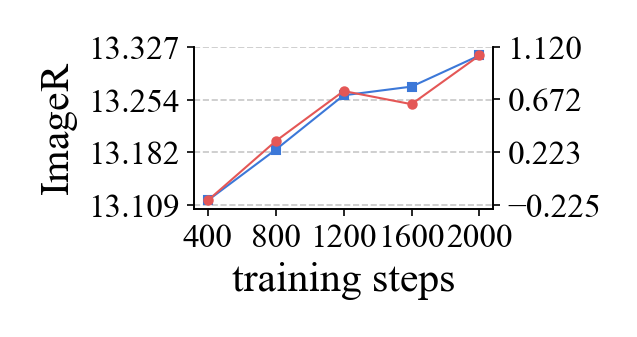}
    \includegraphics[width=0.172\linewidth, trim=5mm 5mm 5mm 5mm, clip]{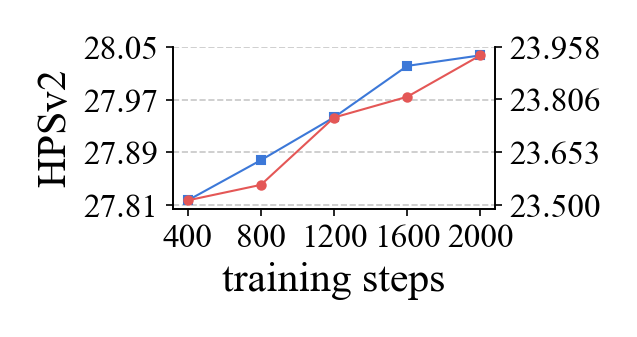}
    \includegraphics[width=0.172\linewidth, trim=5mm 5mm 5mm 5mm, clip]{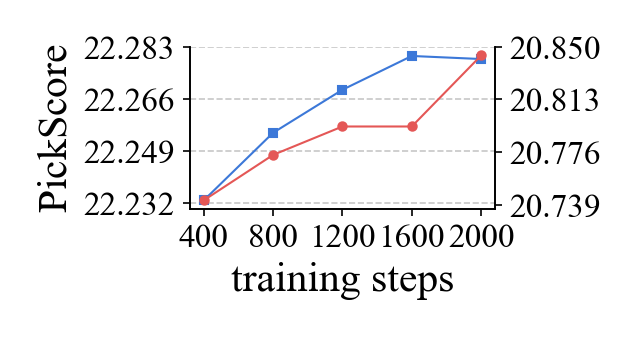}
    \includegraphics[width=0.172\linewidth, trim=5mm 5mm 5mm 5mm, clip]{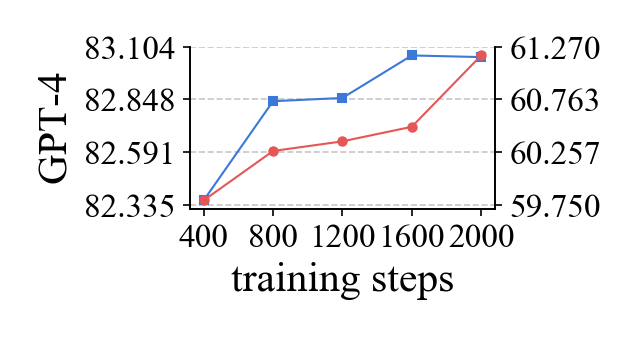}
    \includegraphics[width=0.172\linewidth, trim=5mm 5mm 5mm 5mm, clip]{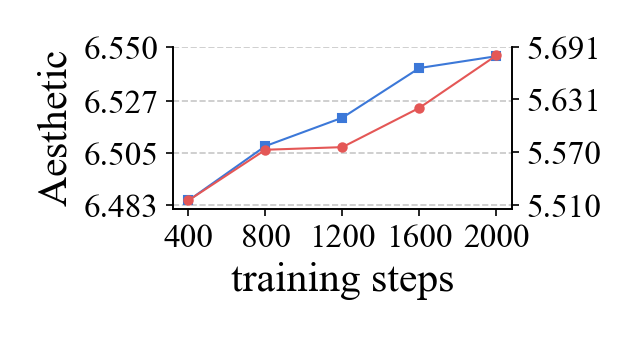}
    \includegraphics[width=0.172\linewidth, trim=5mm 5mm 5mm 5mm, clip]{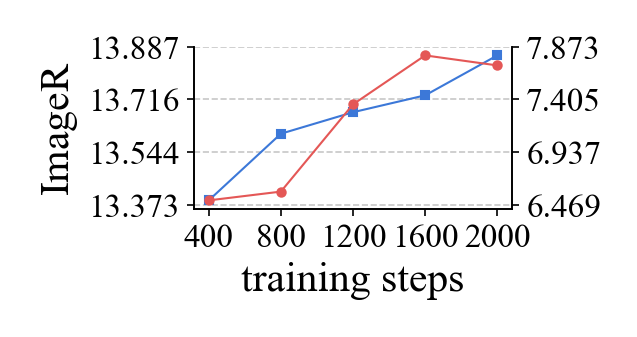}
    \includegraphics[width=0.172\linewidth, trim=5mm 5mm 5mm 5mm, clip]{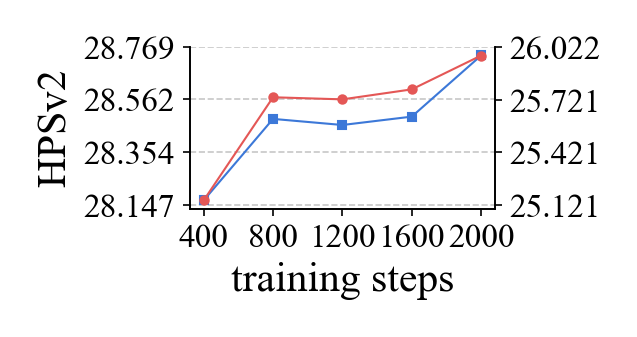}
    \includegraphics[width=0.172\linewidth, trim=5mm 5mm 5mm 5mm, clip]{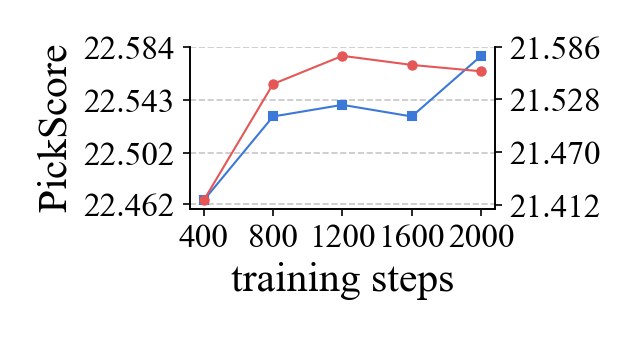}
    \includegraphics[width=0.172\linewidth, trim=5mm 5mm 5mm 5mm, clip]{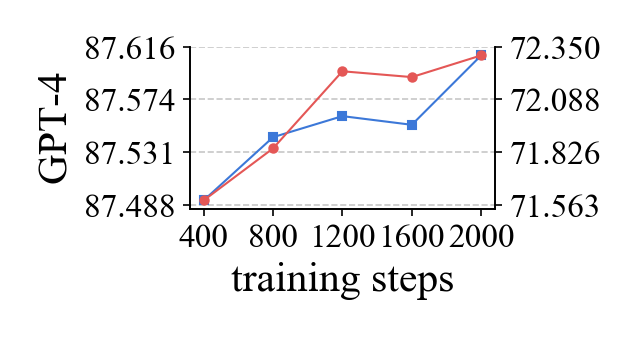}
    \caption{Sample scaling using Ensemble.}
\end{subfigure}
\caption{\textbf{Candidate scaling} (a-c) and \textbf{sample scaling} (d-f) using HPSv2, PickScore, and Ensemble. We employ Aesthetic, ImageR, HPSv2, PickScore, and GPT-4 for evaluation. The first and second row of each sub-figure is based on BrushNet and FLUX.1 Fill, respectively. We use training steps to indicate the consumed samples to align the scaling across models (their batch-sizes are different).}
\label{fig:scaling}
\end{figure*}

The ability of reward models to accurately predict human preferences is critical to the performance of preference alignment algorithms. To evaluate this capability, we apply DPO to the preference data constructed by the reward models and evaluate the model's performance after training. Specifically, based on the popular dataset of BrushData~\cite{BrushNet}, we generate 16 \textbf{candidate} inpainting results with varied random seeds for each prompt and the corresponding masked image. The candidates are scored by the reward models, and the highest-scoring (preferred) and lowest-scoring (dispreferred) samples form preference pairs for DPO training. Following~\cite{Inference-Time-Scaling, unifiedreward}, the reward models are employed to serve two purposes: (1) \textit{providing scores to construct training data}, and (2) \textit{evaluating performance after training}.
All experiments adhere to the same training configurations by default (e.g., a learning rate of 1e-7, $\beta$ of 2000, and 2000 training steps, etc.), with the only variation being the reward model used to score and construct the training data. 
Note that we may encounter an \textbf{oracle reward model}~\cite{Inference-Time-Scaling}, where the same model is used both for data construction and performance evaluation within one experiment. We assess results on two benchmarks, i.e., BrushBench~\cite{BrushNet} and EditBench~\cite{EditBench}.

Here, we introduce a new reward model---\textbf{Ensemble}---that selects preference pairs based on the mean rank across all reward models: {\color{rebuttal_blue}for images generated with different random seeds under the same prompt, we rank the images under each reward model and average the per-model ranks of each image to obtain its Ensemble rank; we then select the highest-ranked image as the winning sample and the lowest-ranked image as the losing sample. The full procedure is summarized in Algorithm~\ref{algo:ensemble}.
{\color{rebuttal_blue}
\begin{algorithm}[t]
\caption{\color{rebuttal_blue}Ensemble Reward via Mean Rank}
\label{algo:ensemble}
\SetKwInOut{KwIn}{Input}
\SetKwInOut{KwOut}{Output}

\KwIn{
  $\mathcal{I}$: set of candidate images generated with different random seeds under the same prompt; \\
  $\mathcal{M}$: set of reward models; \\
  $s[m][i]$: reward score of image $i \in \mathcal{I}$ assigned by reward model $m \in \mathcal{M}$.
}
\KwOut{
  Winning sample $i^{w}$ and losing sample $i^{l}$.
}

\Begin{
  \tcp{Step 1: rank candidates under each reward model (rank $1$ denotes the highest score).}
  \For{$m \in \mathcal{M}$}{
    Sort $\{s[m][i]\}_{i \in \mathcal{I}}$ in descending order and let $\mathrm{rank}[m][i]$ be the position of image $i$ in the sorted list\;
  }
  \tcp{Step 2: average per-model ranks to obtain the Ensemble rank.}
  \For{$i \in \mathcal{I}$}{
    $\mathrm{Ensemble}[i] \leftarrow \dfrac{1}{|\mathcal{M}|} \sum_{m \in \mathcal{M}} \mathrm{rank}[m][i]$\;
  }
  \tcp{Step 3: select the best and worst candidates (a smaller Ensemble value indicates a better candidate).}
  $i^{w} \leftarrow \argmin_{i \in \mathcal{I}} \mathrm{Ensemble}[i]$\;
  $i^{l} \leftarrow \argmax_{i \in \mathcal{I}} \mathrm{Ensemble}[i]$\;
}
\end{algorithm}
}

}

We make \textbf{GPT-4}~\cite{GPT-4} serve as a ``fair'' evaluator by assessing aesthetic quality, structural coherence, and semantic alignment of the results. {\color{rebuttal_blue}
Given GPT-4's~\cite{GPT-4} strong multi-modal understanding capabilities, we use it to evaluate image inpainting results. Specifically, we provide GPT-4 with (1) a system prompt, (2) an input image, (3) the mask to inpaint, (4) an inpainting prompt, (5) and the inpainting result. The prompt for GPT-4 is: \textit{You are an expert in analysis of image inpainting. Please evaluate the image inpainting result based on three criteria: Aesthetic Quality (0–40 points)—visual appeal in color harmony, composition, and style coherence, as well as texture realism and naturalness; Structural Coherence (0–30 points)—preservation of geometric structures and content continuity, and seamlessness at mask boundaries; Semantic Alignment (0–30 points)—faithfulness to the Text Prompt instructions and contextual consistency of added or restored content. For each criterion, provide a sub-score and a 1–2-sentence justification, then compute the total score (0–100).}} 

We report two other results. \textbf{Baseline:} The model's performance prior to DPO training. \textbf{Random:} It involves training with randomly sampled preferred and dispreferred pairs. Results are reported in Table~\ref{tab:brushnet_metrics_8cols_twobench} and Table~\ref{tab:flux_metrics_8col_twobench}. We have the following observations and conclusions.

\textbf{Some reward models are not reliable evaluators.}
It is believed that an accurate and robust reward model should assign high evaluation scores to models trained on its own preference dataset (i.e., the oracle reward model setting). Surprisingly, we find that CLIPScore, VQAScore, and Perception fail to meet this requirement---in~\autoref{tab:flux_metrics_8col_twobench}, their scores can be even lower than those of the baseline or the random-pair baseline.
We hypothesize that the failure of CLIPScore and Perception stems from their large-scale yet potentially coarse contrastive pre-training, and the failure of VQAScore likely arises from its simplistic, VQA-like evaluation approach. In light of this, \textit{we exclude these models from subsequent analyses}.

\textbf{Most reward models provide valid reward scores.} 
Most reward models are capable of offering valid reward scores for preference data construction, as they outperform both the baseline and random selection across most evaluation results---especially GPT-4. Even though CLIPScore and Perception are observed to be less effective at accurately evaluating on small-scale benchmarks, they remain viable when their reward scores are incorporated into larger-scale preference training datasets. In this context, we continue to attribute VQAScore’s limitations to its simple scoring methodology.

\textbf{Reward models share common biases.}
We find that the model trained on HPSv2-constructed data outperforms most competitors when evaluated using public reward models. Specifically, when trained using BrushNet, it ranks first or second in 4 out of 12 evaluations; when trained using FLUX.1 Fill, it ranks first or second in 9 out of 12 evaluations. This pattern aligns with GPT-4’s results when using FLUX.1 Fill but diverges when using BrushNet---under the latter condition, the model is largely outperformed by PickScore. We posit that HPSv2 and many other models may share some common biases, which can potentially lead to reward hacking~\cite{hack}. 

\textbf{Ensemble is an accurate and robust reward model.}
It shows that Ensemble ranks first or second in 11 out of 12 public model evaluations when using BrushNet, and 7 out of 12 when using FLUX.1 Fill. Besides, Ensemble ranks first or second in 3 out of 4 GPT-4's evaluations across both baseline models, demonstrating its robustness in constructing effective preference data. We hypothesize that its versatility arises from the bias of reward models being weakened in Ensemble.


\section{How Scalable are Preference Data?}
\label{sec:scalable}

\begin{figure*}[t]
\begin{subfigure}[t]{\linewidth}
    \centering
    \includegraphics[width=0.11\linewidth, trim=0mm 0mm 0mm 0mm, clip]{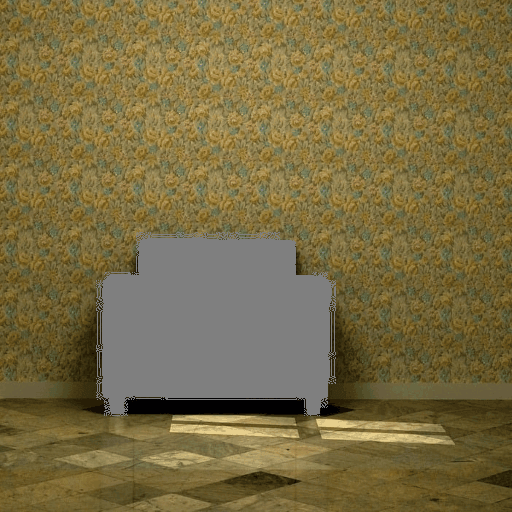}
    \includegraphics[width=0.11\linewidth, trim=0mm 0mm 0mm 0mm, clip]{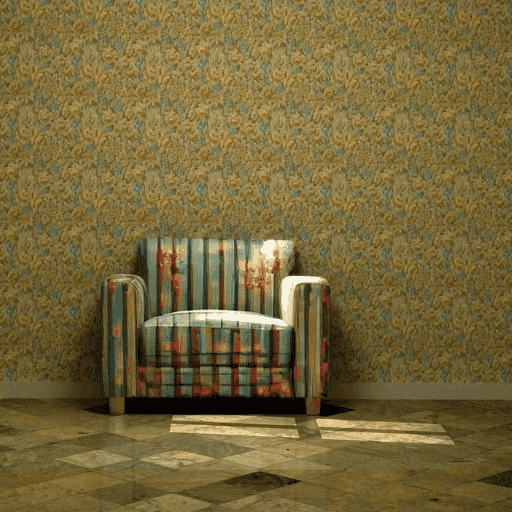}
    \includegraphics[width=0.11\linewidth, trim=0mm 0mm 0mm 0mm, clip]{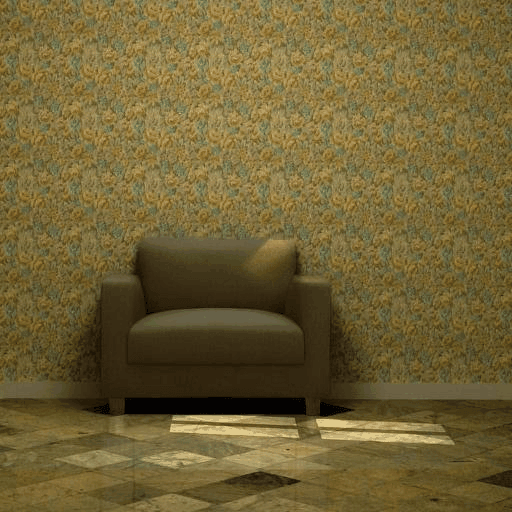}
    \includegraphics[width=0.11\linewidth, trim=0mm 0mm 0mm 0mm, clip]{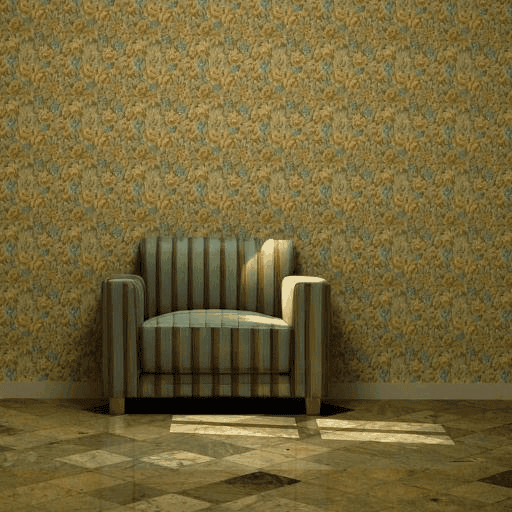}
    \includegraphics[width=0.11\linewidth, trim=0mm 0mm 0mm 0mm, clip]{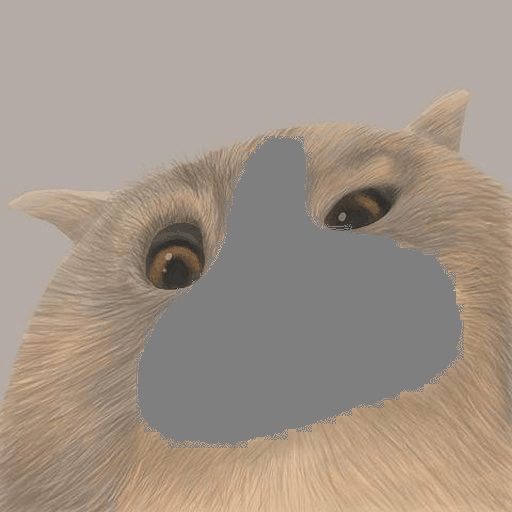}
    \includegraphics[width=0.11\linewidth, trim=0mm 0mm 0mm 0mm, clip]{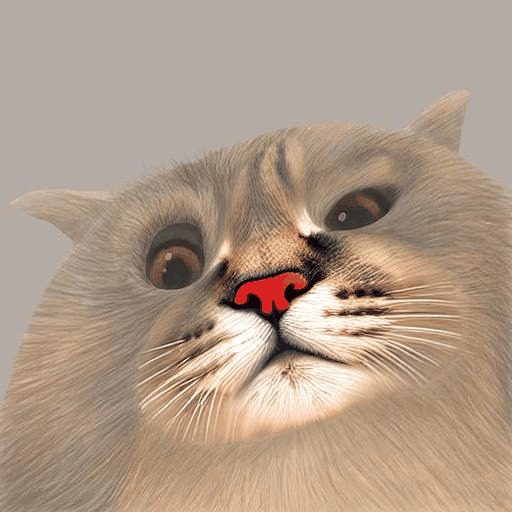}
    \includegraphics[width=0.11\linewidth, trim=0mm 0mm 0mm 0mm, clip]{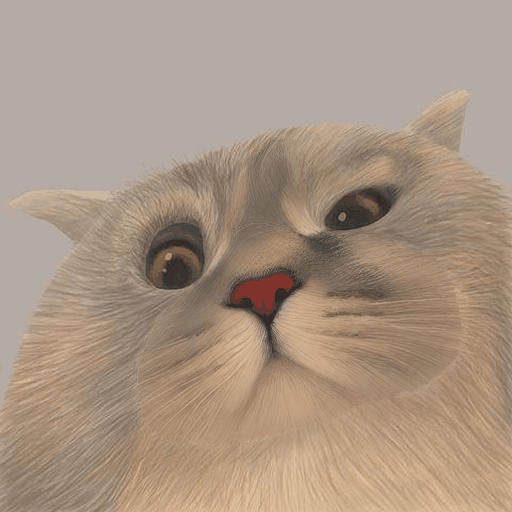}
    \includegraphics[width=0.11\linewidth, trim=0mm 0mm 0mm 0mm, clip]{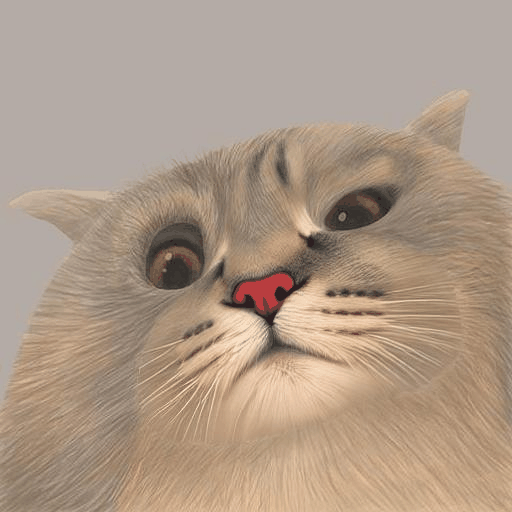}
    \includegraphics[width=0.11\linewidth, trim=0mm 0mm 0mm 0mm, clip]{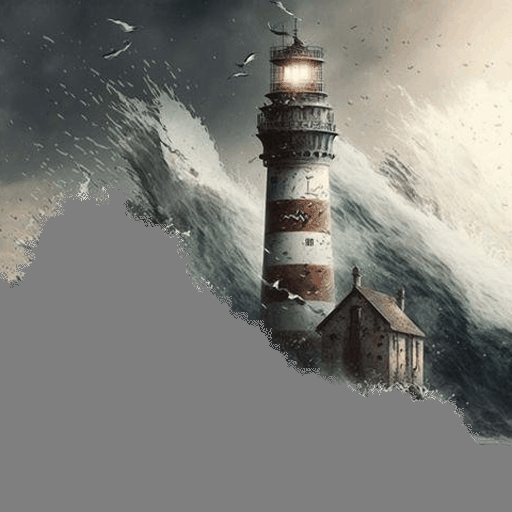}
    \includegraphics[width=0.11\linewidth, trim=0mm 0mm 0mm 0mm, clip]{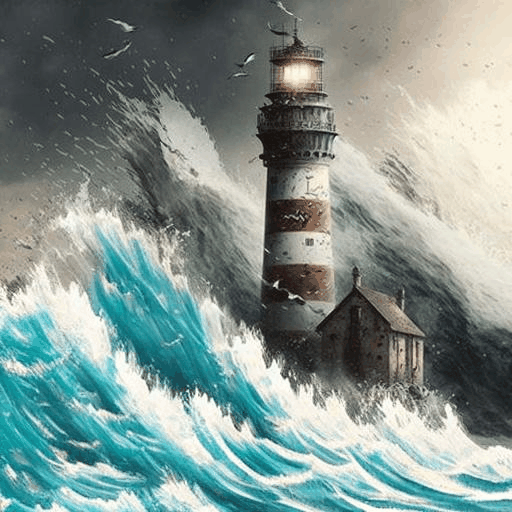}
    \includegraphics[width=0.11\linewidth, trim=0mm 0mm 0mm 0mm, clip]{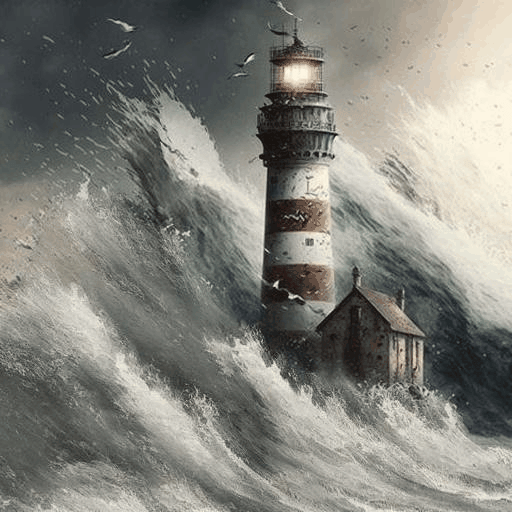}
    \includegraphics[width=0.11\linewidth, trim=0mm 0mm 0mm 0mm, clip]{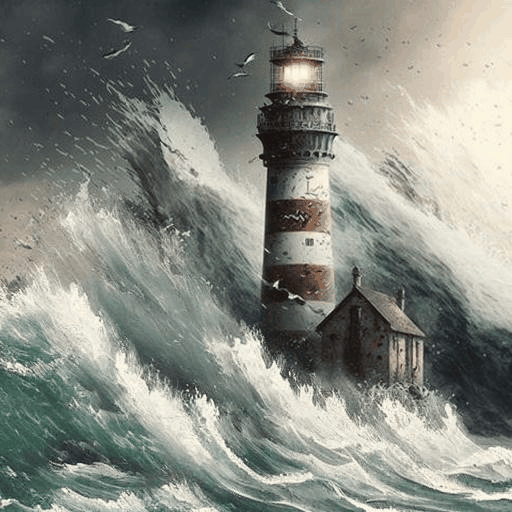}
    \includegraphics[width=0.11\linewidth, trim=0mm 0mm 0mm 0mm, clip]{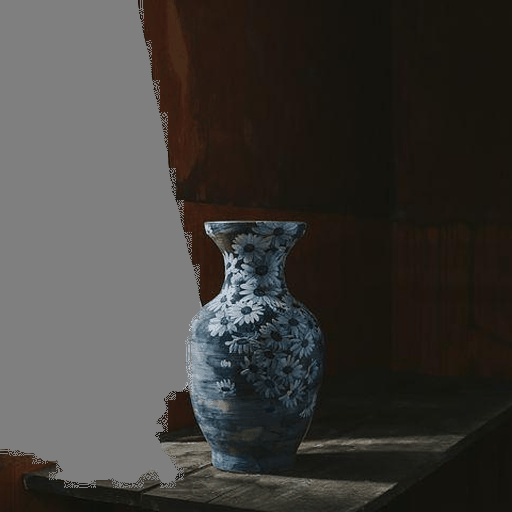}
    \includegraphics[width=0.11\linewidth, trim=0mm 0mm 0mm 0mm, clip]{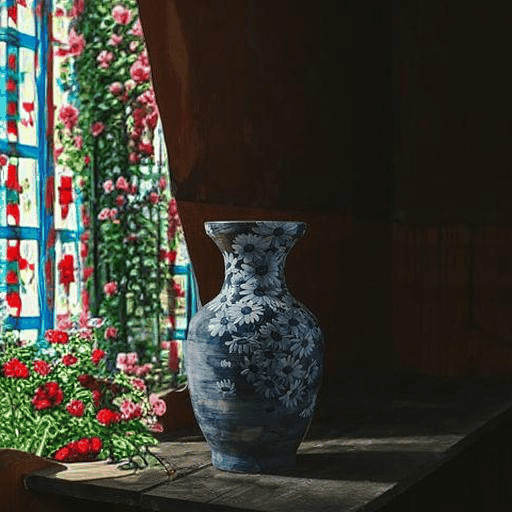}
    \includegraphics[width=0.11\linewidth, trim=0mm 0mm 0mm 0mm, clip]{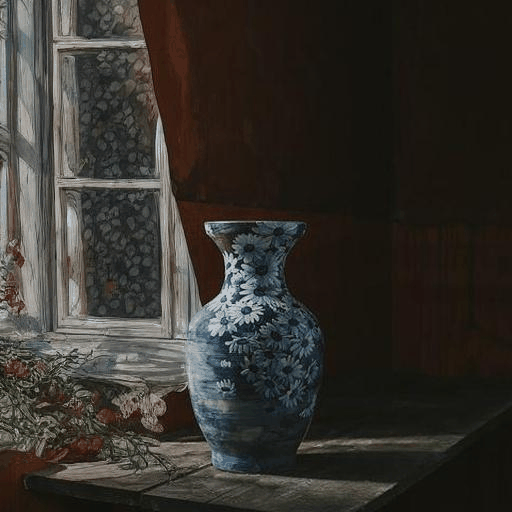}
    \includegraphics[width=0.11\linewidth, trim=0mm 0mm 0mm 0mm, clip]{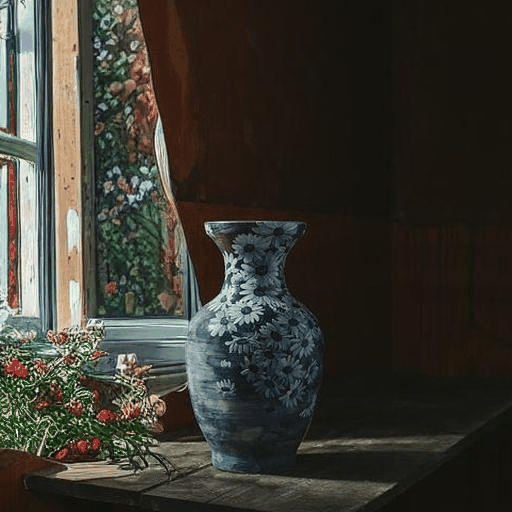}
    \caption{Examples from models trained using \textbf{BrushNet}.}
\end{subfigure}
\begin{subfigure}[t]{\linewidth}
    \centering
    \includegraphics[width=0.11\linewidth, trim=0mm 0mm 0mm 0mm, clip]{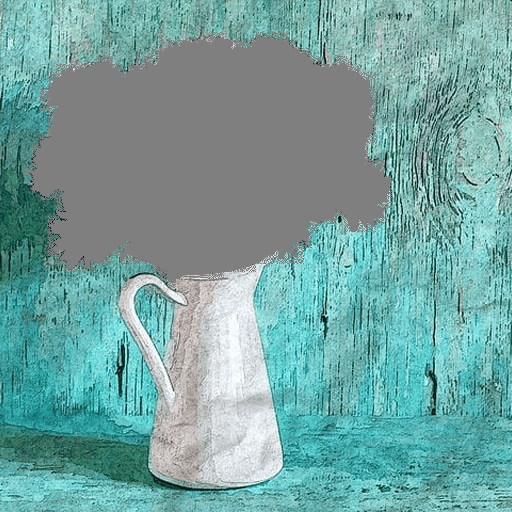}
    \includegraphics[width=0.11\linewidth, trim=0mm 0mm 0mm 0mm, clip]{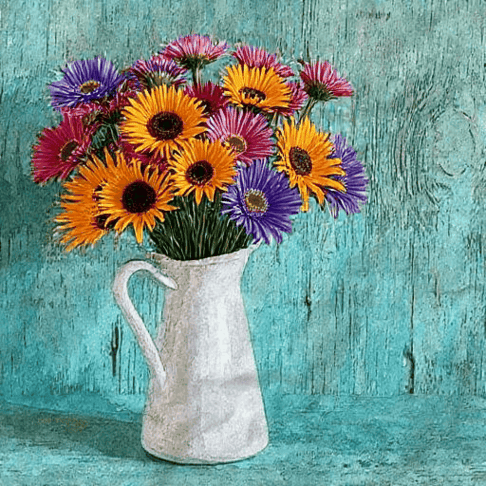}
    \includegraphics[width=0.11\linewidth, trim=0mm 0mm 0mm 0mm, clip]{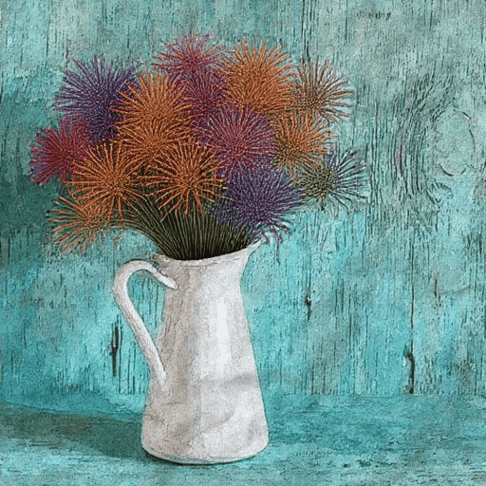}
    \includegraphics[width=0.11\linewidth, trim=0mm 0mm 0mm 0mm, clip]{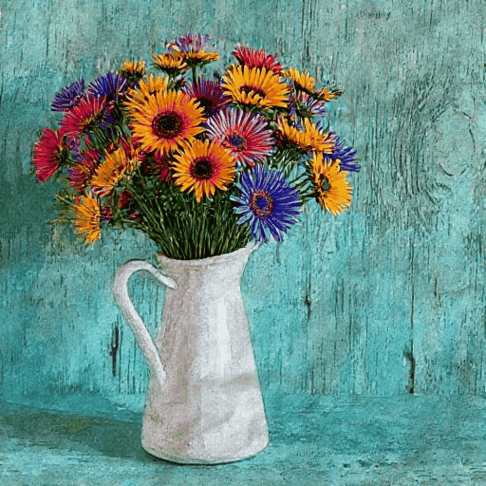}
    \includegraphics[width=0.11\linewidth, trim=0mm 0mm 0mm 0mm, clip]{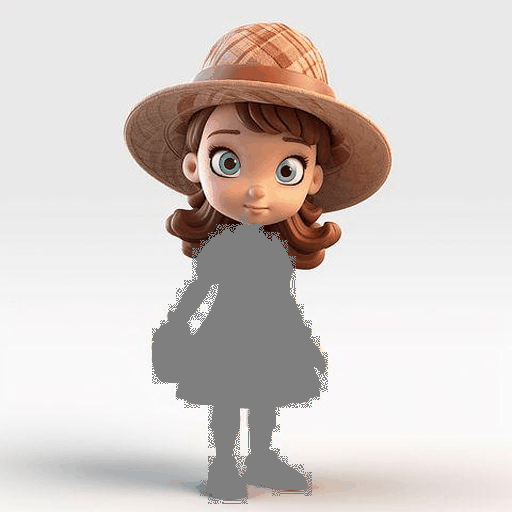}
    \includegraphics[width=0.11\linewidth, trim=0mm 0mm 0mm 0mm, clip]{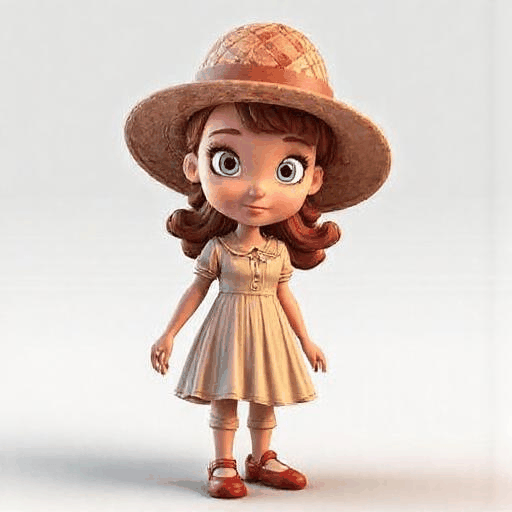}
    \includegraphics[width=0.11\linewidth, trim=0mm 0mm 0mm 0mm, clip]{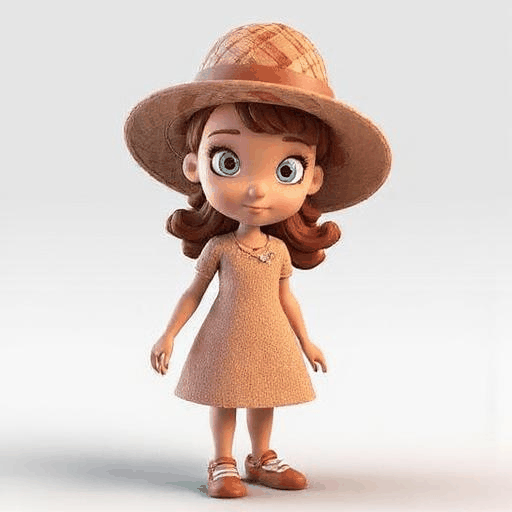}
    \includegraphics[width=0.11\linewidth, trim=0mm 0mm 0mm 0mm, clip]{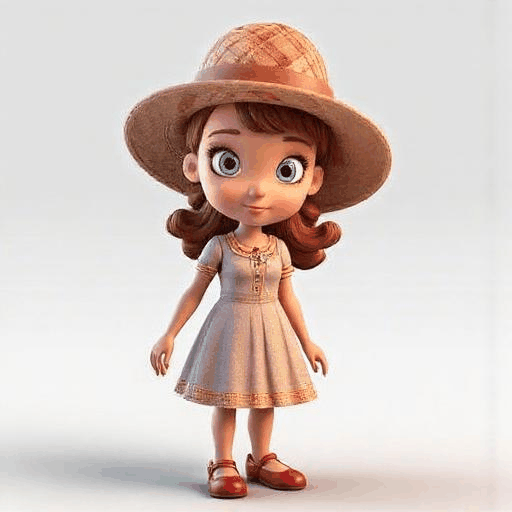}
    \includegraphics[width=0.11\linewidth, trim=0mm 0mm 0mm 0mm, clip]{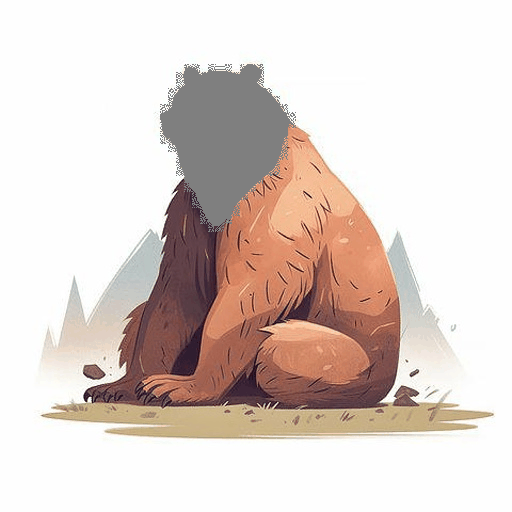}
    \includegraphics[width=0.11\linewidth, trim=0mm 0mm 0mm 0mm, clip]{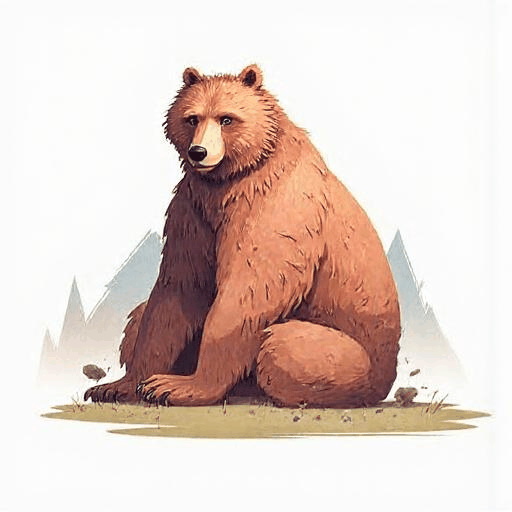}
    \includegraphics[width=0.11\linewidth, trim=0mm 0mm 0mm 0mm, clip]{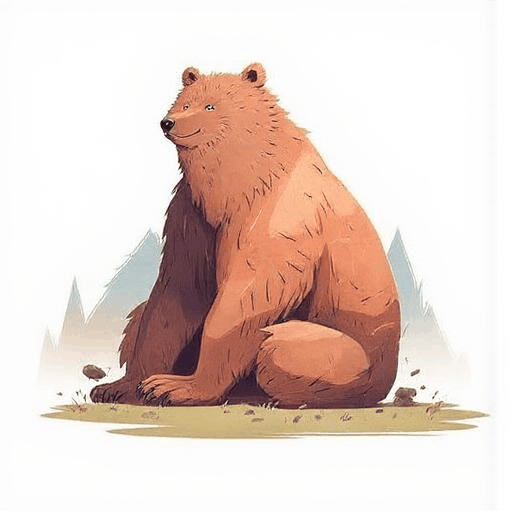}
    \includegraphics[width=0.11\linewidth, trim=0mm 0mm 0mm 0mm, clip]{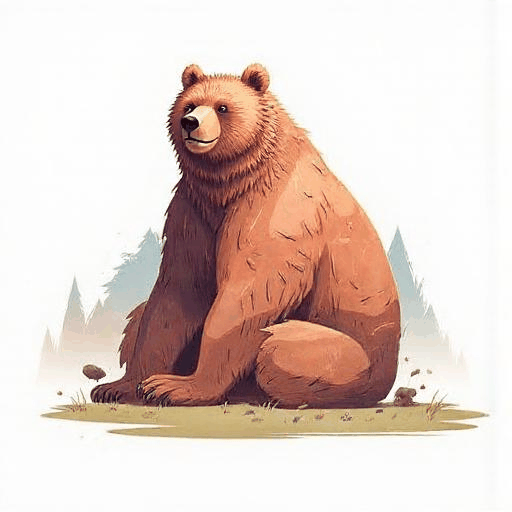}
    \includegraphics[width=0.11\linewidth, trim=0mm 0mm 0mm 0mm, clip]{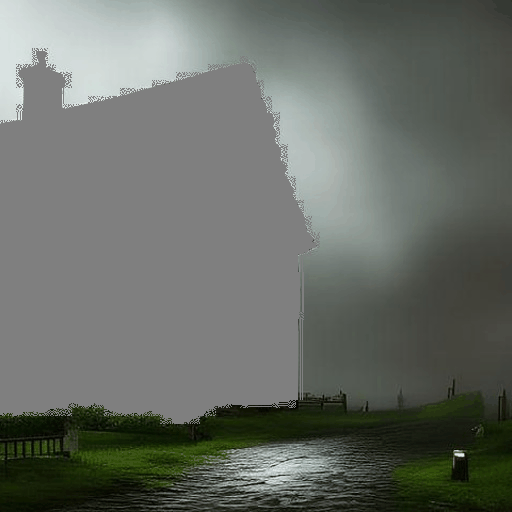}
    \includegraphics[width=0.11\linewidth, trim=0mm 0mm 0mm 0mm, clip]{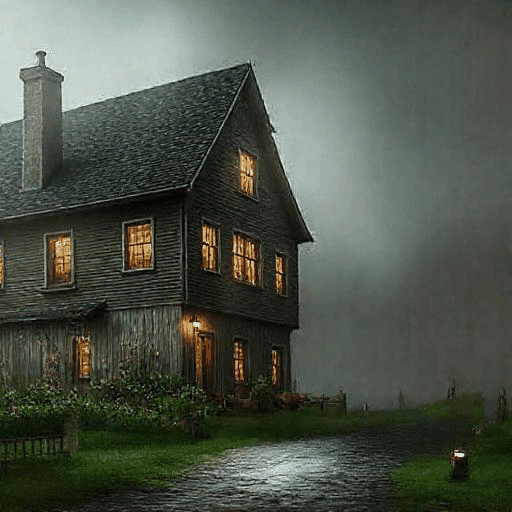}
    \includegraphics[width=0.11\linewidth, trim=0mm 0mm 0mm 0mm, clip]{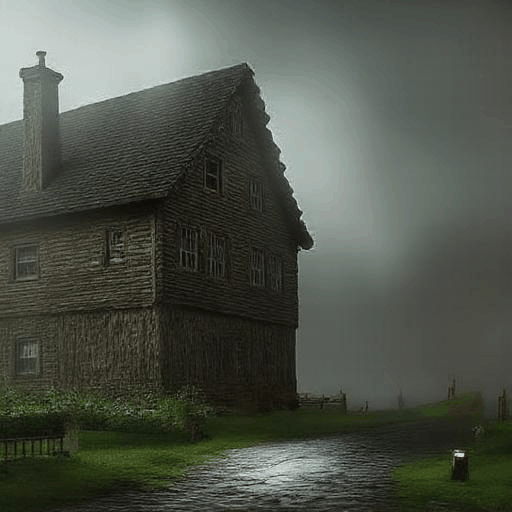}
    \includegraphics[width=0.11\linewidth, trim=0mm 0mm 0mm 0mm, clip]{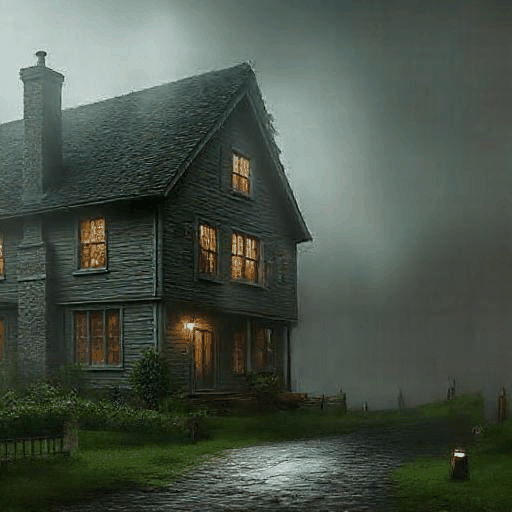}
    \caption{Examples from models trained using \textbf{FLUX.1 Fill}.}
\end{subfigure}
\caption{\textbf{Bias studies.} In each sub-figure, the four images (from left to right) display: the \textit{masked image}, followed by inpainting results from models trained using \textit{HPSv2}, \textit{PickScore}, and \textit{Ensemble}. We omit text prompts for brevity. Zoom in to see details. Find more examples in the Appendix.}
\label{fig:bias}
\end{figure*}

\begin{figure*}[t]
\begin{subfigure}[t]{\linewidth}
\centering
	\includegraphics[width=0.1\linewidth, trim=0mm 0mm 0mm 0mm, clip]{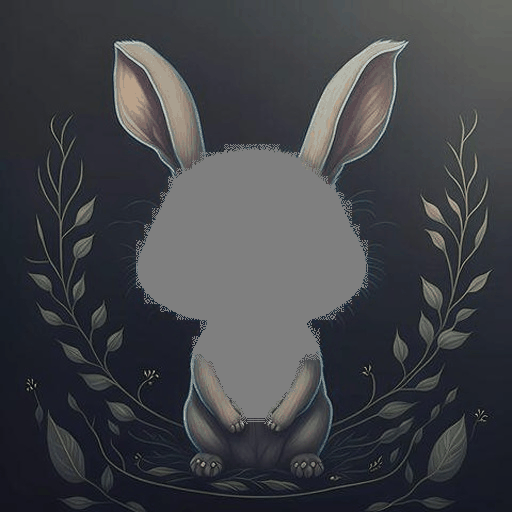}
	\includegraphics[width=0.1\linewidth, trim=0mm 0mm 0mm 0mm, clip]{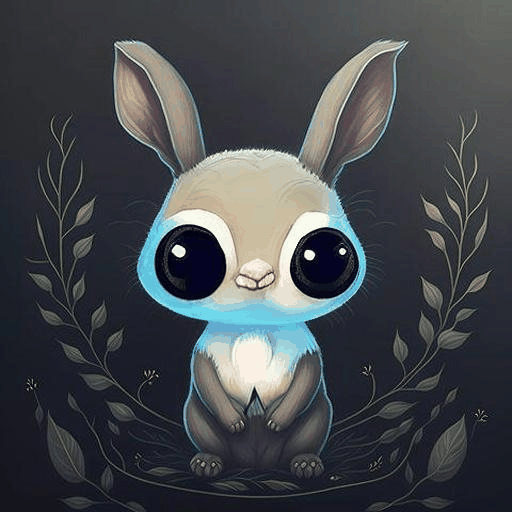}
	\includegraphics[width=0.1\linewidth, trim=0mm 0mm 0mm 0mm, clip]{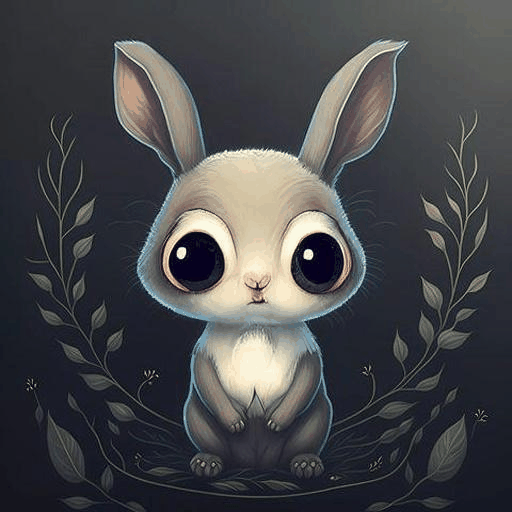}
	\includegraphics[width=0.1\linewidth, trim=0mm 0mm 0mm 0mm, clip]{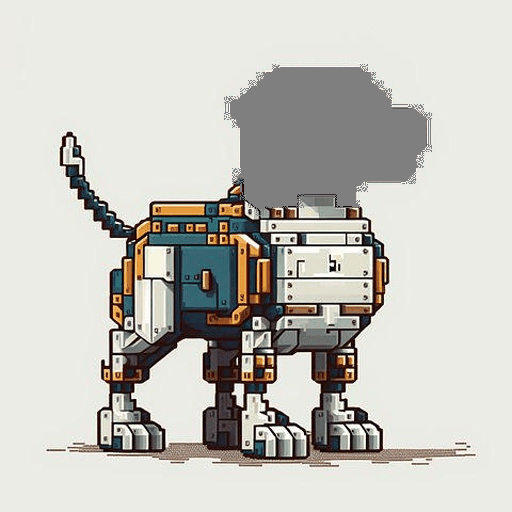}
	\includegraphics[width=0.1\linewidth, trim=0mm 0mm 0mm 0mm, clip]{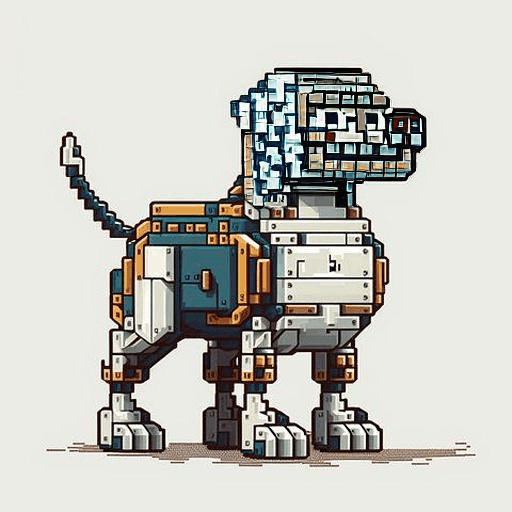}
	\includegraphics[width=0.1\linewidth, trim=0mm 0mm 0mm 0mm, clip]{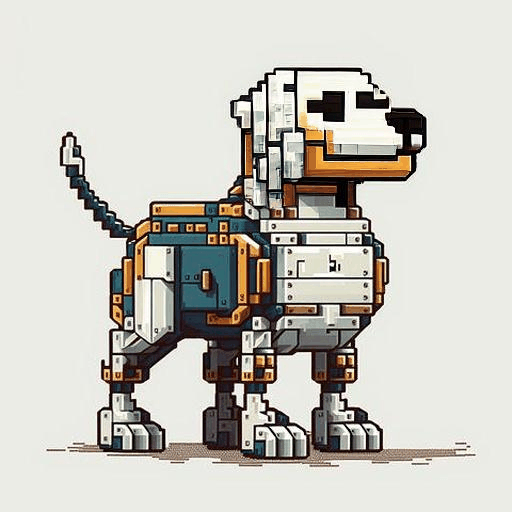}
	\includegraphics[width=0.1\linewidth, trim=0mm 0mm 0mm 0mm, clip]{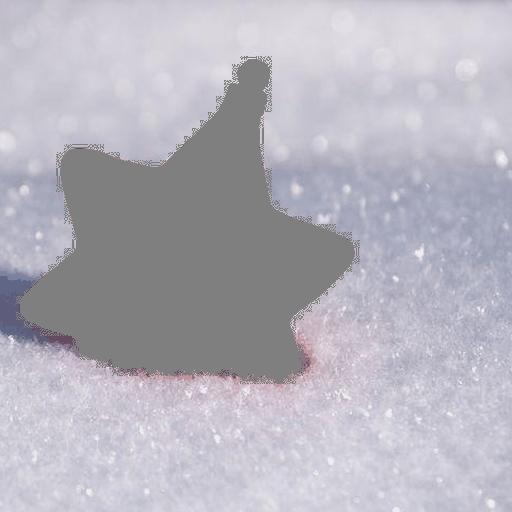}
	\includegraphics[width=0.1\linewidth, trim=0mm 0mm 0mm 0mm, clip]{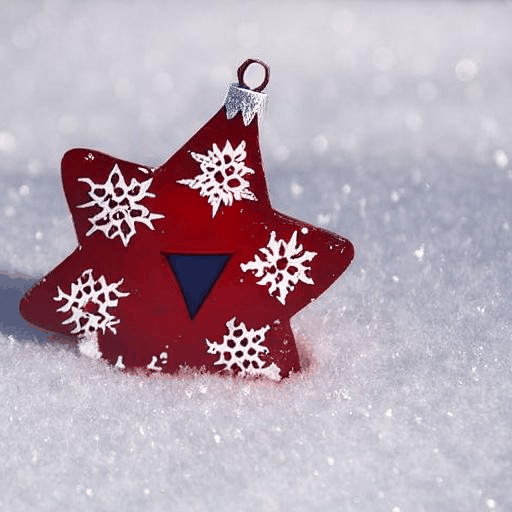}
	\includegraphics[width=0.1\linewidth, trim=0mm 0mm 0mm 0mm, clip]{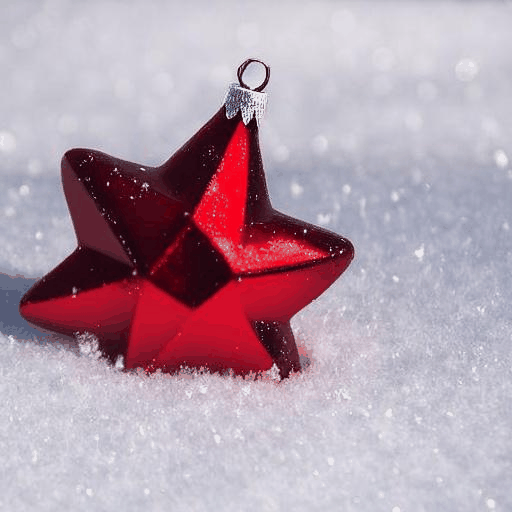}
\caption{Examples from models trained using \textbf{BrushNet}.}
\end{subfigure}
\begin{subfigure}[t]{\linewidth}
\centering
	\includegraphics[width=0.1\linewidth, trim=0mm 0mm 0mm 0mm, clip]{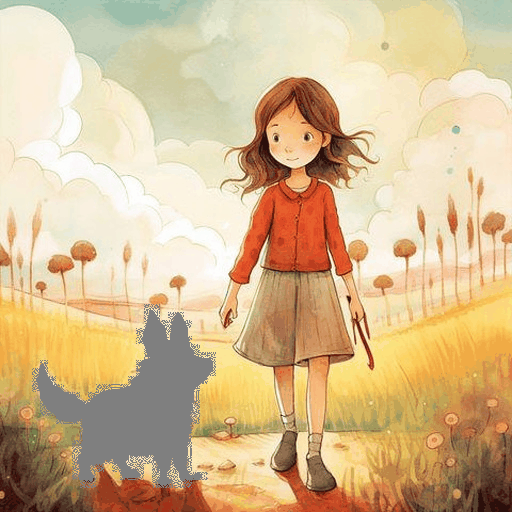}
	\includegraphics[width=0.1\linewidth, trim=0mm 0mm 0mm 0mm, clip]{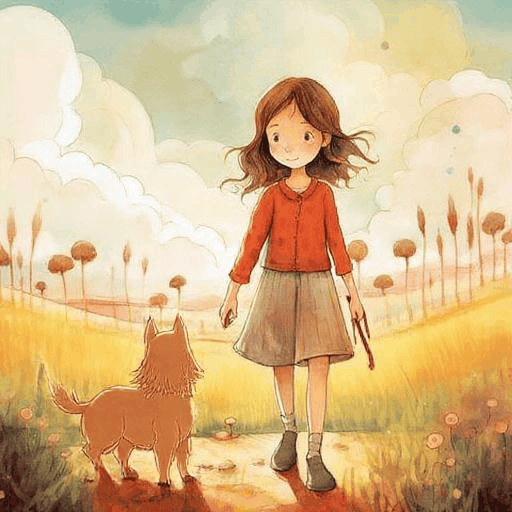}
	\includegraphics[width=0.1\linewidth, trim=0mm 0mm 0mm 0mm, clip]{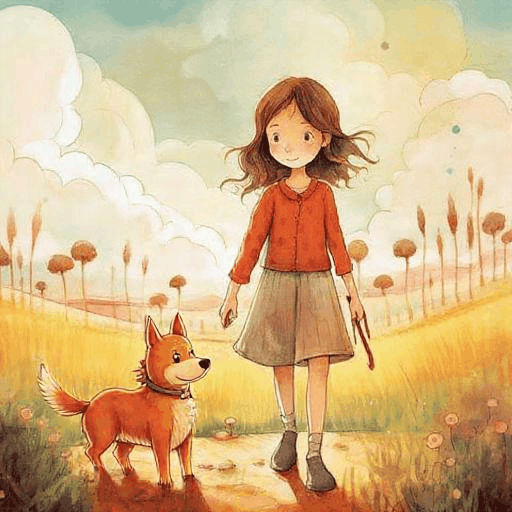}
	\includegraphics[width=0.1\linewidth, trim=0mm 0mm 0mm 0mm, clip]{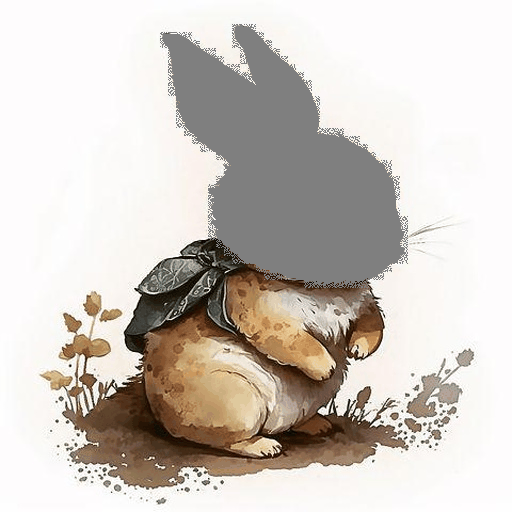}
	\includegraphics[width=0.1\linewidth, trim=0mm 0mm 0mm 0mm, clip]{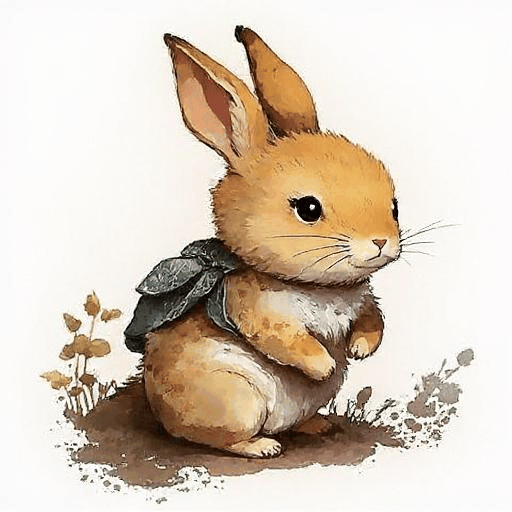}
	\includegraphics[width=0.1\linewidth, trim=0mm 0mm 0mm 0mm, clip]{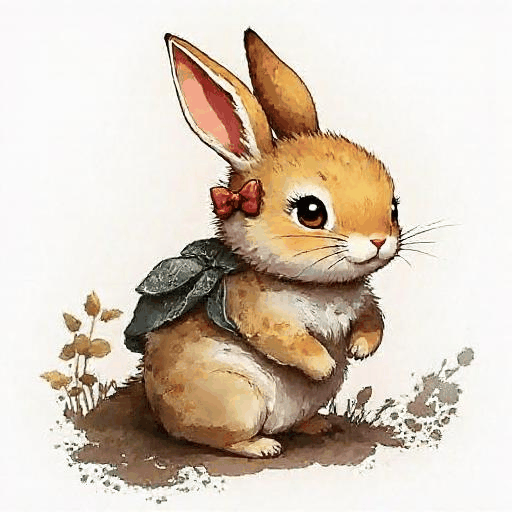}
	\includegraphics[width=0.1\linewidth, trim=0mm 0mm 0mm 0mm, clip]{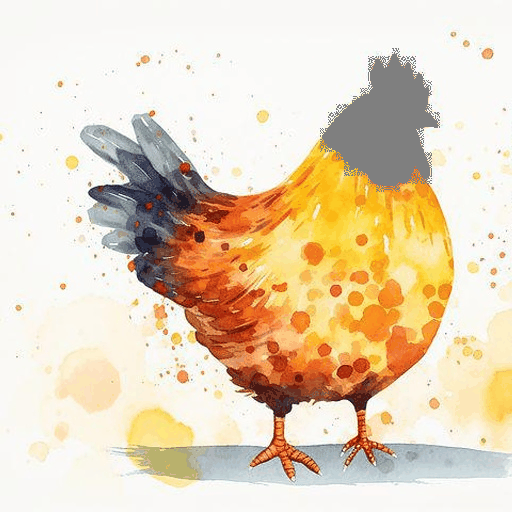}
	\includegraphics[width=0.1\linewidth, trim=0mm 0mm 0mm 0mm, clip]{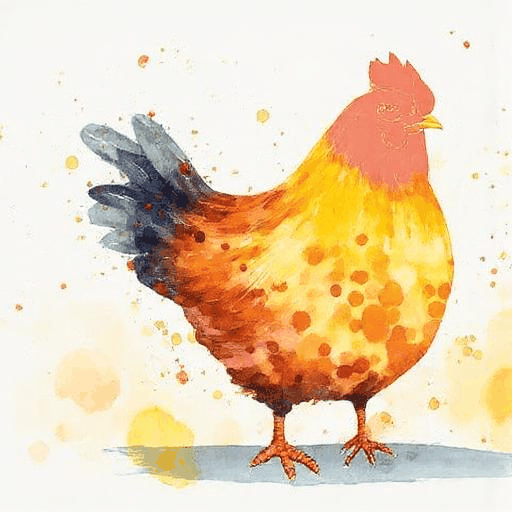}
	\includegraphics[width=0.1\linewidth, trim=0mm 0mm 0mm 0mm, clip]{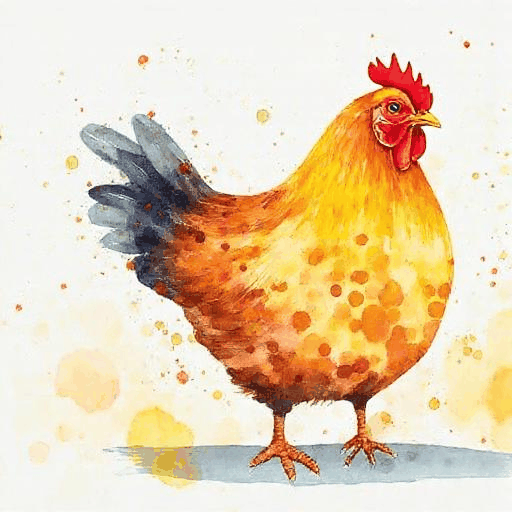}
\caption{Examples from models trained using \textbf{FLUX.1 Fill}.}
\end{subfigure}
\caption{\textbf{Qualitative results of ablations.} In each sub-figure, the three images (from left to right) display: the \textit{masked image}, followed by inpainting results from \textit{baseline models} and \textit{baseline models + preference alignment using Ensemble}. We omit text prompts for brevity. Zoom in to see details.}
\label{fig:baseline}
\end{figure*}

The results in \autoref{sec:effective} have shown that \textit{HPSv2}, \textit{PickScore}, and \textit{Ensemble} are the most effective reward models for preference data construction. Building on this finding, we conduct an investigation into the scalability of preference data using these reward models.
Specifically, we explore along two dimensions: (1) \textbf{Candidate scaling.} As the number of candidate samples generated from different random seeds increases, their diversity expands. Larger candidate diversity widens the quality gap between the highest- and lowest-scoring samples, making the constructed preference pairs more informative. (2) \textbf{Sample scaling.} A larger dataset enables the model to capture nuanced patterns more comprehensively, leading to deeper learning of preferences.
Based on the insights in \autoref{sec:effective}, we select Aesthetic, ImageR, HPSv2, PickScore, and GPT-4 as the evaluation models. To enable the model to achieve optimal performance, we conduct a search over two typical hyper-parameters---$\beta$ and learning rate (see details in the Appendix), before the scaling experiments. For each experiment, we tune one scaling dimension and fix the other dimension. The results are reported in \autoref{fig:scaling}. We have the following results and discoveries.

\textbf{Consistent scaling trends across models and benchmarks.} First, we observe that data scaling demonstrates robust trends regardless of the model used---BrushNet and FLUX.1 Fill, in the first and second rows of each sub-figure, respectively. Second, we find that similar scaling trends emerge when evaluating on different benchmarks, as evidenced by the comparable patterns of the two lines within each sub-figure. These findings indicate that the observed scaling behavior is robust and generalizable. However, we also observe some inconsistent phenomena: when using PickScore as the reward model, ImageR/HPSv2 exhibit opposite trends on BrushNet and FLUX.1 Fill. This issue is caused by the characteristics of both reward models and baseline models, as  analyzed in~\autoref{hacking}.

\textbf{Reward hacking from HPSv2 undermines training.}
When evaluated by Aesthetic, ImageR, HPSv2, and PickScore, using HPSv2 as the reward model shows benefits from both candidate scaling and sample scaling. However, its GPT-4 results deteriorate significantly in the later stages of scaling. This observation aligns with our finding in \autoref{sec:effective}, where HPSv2 achieves good results under public model evaluations but sometimes loses to others when assessed by GPT-4. We hypothesize that this degradation stems from some shared common biases among these reward models.

\textbf{Ensemble offers robust data scaling by resisting hacking.}
Although PickScore demonstrates good scaling behavior, its performance remains sub-optimal. In contrast, the Ensemble approach achieves the best results across benchmarks, model structures, evaluation models, and scaling dimensions. This is likely because Ensemble averages the preference choices of different reward models, which eliminates the biases of the employed reward models and improves its resistance to the hacking.

\begin{table*}[t]
\centering
\caption{Ablation studies on a new dataset of \textit{I Dream My Painting} \cite{idream}.}
\resizebox{1.\linewidth}{!}{
\begin{threeparttable}  
\renewcommand{\arraystretch}{1.2}
\setlength{\tabcolsep}{5pt}
{
     \begin{tabular}{l l C{1.8cm}C{1.8cm}C{1.8cm}C{1.8cm}C{1.8cm}C{1.8cm}C{1.8cm}C{1.8cm}C{1.8cm}C{1.8cm}}
          \toprule
          {inpainting model}  
          & {{CLIPScore}}   & {{Aesthetic}}  & {{ImageR}} & {{PickScore}}  & {{HPSv2}}   & {{VQAScore}} & {{UnifiedR}} & {{Perception}} & {{HPSv3}} & {\cellcolor{sem!15}{GPT-4}}     \\ 
                                     
          \midrule
                {BrushNet} &            24.849 &            5.923 &           -0.246 &            20.550 &            19.749 &            8.503 &            2.317 &            27.317 &           -0.551 &     \cellcolor{sem!15}72.669 \\
  \textbf{BruPA (ours) } &            \textbf{25.460} &            \textbf{6.111} &            \textbf{2.152} &            \textbf{20.735} &            \textbf{21.086} &            \textbf{8.653} &            \textbf{2.463} &            \textbf{28.294} &            \textbf{1.265} &     \cellcolor{sem!15}\textbf{73.739} \\
          \hline
               FLUX.1 Fill &            24.194 &            6.017 &            0.544 &            20.855 &            20.203 &            8.667 &            2.476 &            26.627 &            0.547 &   \cellcolor{sem!15}{76.391} \\
    \textbf{FluPA (ours)} &   \textbf{25.500} &   \textbf{6.448} &   \textbf{5.961} &   \textbf{21.407} &   \textbf{23.770} &   \textbf{9.031} &   \textbf{2.784} &   \textbf{28.868} &   \textbf{5.023} &   \cellcolor{sem!15}\textbf{79.255} \\
          \bottomrule
          
     \end{tabular}
     \begin{tablenotes}
     \item \textbf{Bold} values denote the best results. 
     \end{tablenotes}
}
\end{threeparttable}
}
\label{tab:comparison_on_I_DREAM_MY_PAINT}
\end{table*}

\section{How does Reward Hacking Happen?}
\label{hacking}
\begin{table*}[t]
\centering
\caption{Comparisons of state-of-the-art image inpainting models on BrushBench and EditBench.}
\resizebox{1.\linewidth}{!}{
\begin{threeparttable}  
\renewcommand{\arraystretch}{1.5}
\setlength{\tabcolsep}{0pt}
{
     \begin{tabular}{l l C{1.2cm}C{1.2cm}C{1.2cm}C{1.2cm}C{1.2cm}C{1.2cm}C{1.2cm}C{1.2cm}C{1.2cm}C{1.2cm}C{1.2cm}C{1.2cm}C{1.2cm}C{1.2cm}C{1.2cm}C{1.2cm}C{1.2cm}C{1.2cm}C{1.2cm}C{1.2cm}}
          \toprule
          \multirow{2}{*}{inpainting model}  
          & \multicolumn{2}{c}{{CLIPScore}}   & \multicolumn{2}{c}{{Aesthetic}}  & \multicolumn{2}{c}{{ImageR}} & \multicolumn{2}{c}{{PickScore}}  & \multicolumn{2}{c}{{HPSv2}}   & \multicolumn{2}{c}{{VQAScore}} & \multicolumn{2}{c}{{UnifiedR}} & \multicolumn{2}{c}{{Perception}} & \multicolumn{2}{c}{{HPSv3}} & \multicolumn{2}{c}{\cellcolor{sem!15}{GPT-4}}     \\ 
                                             \cmidrule(lr){2-3} \cmidrule(lr){4-5} \cmidrule(lr){6-7} \cmidrule(lr){8-9} \cmidrule(lr){10-11} \cmidrule(lr){12-13} \cmidrule(lr){14-15} \cmidrule(lr){16-17} \cmidrule(lr){18-19} \cmidrule(lr){20-21}
                                       & Brush. & Edit. & Brush. & Edit. & Brush. & Edit. & Brush. & Edit. & Brush. & Edit. & Brush. & Edit. & Brush. & Edit. & Brush. & Edit. & Brush. & Edit. & \cellcolor{sem!15}Brush. & \cellcolor{sem!15}Edit. \\
          \midrule
          SDI
          &                    26.304 &                  26.526 &              6.368 &                    5.377 &              12.026 &               -1.100 &                 22.105 &                20.791 &                   27.079 &               23.203 &                8.981 &             6.923 &                3.268 &              2.069 &                26.190 &                25.382 &                  5.320 &                  0.849 &                    \cellcolor{sem!15}79.004 &                    \cellcolor{sem!15}60.751 \\
          CNI
          &                    26.341 &                  26.972 &              6.305 &                    5.382 &              11.421 &               -1.044 &                 21.953 &                20.874 &                   26.633 &               23.076 &                8.890 &             6.906 &                3.218 &              2.125 &                26.150 &                25.894 &                  4.546 &                  0.894 &                    \cellcolor{sem!15}74.173 &                    \cellcolor{sem!15}63.921 \\
          BLD
          &                    26.337 &      27.666 &              6.262 &                    5.372 &              11.161 &                0.563 &                 21.901 &                20.980 &                   26.723 &               23.839 &                8.852 &           {7.467} &                3.202 &            {2.228} &                26.128 &    \underline{27.093} &                  4.559 &                  1.114 &                    \cellcolor{sem!15}71.794 &                    \cellcolor{sem!15}62.690 \\
          PowerPaint
          &                  {26.265} &                {27.291} &              6.312 &                    5.448 &              11.771 &              {0.720} &                 22.089 &                20.912 &                   27.065 &               23.347 &                8.931 &           {7.238} &                3.271 &            {2.219} &                26.123 &                26.264 &                  5.112 &                  1.068 &                    \cellcolor{sem!15}78.241 &                  \cellcolor{sem!15}63.092   \\
          PrefPaint
          &                    26.268 &                  25.569 &              6.377 &                    5.296 &              11.798 &               -3.023 &                 22.125 &                20.666 &                   26.855 &               22.241 &                8.925 &             6.226 &                3.271 &              1.951 &                26.116 &                24.264 &                  5.208 &                 -0.336 &                    \cellcolor{sem!15}80.327 &                  \cellcolor{sem!15}60.815   \\
          StrDiffusion
          &                    23.872 &                  21.398 &              5.330 &                    4.405 &              -0.063 &              -16.282 &                 20.342 &                19.147 &                   21.431 &               16.616 &                7.281 &             3.662 &                2.417 &              1.236 &                23.381 &                20.102 &                 -2.243 &                 -7.131 &                    \cellcolor{sem!15}34.255 &                    \cellcolor{sem!15}25.200 \\
          HD-Painter
          &           26.367 &                  26.934 &            {6.480} &        \underline{5.640} &            {12.913} &                0.046 &               {22.314} &              {21.016} &                 {27.931} &             {23.951} &                9.019 &             6.682 &              {3.349} &              2.136 &              {26.214} &                25.721 &                {6.224} &                {1.983} &        \cellcolor{sem!15}\underline{85.016} &      \cellcolor{sem!15}\underline{69.087}   \\
          ASUKA
          &                    24.387 &                  20.842 &              6.294 &                    5.078 &               5.208 &              -14.110 &                 21.601 &                19.603 &                   25.285 &               18.702 &                7.681 &             3.631 &                2.862 &              1.282 &                23.959 &                19.017 &                  3.556 &                 -3.506 &                    \cellcolor{sem!15}75.140 &                  \cellcolor{sem!15}58.686   \\
          \hline
          {BrushNet
          }                            
          &                    26.415 &                  27.337 &              6.425 &                    5.392 &              12.717 &               -1.296 &                 22.133 &                20.616 &                   27.509 &               23.076 &                9.060 &             6.770 &                3.303 &              2.100 &    \underline{26.290} &              {26.410} &                  5.749 &                  0.403 &                    \cellcolor{sem!15}79.391 &                  \cellcolor{sem!15}57.046   \\
          \textbf{BruPA (ours) }                            
          &        \textbf{26.547} &                \underline{27.694} &  \underline{6.516} &                  {5.577} &  \underline{13.315} &      \textbf{10.463} &                 22.279 &                20.844 &       \underline{28.037} &               23.933 &    \underline{9.093} &             7.043 &    \underline{3.371} &              2.193 &       \textbf{26.390} &              {26.881} &      \underline{6.276} &                  1.398 &                    \cellcolor{sem!15}83.054 &                  \cellcolor{sem!15}61.186   \\
          FLUX.1 Fill
          &                    26.244 &                  27.103 &              6.429 &                    5.458 &              12.760 &                4.910 &     \underline{22.327} &    \underline{21.211} &                   27.476 &   \underline{24.076} &                9.081 & \underline{8.021} &                3.360 &  \underline{2.485} &                25.945 &                26.834 &                  6.055 &      \underline{2.470} &                  \cellcolor{sem!15}{83.935} &      \cellcolor{sem!15}{         {66.979}}  \\
          \textbf{FluPA (ours)}                                   
          &                  \underline{26.436} &         \textbf{27.813} &     \textbf{6.546} &           \textbf{5.681} &     \textbf{13.859} &    \underline{7.707} &        \textbf{22.577} &       \textbf{21.559} &          \textbf{28.735} &      \textbf{25.972} &       \textbf{9.152} &    \textbf{8.434} &       \textbf{3.457} &     \textbf{2.649} &                26.096 &       \textbf{27.617} &         \textbf{7.000} &         \textbf{4.230} &           \cellcolor{sem!15}\textbf{87.609} &         \cellcolor{sem!15}\textbf{72.307}   \\
          \bottomrule
          
     \end{tabular}
     \begin{tablenotes}
     \item \textbf{Bold} values denote the best results. \underline{Underlined} values denote the second-best results. All methods are evaluated using official implementations with blending~\cite{BrushNet}. 
     \end{tablenotes}
}
\end{threeparttable}
}
\label{tab:comparison_with_sota}
\end{table*}

We identify potential biases in reward models that may lead to reward hacking, as discussed in \autoref{sec:effective} and \autoref{sec:scalable}. In this section, we delve deeper into exploring these intriguing biases---examining their nature and how they make reward hacking happen. To investigate it, we sample inpainting examples in~\autoref{fig:bias} and~\autoref{fig:baseline}. We report the following findings and insights.

\textbf{Reward models exhibit biases in brightness, composition, and color scheme.}
As evidenced by the results from HPSv2 and PickScore---the second and third images in each sub-figure of~\autoref{fig:bias} respectively, we observe notable biases in their preferences. HPSv2 tends to favor images with bright lighting, complex composition with rich details, and vivid colors. In contrast, PickScore shows a preference for dim lighting, simple composition with few details, and muted colors. {\color{rebuttal_blue}We also make an attempt to leverage GPT-4 for evaluating generation biases quantitatively. We report the results on BrushBench in \autoref{tab:bcc_4o_single}. The quantitative results well validate our analyses that HPSv2 favors images with bright lighting, complex composition, and vivid colors; PickScore shows the opposite tendency; and our Ensemble falls squarely between the two.
\begin{table}[h]
\centering
\caption{\color{rebuttal_blue}Quantitative results of biases evaluated by GPT-4.}
\centering
\resizebox{\linewidth}{!}{
\begin{threeparttable}
\renewcommand{\arraystretch}{1.3}
\setlength{\tabcolsep}{5pt}
{
     \begin{tabular}{l l c c c}
          \toprule
          {model}  & reward model
          & {{brightness}}   & {{composition}}  & {{color}}      \\
          \midrule
          \multirow{3}{*}{BrushNet}
   &       HPSv2 &   \textbf{79.221} &   \textbf{82.979} &   \textbf{81.267} \\
   &    Ensemble &   \underline{78.180} &   \underline{80.922} &   \underline{78.513} \\
   &   PickScore &            73.545 &            75.019 &            70.641 \\
          \hline
          \multirow{3}{*}{FLUX.1 Fill}
   &       HPSv2 &   \textbf{76.617} &   \textbf{74.547} &   \textbf{70.978} \\
   &    Ensemble &   \underline{76.412} &   \underline{74.057} &   \underline{70.496} \\
   &   PickScore &            76.124 &            73.512 &            69.824 \\
          \bottomrule
     \end{tabular}
}
\end{threeparttable}
}
\label{tab:bcc_4o_single}
\end{table}
} 

\textbf{Biases in reward models affect different baseline models in distinct ways.}
Although each reward model has its own inherent biases, we find that their influence varies across baseline models. For instance, BrushNet trained using HPSv2 produces inpainting outputs characterized by excessively bright lighting, overly intricate details, and unnaturally vivid colors---they seem to deviate from human aesthetic preferences. In contrast, FLUX.1 Fill trained using HPSv2 generates visually pleasing results. PickScore shows a similar disparity in performance. 
It stems from the characteristics of baseline models as shown in~\autoref{fig:baseline}: BrushNet generates vibrant images, making PickScore particularly suitable for it; while FLUX.1 Fill produces plain images, aligning with HPSv2's property.

\textbf{Ensemble shows generality and generalization by mitigating biases.}
Ensemble, a simple and straightforward method implemented through reward ensembling, exhibits strong versatility across models by producing balanced and aesthetically pleasing inpainting results, as shown in~\autoref{fig:bias} and~\autoref{fig:baseline}. It likely stems from Ensemble’s ability to mitigate biases inherent in reward models.

\begin{figure*}[t]
\centering
\includegraphics[width=\linewidth, trim=0mm 0mm 0mm 0mm, clip]{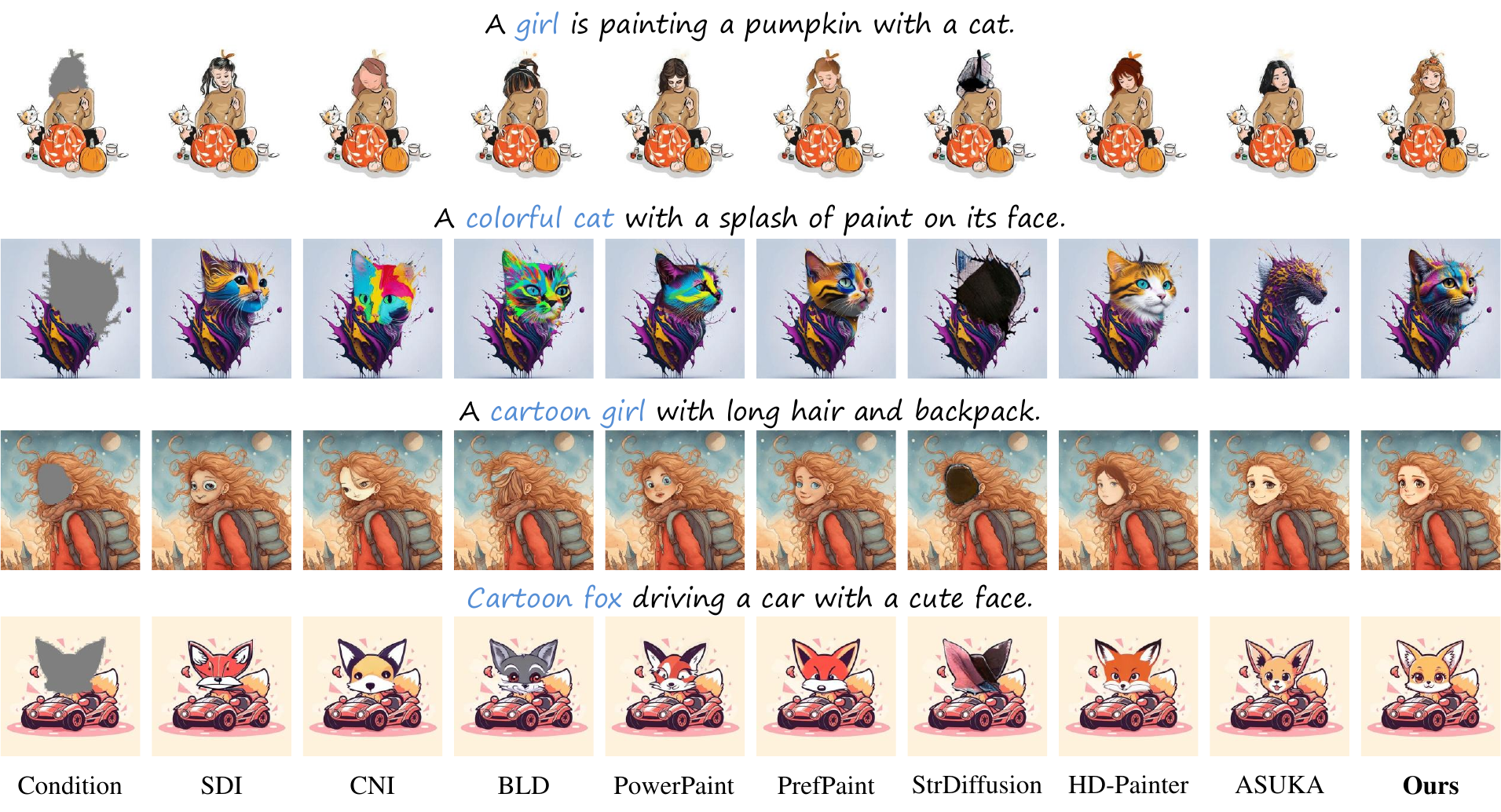}
\caption{Qualitative comparisons with state-of-the-art image inpainting models.}
\label{fig:sota}
\end{figure*}

{\color{rebuttal_blue}
\section{What is the Origin of the Biases?}
To better understand the source of the biases identified above, we examine the training data of two representative reward models---HPSv2 and PickScore---whose biases run in opposite directions. Specifically, we sample $1{,}000$ preference pairs (each consisting of one winning and one losing sample) from HPDv2~\cite{hpsv2} and Pick-a-Pic~\cite{pickscore}, the datasets used to train HPSv2 and PickScore, respectively. We then prompt GPT-4 to score these images along three attributes---brightness, composition, and color---and report the results in~\autoref{tab:dataset_diff}. As expected, HPDv2 exhibits substantial gaps between its winning and losing samples across all three attributes, whereas Pick-a-Pic shows only marginal differences. These dataset characteristics offer a natural explanation for the model-level behavior observed in previous sections: HPSv2 tends to favor bright, compositionally complex, and vividly colored images, whereas PickScore exhibits the opposite tendency.
}
\vspace{-0.3em}
\begin{table}[h]
\centering
\caption{\color{rebuttal_blue}Bias analysis on HPDv2 and Pick-a-Pic.}
\resizebox{1.\linewidth}{!}{
\begin{threeparttable}  
\renewcommand{\arraystretch}{1.2}
\setlength{\tabcolsep}{2pt}
{
     \begin{tabular}{l ccc ccc ccc}
          \toprule
          \multirow{2}{*}{dataset}  
          & \multicolumn{3}{c}{brightness} & \multicolumn{3}{c}{composition} & \multicolumn{3}{c}{color} \\
          \cmidrule(lr){2-4} \cmidrule(lr){5-7} \cmidrule(lr){8-10}
          
          & win & loss & win$-$lose & win & loss & win$-$lose & win & loss & win$-$lose \\ 
          \midrule
          HPDv2 
          & 40.641 & 39.236 & \textbf{1.405} & 45.208 & 41.560 & \textbf{3.648} & 42.340 & 38.269 & \textbf{4.071} \\
          Pick-a-Pic 
          & \textbf{42.421} & \textbf{42.415} & 0.006 & \textbf{46.144} & \textbf{45.899} & 0.245 & \textbf{43.943} & \textbf{44.137} & -0.194 \\
          \bottomrule
     \end{tabular}
}
\end{threeparttable}
}
\label{tab:dataset_diff}
\end{table}

\section{Ablation Studies and Comparisons with State-of-the-Art}
\label{ablation_comparison}

\begin{figure*}[t]
\centering
\includegraphics[width=\linewidth, trim=7mm 7mm 7mm 7mm, clip]{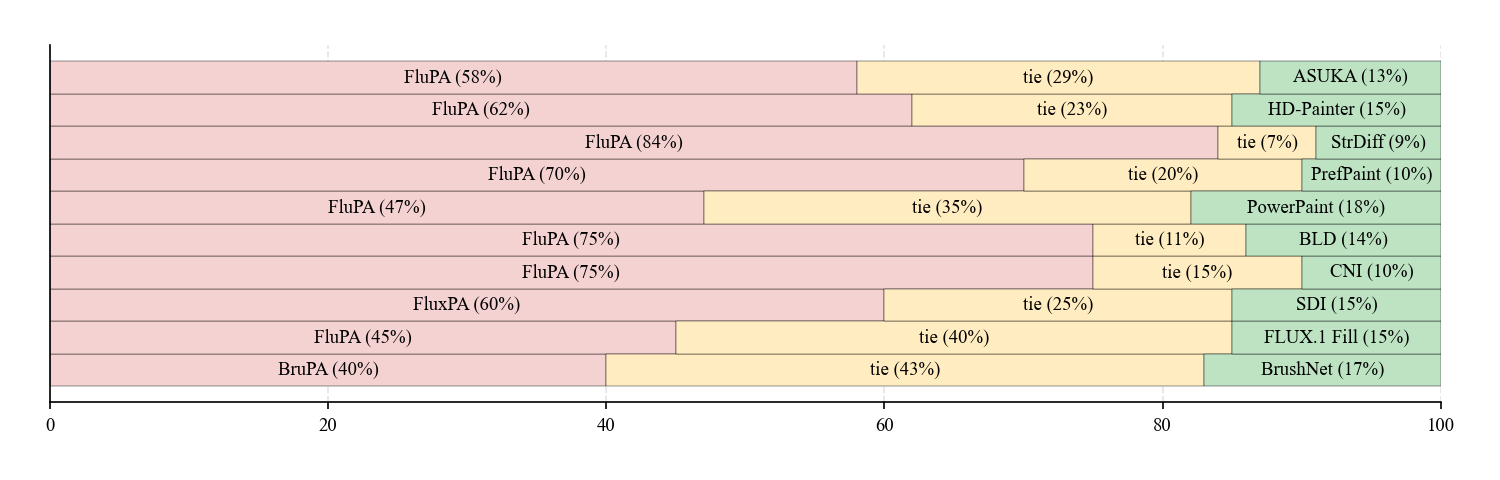}
\caption{\textbf{User studies.}
We compare each pair of models by randomly sampling 100 pairs from their inpainting results. We invite 30 volunteers to participate in a blind assessment to determine which one is better (``A win'', ``B win'', or ``tie'') based on their preferences. We report the \textbf{winning rates}.}
\label{fig:user}
\end{figure*}

We name our methods \textbf{BruPA} and \textbf{FluPA}---BrushNet and FLUX.1 Fill with Ensemble-based preference alignment. We compare them with state-of-the-art image inpainting models. Specifically, the following methods are compared (they are introduced in~\autoref{sec:related-work}): SDI~\cite{sd}, CNI~\cite{CNI}, BLD~\cite{BLD}, PowerPaint~\cite{powerpaint}, BrushNet~\cite{BrushNet}, PrefPaint~\cite{Prefpaint}, StrDiffusion~\cite{StructureMatters}, FLUX.1 Fill~\cite{flux1filldev}, HD-Painter~\cite{HD-Painter}, and ASUKA~\cite{ASUKA}, where \textit{BrushNet and FLUX.1 Fill are also the baseline models for ablation studies}.

\begin{table}[h]
\centering
\caption{{\color{rebuttal_blue}Details of GPT-4 evaluations.}}
\resizebox{\linewidth}{!}{
\begin{threeparttable}  
\renewcommand{\arraystretch}{1.2}
\setlength{\tabcolsep}{5pt}
{
     \begin{tabular}{l c c c c}
          \toprule          
          {inpainting model}  & overall 
          & {{aesthetics}}   & {{structure}}  & {{semantics}}      \\ 
                                     
          \midrule
  {BrushNet} &            79.391 &            30.301 &   24.035 &   25.055 \\
     {BruPA} &   \textbf{83.054} &   \textbf{33.516} &            \textbf{24.069} &            \textbf{25.469} \\
          \hline
        {FLUX.1 Fill} &            83.935 &            32.617 &   24.997 &   26.321 \\
  {FluPA} &   \textbf{87.609} &   \textbf{35.100} &            \textbf{25.627} &           \textbf{26.883} \\
          \bottomrule
          
     \end{tabular}
}
\end{threeparttable}
}
\label{tab:sub_gpt}
\end{table}

\begin{table*}[t]
\centering
\caption{{\color{rebuttal_blue}Evaluation by GPT-4 and Qwen3-VL}}
\resizebox{1.\linewidth}{!}{
\begin{threeparttable}  
\renewcommand{\arraystretch}{1.3}
\setlength{\tabcolsep}{24pt}
{
     \begin{tabular}{l C{2.5cm} C{2.5cm}C{2.5cm}C{2.5cm}}
          \toprule
          \multirow{2}{*}{inpainting model}  
          &  \multicolumn{2}{c}{{GPT-4}} & \multicolumn{2}{c}{{Qwen3-VL}}     \\ 
                                             \cmidrule(lr){2-3} \cmidrule(lr){4-5}
                                        & Brush. & Edit. & Brush. & Edit. \\
          \midrule
           SDI &                 79.004 $\pm$ 0.203 &               63.921 $\pm$ 0.410 &           90.323 $\pm$ 0.245 &           88.300 $\pm$ 0.310 \\
           CNI &                 74.173 $\pm$ 0.817 &               60.751 $\pm$ 0.590 &           88.101 $\pm$ 0.277 &           85.519 $\pm$ 0.337 \\
           BLD &                 71.794 $\pm$ 0.657 &               62.690 $\pm$ 0.736 &           85.719 $\pm$ 0.461 &           86.886 $\pm$ 0.336 \\
    PowerPaint &                 78.241 $\pm$ 0.912 &               63.092 $\pm$ 0.738 &           89.763 $\pm$ 0.250 &           85.682 $\pm$ 0.425 \\
     PrefPaint &                 80.327 $\pm$ 0.293 &               60.815 $\pm$ 0.392 &           91.249 $\pm$ 0.220 &           83.269 $\pm$ 0.449 \\
  StrDiffusion &                 34.255 $\pm$ 0.159 &               25.200 $\pm$ 0.236 &           49.560 $\pm$ 0.394 &           40.867 $\pm$ 0.336 \\
    HD-Painter &     \underline{85.016} $\pm$ 0.626 &   \underline{69.089} $\pm$ 0.628 &           92.361 $\pm$ 0.026 &           87.389 $\pm$ 0.325 \\
         ASUKA &                 75.140 $\pm$ 1.022 &               58.686 $\pm$ 0.477 &           84.597 $\pm$ 0.458 &           73.614 $\pm$ 0.980 \\
          \hline
              {BrushNet} &            79.391 $\pm$ 0.375 &            57.046 $\pm$ 1.151 &             89.930 $\pm$ 0.396 &             85.064 $\pm$ 0.450 \\
  \textbf{BruPA (ours) } &            83.054 $\pm$ 0.886 &            61.081 $\pm$ 0.764 &             92.255 $\pm$ 0.163 & \underline{88.686} $\pm$ 0.248 \\
             FLUX.1 Fill &            83.935 $\pm$ 0.414 &            67.465 $\pm$ 0.463 & \underline{92.752} $\pm$ 0.145 &             88.649 $\pm$ 0.306 \\
   \textbf{FluPA (ours)} &   \textbf{87.609} $\pm$ 0.291 &   \textbf{72.307} $\pm$ 0.560 &    \textbf{94.501} $\pm$ 0.154 &    \textbf{91.603} $\pm$ 0.220 \\
          \bottomrule
          
     \end{tabular}
}
\end{threeparttable}
}
\label{tab:cross-validation}
\end{table*}

{\color{rebuttal_blue}\textbf{Ablation studies on preference alignment}.}
We report quantitative and qualitative ablation studies, before and after preference alignment training, in~\autoref{tab:comparison_with_sota} and~\autoref{fig:baseline}, respectively. After preference alignment using Ensemble, our method significantly surpasses the baseline models by achieving much better results and yielding visually appealing results. {\color{rebuttal_blue}We also report the details of GPT-4 evaluation in~\autoref{tab:sub_gpt}. According to GPT-4 on each sub-dimension, preference alignment consistently improves image quality across all aspects---aesthetic appeal, structure accuracy, and semantic alignment.} We further conduct ablation studies on a new dataset, which are reported in~\autoref{tab:comparison_on_I_DREAM_MY_PAINT}. It also confirms that our improvement is generalizable across data distributions.

\textbf{Comparisons with state-of-the-art.}
\autoref{tab:comparison_with_sota} reports the results. 
Our BruPA and FluPA set new state-of-the-art results, attaining the best scores on all evaluations and the second-best scores in nearly half of the cases. Notably, even on coarser metrics—CLIPScore, VQAScore, and Perception (analyzed in \autoref{sec:effective})—our methods still outperform competitors. Besides, BruPA and FluPA significantly outperform BrushNet and FLUX.1 Fill, i.e., the baselines before applying preference alignment. The qualitative results are reported in~\autoref{fig:sota}, and our models generate images with better aesthetics.

\begin{figure*}[t]
  \begin{subfigure}[t]{\linewidth}
    \centering
    \includegraphics[width=0.18\linewidth, trim=5mm 5mm 5mm 5mm, clip]{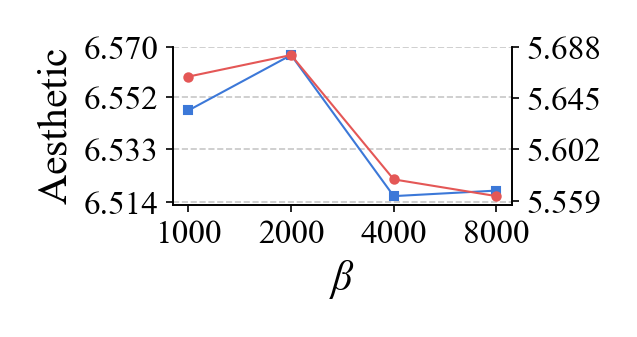}
    \includegraphics[width=0.18\linewidth, trim=5mm 5mm 5mm 5mm, clip]{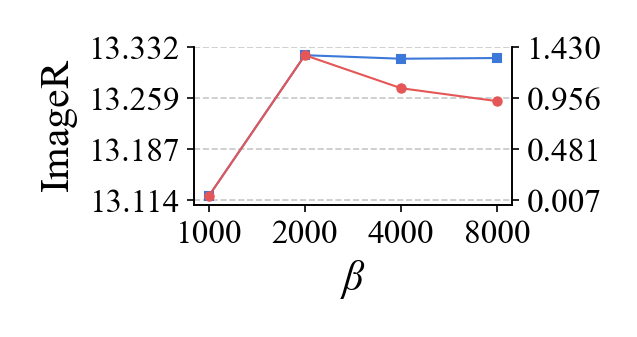}
    \includegraphics[width=0.18\linewidth, trim=5mm 5mm 5mm 5mm, clip]{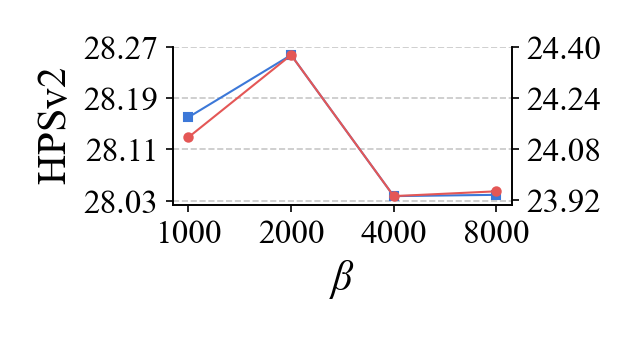}
    \includegraphics[width=0.18\linewidth, trim=5mm 5mm 5mm 5mm, clip]{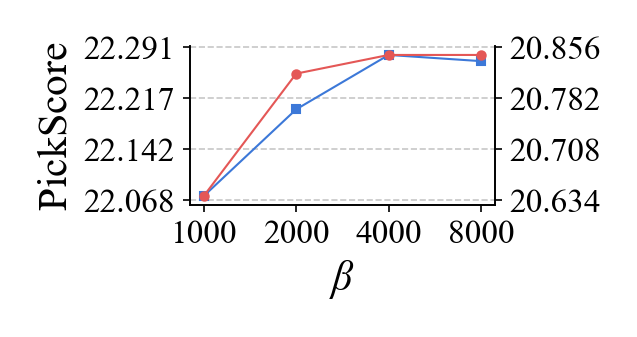}
    \includegraphics[width=0.18\linewidth, trim=5mm 5mm 5mm 5mm, clip]{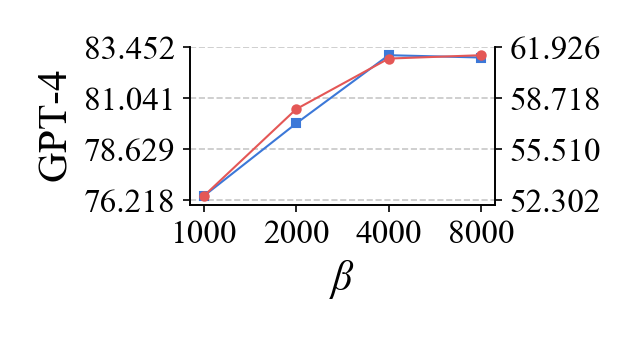}
    \includegraphics[width=0.18\linewidth, trim=5mm 5mm 5mm 5mm, clip]{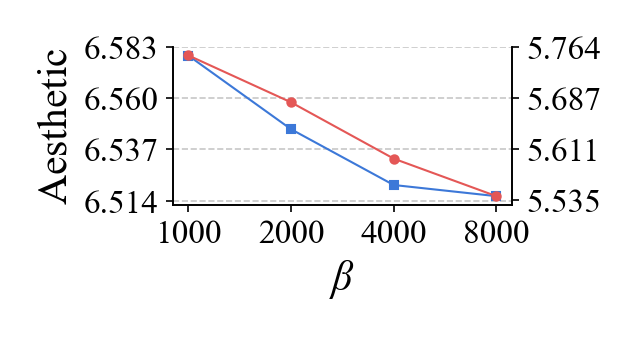}
    \includegraphics[width=0.18\linewidth, trim=5mm 5mm 5mm 5mm, clip]{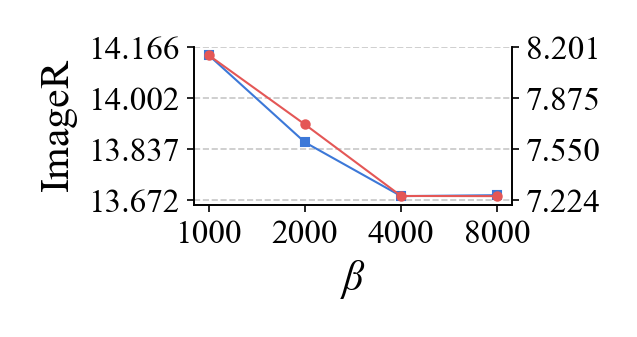}
    \includegraphics[width=0.18\linewidth, trim=5mm 5mm 5mm 5mm, clip]{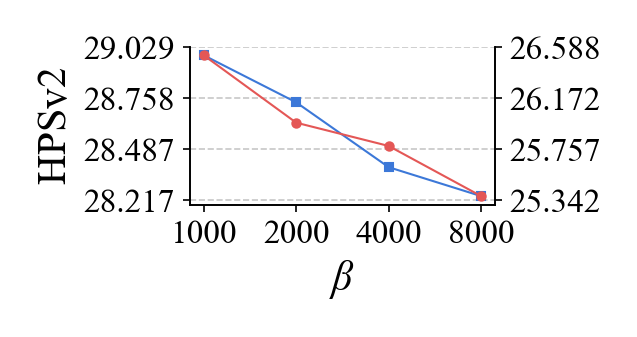}
    \includegraphics[width=0.18\linewidth, trim=5mm 5mm 5mm 5mm, clip]{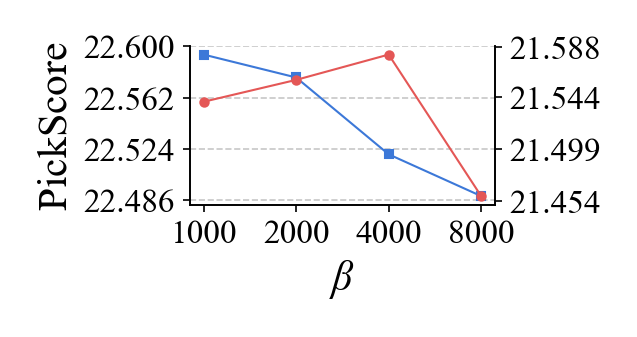}
    \includegraphics[width=0.18\linewidth, trim=5mm 5mm 5mm 5mm, clip]{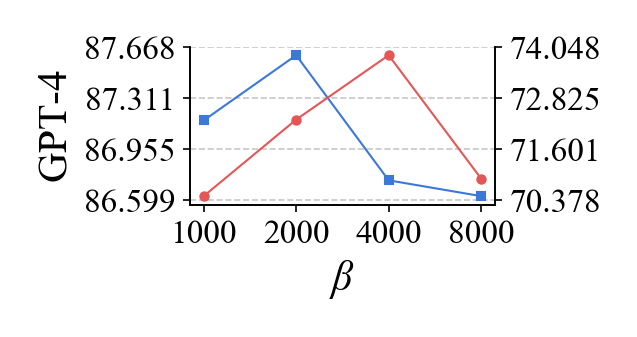}
    \caption{Sensitivity analysis on $\beta$.}
\end{subfigure}
\begin{subfigure}[t]{\linewidth}
    \centering
    \includegraphics[width=0.18\linewidth, trim=5mm 5mm 5mm 5mm, clip]{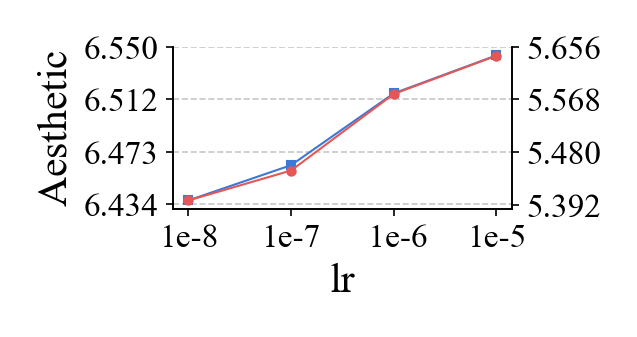}
    \includegraphics[width=0.18\linewidth, trim=5mm 5mm 5mm 5mm, clip]{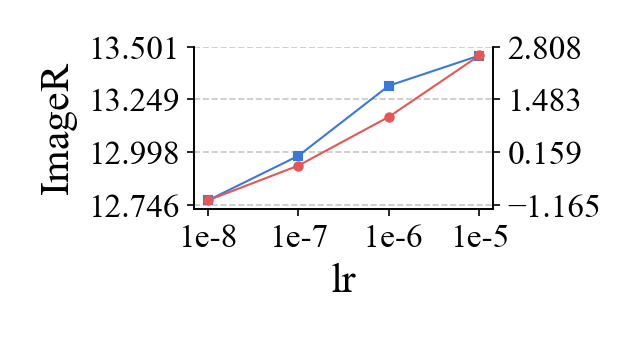}
    \includegraphics[width=0.18\linewidth, trim=5mm 5mm 5mm 5mm, clip]{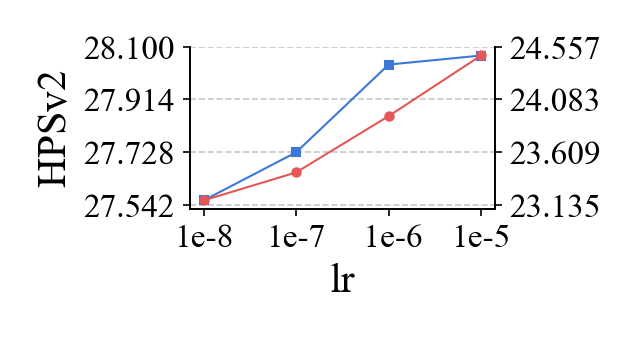}
    \includegraphics[width=0.18\linewidth, trim=5mm 5mm 5mm 5mm, clip]{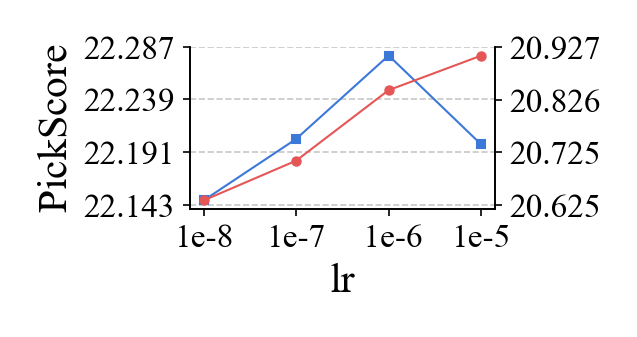}
    \includegraphics[width=0.18\linewidth, trim=5mm 5mm 5mm 5mm, clip]{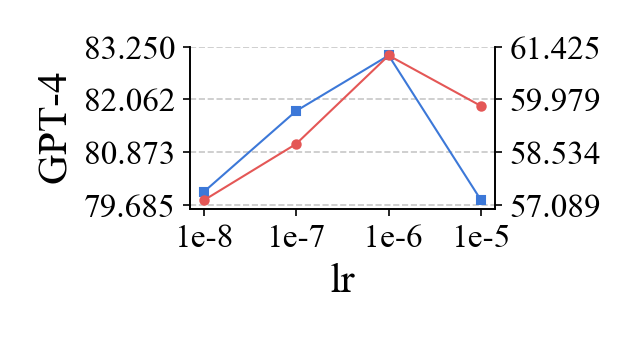}
    \includegraphics[width=0.18\linewidth, trim=5mm 5mm 5mm 5mm, clip]{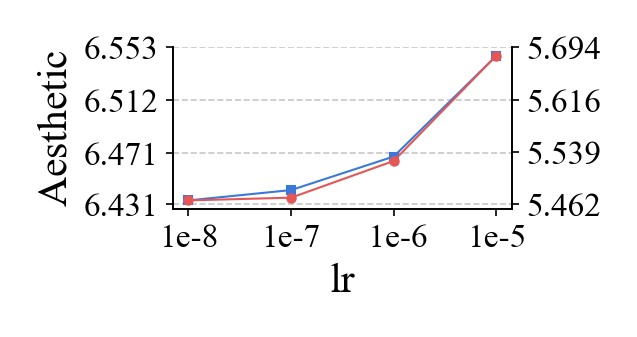}
    \includegraphics[width=0.18\linewidth, trim=5mm 5mm 5mm 5mm, clip]{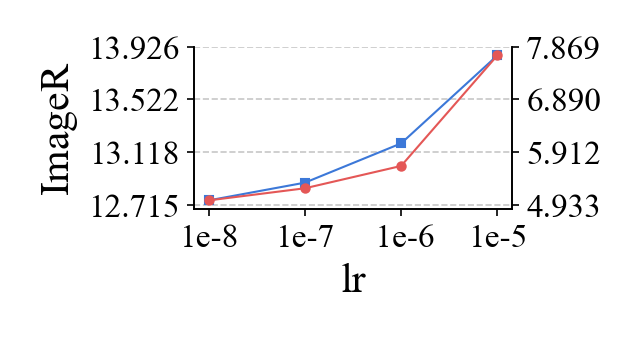}
    \includegraphics[width=0.18\linewidth, trim=5mm 5mm 5mm 5mm, clip]{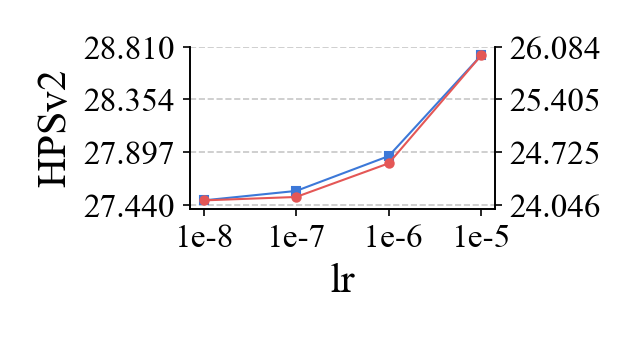}
    \includegraphics[width=0.18\linewidth, trim=5mm 5mm 5mm 5mm, clip]{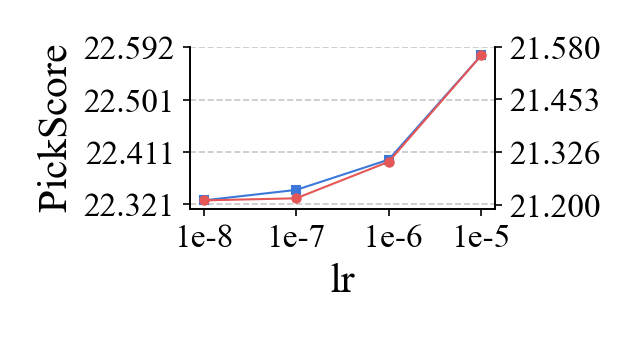}
    \includegraphics[width=0.18\linewidth, trim=5mm 5mm 5mm 5mm, clip]{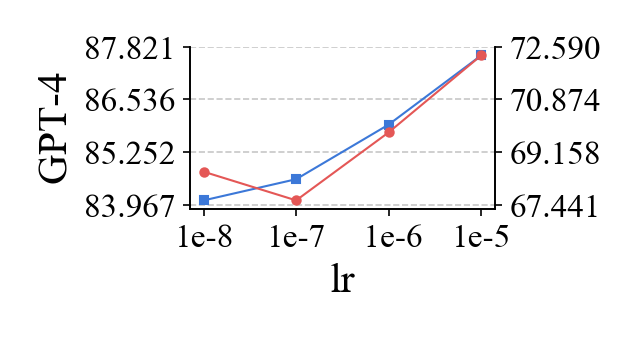}
    \caption{Sensitivity analysis on lr.}
\end{subfigure}
\caption{{\color{rebuttal_blue}\textbf{Searches} of $\beta$, i.e., sub-figure (a), and learning rate (lr), i.e., sub-figure (b). The first and second row of each sub-figure is based on \textbf{BrushNet} and \textbf{FLUX.1 Fill}, respectively.}}
\label{fig:lr-beta}
\vspace{-4mm}
\end{figure*}

{\color{rebuttal_blue}
\textbf{Effectiveness of GPT-4.}
To validate the effectiveness of GPT-4, we conduct user studies. All participants were informed of the purpose and procedure of the user study and provided informed consent before participation. The study collected only anonymized preference judgments and did not collect personally identifiable information. We randomly selected 500 inpainting pairs and invited volunteers to judge which result is better. Our human raters were drawn completely at random from a large and diverse volunteer pool. This random selection ensures that the evaluation is representative. To quantify agreement, we compute the alignment accuracy between the reward models and human raters, defined as the average proportion of cases in which both the raters and the model prefer the same result. In our evaluation, GPT-4 achieves 86\% accuracy; HPSv2: 82\%, PickScore: 80\%, ImageReward: 77\%, and Aesthetic: 80\%. To further evaluate the results, we include an open-source vision-language foundation model---Qwen3-VL~\cite{Qwen3-VL}---for evaluation. The results with means and standard deviations are reported in \autoref{tab:cross-validation}. Evaluation results of GPT-4 and Qwen3-VL are broadly stable, with our BruPA and FluPA delivering consistently remarkable performance gains. For completeness, we also report the Qwen3-VL counterparts of \autoref{tab:bcc_4o_single} and \autoref{tab:sub_gpt} in \autoref{tab:bcc_qwen_single} and \autoref{tab:sub_qwen}, respectively. The conclusions remain highly consistent across different vision-language foundation models.

To eliminate the potential biases in our design of prompts, we provide more results on the weights of the criteria---aesthetic quality, structural coherence, and semantic alignment---in \autoref{tab:diff_weight}. Our BruPA and FluPA consistently outperform the baseline models by a large margin, showing the significant and robust improvement of our method in GPT-4 evaluations.
}

{\color{rebuttal_blue}
\textbf{Analysis of Hyper-Parameters.} We conduct hyper-parameter searches for Ensemble, as shown in~\autoref{fig:lr-beta}. For Ensemble, we finally adopt a learning rate of 1e-6, and set $\beta=4000$ for BrushNet; while using a learning rate of 1e-5 and setting $\beta=2000$ for FLUX.1 Fill. For HPSv2, we use a learning rate of 1e-5 with $\beta=4000$ for BrushNet; and a learning rate of 1e-6 with $\beta=8000$ for FLUX.1 Fill. For PickScore, we set the learning rate to 1e-7 and use $\beta=2000$ for both BrushNet and FLUX.1 Fill.

{We also find that fine-tuning these hyper-parameters is not critical. Even when using fixed moderate values ($\beta=4000$ and $\mathrm{lr}=1e-6$, as suggested by~\cite{Diffusion-DPO}) across all experiments, our method still far exceeds the baseline w/o DPO: BruPA on BrushBench (83.133 vs. 79.391), BruPA on EditBench (72.442 vs. 57.046), FluPA on BrushBench (86.337 vs. 83.935), FluPA on EditBench (77.908 vs. 66.979). The new results remain the SOTA by outperforming previous best methods: BrushBench (86.337 vs. 85.016) and EditBench (77.908 vs. 69.087). Besides, Ensemble is less sensitive to hyper-parameter variations than single reward models like HPSv2 and PickScore. We hypothesize that this robustness arises because the Ensemble integrates different reward models, thereby mitigating the sensitivity inherent to any individual reward model.}
\begin{table}[t]
\centering
\caption{{\color{rebuttal_blue}Quantitative results of biases evaluated by Qwen3-VL.}}
\vspace{-2mm}
\resizebox{\linewidth}{!}{
\begin{threeparttable}  
\renewcommand{\arraystretch}{1.3}
\setlength{\tabcolsep}{5pt}
{
     \begin{tabular}{l l c c c}
          \toprule
          {model}  & reward model
          & {{brightness}}   & {{composition}}  & {{color}}      \\ 
          \midrule
          \multirow{3}{*}{BrushNet}
   &       HPSv2 &   \textbf{83.843} &   \textbf{88.310} &   \textbf{85.565} \\
   &    Ensemble &   \underline{83.284} &   \underline{88.180} &   \underline{84.644} \\
   &   PickScore &            82.615 &            86.530 &            82.516 \\
          \hline
          \multirow{3}{*}{FLUX.1 Fill}
   &       HPSv2 &   \textbf{82.750} &   \textbf{85.832} &   \textbf{81.695} \\
   &    Ensemble &   \underline{82.527} &   \underline{85.611} &   \underline{81.454} \\
   &   PickScore &            82.508 &            85.481 &            81.389 \\
          \bottomrule
     \end{tabular}
}
\end{threeparttable}
}
\label{tab:bcc_qwen_single}
\vspace{-2mm}
\end{table}

\begin{table}[t]
\centering
\centering
\caption{{\color{rebuttal_blue}Details of Qwen3-VL evaluations.}}
\vspace{-2mm}
\resizebox{\linewidth}{!}{
\begin{threeparttable}  
\renewcommand{\arraystretch}{1.2}
\setlength{\tabcolsep}{5pt}
{
     \begin{tabular}{l c c c c}
          \toprule          
          {inpainting model}  & overall 
          & {{aesthetics }}   & {{structure }}  & {{semantics}}      \\ 
                                     
          \midrule
  {BrushNet} &            89.930 &            35.141 &            27.395 &            27.393 \\
     {BruPA} &   \textbf{92.255} &   \textbf{36.032} &   \textbf{28.018} &   \textbf{28.204} \\
          \hline
        {FLUX.1 Fill} &            92.752 &            36.244 &            28.080 &            28.426 \\
  {FluPA} &   \textbf{94.501} &   \textbf{36.972} &   \textbf{28.477} &   \textbf{29.052} \\
          \bottomrule
          
     \end{tabular}
}
\end{threeparttable}
}
\label{tab:sub_qwen}
\vspace{-4mm}
\end{table}

\vspace{-2mm}
\begin{table}[h]
\centering
\caption{{\color{rebuttal_blue}GPT-4 evaluation results with different weight combinations: [aesthetic quality]-[structural coherence]-[semantic alignment]. ``auto'' represents the weight is determined by GPT-4 itself.}}
\vspace{-2mm}
\resizebox{\linewidth}{!}{
\begin{threeparttable}  
\renewcommand{\arraystretch}{1.3}
\setlength{\tabcolsep}{3pt}
{
     \begin{tabular}{l c c c c c}
          \toprule          
          {inpainting model}  & 40-30-30
          & {{30-40-30}}   & {{30-30-40}}  & {{33-33-33}} &  {{auto}}     \\ 
                                     
          \midrule
  {BrushNet} &    79.391 &   78.423 &      77.937 &     78.169 &   77.388 \\
     {BruPA} &    \textbf{83.054} &   \textbf{82.538} &      \textbf{82.388} &     \textbf{82.226} &   \textbf{81.036} \\
          \hline
    {FLUX.1 Fill} &   83.935 &   83.746 &   83.229 &   83.217 &   81.649 \\
          {FluPA} &   \textbf{87.609} &   \textbf{87.559} &   \textbf{87.368} &   \textbf{87.243} &   \textbf{85.309} \\
          \bottomrule
          
     \end{tabular}
}
\end{threeparttable}
}
\label{tab:diff_weight_gpt}

\vspace{-2mm}
\label{tab:diff_weight}
\end{table}

\begin{table*}[t]
\centering
\caption{{\color{rebuttal_blue}Compute cost for giving score to a single sample.}}
\resizebox{1.\linewidth}{!}{
\begin{threeparttable}  
\renewcommand{\arraystretch}{1.5}
\setlength{\tabcolsep}{5pt}
{
     \begin{tabular}{l c ccccccccc}
          \toprule
          {reward model}  
          & {{CLIPScore}}   & {{Aesthetic}}  & {{ImageR}} & {{PickScore}}  & {{HPSv2}}   & {{VQAScore}} & {{UnifiedR}} & {{Perception}} & {{HPSv3}}    \\ 
                                     
          \midrule
              {cost (seconds)} &   0.154 &           0.048  &        0.0759   &  0.0319 &  0.219  &           0.288 & 3.051  &   0.155 &  0.379 \\   
          
          \bottomrule
          
     \end{tabular}
}
\end{threeparttable}
}
\label{tab:latency}
\end{table*}

\begin{figure*}[t]
\begin{subfigure}[t]{\linewidth}
    \centering
    \includegraphics[width=0.11\linewidth, trim=0mm 0mm 0mm 0mm, clip]{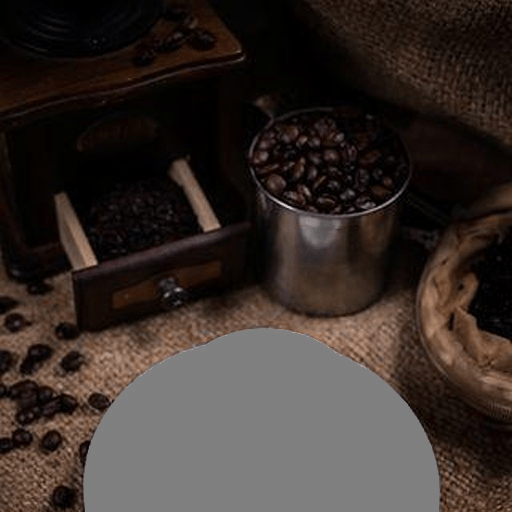}
    \includegraphics[width=0.11\linewidth, trim=0mm 0mm 0mm 0mm, clip]{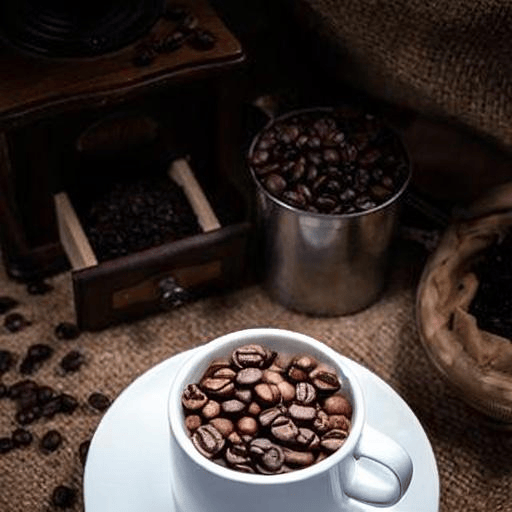}
    \includegraphics[width=0.11\linewidth, trim=0mm 0mm 0mm 0mm, clip]{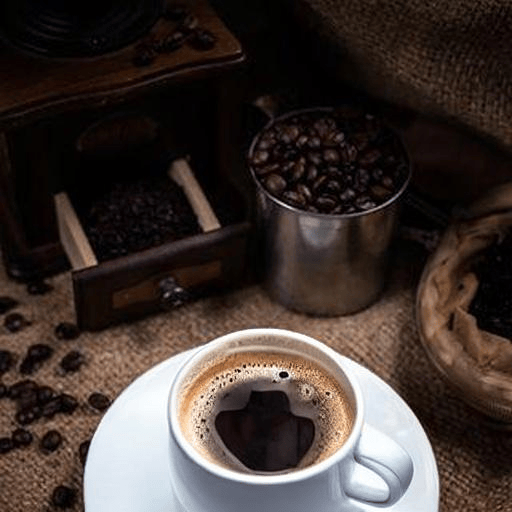}
    \includegraphics[width=0.11\linewidth, trim=0mm 0mm 0mm 0mm, clip]{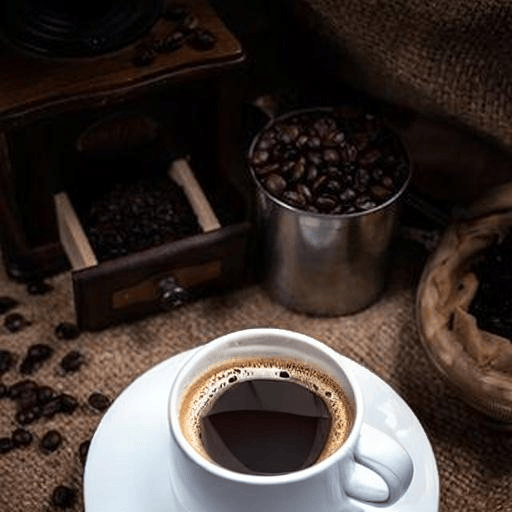}
    \includegraphics[width=0.11\linewidth, trim=0mm 0mm 0mm 0mm, clip]{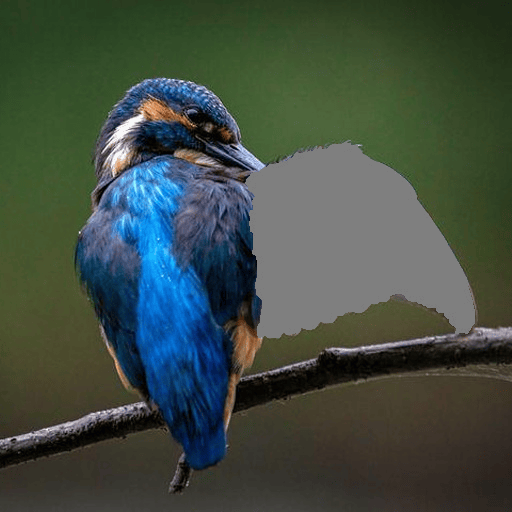}
    \includegraphics[width=0.11\linewidth, trim=0mm 0mm 0mm 0mm, clip]{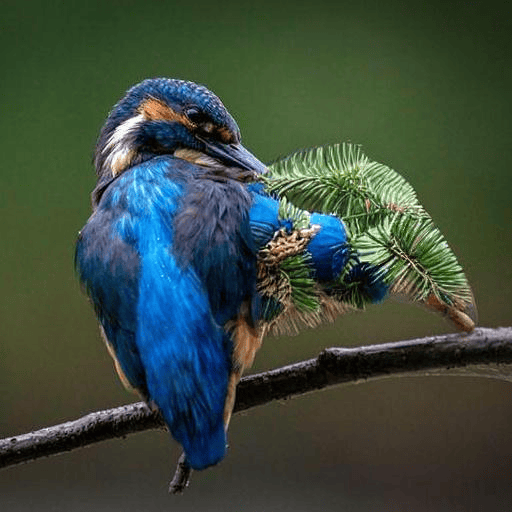}
    \includegraphics[width=0.11\linewidth, trim=0mm 0mm 0mm 0mm, clip]{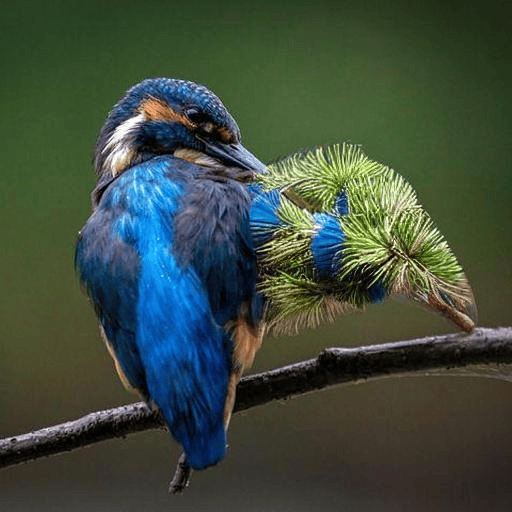}
    \includegraphics[width=0.11\linewidth, trim=0mm 0mm 0mm 0mm, clip]{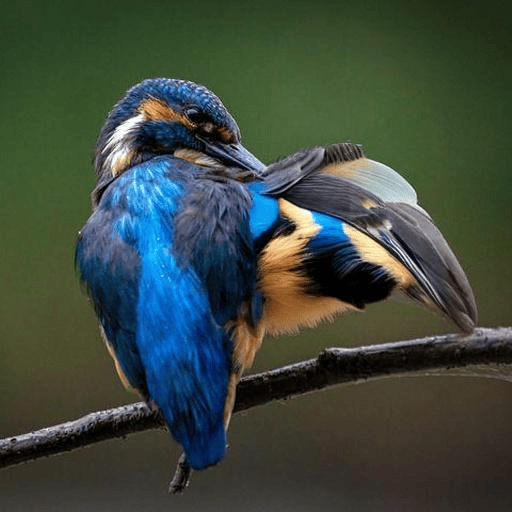}
    \includegraphics[width=0.11\linewidth, trim=0mm 0mm 0mm 0mm, clip]{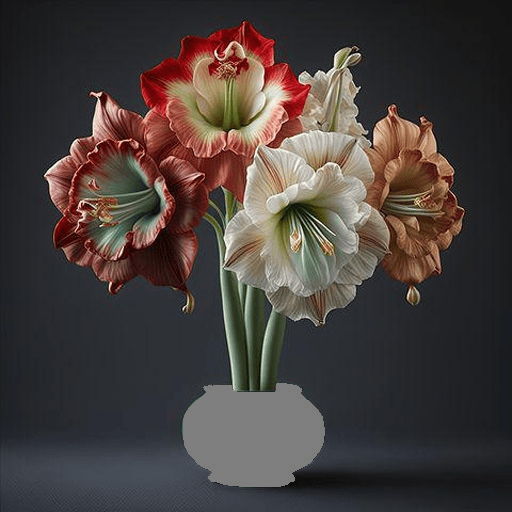}
    \includegraphics[width=0.11\linewidth, trim=0mm 0mm 0mm 0mm, clip]{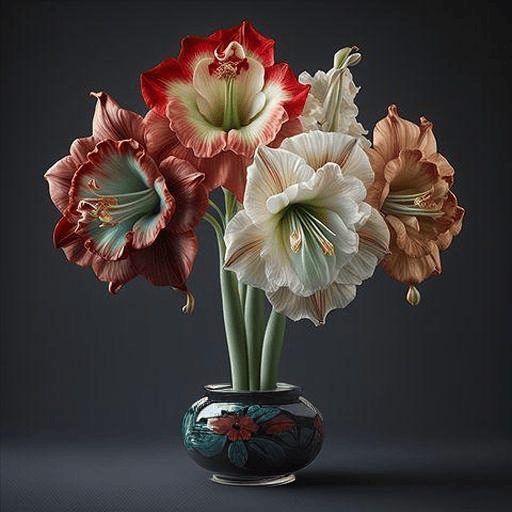}
    \includegraphics[width=0.11\linewidth, trim=0mm 0mm 0mm 0mm, clip]{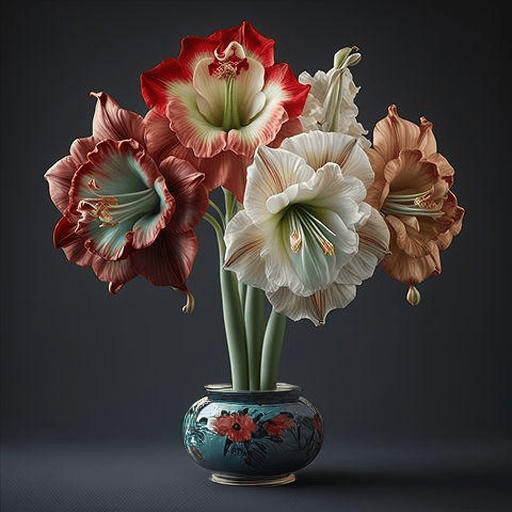}
    \includegraphics[width=0.11\linewidth, trim=0mm 0mm 0mm 0mm, clip]{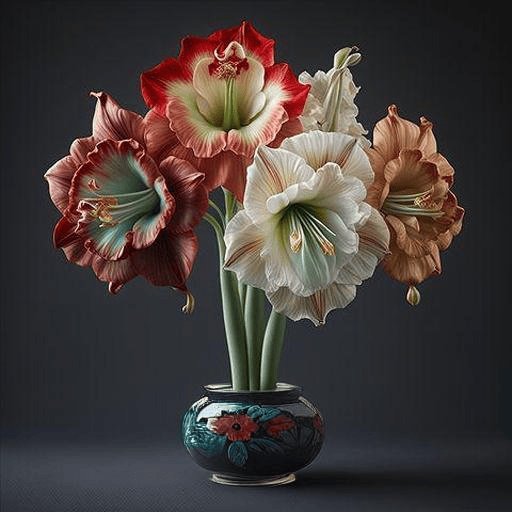}
    \includegraphics[width=0.11\linewidth, trim=0mm 0mm 0mm 0mm, clip]{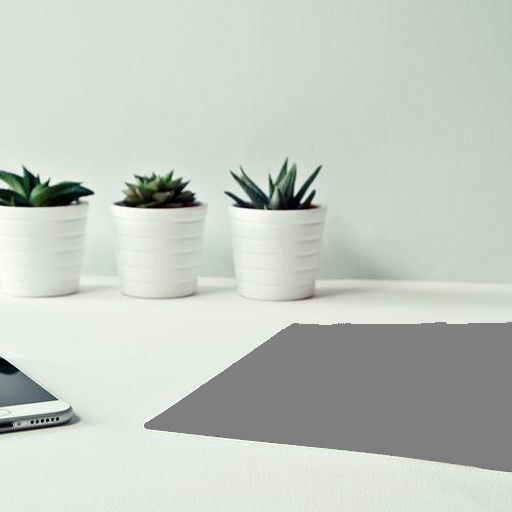}
    \includegraphics[width=0.11\linewidth, trim=0mm 0mm 0mm 0mm, clip]{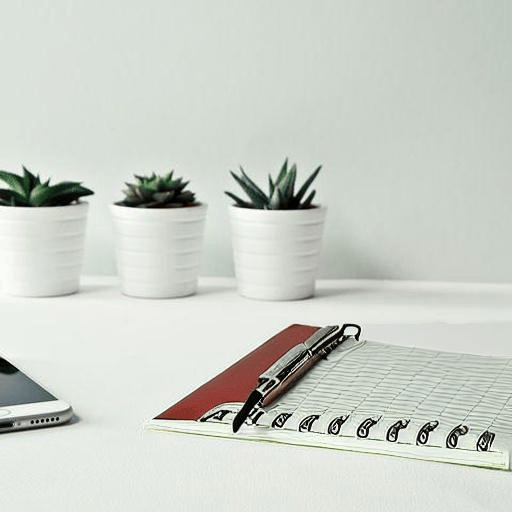}
    \includegraphics[width=0.11\linewidth, trim=0mm 0mm 0mm 0mm, clip]{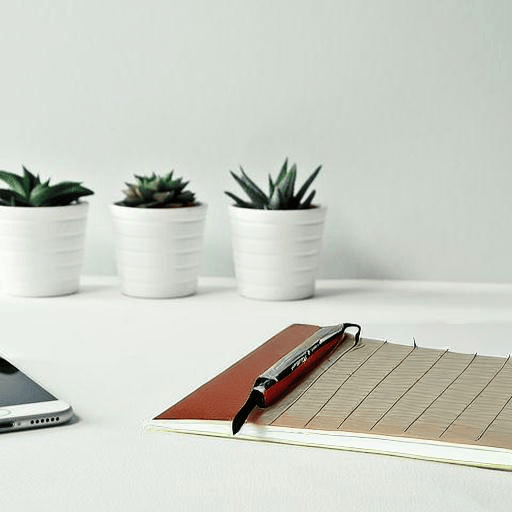}
    \includegraphics[width=0.11\linewidth, trim=0mm 0mm 0mm 0mm, clip]{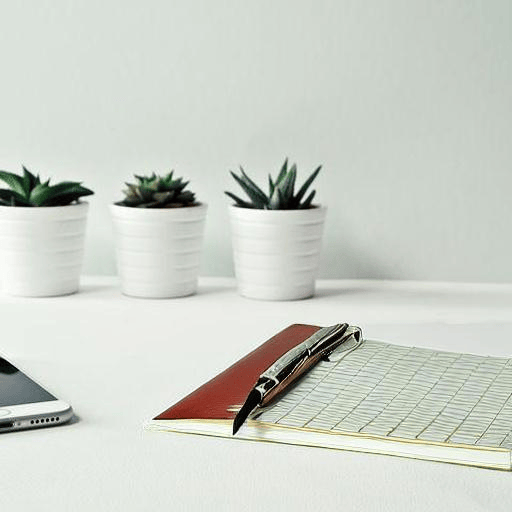}
    \caption{Examples from models trained using \textbf{BrushNet}.}
\end{subfigure}
\begin{subfigure}[t]{\linewidth}
    \centering
  \includegraphics[width=0.11\linewidth, trim=0mm 0mm 0mm 0mm, clip]{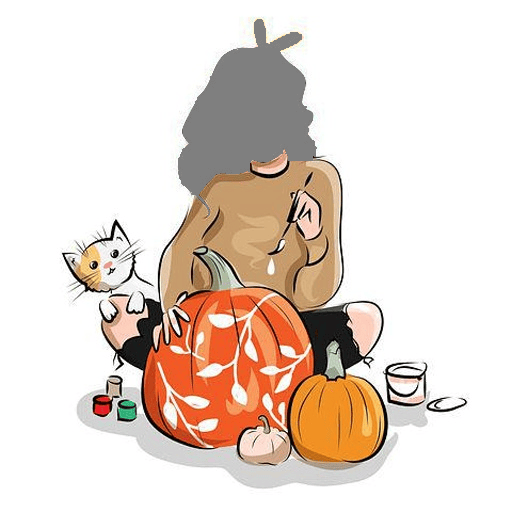}
  \includegraphics[width=0.11\linewidth, trim=0mm 0mm 0mm 0mm, clip]{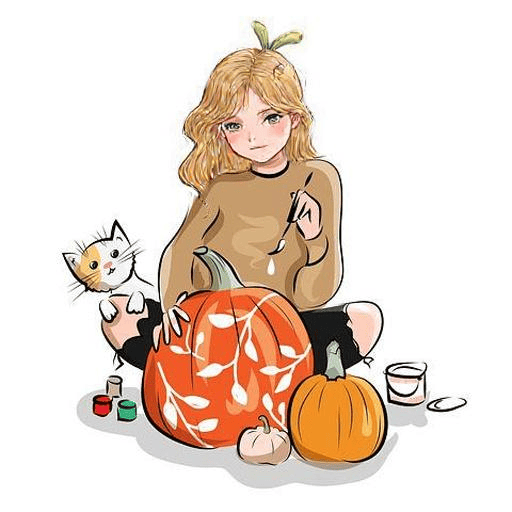}
  \includegraphics[width=0.11\linewidth, trim=0mm 0mm 0mm 0mm, clip]{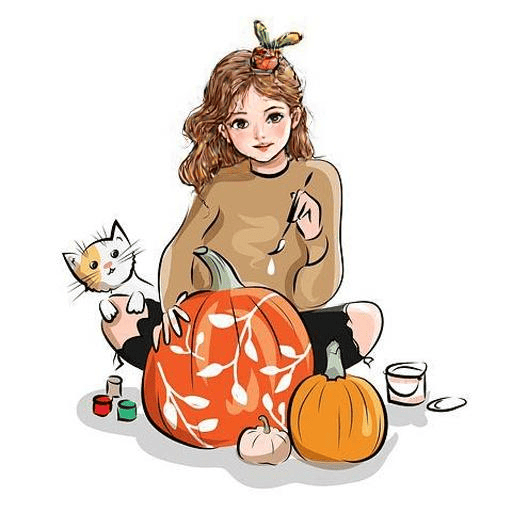}
  \includegraphics[width=0.11\linewidth, trim=0mm 0mm 0mm 0mm, clip]{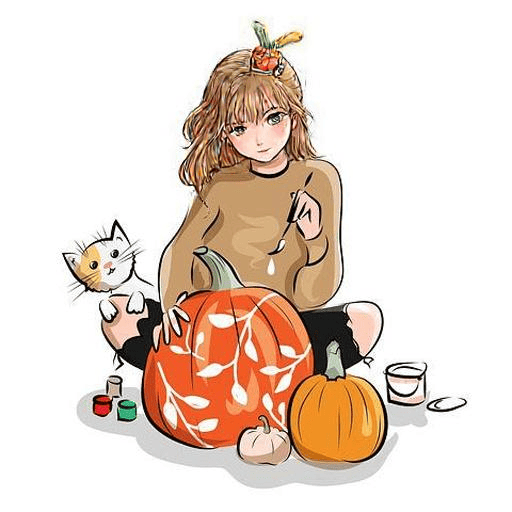}
  \includegraphics[width=0.11\linewidth, trim=0mm 0mm 0mm 0mm, clip]{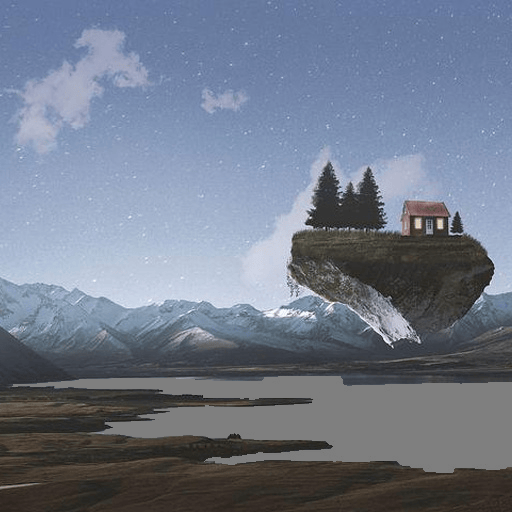}
  \includegraphics[width=0.11\linewidth, trim=0mm 0mm 0mm 0mm, clip]{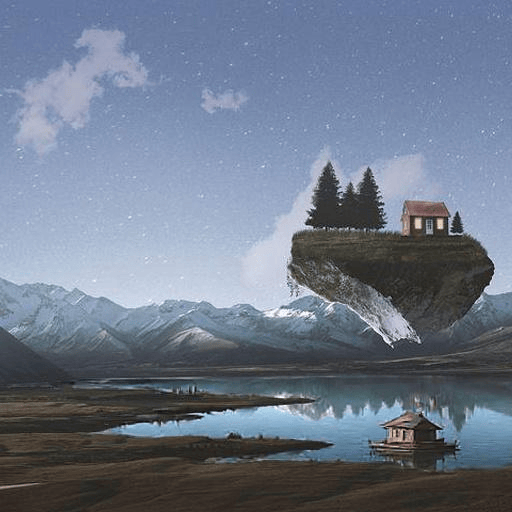}
  \includegraphics[width=0.11\linewidth, trim=0mm 0mm 0mm 0mm, clip]{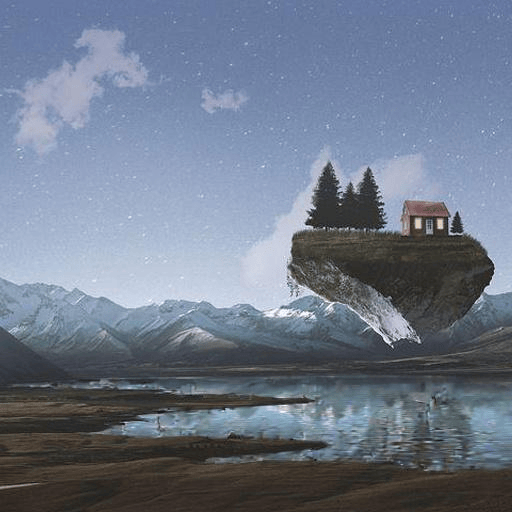}
  \includegraphics[width=0.11\linewidth, trim=0mm 0mm 0mm 0mm, clip]{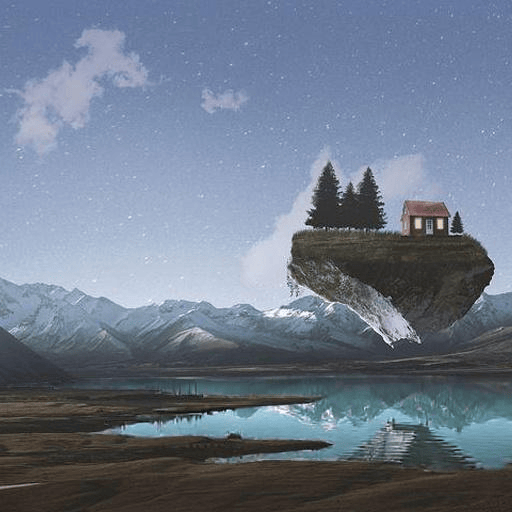}
  \includegraphics[width=0.11\linewidth, trim=0mm 0mm 0mm 0mm, clip]{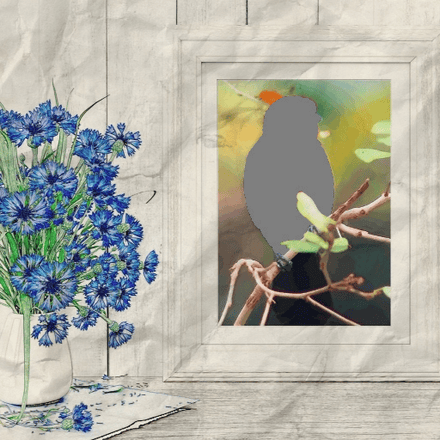}
  \includegraphics[width=0.11\linewidth, trim=0mm 0mm 0mm 0mm, clip]{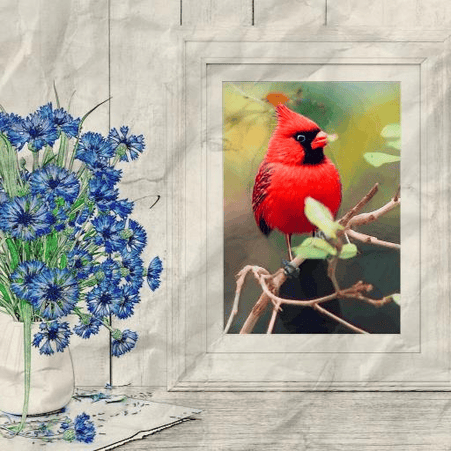}
  \includegraphics[width=0.11\linewidth, trim=0mm 0mm 0mm 0mm, clip]{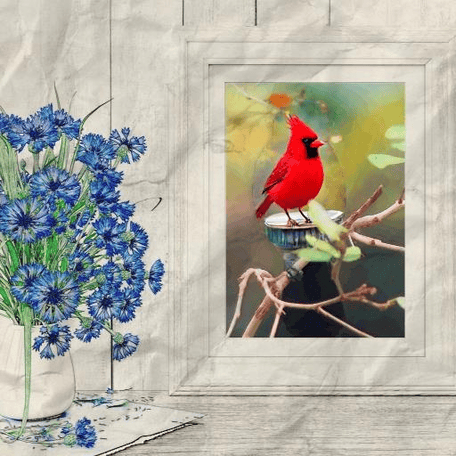}
  \includegraphics[width=0.11\linewidth, trim=0mm 0mm 0mm 0mm, clip]{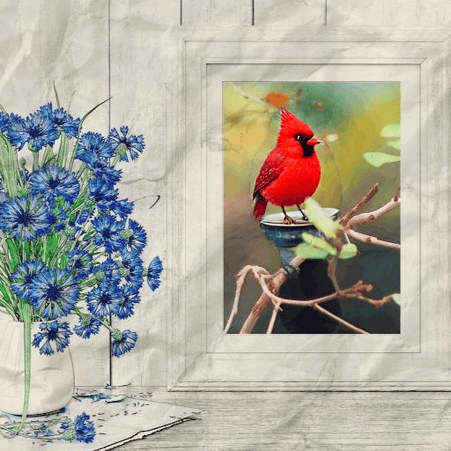}
  \includegraphics[width=0.11\linewidth, trim=0mm 0mm 0mm 0mm, clip]{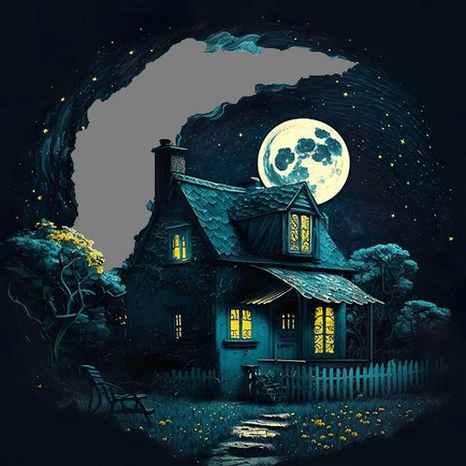}
  \includegraphics[width=0.11\linewidth, trim=0mm 0mm 0mm 0mm, clip]{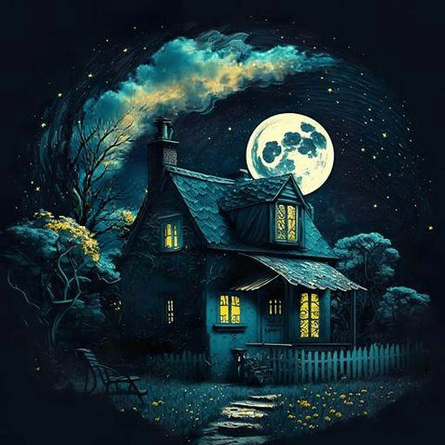}
  \includegraphics[width=0.11\linewidth, trim=0mm 0mm 0mm 0mm, clip]{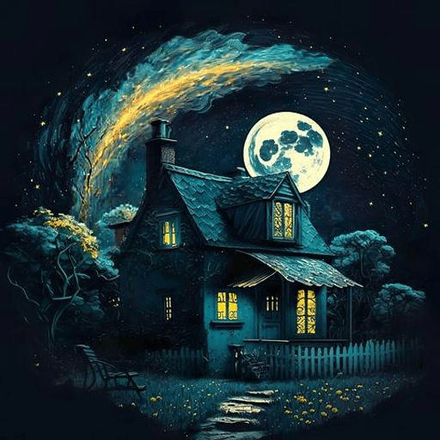}
  \includegraphics[width=0.11\linewidth, trim=0mm 0mm 0mm 0mm, clip]{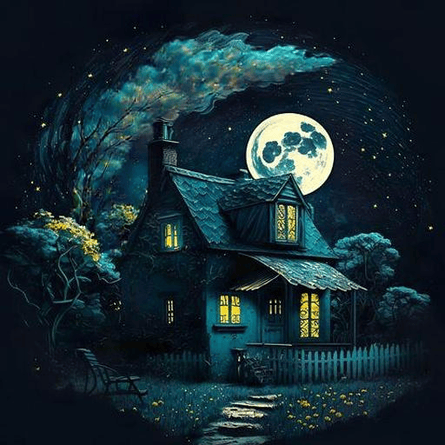}
    \caption{Examples from models trained using \textbf{FLUX.1 Fill}.}
\end{subfigure}
\vspace{-2mm}
\caption{\color{rebuttal_blue}{\textbf{Different Ensemble variants.} In each sub-figure, the four images (from left to right) display: the \textit{masked image}, followed by inpainting results from models trained using \textit{Ensemble (vanilla)}, \textit{Ensemble (w/o VQAScore)}, and \textit{Ensemble (Linear)}. We omit text prompts for brevity. Zoom in to see details.}}
\label{fig:variants}
\end{figure*}

}

{\color{rebuttal_blue}
\textbf{Compute Costs of Each Reward Model.} We report the compute cost of each reward model for giving a score to a single sample in \autoref{tab:latency}. Note that our Ensemble method only introduces additional cost for scoring samples and constructing preference pairs; the cost for training remains the same as in standard methods.
}

\textbf{User studies.} As shown in~\autoref{fig:user}, our models align with human preferences better.

{\color{rebuttal_blue}

\section{Stronger Ensemble Variants}
\label{sec:caen}

We have identified biases across reward models and proposed a new reward model, \textbf{Ensemble}, to mitigate reward hacking. In this section, we further explore how to push the limit of \textbf{Ensemble}. To this end, we introduce three complementary strategies: filtering out weaker reward models, weighted ensembling, and calibrated ensembling.

\textbf{Removing the weaker reward models.} As discussed in~\autoref{sec:effective}, we find that VQAScore is neither a reliable evaluator nor an effective reward signal provider. Therefore, we conduct a new experiment by removing this weak reward model from \textbf{Ensemble}. The quantitative results on BrushBench are reported in \autoref{tab:without_vqascore} and the qualitative results are reported in \autoref{fig:variants}. We observe a significant improvement in the GPT-4 evaluation scores of BruPA and FluPA after excluding VQAScore. This result preliminarily suggests that removing weaker reward models, although sacrificing the diversity of reward signals, can further enhance the results of preference alignment. There seems to be a trade-off between the diversity and effectiveness of reward signals when weaker reward models are present. 
\begin{table*}[!h]
\centering
\caption{\color{rebuttal_blue}Comparisons of different Ensemble variants on removing the weaker reward models.}
\resizebox{1.\linewidth}{!}{
\begin{threeparttable}  
\renewcommand{\arraystretch}{1.4}
\setlength{\tabcolsep}{5pt}
{
     \begin{tabular}{l c c c c c c c c c c c }
          \toprule
          {inpainting model}  
          & {{CLIPScore}}   & {{Aesthetic}}  & {{ImageR}} & {{PickScore}}  & {{HPSv2}}   & {{VQAScore}} & {{UnifiedR}} & {{Perception}} & {{HPSv3}} & {\cellcolor{sem!15}{GPT-4}}     \\ 
                                     
          \midrule
                {BruPA} &            26.547 &            6.516 &            13.315 &   \textbf{22.279} &            28.037 &   \textbf{9.093} &   \textbf{3.371} &   \textbf{26.390} &            6.276 &            \cellcolor{sem!15}83.054 \\
          {BruPA w/o VQAScore } &   \textbf{26.555} &   \textbf{6.524} &   \textbf{13.337} &            22.278 &   \textbf{28.055} &            9.082 &            3.362 &            26.385 &   \textbf{6.306} &   \textbf{\cellcolor{sem!15}83.659} \\
          \hline
               
          {FluPA} &            26.436 &            6.546 &            13.859 &            22.577 &            28.735 &            9.152 &            3.457 &            26.096 &            7.000 &            \cellcolor{sem!15}87.609 \\
             FluPA w/o VQAScore &   \textbf{26.438} &   \textbf{6.576} &   \textbf{14.017} &   \textbf{22.605} &   \textbf{28.747} &   \textbf{9.186} &   \textbf{3.466} &   \textbf{26.113} &   \textbf{7.068} &   \textbf{\cellcolor{sem!15}88.972} \\
          \bottomrule
          
     \end{tabular}
}
\end{threeparttable}
}
\label{tab:without_vqascore}
\end{table*}

\begin{table*}[!h]
\centering
\caption{\color{rebuttal_blue}Comparisons of different Ensemble variants with different weights applied to reward models.}
\resizebox{1.\linewidth}{!}{
\begin{threeparttable}  
\renewcommand{\arraystretch}{1.3}
\setlength{\tabcolsep}{7pt}
{
     \begin{tabular}{l c c c c c c c c c c c}
          \toprule
          {variant}  
          & {{CLIPScore}}   & {{Aesthetic}}  & {{ImageR}} & {{PickScore}}  & {{HPSv2}}   & {{VQAScore}} & {{UnifiedR}} & {{Perception}} & {{HPSv3}} & \cellcolor{sem!15}{GPT-4}     \\ 
                                     
          \midrule
    {BruPA } &               26.547 &   \underline{6.516} &   \underline{{13.315}} &      \textbf{22.279} &             {28.037} &   \underline{{9.093}} &   \underline{{3.371}} &             {26.390} &               6.276 &               \cellcolor{sem!15}83.054 \\
   {BruPA (linear)} &   \underline{26.568} &      \textbf{6.521} &        \textbf{13.342} &   \underline{22.274} &   \underline{28.050} &        \textbf{9.098} &        \textbf{3.372} &   \underline{26.400} &   \underline{6.288} &      \cellcolor{sem!15}\textbf{83.479} \\
  {BruPA (softmax)} &      \textbf{26.573} &   \underline{6.516} &                 13.136 &               22.265 &      \textbf{28.073} &                 9.092 &                 3.359 &      \textbf{26.409} &      \textbf{6.310} &   \underline{\cellcolor{sem!15}83.136} \\
  \hline
             {FluPA } &   \underline{26.436} &               6.546 &   \underline{13.859} &   \underline{22.577} &   \underline{28.735} &               9.152 &      \textbf{3.457} &   \underline{26.096} &   \underline{7.000} &               \cellcolor{sem!15}87.609 \\
   {FluPA (linear)} &      \textbf{26.513} &      \textbf{6.556} &      \textbf{14.030} &      \textbf{22.605} &      \textbf{28.815} &      \textbf{9.172} &   \underline{3.452} &      \textbf{26.171} &      \textbf{7.136} &      \cellcolor{sem!15}\textbf{89.190} \\
  {FluPA (softmax)} &               26.434 &   \underline{6.553} &               13.834 &               22.538 &               28.609 &   \underline{9.166} &               3.443 &               26.080 &               6.984 &   \underline{\cellcolor{sem!15}88.145} \\
          \bottomrule
          
     \end{tabular}
}
\end{threeparttable}
}
\label{tab:weighted}
\end{table*}

\begin{table*}[!h]
\centering
\caption{\color{rebuttal_blue}Comparisons of different Ensemble variants with different training objectives.}
\resizebox{1.\linewidth}{!}{
\begin{threeparttable}  
\renewcommand{\arraystretch}{1.3}
\setlength{\tabcolsep}{7pt}
{
     \begin{tabular}{l c c c c c c c c c c c}
          \toprule
          {variant}  
          & {{CLIPScore}}   & {{Aesthetic}}  & {{ImageR}} & {{PickScore}}  & {{HPSv2}}   & {{VQAScore}} & {{UnifiedR}} & {{Perception}} & {{HPSv3}} & \cellcolor{sem!15}{GPT-4}     \\ 
                                     
          \midrule
         {BruPA} &   \underline{26.547} &   \underline{6.516} &      \textbf{13.315} &               22.279 &      \textbf{28.037} &      \textbf{9.093} &      \textbf{3.371} &               26.390 &      \textbf{6.276} &               \cellcolor{sem!15}83.054 \\
  {BruPA (CaPO)} &      \textbf{26.567} &      \textbf{6.522} &               13.269 &   \underline{22.286} &               27.962 &   \underline{9.084} &               3.350 &   \underline{26.401} &               6.242 &   \cellcolor{sem!15}\underline{83.659} \\
  {BruPA (CaEN)} &               26.537 &               6.515 &   \underline{13.288} &      \textbf{22.293} &   \underline{28.032} &               9.062 &   \underline{3.358} &      \textbf{26.410} &   \underline{6.259} &      \cellcolor{sem!15}\textbf{83.717} \\
  \hline
         {FluPA} &      \textbf{26.436} &   \underline{6.546} &   \underline{13.859} &   \underline{22.577} &      \textbf{28.735} &   \underline{9.152} &   \underline{3.457} &   \underline{26.096} &      \textbf{7.000} &               \cellcolor{sem!15}87.609 \\
  {FluPA (CaPO)} &               26.398 &               6.498 &               13.773 &               22.559 &               28.398 &      \textbf{9.161} &               3.433 &               26.075 &               6.808 &               \cellcolor{sem!15}\underline{88.499} \\
  {FluPA (CaEN)} &   \underline{26.421} &      \textbf{6.557} &      \textbf{13.934} &      \textbf{22.616} &   \underline{28.642} &      \textbf{9.161} &      \textbf{3.474} &      \textbf{26.114} &   \underline{6.993} &      \cellcolor{sem!15}\textbf{88.541} \\
          \bottomrule
          
     \end{tabular}
}
\end{threeparttable}
}
\label{tab:caen}
\end{table*}

\textbf{Weighted Ensemble}. We find that assigning different weights to reward models when constructing ensemble scores significantly affects the performance. We report the results of two new variants—\textit{Linear Weighted Ensemble} (named \textbf{BruPA (linear)} and \textbf{FluPA (linear)}), and \textit{Softmax Weighted Ensemble} (named \textbf{BruPA (softmax)} and \textbf{FluPA (softmax)})—in \autoref{tab:weighted} and \autoref{fig:variants}. 
In \textbf{BruPA (vanilla)} and \textbf{FluPA (vanilla)}, each reward model is given equal weight.
In comparison, in BruPA (linear) and FluPA (linear), each reward model is weighted by its corresponding GPT-4 score (from \autoref{tab:brushnet_metrics_8cols_twobench} and \autoref{tab:flux_metrics_8col_twobench}), thereby granting greater influence to higher-performing models. The BruPA (softmax) and FluPA (softmax) further normalize these weights via a softmax transformation. Based on GPT-4 evaluations, both linear and softmax variants outperform the vanilla Ensemble approach. However, these weighting strategies remain overly simple and may further exacerbate reward hacking.

\begin{figure*}[t]
\begin{subfigure}[t]{\linewidth}
    \centering
    \includegraphics[width=0.11\linewidth, trim=0mm 0mm 0mm 0mm, clip]{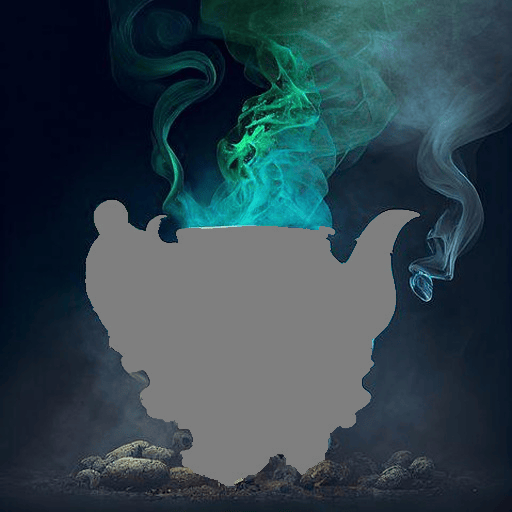}
    \includegraphics[width=0.11\linewidth, trim=0mm 0mm 0mm 0mm, clip]{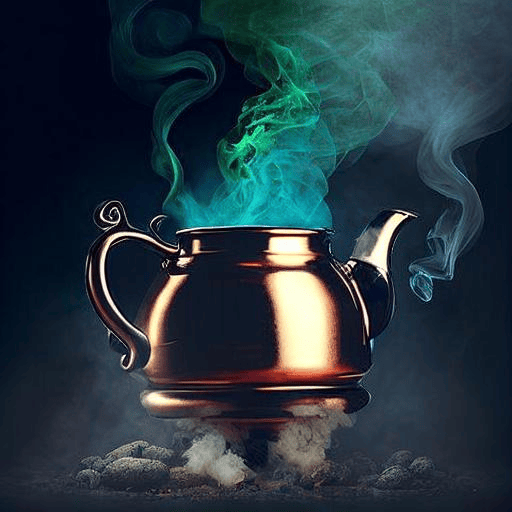}
    \includegraphics[width=0.11\linewidth, trim=0mm 0mm 0mm 0mm, clip]{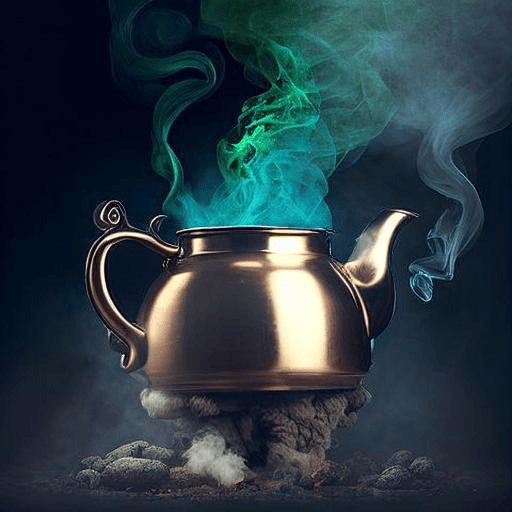}
    \includegraphics[width=0.11\linewidth, trim=0mm 0mm 0mm 0mm, clip]{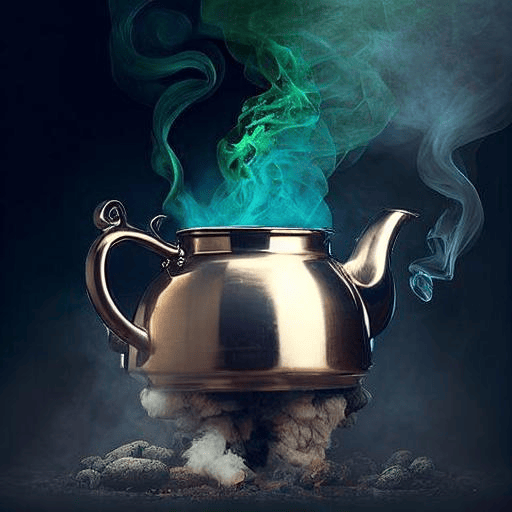}
    \includegraphics[width=0.11\linewidth, trim=0mm 0mm 0mm 0mm, clip]{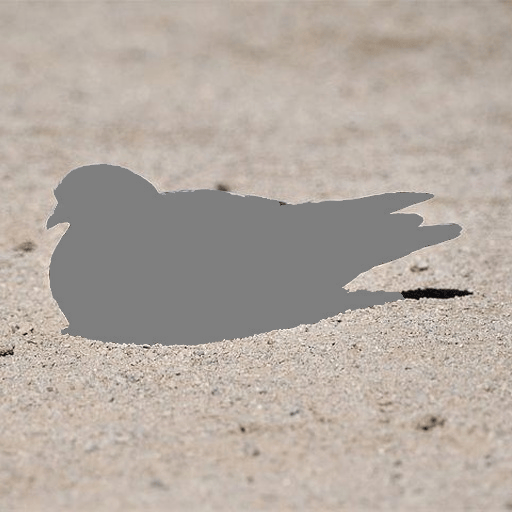}
    \includegraphics[width=0.11\linewidth, trim=0mm 0mm 0mm 0mm, clip]{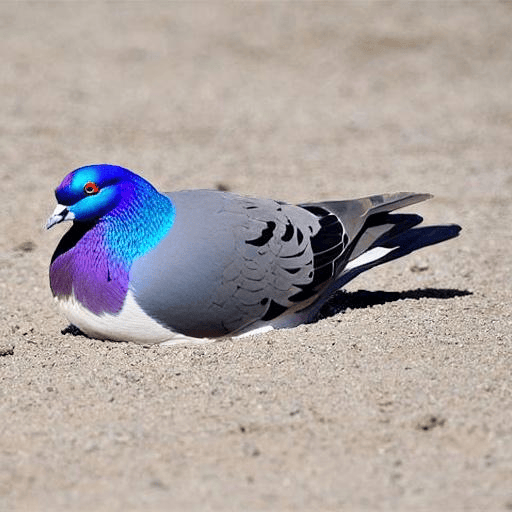}
    \includegraphics[width=0.11\linewidth, trim=0mm 0mm 0mm 0mm, clip]{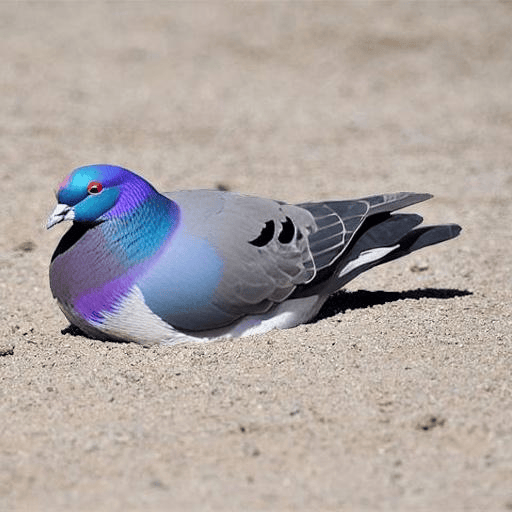}
    \includegraphics[width=0.11\linewidth, trim=0mm 0mm 0mm 0mm, clip]{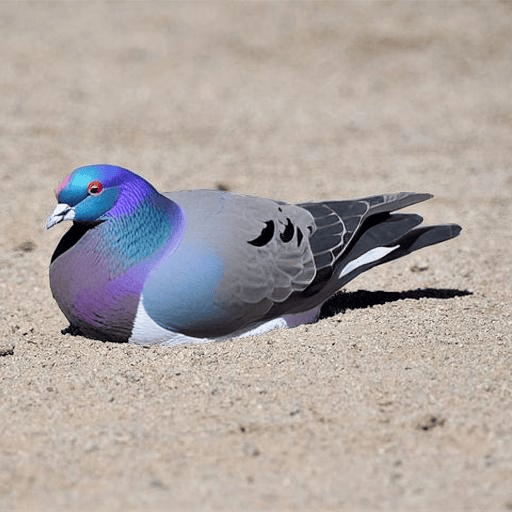}
    \includegraphics[width=0.11\linewidth, trim=0mm 0mm 0mm 0mm, clip]{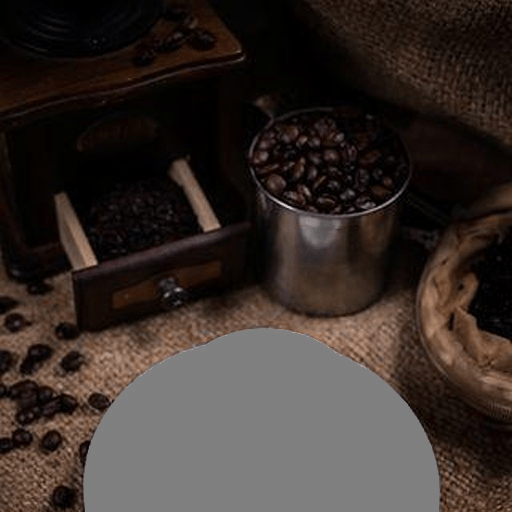}
    \includegraphics[width=0.11\linewidth, trim=0mm 0mm 0mm 0mm, clip]{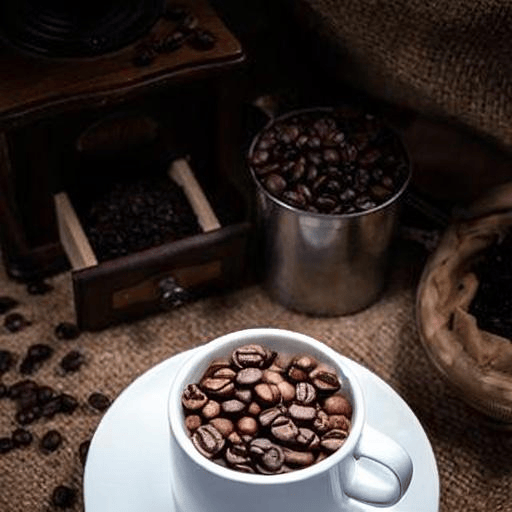}
    \includegraphics[width=0.11\linewidth, trim=0mm 0mm 0mm 0mm, clip]{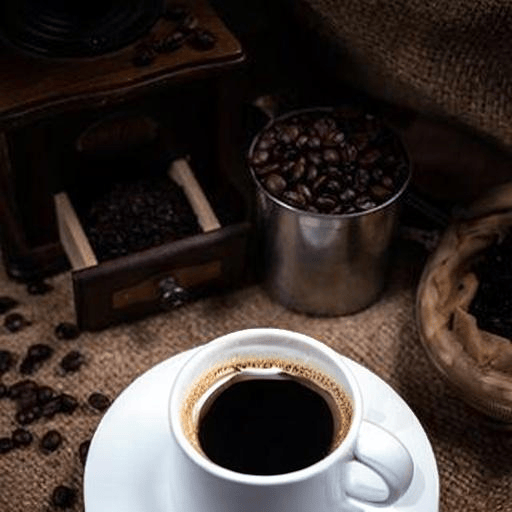}
    \includegraphics[width=0.11\linewidth, trim=0mm 0mm 0mm 0mm, clip]{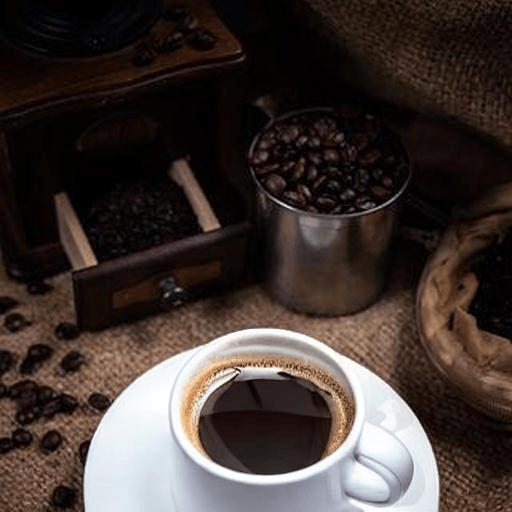}
    \includegraphics[width=0.11\linewidth, trim=0mm 0mm 0mm 0mm, clip]{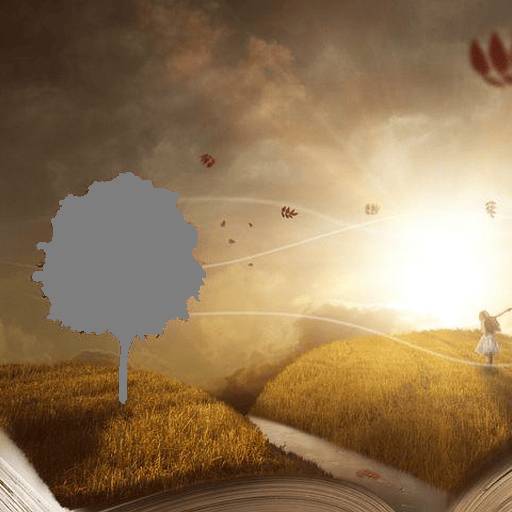}
    \includegraphics[width=0.11\linewidth, trim=0mm 0mm 0mm 0mm, clip]{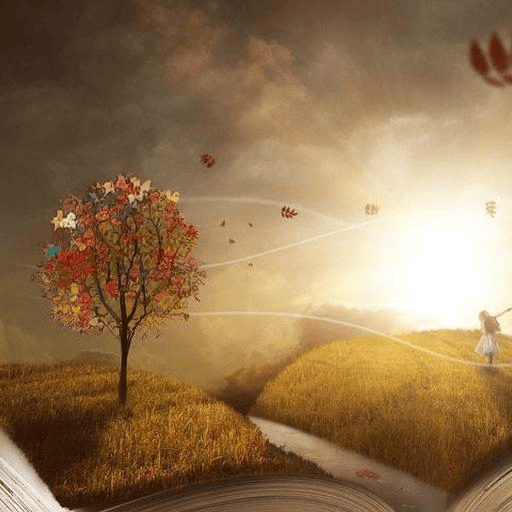}
    \includegraphics[width=0.11\linewidth, trim=0mm 0mm 0mm 0mm, clip]{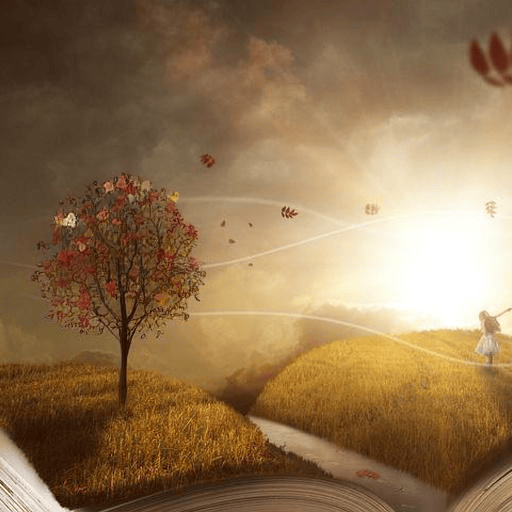}
    \includegraphics[width=0.11\linewidth, trim=0mm 0mm 0mm 0mm, clip]{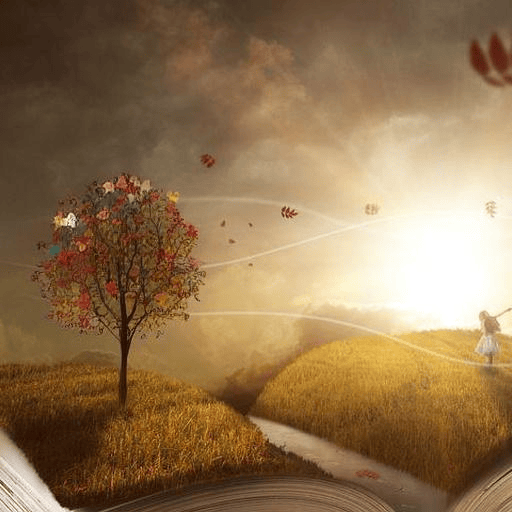}
    \caption{Examples from models trained using \textbf{BrushNet}.}
\end{subfigure}
\begin{subfigure}[t]{\linewidth}
    \centering
    \includegraphics[width=0.11\linewidth, trim=0mm 0mm 0mm 0mm, clip]{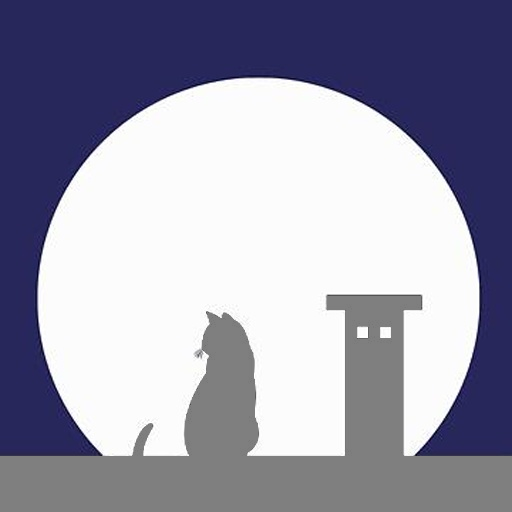}
    \includegraphics[width=0.11\linewidth, trim=0mm 0mm 0mm 0mm, clip]{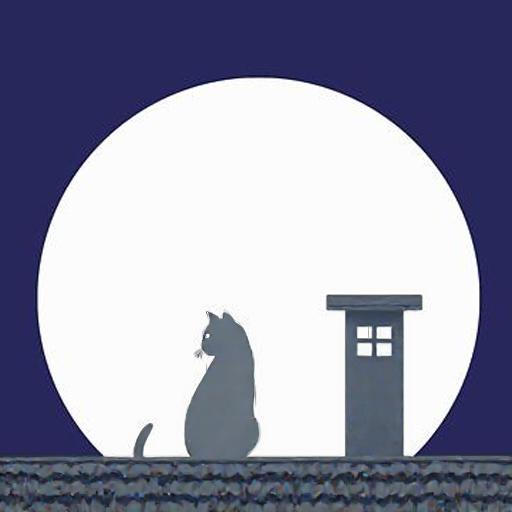}
    \includegraphics[width=0.11\linewidth, trim=0mm 0mm 0mm 0mm, clip]{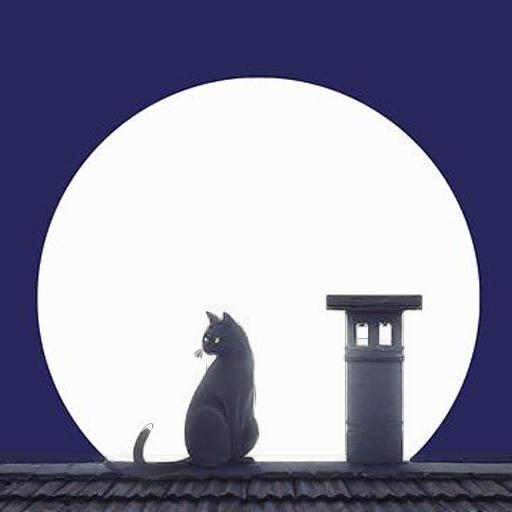}
    \includegraphics[width=0.11\linewidth, trim=0mm 0mm 0mm 0mm, clip]{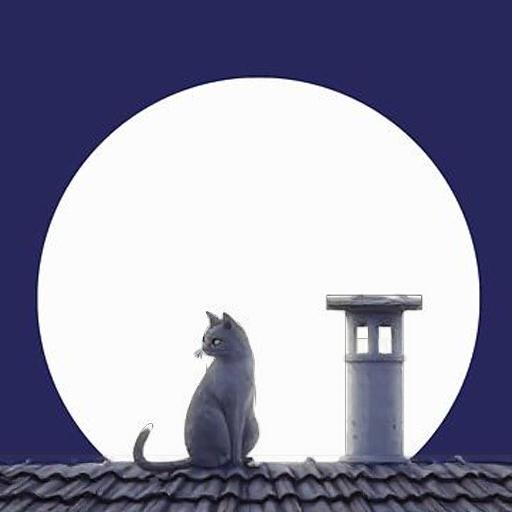}
    \includegraphics[width=0.11\linewidth, trim=0mm 0mm 0mm 0mm, clip]{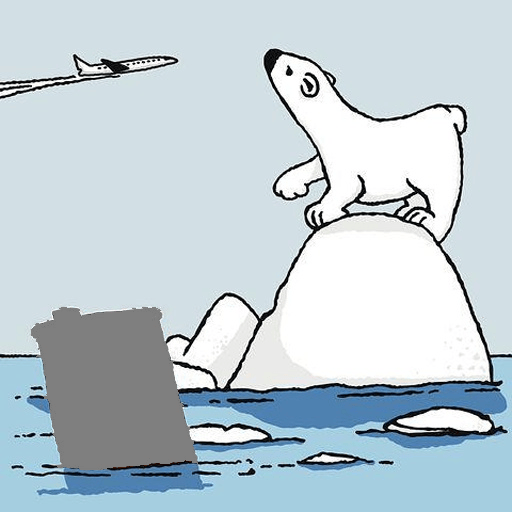}
    \includegraphics[width=0.11\linewidth, trim=0mm 0mm 0mm 0mm, clip]{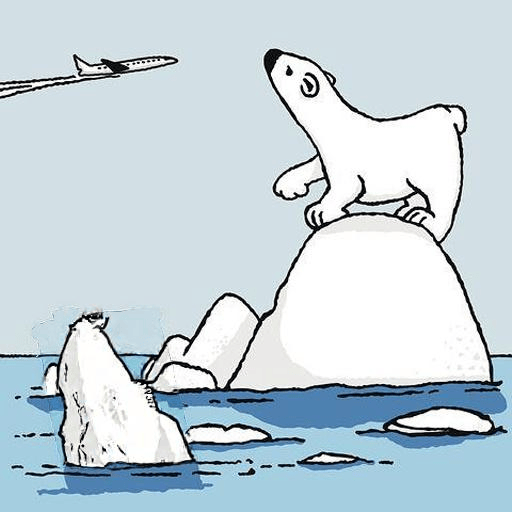}
    \includegraphics[width=0.11\linewidth, trim=0mm 0mm 0mm 0mm, clip]{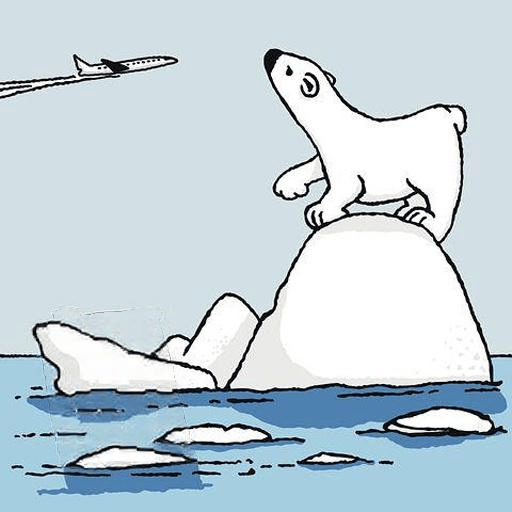}
    \includegraphics[width=0.11\linewidth, trim=0mm 0mm 0mm 0mm, clip]{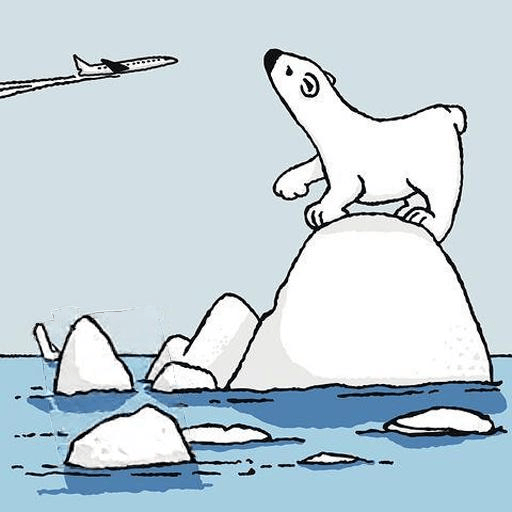}
    \includegraphics[width=0.11\linewidth, trim=0mm 0mm 0mm 0mm, clip]{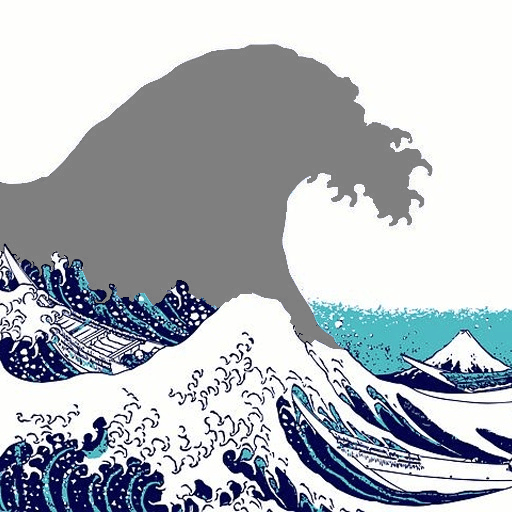}
    \includegraphics[width=0.11\linewidth, trim=0mm 0mm 0mm 0mm, clip]{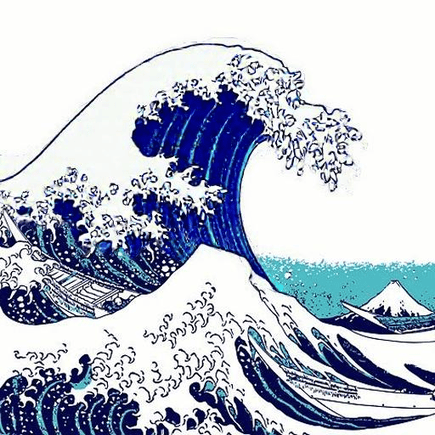}
    \includegraphics[width=0.11\linewidth, trim=0mm 0mm 0mm 0mm, clip]{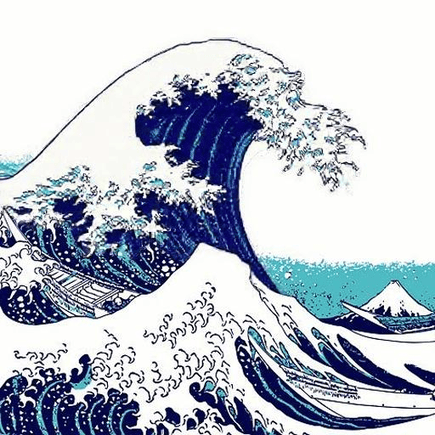}
    \includegraphics[width=0.11\linewidth, trim=0mm 0mm 0mm 0mm, clip]{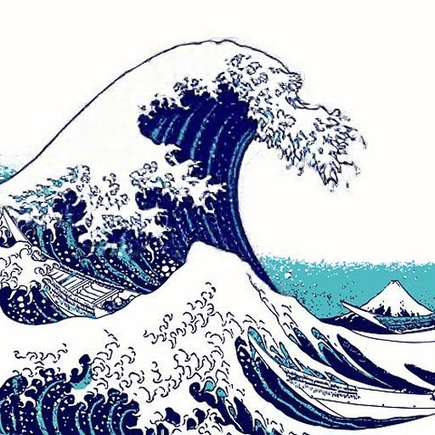}
    \includegraphics[width=0.11\linewidth, trim=0mm 0mm 0mm 0mm, clip]{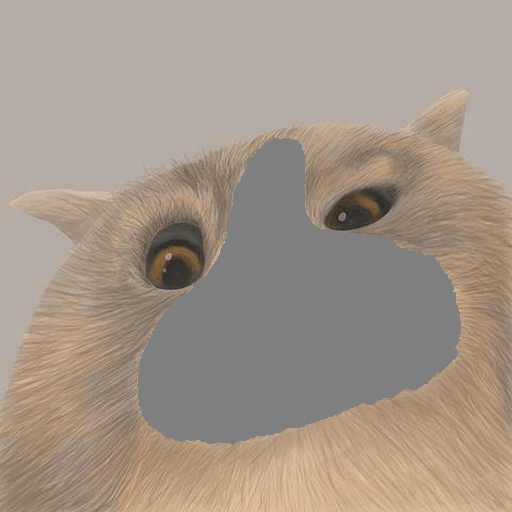}
    \includegraphics[width=0.11\linewidth, trim=0mm 0mm 0mm 0mm, clip]{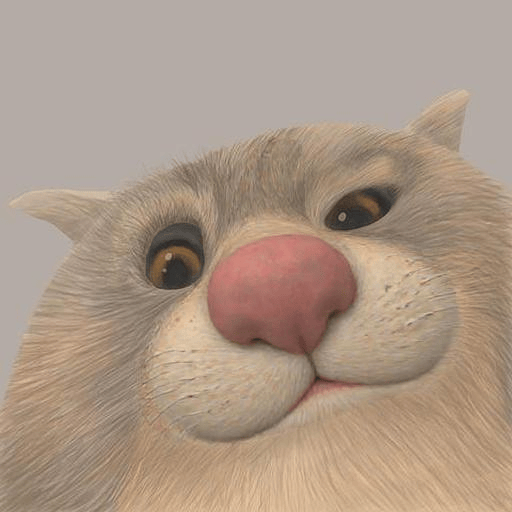}
    \includegraphics[width=0.11\linewidth, trim=0mm 0mm 0mm 0mm, clip]{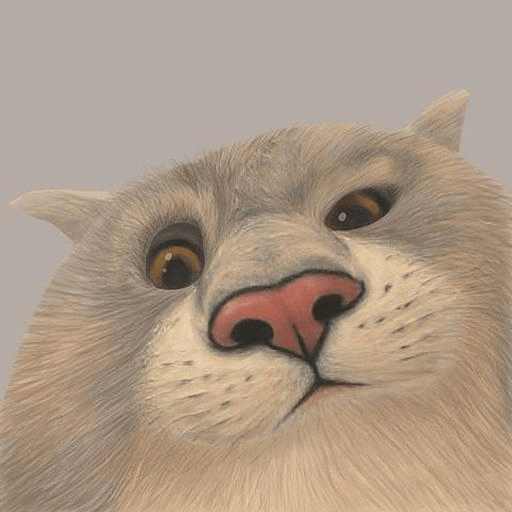}
    \includegraphics[width=0.11\linewidth, trim=0mm 0mm 0mm 0mm, clip]{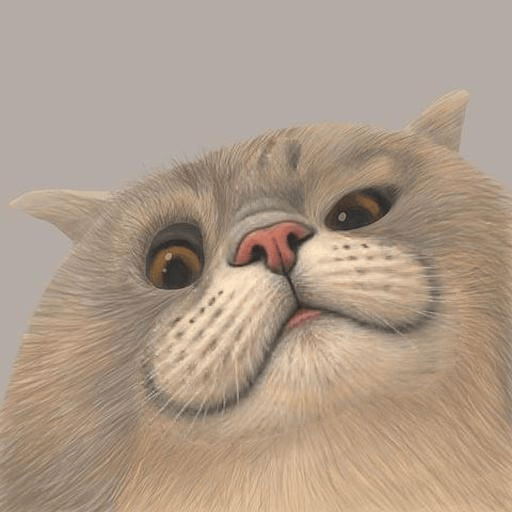}
    \caption{Examples from models trained using \textbf{FLUX.1 Fill}.}
\end{subfigure}
\caption{\color{rebuttal_blue}\textbf{Qualitative results.} In each sub-figure, the four images (from left to right) display: the \textit{masked image}, followed by inpainting results from models trained using \textit{Ensemble (vanilla)}, \textit{CaPO}, and \textit{CaEN}. We omit text prompts for brevity. Zoom in to see details. Find more examples in the Appendix.}
\label{fig:ensemble_bias}
\end{figure*}

\textbf{{Ensemble as a calibrated reward.}} One core difficulty in multi-reward preference alignment lies in the inconsistency across the reward distributions. CaPO~\cite{capo} addresses this challenge by approximating the advantage of sample $x_i$ using the average pairwise win rate between $x_i$ and $x_j$ in the batch $\{\mathbf{x}_i\}_{i=1}^N$ sampled from the reference policy $p_{\mathrm{ref}}(\cdot|\mathbf{c})$ under the conditions $\mathbf{c}$ for a given reward model $R$:
\begin{equation}
R_\mathrm{ca}(\mathbf{x}_i,\mathbf{c})=\frac{1}{N-1}\sum_{j\neq i}\sigma\left(R(\mathbf{x}_i,\mathbf{c})-R(\mathbf{x}_j,\mathbf{c})\right).
\label{eq:capo_reward}
\end{equation}
For an ensemble of $L$ reward models, the calibrated reward is given by
\begin{equation}
R_\mathrm{ca}(\mathbf{x},\mathbf{c})=\frac{1}{L}\sum_{j=1}^LR_\mathrm{ca}^{(j)}(\mathbf{x},\mathbf{c}).
\label{eq:capo_loss}
\end{equation}
Even though \autoref{eq:capo_reward} and \autoref{eq:capo_loss} align the scales of different reward models, they ignore the differences in variance. To this end, we revisit the \textbf{Ensemble} and reformulate it as
\begin{equation}
R_{\mathrm{En}}(\mathbf{x}_i, c) = \frac{1}{N-1}\sum_{j \ne i}\!\mathbf{1}\{R(\mathbf{x}_i, c)>R(\mathbf{x}_j, c)\},
\label{eq:caen_reward}
\end{equation}
where $\mathbf{1}\{\cdot\}$ denotes the indicator function. We further replace the calibrated reward in CaPO with \autoref{eq:caen_reward} , which yields the CaEN loss:
\begin{equation}
R_\mathrm{En}(\mathbf{x},\mathbf{c})=\frac{1}{L}\sum_{j=1}^LR_\mathrm{En}^{(j)}(\mathbf{x},\mathbf{c}).
\label{eq:caen_loss}
\end{equation}
\autoref{eq:caen_reward} and \autoref{eq:caen_loss} not only account for scale alignment, but also eliminate the influence of variance differences. Following CaPO, we obtain the {CaEN} objective:
\begin{equation}
\begin{aligned}
\mathcal{L}_{\mathrm{CaEN}}(\theta)
&=\mathbb{E}_{t,\boldsymbol{\epsilon}^{w},\boldsymbol{\epsilon}^{l}}\bigg[(R_{\mathrm{En}}(\mathbf{x}^{w},\mathbf{c})-R_{\mathrm{En}}(\mathbf{x}^{l},\mathbf{c}))
\\&-(-\beta((\mathcal{L}_{\theta}^{w}-\mathcal{L}_{\mathrm{ref}}^{w})-(\mathcal{L}_{\theta}^{l}-\mathcal{L}_{\mathrm{ref}}^{l})))\bigg]^2,
\end{aligned}
\label{eq:caen_obj}
\end{equation}
where we assign different noises $\epsilon^{w},\epsilon^{l}\sim\mathcal{N}(0,I)$ to $(\mathcal{L}_{\theta}^{w},\mathcal{L}_{\mathrm{ref}}^{w})$ and $(\mathcal{L}_{\theta}^{l},\mathcal{L}_{\mathrm{ref}}^{l})$, following CaPO. We train FLUX.1 fill using a learning rate of 1e-5 and choose $\beta$ by sweeping over \{30, 50, 100\} for both CaPO and CaEN. For BrushNet, we use a learning rate of 1e-6 and choose $\beta$ by sweeping over \{300, 500, 1000\} for both CaPO and CaEN, and we report the comparison among vanilla Ensemble, CaPO and CaEN in \autoref{tab:caen}. Similar to the observations in \autoref{sec:effective}, we find that models trained on vanilla Ensemble-constructed data, including both BruPA and FluPA, outperform those trained on CaPO-constructed data when evaluated by public reward models but fail under GPT-4 evaluation. On top of this, we posit that reward hacking still undermines Ensemble while CaPO and CaEN further mitigate the hacking. To further understand the potential biases in Ensemble, we sample inpainting examples and present them in \autoref{fig:ensemble_bias} and the appendix. Empirically, we find that CaEN amplifies the biases shared across reward models, as discussed in \autoref{sec:effective}. To offset these biases, we assign higher weights to CLIPScore and Perception. Besides, compared with CaPO, CaEN attains the best results in 8 out of 10 evaluations and the second-best results in more than half of the cases. Both qualitative and quantitative results consistently demonstrate that both CaPO and CaEN mitigate the biases inherent in the vanilla Ensemble, with CaEN exhibiting better performance.

}

{\color{rebuttal_blue}
\section{Application to Object Removal}
\label{sec:removal}
\begin{figure*}[t]
\centering
	\includegraphics[width=0.1\linewidth, trim=0mm 0mm 0mm 0mm, clip]{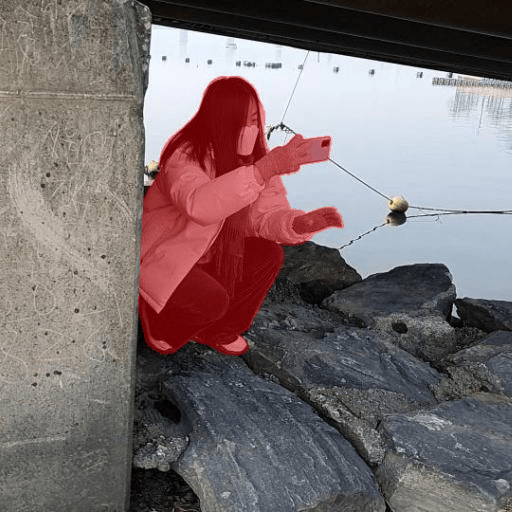}
  \includegraphics[width=0.1\linewidth, trim=0mm 0mm 0mm 0mm, clip]{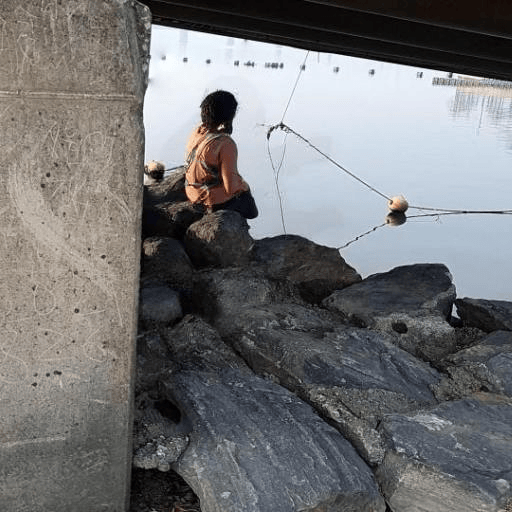}
  \includegraphics[width=0.1\linewidth, trim=0mm 0mm 0mm 0mm, clip]{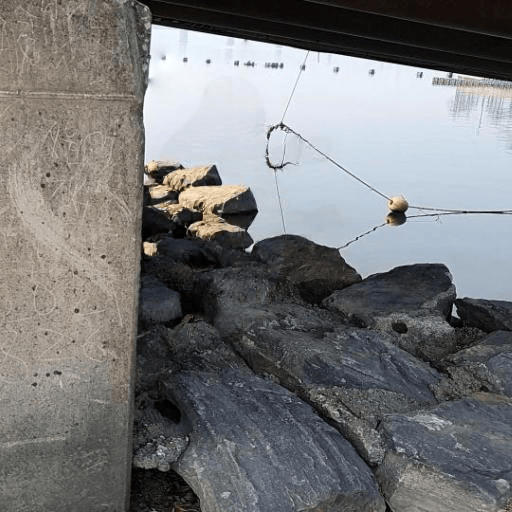}
  \includegraphics[width=0.1\linewidth, trim=0mm 0mm 0mm 0mm, clip]{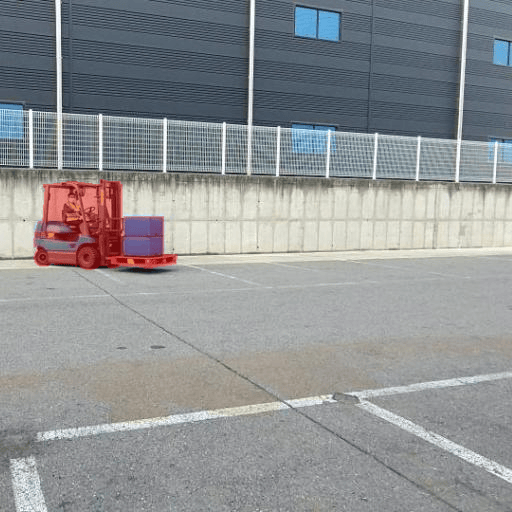}
  \includegraphics[width=0.1\linewidth, trim=0mm 0mm 0mm 0mm, clip]{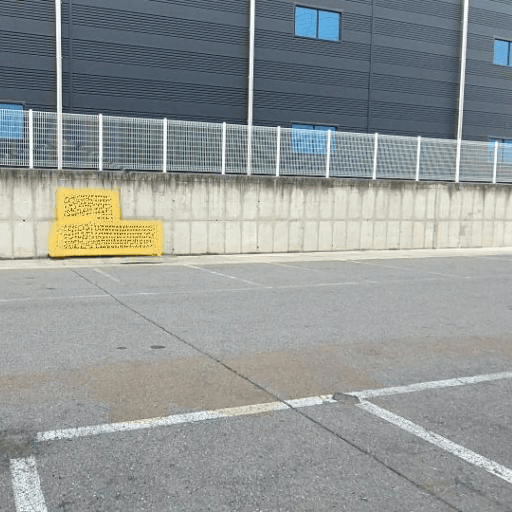}
  \includegraphics[width=0.1\linewidth, trim=0mm 0mm 0mm 0mm, clip]{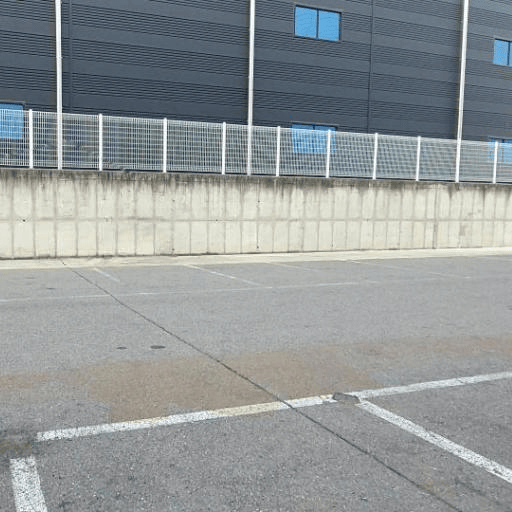}
  \includegraphics[width=0.1\linewidth, trim=0mm 0mm 0mm 0mm, clip]{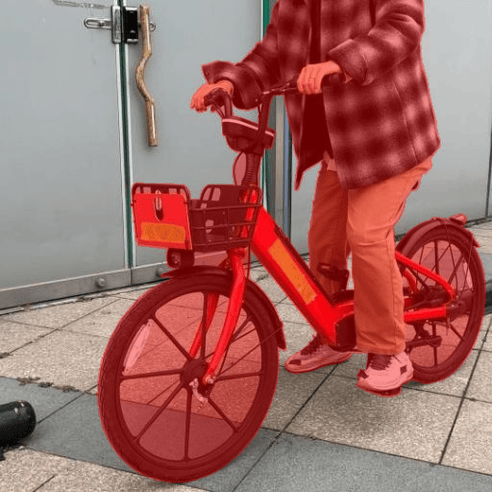}
  \includegraphics[width=0.1\linewidth, trim=0mm 0mm 0mm 0mm, clip]{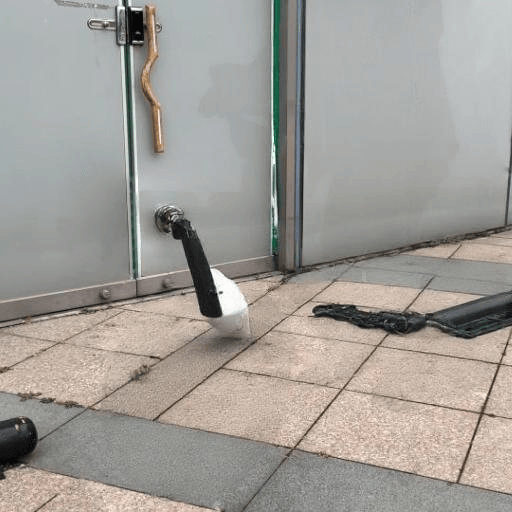}
  \includegraphics[width=0.1\linewidth, trim=0mm 0mm 0mm 0mm, clip]{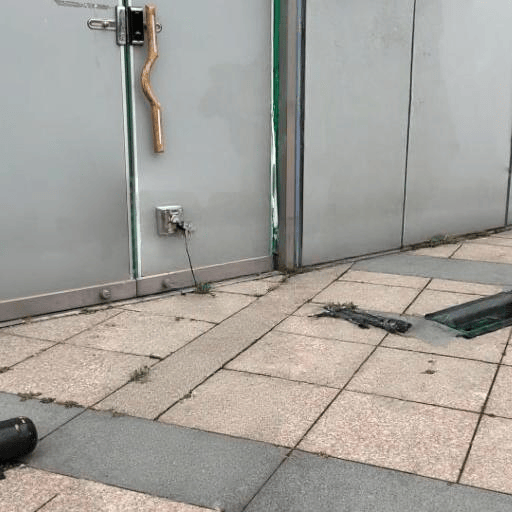}
  \includegraphics[width=0.1\linewidth, trim=0mm 0mm 0mm 0mm, clip]{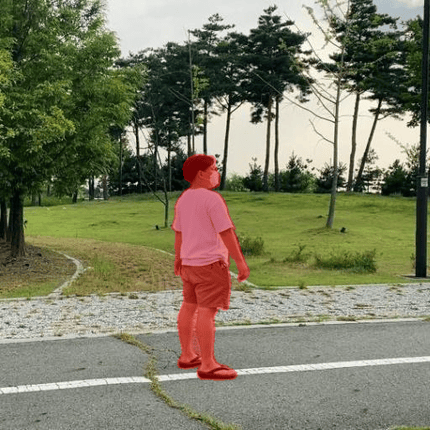}
  \includegraphics[width=0.1\linewidth, trim=0mm 0mm 0mm 0mm, clip]{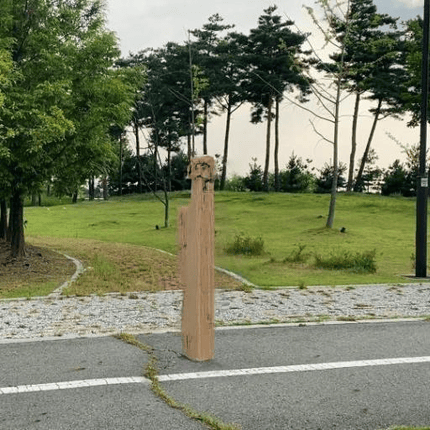}
  \includegraphics[width=0.1\linewidth, trim=0mm 0mm 0mm 0mm, clip]{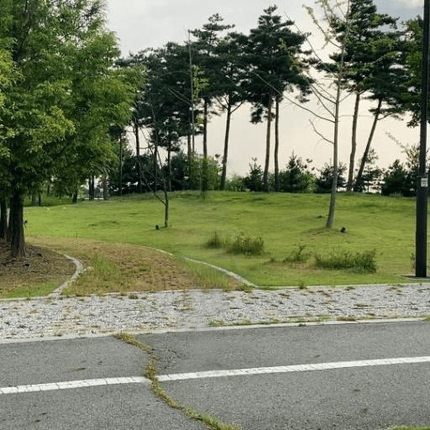}
  \includegraphics[width=0.1\linewidth, trim=0mm 0mm 0mm 0mm, clip]{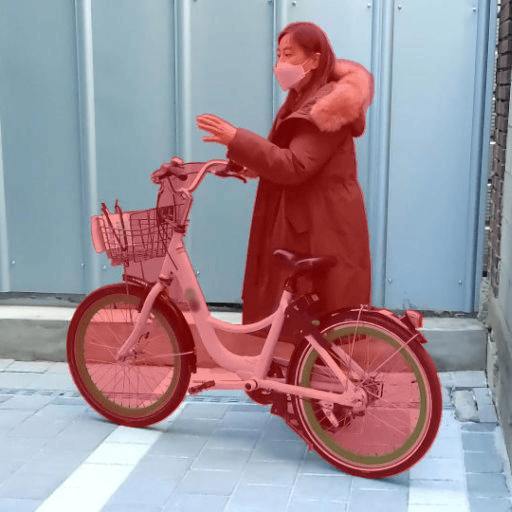}
  \includegraphics[width=0.1\linewidth, trim=0mm 0mm 0mm 0mm, clip]{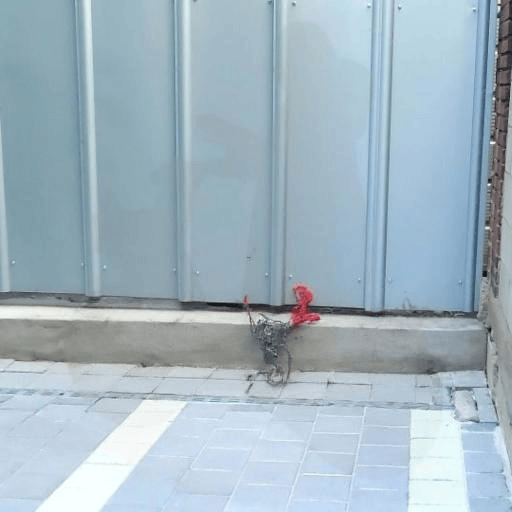}
  \includegraphics[width=0.1\linewidth, trim=0mm 0mm 0mm 0mm, clip]{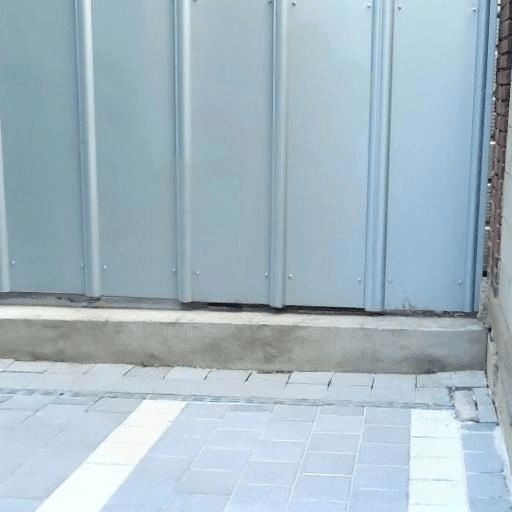}
  \includegraphics[width=0.1\linewidth, trim=0mm 0mm 0mm 0mm, clip]{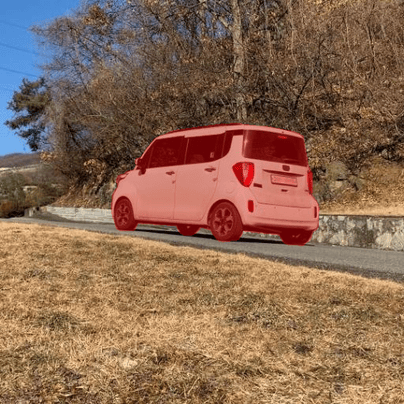}
  \includegraphics[width=0.1\linewidth, trim=0mm 0mm 0mm 0mm, clip]{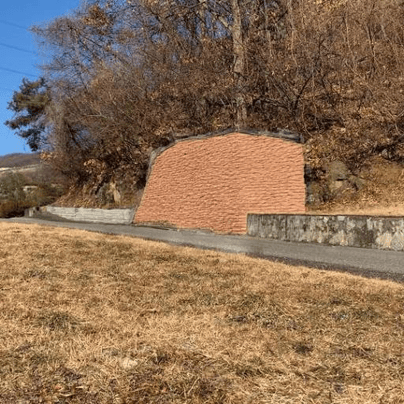}
  \includegraphics[width=0.1\linewidth, trim=0mm 0mm 0mm 0mm, clip]{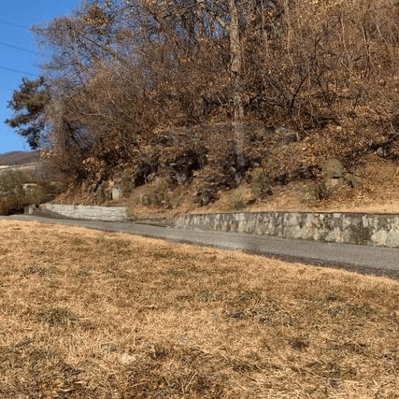}

\caption{\color{rebuttal_blue}\textbf{Qualitative results of ablations in object removal.} In each sub-figure, the three images (from left to right) display: the \textit{masked image}, followed by inpainting results from \textit{baseline models} and \textit{baseline models + preference alignment}.}
\label{fig:removal_sample}
\end{figure*}
In this section, we evaluate preference alignment with the proposed method on the task of object removal.

\subsection{Experiment Setup}
\textbf{Training details}. We generate preference data on the high-quality object removal dataset RORem~\cite{RORem}. Specifically, we generate 16 candidates for each image--mask pair and leave the text prompt empty. We train BruPA and FluPA for 800 steps, following the same training recipes.

\textbf{Evaluation details}. We construct the test set by randomly sampling 600 image--mask pairs from RORD-Val~\cite{RORem}. We evaluate the efficacy of our methods using three widely adopted metrics: PSNR, CLIPScore~\cite{clipscore}, and LPIPS~\cite{lpips}.

\subsection{Comparison with the Baselines}
\begin{table}[t]
\centering
\caption{\color{rebuttal_blue}Results for object removal.}
\label{tab:removal}
\resizebox{1.\linewidth}{!}{
\begin{threeparttable}  
\renewcommand{\arraystretch}{1.3}
\setlength{\tabcolsep}{4pt}
{
     \begin{tabular}{l c c c c}
          \toprule
          {inpainting model}  
          & {{CLIPScore}}   & {{PSNR}}  & {{PSNR-BG}} & {{LPIPS$\downarrow$}}      \\ 
                                     
          \midrule
          {BrushNet} &            87.640 &            0.116 &            43.445 &            0.082 \\
          {BruPA} &   \textbf{89.896} &   \textbf{1.342} &   \textbf{43.791} &   \textbf{0.071} \\
          \midrule
          {FLUX.1 Fill} &            92.736 &            3.615 &            43.820 &            0.051 \\
          {FluPA} &   \textbf{93.549} &   \textbf{4.353} &   \textbf{44.072} &   \textbf{0.047} \\
          \bottomrule
     \end{tabular}
}
\end{threeparttable}
}
\end{table}

To evaluate the efficacy of preference alignment, we report the results in~\autoref{tab:removal}. The results demonstrate that BruPA and FluPA significantly outperform their baseline models on the object-removal task. This observation suggests that our method may generalize to other image inpainting and editing tasks. We attribute this improvement to the fact that preference alignment training suppressing the model's tendency to generate spurious or hallucinated content.
}

\section{Conclusion and Discussion}
We conduct extensive studies on image inpainting with preference alignment and obtain key insights into the effectiveness, scalability, and challenges in achieving alignment. We find that a simple ensemble method mitigates biases and achieves non-trivial results {\color{rebuttal_blue}in the conference version of this work. We further generalize preference alignment from creative image inpainting to object-removal inpainting, a more challenging setting that emphasizes coherent and visually plausible background completion rather than open-ended generation. We provide a substantially more comprehensive investigation of the ensemble strategy by analyzing its limitations under reward hacking, studying weighting schemes across reward models, and examining the efficiency gains enabled by filtering out invalid reward signals. Building upon the simple ensemble method, we find it still vulnerable to reward hacking, and we therefore propose a calibrated ensemble that mitigates bias-induced hacking and establishes a new state of the art}. 

{\color{rebuttal_blue}\section{Limitations and Potential Negative Social Impacts}
Our work has the following limitations. First, our work is confined to image data, and we leave the extension of these findings to video and 3D data as future work. Second, our approach relies on off-the-shelf reward models to construct preference training data. As shown in our analysis, these reward models can exhibit biases and may induce reward hacking. While our ensemble and calibrated ensemble methods improve robustness, they may not fully eliminate this issue. Finally, our evaluation still depends partly on automatic evaluators, including vision-language model-based judges, which cannot fully replace human judgments. Researchers should also be aware of the potential negative social impacts arising from the misuse of inpainting models. For example, such models can generate highly realistic content that may be indistinguishable from real photographs.}
\FloatBarrier

{
\small
\bibliographystyle{ieeenat_fullname}
\bibliography{iclr2026_conference}

@article{clipscore,
  title={Clipscore: A reference-free evaluation metric for image captioning},
  author={Hessel, Jack and Holtzman, Ari and Forbes, Maxwell and Bras, Ronan Le and Choi, Yejin},
  journal={arXiv preprint arXiv:2104.08718},
  year={2021}
}

@article{aesthetic,
  title={Laion-5b: An open large-scale dataset for training next generation image-text models},
  author={Schuhmann, Christoph and Beaumont, Romain and Vencu, Richard and Gordon, Cade and Wightman, Ross and Cherti, Mehdi and Coombes, Theo and Katta, Aarush and Mullis, Clayton and Wortsman, Mitchell and others},
  journal={Advances in neural information processing systems},
  year={2022}
}

@article{imagereward,
  title={Imagereward: Learning and evaluating human preferences for text-to-image generation},
  author={Xu, Jiazheng and Liu, Xiao and Wu, Yuchen and Tong, Yuxuan and Li, Qinkai and Ding, Ming and Tang, Jie and Dong, Yuxiao},
  journal={Advances in Neural Information Processing Systems},
  year={2023}
}

@article{pickscore,
  title={Pick-a-pic: An open dataset of user preferences for text-to-image generation},
  author={Kirstain, Yuval and Polyak, Adam and Singer, Uriel and Matiana, Shahbuland and Penna, Joe and Levy, Omer},
  journal={Advances in neural information processing systems},
  year={2023}
}

@article{hpsv2,
  title={Human preference score v2: A solid benchmark for evaluating human preferences of text-to-image synthesis},
  author={Wu, Xiaoshi and Hao, Yiming and Sun, Keqiang and Chen, Yixiong and Zhu, Feng and Zhao, Rui and Li, Hongsheng},
  journal={arXiv preprint arXiv:2306.09341},
  year={2023}
}

@article{rl-in-mm-survey,
  title={Reinforcement Learning in Generative Multimodal AI: A Survey},
  author={Hu, Zijing and Yuan, Junkun and Han, Kairong and Tong, Yunze and Zhang, Shengyu and Wu, Fei and Kuang, Kun},
  journal={Authorea Preprints},
  year={2026},
  publisher={Authorea}
}

@inproceedings{vqascore,
  title={Evaluating text-to-visual generation with image-to-text generation},
  author={Lin, Zhiqiu and Pathak, Deepak and Li, Baiqi and Li, Jiayao and Xia, Xide and Neubig, Graham and Zhang, Pengchuan and Ramanan, Deva},
  booktitle={European Conference on Computer Vision},
  year={2024},
}

@article{unifiedreward,
  title={Unified reward model for multimodal understanding and generation},
  author={Wang, Yibin and Zang, Yuhang and Li, Hao and Jin, Cheng and Wang, Jiaqi},
  journal={arXiv preprint arXiv:2503.05236},
  year={2025}
}

@article{perception,
  title={Perception encoder: The best visual embeddings are not at the output of the network},
  author={Bolya, Daniel and Huang, Po-Yao and Sun, Peize and Cho, Jang Hyun and Madotto, Andrea and Wei, Chen and Ma, Tengyu and Zhi, Jiale and Rajasegaran, Jathushan and Rasheed, Hanoona and others},
  journal={arXiv preprint arXiv:2504.13181},
  year={2025}
}

@inproceedings{thermodynamics,
  title={Deep unsupervised learning using nonequilibrium thermodynamics},
  author={Sohl-Dickstein, Jascha and Weiss, Eric and Maheswaranathan, Niru and Ganguli, Surya},
  booktitle={International conference on machine learning},
  year={2015},
}

@article{DDPM,
  title={Denoising diffusion probabilistic models},
  author={Ho, Jonathan and Jain, Ajay and Abbeel, Pieter},
  journal={Advances in neural information processing systems},
  year={2020}
}

@article{hap,
  title={Hap: Structure-aware masked image modeling for human-centric perception},
  author={Yuan, Junkun and Zhang, Xinyu and Zhou, Hao and Wang, Jian and Qiu, Zhongwei and Shao, Zhiyin and Zhang, Shaofeng and Long, Sifan and Kuang, Kun and Yao, Kun and others},
  journal={Advances in Neural Information Processing Systems},
  year={2023}
}

@inproceedings{follow-your-emoji,
  title={Follow-your-emoji: Fine-controllable and expressive freestyle portrait animation},
  author={Ma, Yue and Liu, Hongyu and Wang, Hongfa and Pan, Heng and He, Yingqing and Yuan, Junkun and Zeng, Ailing and Cai, Chengfei and Shum, Heung-Yeung and Liu, Wei and others},
  booktitle={SIGGRAPH Asia 2024 Conference Papers},
  year={2024}
}

@article{follow-your-canvas,
  title={Follow-your-canvas: Higher-resolution video outpainting with extensive content generation},
  author={Chen, Qihua and Ma, Yue and Wang, Hongfa and Yuan, Junkun and Zhao, Wenzhe and Tian, Qi and Wang, Hongmei and Min, Shaobo and Chen, Qifeng and Liu, Wei},
  journal={arXiv preprint arXiv:2409.01055},
  year={2024}
}

@article{DDIM,
  title={Denoising diffusion implicit models},
  author={Song, Jiaming and Meng, Chenlin and Ermon, Stefano},
  journal={arXiv preprint arXiv:2010.02502},
  year={2020}
}

@article{FlowMatching,
  title={Flow matching for generative modeling},
  author={Lipman, Yaron and Chen, Ricky TQ and Ben-Hamu, Heli and Nickel, Maximilian and Le, Matt},
  journal={arXiv preprint arXiv:2210.02747},
  year={2022}
}

@inproceedings{Sit,
  title={Sit: Exploring flow and diffusion-based generative models with scalable interpolant transformers},
  author={Ma, Nanye and Goldstein, Mark and Albergo, Michael S and Boffi, Nicholas M and Vanden-Eijnden, Eric and Xie, Saining},
  booktitle={European Conference on Computer Vision},
  year={2024},
}

@article{PreAliSurvey,
  title={Preference Alignment on Diffusion Model: A Comprehensive Survey for Image Generation and Editing},
  author={Wu, Sihao and Si, Xiaonan and Xing, Chi and Wang, Jianhong and Jin, Gaojie and Cheng, Guangliang and Zhang, Lijun and Huang, Xiaowei},
  journal={arXiv preprint arXiv:2502.07829},
  year={2025}
}

@article{DDPO,
  title={Training diffusion models with reinforcement learning},
  author={Black, Kevin and Janner, Michael and Du, Yilun and Kostrikov, Ilya and Levine, Sergey},
  journal={arXiv preprint arXiv:2305.13301},
  year={2023}
}

@inproceedings{BrushNet,
  title={Brushnet: A plug-and-play image inpainting model with decomposed dual-branch diffusion},
  author={Ju, Xuan and Liu, Xian and Wang, Xintao and Bian, Yuxuan and Shan, Ying and Xu, Qiang},
  booktitle={European Conference on Computer Vision},
  year={2024},
}

@inproceedings{EditBench,
  title={Imagen editor and editbench: Advancing and evaluating text-guided image inpainting},
  author={Wang, Su and Saharia, Chitwan and Montgomery, Ceslee and Pont-Tuset, Jordi and Noy, Shai and Pellegrini, Stefano and Onoe, Yasumasa and Laszlo, Sarah and Fleet, David J and Soricut, Radu and others},
  booktitle={IEEE/CVF Conference on Computer Vision and Pattern Recognition},
  year={2023}
}

@article{GPT-4,
  title={Gpt-4 technical report},
  author={Achiam, Josh and Adler, Steven and Agarwal, Sandhini and Ahmad, Lama and Akkaya, Ilge and Aleman, Florencia Leoni and Almeida, Diogo and Altenschmidt, Janko and Altman, Sam and Anadkat, Shyamal and others},
  journal={arXiv preprint arXiv:2303.08774},
  year={2023}
}

@misc{FLUX,
  title={Flux.1-dev},
  author={BlackForestLabs},
  howpublished = {\url{https://huggingface.co/black-forest-labs/FLUX.1-dev}},
  year={2024},
  note={Accessed: 2025-07-12}
}

@misc{Qwen3-VL,
  title={Qwen3-VL},
  author={QwenTeam},
  howpublished = {\url{https://qwen.ai/blog?id=qwen3-vl}},
  year={2025},
  note={Accessed: 2025-11-20}
}

@inproceedings{Diffusion-DPO,
  title={Diffusion model alignment using direct preference optimization},
  author={Wallace, Bram and Dang, Meihua and Rafailov, Rafael and Zhou, Linqi and Lou, Aaron and Purushwalkam, Senthil and Ermon, Stefano and Xiong, Caiming and Joty, Shafiq and Naik, Nikhil},
  booktitle={IEEE/CVF Conference on Computer Vision and Pattern Recognition},
  pages={8228--8238},
  year={2024}
}

@article{Prefpaint,
  title={Prefpaint: Aligning image inpainting diffusion model with human preference},
  author={Liu, Kendong and Zhu, Zhiyu and Li, Chuanhao and Liu, Hui and Zeng, Huanqiang and Hou, Junhui},
  journal={Advances in Neural Information Processing Systems},
  year={2024}
}

@article{HunyuanVideo,
  title={Hunyuanvideo: A systematic framework for large video generative models},
  author={Kong, Weijie and Tian, Qi and Zhang, Zijian and Min, Rox and Dai, Zuozhuo and Zhou, Jin and Xiong, Jiangfeng and Li, Xin and Wu, Bo and Zhang, Jianwei and others},
  journal={arXiv preprint arXiv:2412.03603},
  year={2024}
}

@book{rl,
  title={Reinforcement learning: An introduction},
  author={Sutton, Richard S and Barto, Andrew G and others},
  year={1998},
  publisher={MIT press Cambridge}
}

@article{Wan,
  title={Wan: Open and advanced large-scale video generative models},
  author={Wan, Team and Wang, Ang and Ai, Baole and Wen, Bin and Mao, Chaojie and Xie, Chen-Wei and Chen, Di and Yu, Feiwu and Zhao, Haiming and Yang, Jianxiao and others},
  journal={arXiv preprint arXiv:2503.20314},
  year={2025}
}

@inproceedings{capo,
  title={Calibrated multi-preference optimization for aligning diffusion models},
  author={Lee, Kyungmin and Li, Xiahong and Wang, Qifei and He, Junfeng and Ke, Junjie and Yang, Ming-Hsuan and Essa, Irfan and Shin, Jinwoo and Yang, Feng and Li, Yinxiao},
  booktitle={IEEE/CVF Conference on Computer Vision and Pattern Recognition},
  year={2025}
}

@article{Seedream,
  title={Seedream 3.0 technical report},
  author={Gao, Yu and Gong, Lixue and Guo, Qiushan and Hou, Xiaoxia and Lai, Zhichao and Li, Fanshi and Li, Liang and Lian, Xiaochen and Liao, Chao and Liu, Liyang and others},
  journal={arXiv preprint arXiv:2504.11346},
  year={2025}
}

@article{Seedance,
  title={Seedance 1.0: Exploring the Boundaries of Video Generation Models},
  author={Gao, Yu and Guo, Haoyuan and Hoang, Tuyen and Huang, Weilin and Jiang, Lu and Kong, Fangyuan and Li, Huixia and Li, Jiashi and Li, Liang and Li, Xiaojie and others},
  journal={arXiv preprint arXiv:2506.09113},
  year={2025}
}

@inproceedings{SD3,
  title={Scaling rectified flow transformers for high-resolution image synthesis},
  author={Esser, Patrick and Kulal, Sumith and Blattmann, Andreas and Entezari, Rahim and M{\"u}ller, Jonas and Saini, Harry and Levi, Yam and Lorenz, Dominik and Sauer, Axel and Boesel, Frederic and others},
  booktitle={International conference on machine learning},
  year={2024}
}

@inproceedings{UNet,
  title={U-net: Convolutional networks for biomedical image segmentation},
  author={Ronneberger, Olaf and Fischer, Philipp and Brox, Thomas},
  booktitle={International Conference on Medical image computing and computer-assisted intervention},
  year={2015},
}

@article{Transformer,
  title={Attention is all you need},
  author={Vaswani, Ashish and Shazeer, Noam and Parmar, Niki and Uszkoreit, Jakob and Jones, Llion and Gomez, Aidan N and Kaiser, {\L}ukasz and Polosukhin, Illia},
  journal={Advances in neural information processing systems},
  year={2017}
}

@article{PPO,
  title={Proximal policy optimization algorithms},
  author={Schulman, John and Wolski, Filip and Dhariwal, Prafulla and Radford, Alec and Klimov, Oleg},
  journal={arXiv preprint arXiv:1707.06347},
  year={2017}
}

@article{GRPO,
  title={Deepseek-r1: Incentivizing reasoning capability in llms via reinforcement learning},
  author={Guo, Daya and Yang, Dejian and Zhang, Haowei and Song, Junxiao and Zhang, Ruoyu and Xu, Runxin and Zhu, Qihao and Ma, Shirong and Wang, Peiyi and Bi, Xiao and others},
  journal={arXiv preprint arXiv:2501.12948},
  year={2025}
}

@article{RLHF,
  title={Training a helpful and harmless assistant with reinforcement learning from human feedback},
  author={Bai, Yuntao and Jones, Andy and Ndousse, Kamal and Askell, Amanda and Chen, Anna and DasSarma, Nova and Drain, Dawn and Fort, Stanislav and Ganguli, Deep and Henighan, Tom and others},
  journal={arXiv preprint arXiv:2204.05862},
  year={2022}
}

@article{DPO,
  title={Direct preference optimization: Your language model is secretly a reward model},
  author={Rafailov, Rafael and Sharma, Archit and Mitchell, Eric and Manning, Christopher D and Ermon, Stefano and Finn, Chelsea},
  journal={Advances in neural information processing systems},
  year={2023}
}

@inproceedings{inpaint,
  title={Image inpainting},
  author={Bertalmio, Marcelo and Sapiro, Guillermo and Caselles, Vincent and Ballester, Coloma},
  booktitle={Proceedings of the 27th annual conference on Computer graphics and interactive techniques},
  year={2000}
}

@article{ma2026group,
  title={Group Editing: Edit Multiple Images in One Go},
  author={Ma, Yue and Wang, Xinyu and Ma, Qianli and Wang, Qinghe and Zheng, Mingzhe and Yang, Xiangpeng and Li, Hao and Zhao, Chongbo and Ying, Jixuan and Yang, Harry and others},
  journal={arXiv preprint arXiv:2603.22883},
  year={2026}
}

@inproceedings{feng2025dit4edit,
  title={Dit4edit: Diffusion transformer for image editing},
  author={Feng, Kunyu and Ma, Yue and Wang, Bingyuan and Qi, Chenyang and Chen, Haozhe and Chen, Qifeng and Wang, Zeyu},
  booktitle={Proceedings of the AAAI Conference on Artificial Intelligence},
  volume={39},
  number={3},
  pages={2969--2977},
  year={2025}
}

@article{wang2024taming,
  title={Taming rectified flow for inversion and editing},
  author={Wang, Jiangshan and Pu, Junfu and Qi, Zhongang and Guo, Jiayi and Ma, Yue and Huang, Nisha and Chen, Yuxin and Li, Xiu and Shan, Ying},
  journal={arXiv preprint arXiv:2411.04746},
  year={2024}
}

@inproceedings{ma2024followpose,
  title={Follow your pose: Pose-guided text-to-video generation using pose-free videos},
  author={Ma, Yue and He, Yingqing and Cun, Xiaodong and Wang, Xintao and Chen, Siran and Li, Xiu and Chen, Qifeng},
  booktitle={Proceedings of the AAAI Conference on Artificial Intelligence},
  volume={38},
  number={5},
  pages={4117--4125},
  year={2024}
}

@article{ma2025followcreation,
  title={Follow-Your-Creation: Empowering 4D Creation through Video Inpainting},
  author={Ma, Yue and Feng, Kunyu and Zhang, Xinhua and Liu, Hongyu and Zhang, David Junhao and Xing, Jinbo and Zhang, Yinhan and Yang, Ayden and Wang, Zeyu and Chen, Qifeng},
  journal={arXiv preprint arXiv:2506.04590},
  year={2025}
}

@article{ma2026fastvmt,
  title={FastVMT: Eliminating Redundancy in Video Motion Transfer},
  author={Ma, Yue and Wang, Zhikai and Ren, Tianhao and Zheng, Mingzhe and Liu, Hongyu and Guo, Jiayi and Fong, Mark and Xue, Yuxuan and Zhao, Zixiang and Schindler, Konrad and others},
  journal={arXiv preprint arXiv:2602.05551},
  year={2026}
}

@article{ma2025followyourmotion,
  title={Follow-Your-Motion: Video Motion Transfer via Efficient Spatial-Temporal Decoupled Finetuning},
  author={Ma, Yue and Liu, Yulong and Zhu, Qiyuan and Yang, Ayden and Feng, Kunyu and Zhang, Xinhua and Li, Zhifeng and Han, Sirui and Qi, Chenyang and Chen, Qifeng},
  journal={arXiv preprint arXiv:2506.05207},
  year={2025}
}

@article{ma2025controllable,
  title={Controllable Video Generation: A Survey},
  author={Ma, Yue and Feng, Kunyu and Hu, Zhongyuan and Wang, Xinyu and Wang, Yucheng and Zheng, Mingzhe and He, Xuanhua and Zhu, Chenyang and Liu, Hongyu and He, Yingqing and others},
  journal={arXiv preprint arXiv:2507.16869},
  year={2025}
}

@article{qwen2-vl,
  title={Qwen2-vl: Enhancing vision-language model's perception of the world at any resolution},
  author={Wang, Peng and Bai, Shuai and Tan, Sinan and Wang, Shijie and Fan, Zhihao and Bai, Jinze and Chen, Keqin and Liu, Xuejing and Wang, Jialin and Ge, Wenbin and others},
  journal={arXiv preprint arXiv:2409.12191},
  year={2024}
}

@article{hpsv3,
  title={HPSv3: Towards Wide-Spectrum Human Preference Score},
  author={Ma, Yuhang and Wu, Xiaoshi and Sun, Keqiang and Li, Hongsheng},
  journal={arXiv preprint arXiv:2508.03789},
  year={2025}
}

@article{dancegrpo,
  title={DanceGRPO: Unleashing GRPO on Visual Generation},
  author={Xue, Zeyue and Wu, Jie and Gao, Yu and Kong, Fangyuan and Zhu, Lingting and Chen, Mengzhao and Liu, Zhiheng and Liu, Wei and Guo, Qiushan and Huang, Weilin and others},
  journal={arXiv preprint arXiv:2505.07818},
  year={2025}
}

@article{flow-dpo,
  title={Improving video generation with human feedback},
  author={Liu, Jie and Liu, Gongye and Liang, Jiajun and Yuan, Ziyang and Liu, Xiaokun and Zheng, Mingwu and Wu, Xiele and Wang, Qiulin and Qin, Wenyu and Xia, Menghan and others},
  journal={arXiv preprint arXiv:2501.13918},
  year={2025}
}

@article{SimpleAR,
  title={Simplear: Pushing the frontier of autoregressive visual generation through pretraining, sft, and rl},
  author={Wang, Junke and Tian, Zhi and Wang, Xun and Zhang, Xinyu and Huang, Weilin and Wu, Zuxuan and Jiang, Yu-Gang},
  journal={arXiv preprint arXiv:2504.11455},
  year={2025}
}

@inproceedings{CLIP,
  title={Learning transferable visual models from natural language supervision},
  author={Radford, Alec and Kim, Jong Wook and Hallacy, Chris and Ramesh, Aditya and Goh, Gabriel and Agarwal, Sandhini and Sastry, Girish and Askell, Amanda and Mishkin, Pamela and Clark, Jack and others},
  booktitle={International conference on machine learning},
  year={2021},
}

@inproceedings{BLIP,
  title={Blip: Bootstrapping language-image pre-training for unified vision-language understanding and generation},
  author={Li, Junnan and Li, Dongxu and Xiong, Caiming and Hoi, Steven},
  booktitle={International conference on machine learning},
  year={2022},
}

@inproceedings{Inference-Time-Scaling,
  title={Scaling Inference Time Compute for Diffusion Models},
  author={Ma, Nanye and Tong, Shangyuan and Jia, Haolin and Hu, Hexiang and Su, Yu-Chuan and Zhang, Mingda and Yang, Xuan and Li, Yandong and Jaakkola, Tommi and Jia, Xuhui and others},
  booktitle={IEEE/CVF Conference on Computer Vision and Pattern Recognition},
  year={2025}
}

@inproceedings{CNI,
  title={Adding conditional control to text-to-image diffusion models},
  author={Zhang, Lvmin and Rao, Anyi and Agrawala, Maneesh},
  booktitle={IEEE/CVF Conference on Computer Vision and Pattern Recognition},
  year={2023}
}

@inproceedings{ASUKA,
  title={Towards Enhanced Image Inpainting: Mitigating Unwanted Object Insertion and Preserving Color Consistency},
  author={Wang, Yikai and Cao, Chenjie and Yu, Junqiu and Fan, Ke and Xue, Xiangyang and Fu, Yanwei},
  booktitle={IEEE/CVF Conference on Computer Vision and Pattern Recognition},
  year={2025}
}

@article{hack,
  title={The effects of reward misspecification: Mapping and mitigating misaligned models},
  author={Pan, Alexander and Bhatia, Kush and Steinhardt, Jacob},
  journal={arXiv preprint arXiv:2201.03544},
  year={2022}
}

@article{BLD,
  title={Blended latent diffusion},
  author={Avrahami, Omri and Fried, Ohad and Lischinski, Dani},
  journal={ACM transactions on graphics},
  year={2023},
}

@inproceedings{StructureMatters,
  title={Structure matters: Tackling the semantic discrepancy in diffusion models for image inpainting},
  author={Liu, Haipeng and Wang, Yang and Qian, Biao and Wang, Meng and Rui, Yong},
  booktitle={IEEE/CVF Conference on Computer Vision and Pattern Recognition},
  year={2024}
}

@inproceedings{HD-Painter,
  title={Hd-painter: high-resolution and prompt-faithful text-guided image inpainting with diffusion models},
  author={Manukyan, Hayk and Sargsyan, Andranik and Atanyan, Barsegh and Wang, Zhangyang and Navasardyan, Shant and Shi, Humphrey},
  booktitle={International Conference on Learning Representations},
  year={2023}
}

@inproceedings{sd,
  title={High-resolution image synthesis with latent diffusion models},
  author={Rombach, Robin and Blattmann, Andreas and Lorenz, Dominik and Esser, Patrick and Ommer, Bj{\"o}rn},
  booktitle={IEEE/CVF Conference on Computer Vision and Pattern Recognition},
  year={2022}
}

@article{dpok,
  title={Dpok: Reinforcement learning for fine-tuning text-to-image diffusion models},
  author={Fan, Ying and Watkins, Olivia and Du, Yuqing and Liu, Hao and Ryu, Moonkyung and Boutilier, Craig and Abbeel, Pieter and Ghavamzadeh, Mohammad and Lee, Kangwook and Lee, Kimin},
  journal={Advances in Neural Information Processing Systems},
  year={2023}
}

@misc{flux1filldev,
  author       = "BlackForestLabs",
  title        = {FLUX.1-Fill-dev},
  year         = {2024},
  howpublished = {\url{https://huggingface.co/black-forest-labs/FLUX.1-Fill-dev}},
  note         = {Accessed: 2025-07-12}
}

@inproceedings{restoration,
  title={Swinir: Image restoration using swin transformer},
  author={Liang, Jingyun and Cao, Jiezhang and Sun, Guolei and Zhang, Kai and Van Gool, Luc and Timofte, Radu},
  booktitle={IEEE/CVF Conference on Computer Vision and Pattern Recognition},
  year={2021}
}

@inproceedings{label-eff,
  title={Label-efficient domain generalization via collaborative exploration and generalization},
  author={Yuan, Junkun and Ma, Xu and Chen, Defang and Kuang, Kun and Wu, Fei and Lin, Lanfen},
  booktitle={Proceedings of the 30th ACM international conference on multimedia},
  pages={2361--2370},
  year={2022}
}

@inproceedings{powerpaint,
  title={A task is worth one word: Learning with task prompts for high-quality versatile image inpainting},
  author={Zhuang, Junhao and Zeng, Yanhong and Liu, Wenran and Yuan, Chun and Chen, Kai},
  booktitle={European Conference on Computer Vision},
  year={2024},
}

@article{magicbrush,
  title={Magicbrush: A manually annotated dataset for instruction-guided image editing},
  author={Zhang, Kai and Mo, Lingbo and Chen, Wenhu and Sun, Huan and Su, Yu},
  journal={Advances in Neural Information Processing Systems},
  year={2023}
}

@article{domain-specific,
  title={Domain-specific bias filtering for single labeled domain generalization},
  author={Yuan, Junkun and Ma, Xu and Chen, Defang and Kuang, Kun and Wu, Fei and Lin, Lanfen},
  journal={International Journal of Computer Vision},
  volume={131},
  number={2},
  pages={552--571},
  year={2023},
  publisher={Springer}
}

@inproceedings{idream,
  title={I Dream My Painting: Connecting MLLMs and Diffusion Models via Prompt Generation for Text-Guided Multi-Mask Inpainting},
  author={Fanelli, Nicola and Vessio, Gennaro and Castellano, Giovanna},
  booktitle={IEEE/CVF Winter Conference on Applications of Computer Vision},
  year={2025},
}

@inproceedings{RORem,
  title={RORem: Training a Robust Object Remover with Human-in-the-Loop},
  author={Li, Ruibin and Yang, Tao and Guo, Song and Zhang, Lei},
  booktitle={IEEE/CVF Conference on Computer Vision and Pattern Recognition},
  year={2025}
}

@article{collaborative,
  title={Collaborative semantic aggregation and calibration for federated domain generalization},
  author={Yuan, Junkun and Ma, Xu and Chen, Defang and Wu, Fei and Lin, Lanfen and Kuang, Kun},
  journal={IEEE Transactions on Knowledge and Data Engineering},
  volume={35},
  number={12},
  pages={12528--12541},
  year={2023},
  publisher={IEEE}
}

@article{follow-your-preference,
  title={Follow-Your-Preference: Towards Preference-Aligned Image Inpainting},
  author={Shen, Yutao and Yuan, Junkun and Aonishi, Toru and Nakayama, Hideki and Ma, Yue},
  journal={arXiv preprint arXiv:2509.23082},
  year={2025}
}

@article{prompt2prompt,
  title={Prompt-to-prompt image editing with cross attention control},
  author={Hertz, Amir and Mokady, Ron and Tenenbaum, Jay and Aberman, Kfir and Pritch, Yael and Cohen-Or, Daniel},
  journal={arXiv preprint arXiv:2208.01626},
  year={2022}
}

@inproceedings{repaint,
  title={Repaint: Inpainting using denoising diffusion probabilistic models},
  author={Lugmayr, Andreas and Danelljan, Martin and Romero, Andres and Yu, Fisher and Timofte, Radu and Van Gool, Luc},
  booktitle={IEEE/CVF Conference on Computer Vision and Pattern Recognition},
  year={2022}
}

@inproceedings{instructpix2pix,
  title={Instructpix2pix: Learning to follow image editing instructions},
  author={Brooks, Tim and Holynski, Aleksander and Efros, Alexei A},
  booktitle={IEEE/CVF Conference on Computer Vision and Pattern Recognition},
  year={2023}
}

@inproceedings{rad,
  title={Rad: Region-aware diffusion models for image inpainting},
  author={Kim, Sora and Suh, Sungho and Lee, Minsik},
  booktitle={IEEE/CVF Conference on Computer Vision and Pattern Recognition},
  year={2025}
}

@inproceedings{lpips,
  title={The unreasonable effectiveness of deep features as a perceptual metric},
  author={Zhang, Richard and Isola, Phillip and Efros, Alexei A and Shechtman, Eli and Wang, Oliver},
  booktitle={IEEE/CVF Conference on Computer Vision and Pattern Recognition},
  year={2018}
}

@inproceedings{ipo,
  title={A general theoretical paradigm to understand learning from human preferences},
  author={Azar, Mohammad Gheshlaghi and Guo, Zhaohan Daniel and Piot, Bilal and Munos, Remi and Rowland, Mark and Valko, Michal and Calandriello, Daniele},
  booktitle={International Conference on Artificial Intelligence and Statistics},
  year={2024},
}
}

\clearpage
\onecolumn
\maketitlesupplementary
\appendix
\section{More Results on Reward Model Bias Studies}
\begin{figure}[!htbp]
\begin{subfigure}[t]{.49\linewidth}
\centering
	\includegraphics[width=0.22\linewidth, trim=0mm 0mm 0mm 0mm, clip]{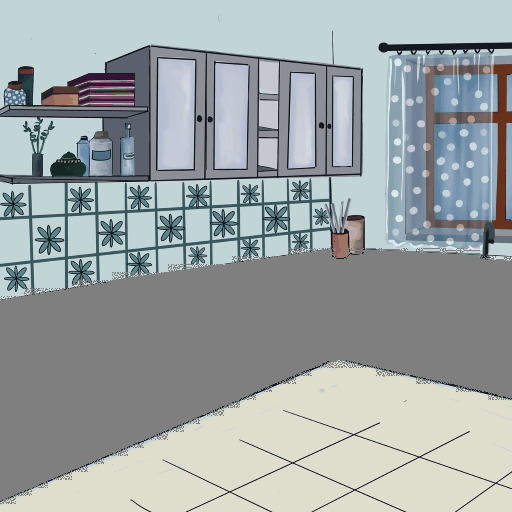}
	\includegraphics[width=0.22\linewidth, trim=0mm 0mm 0mm 0mm, clip]{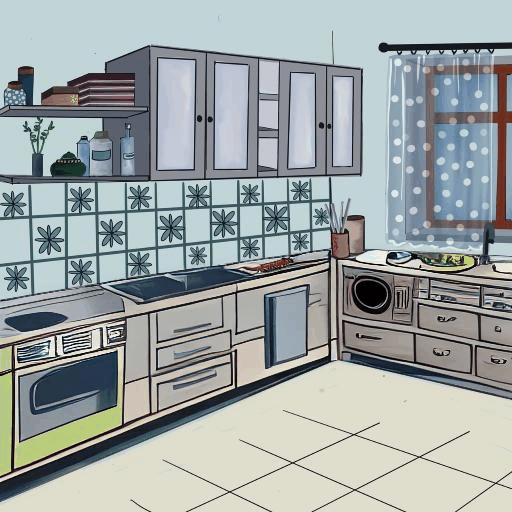}
	\includegraphics[width=0.22\linewidth, trim=0mm 0mm 0mm 0mm, clip]{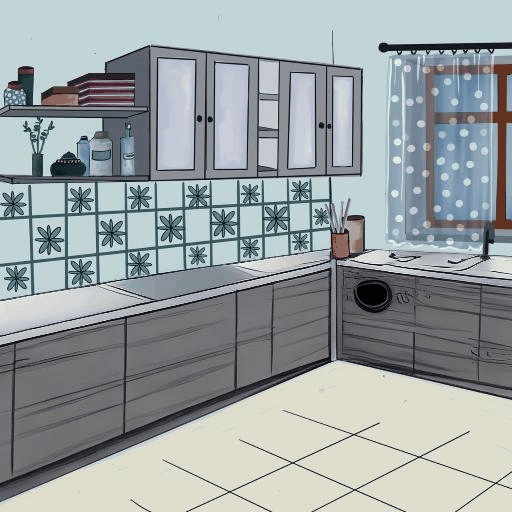}
	\includegraphics[width=0.22\linewidth, trim=0mm 0mm 0mm 0mm, clip]{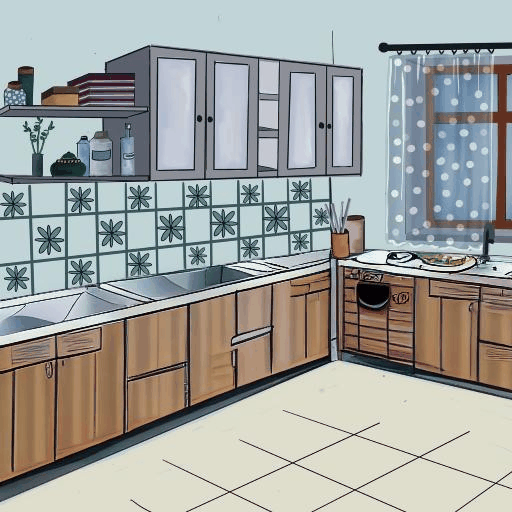}
\caption{A cartoon drawing of a kitchen.}
\end{subfigure}
\begin{subfigure}[t]{.49\linewidth}
\centering
	\includegraphics[width=0.22\linewidth, trim=0mm 0mm 0mm 0mm, clip]{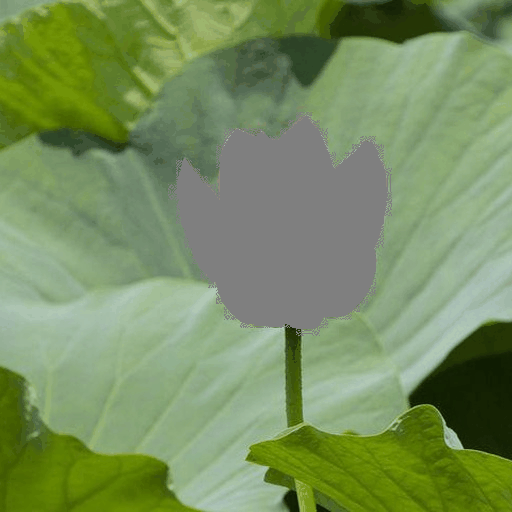}
	\includegraphics[width=0.22\linewidth, trim=0mm 0mm 0mm 0mm, clip]{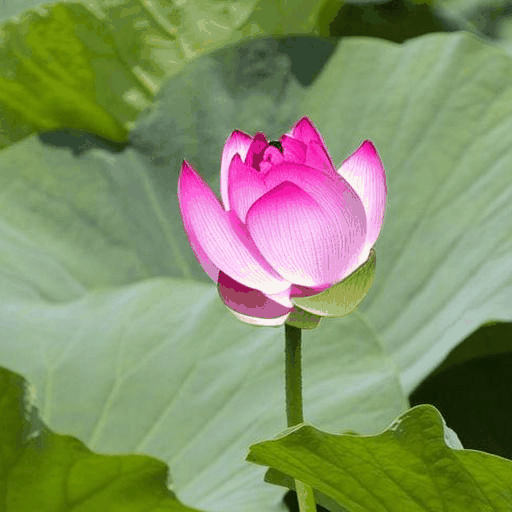}
	\includegraphics[width=0.22\linewidth, trim=0mm 0mm 0mm 0mm, clip]{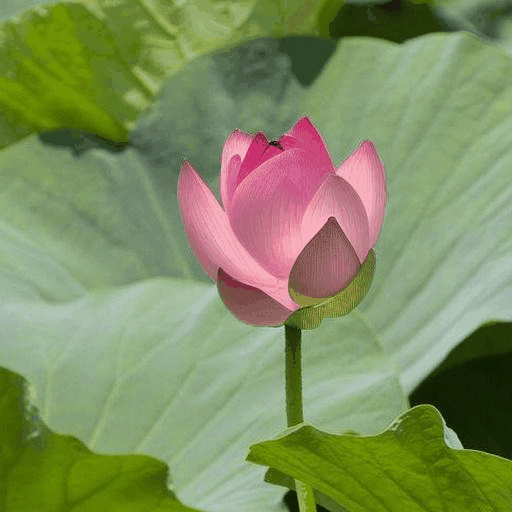}
	\includegraphics[width=0.22\linewidth, trim=0mm 0mm 0mm 0mm, clip]{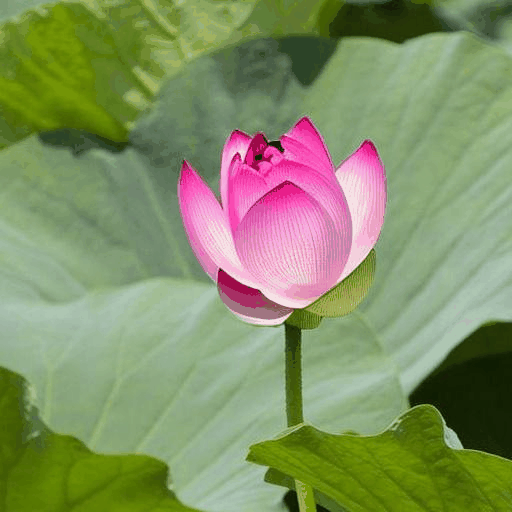}
\caption{A pink lotus flower blooming with green leaves.}
\end{subfigure} 
\begin{subfigure}[t]{.49\linewidth}
\centering
	\includegraphics[width=0.22\linewidth, trim=0mm 0mm 0mm 0mm, clip]{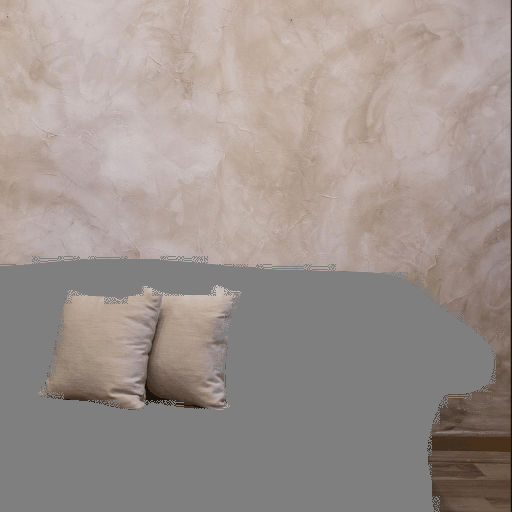}
	\includegraphics[width=0.22\linewidth, trim=0mm 0mm 0mm 0mm, clip]{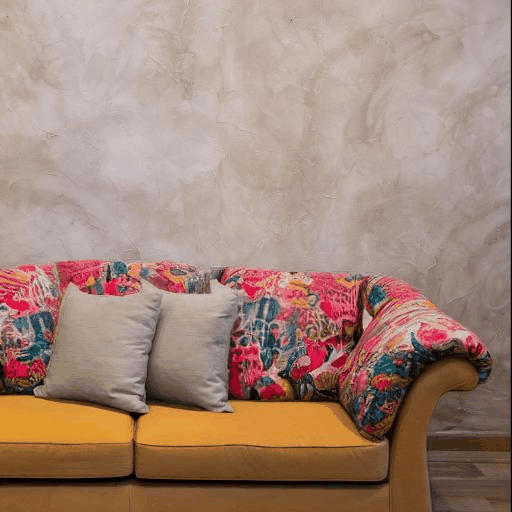}
	\includegraphics[width=0.22\linewidth, trim=0mm 0mm 0mm 0mm, clip]{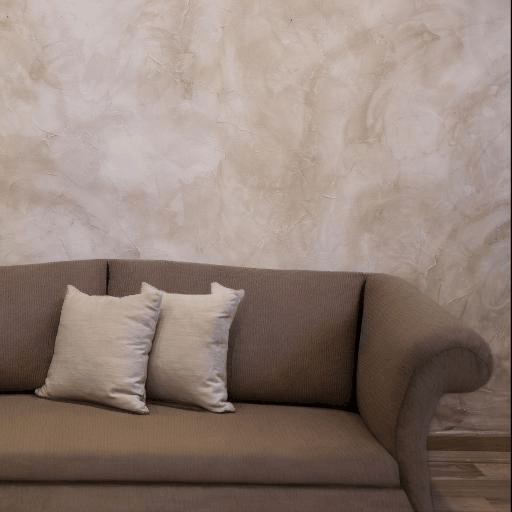}
	\includegraphics[width=0.22\linewidth, trim=0mm 0mm 0mm 0mm, clip]{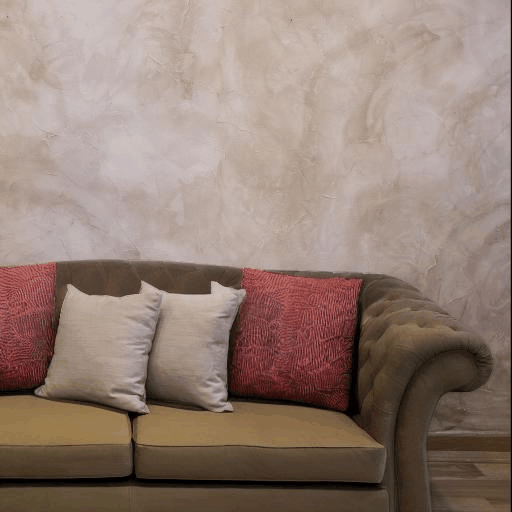}
\caption{A couch with pillows and a wall behind it.}
\end{subfigure} 
\begin{subfigure}[t]{.49\linewidth}
\centering
	\includegraphics[width=0.22\linewidth, trim=0mm 0mm 0mm 0mm, clip]{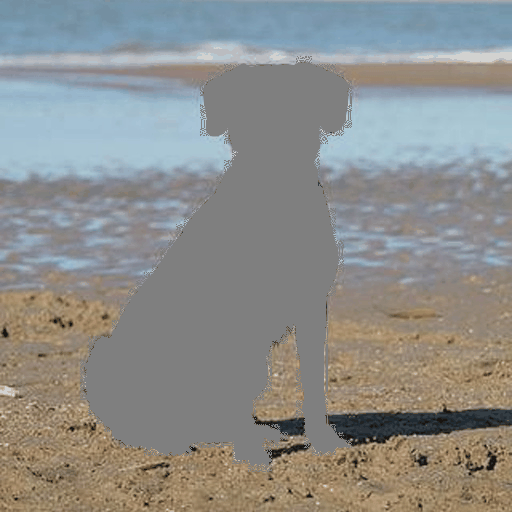}
	\includegraphics[width=0.22\linewidth, trim=0mm 0mm 0mm 0mm, clip]{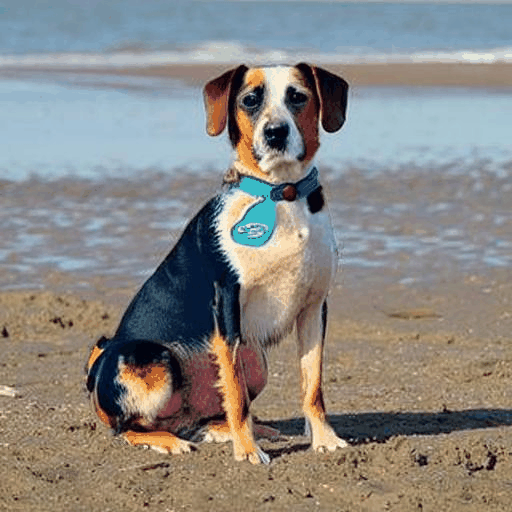}
	\includegraphics[width=0.22\linewidth, trim=0mm 0mm 0mm 0mm, clip]{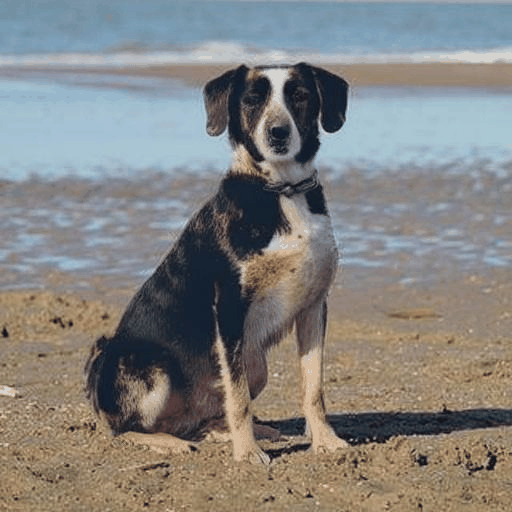}
	\includegraphics[width=0.22\linewidth, trim=0mm 0mm 0mm 0mm, clip]{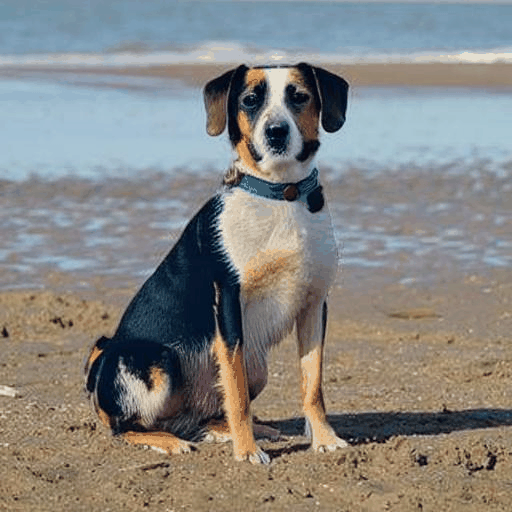}
\caption{A dog sitting on the beach.}
\end{subfigure} 
\begin{subfigure}[t]{.49\linewidth}
\centering
	\includegraphics[width=0.22\linewidth, trim=0mm 0mm 0mm 0mm, clip]{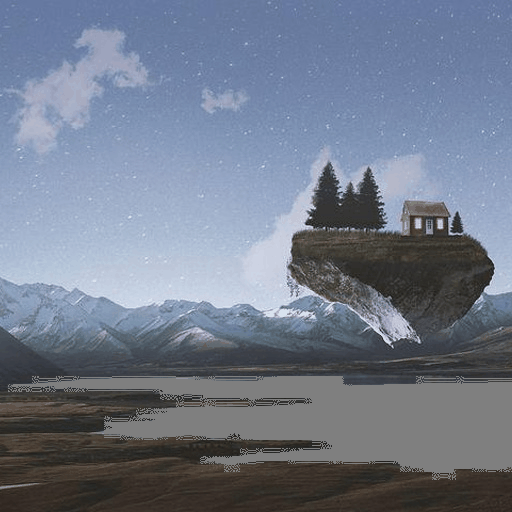}
	\includegraphics[width=0.22\linewidth, trim=0mm 0mm 0mm 0mm, clip]{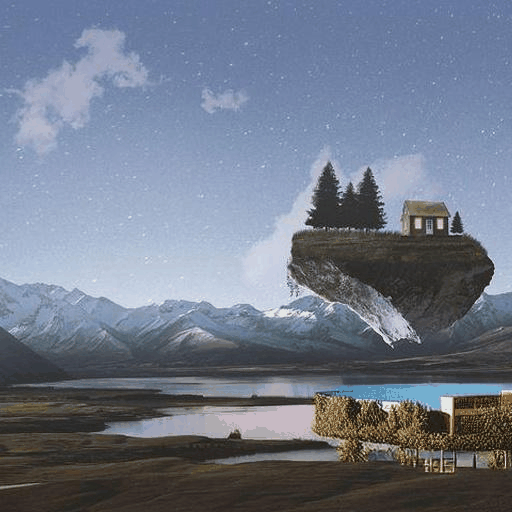}
	\includegraphics[width=0.22\linewidth, trim=0mm 0mm 0mm 0mm, clip]{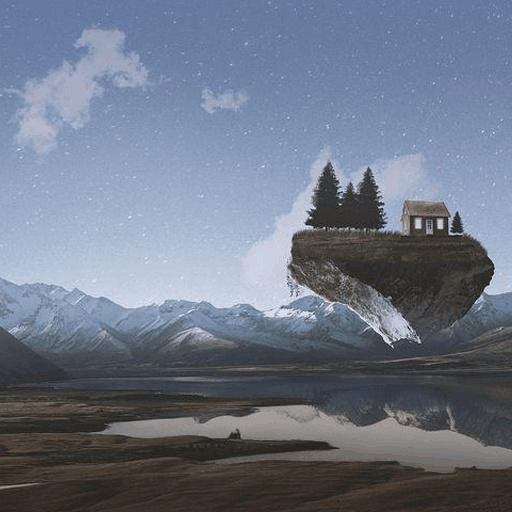}
	\includegraphics[width=0.22\linewidth, trim=0mm 0mm 0mm 0mm, clip]{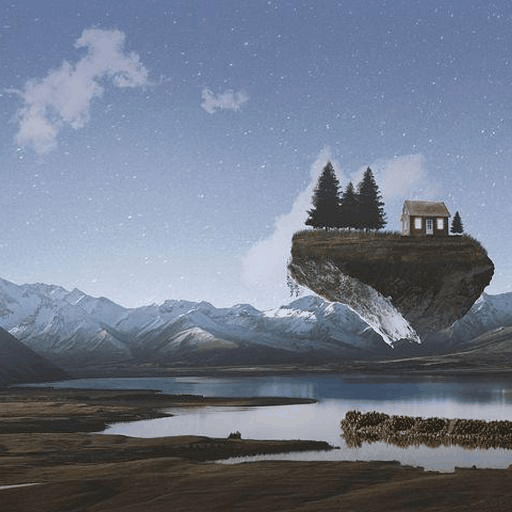}
\caption{A house floating in the air over a lake.}
\end{subfigure}
\begin{subfigure}[t]{.49\linewidth}
\centering
	\includegraphics[width=0.22\linewidth, trim=0mm 0mm 0mm 0mm, clip]{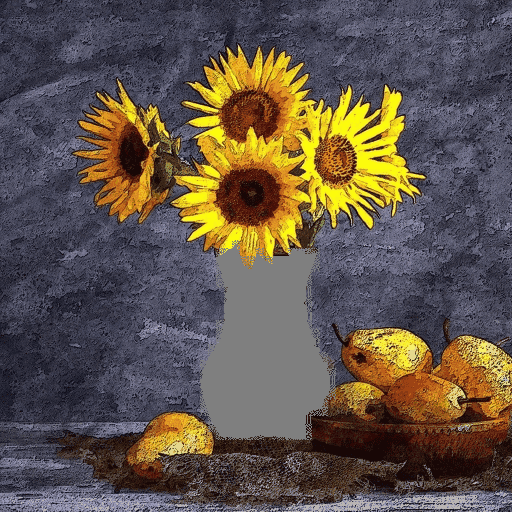}
	\includegraphics[width=0.22\linewidth, trim=0mm 0mm 0mm 0mm, clip]{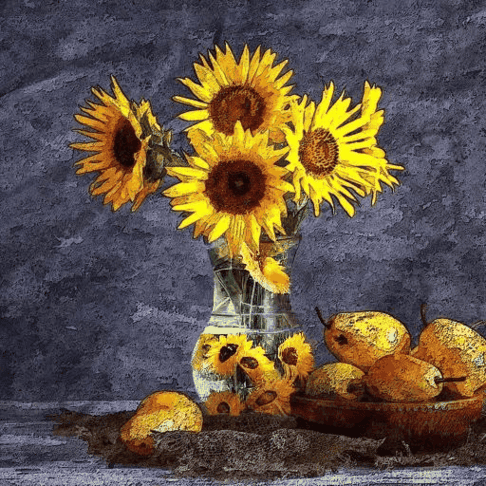}
	\includegraphics[width=0.22\linewidth, trim=0mm 0mm 0mm 0mm, clip]{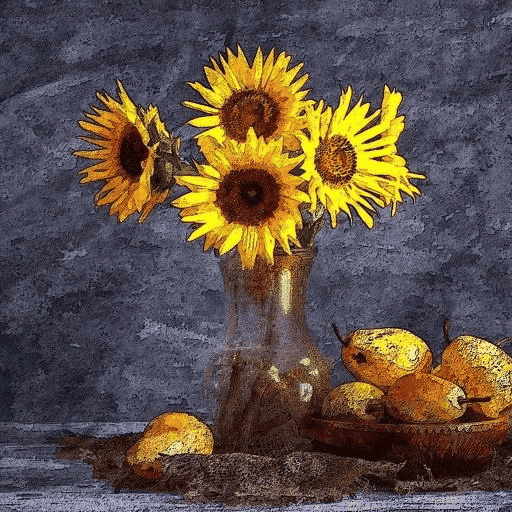}
	\includegraphics[width=0.22\linewidth, trim=0mm 0mm 0mm 0mm, clip]{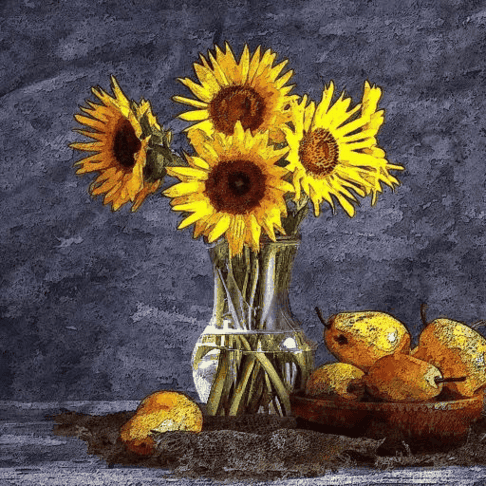}
\caption{Sunflowers in a vase with pears on a table.}
\end{subfigure}
\begin{subfigure}[t]{.49\linewidth}
\centering
	\includegraphics[width=0.22\linewidth, trim=0mm 0mm 0mm 0mm, clip]{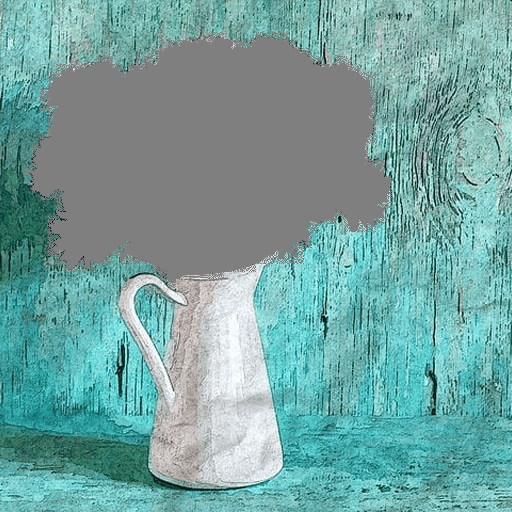}
	\includegraphics[width=0.22\linewidth, trim=0mm 0mm 0mm 0mm, clip]{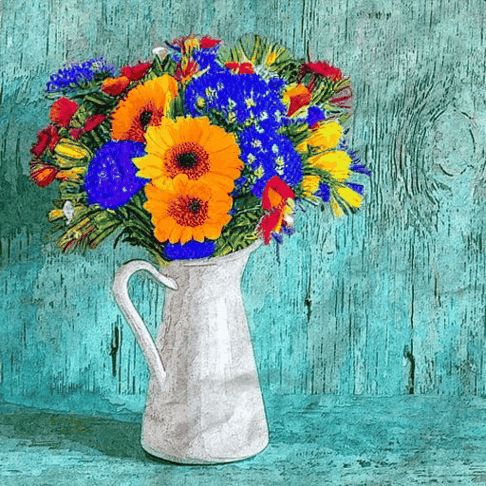}
	\includegraphics[width=0.22\linewidth, trim=0mm 0mm 0mm 0mm, clip]{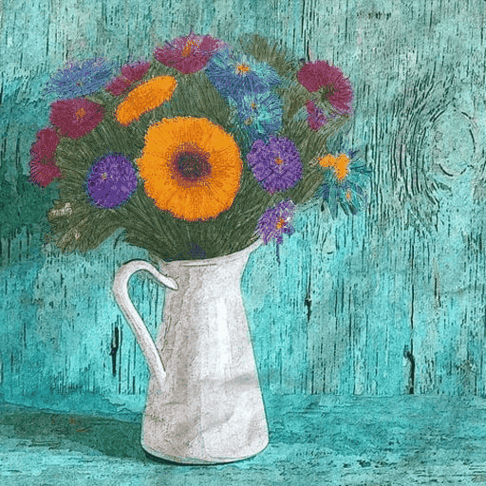}
	\includegraphics[width=0.22\linewidth, trim=0mm 0mm 0mm 0mm, clip]{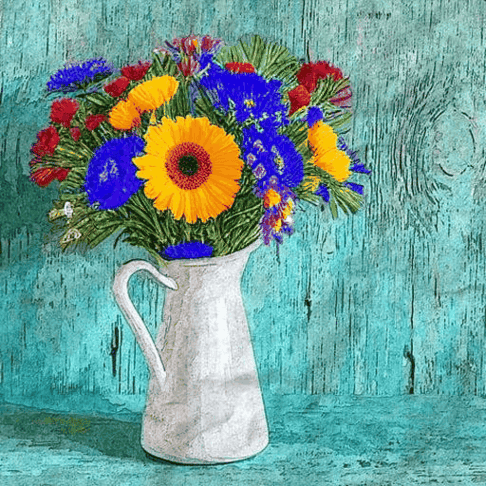}
\caption{A vase filled with colorful flowers on a table.}
\end{subfigure}
\begin{subfigure}[t]{.49\linewidth}
\centering
	\includegraphics[width=0.22\linewidth, trim=0mm 0mm 0mm 0mm, clip]{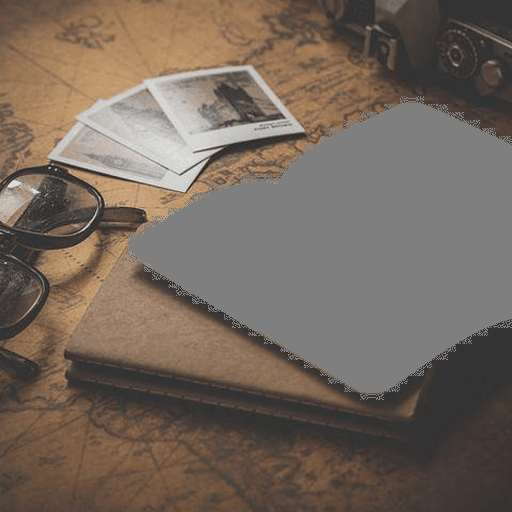}
	\includegraphics[width=0.22\linewidth, trim=0mm 0mm 0mm 0mm, clip]{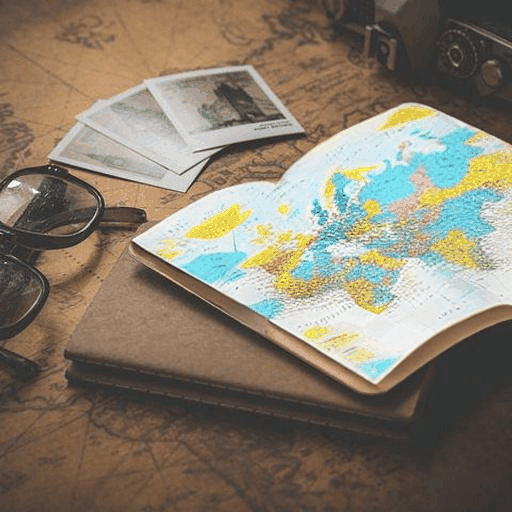}
	\includegraphics[width=0.22\linewidth, trim=0mm 0mm 0mm 0mm, clip]{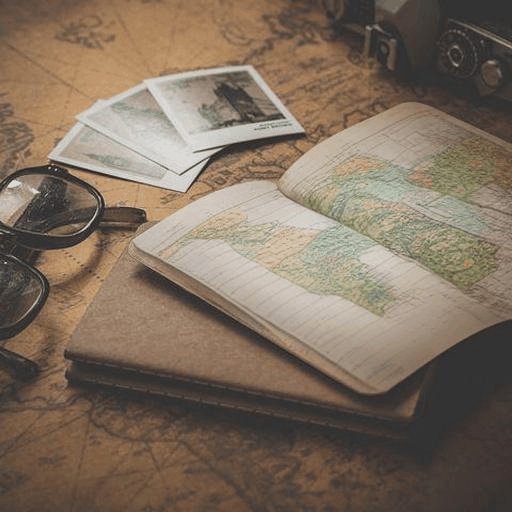}
	\includegraphics[width=0.22\linewidth, trim=0mm 0mm 0mm 0mm, clip]{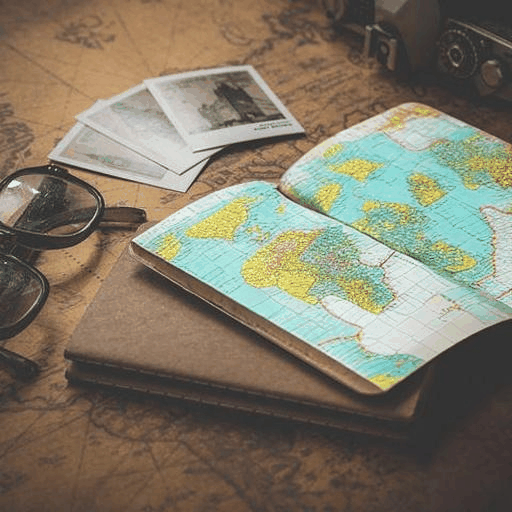}
\caption{A notebook, glasses and a camera on a map.}
\end{subfigure} 
\begin{subfigure}[t]{.49\linewidth}
\centering
	\includegraphics[width=0.22\linewidth, trim=0mm 0mm 0mm 0mm, clip]{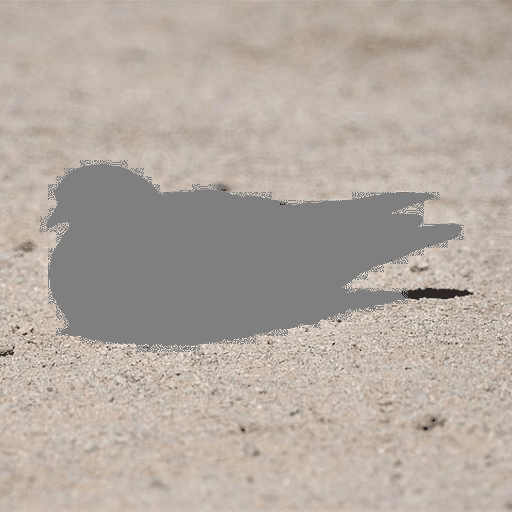}
	\includegraphics[width=0.22\linewidth, trim=0mm 0mm 0mm 0mm, clip]{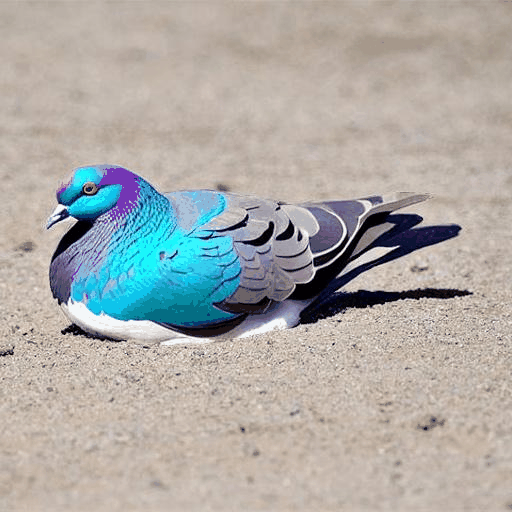}
	\includegraphics[width=0.22\linewidth, trim=0mm 0mm 0mm 0mm, clip]{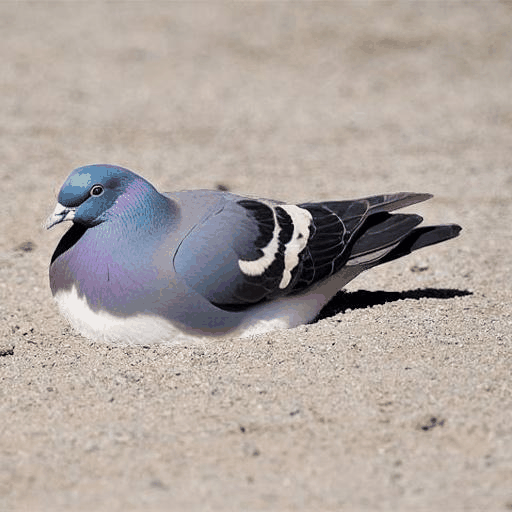}
	\includegraphics[width=0.22\linewidth, trim=0mm 0mm 0mm 0mm, clip]{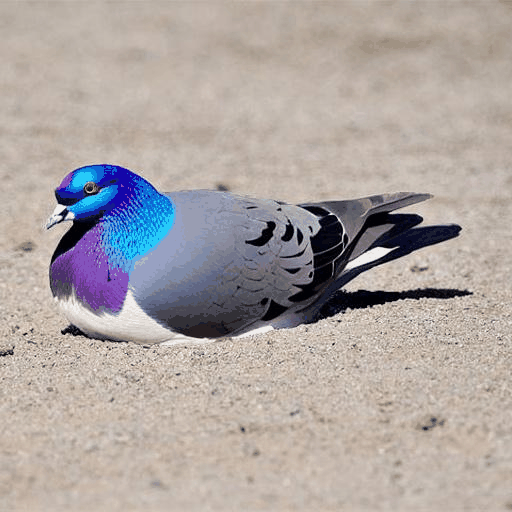}
\caption{A pigeon is sitting on the ground.}
\end{subfigure} 
\begin{subfigure}[t]{.49\linewidth}
\centering
	\includegraphics[width=0.22\linewidth, trim=0mm 0mm 0mm 0mm, clip]{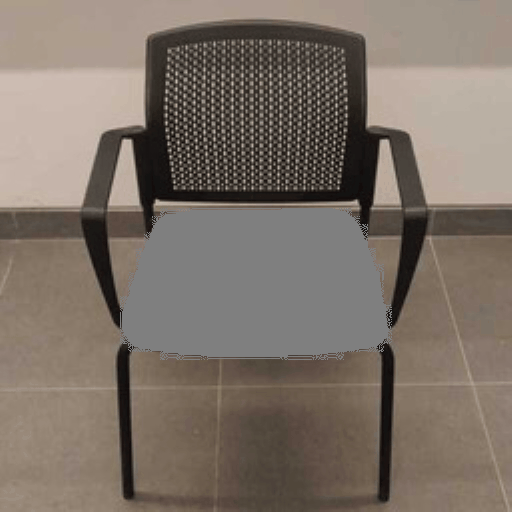}
	\includegraphics[width=0.22\linewidth, trim=0mm 0mm 0mm 0mm, clip]{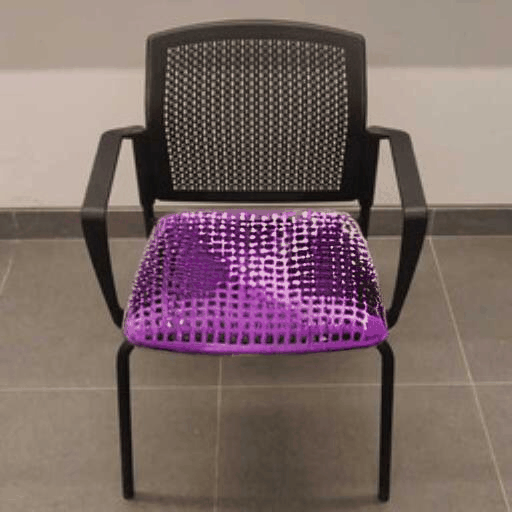}
	\includegraphics[width=0.22\linewidth, trim=0mm 0mm 0mm 0mm, clip]{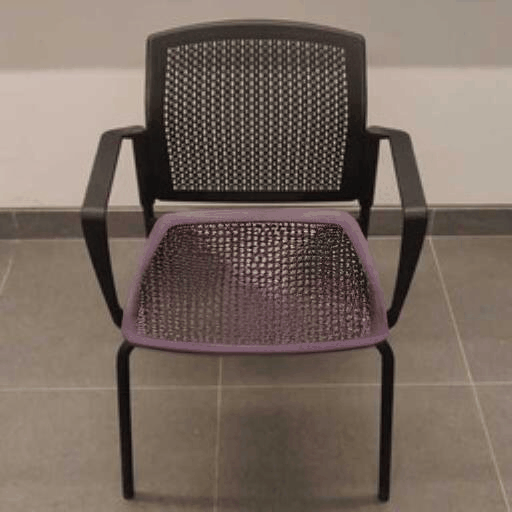}
	\includegraphics[width=0.22\linewidth, trim=0mm 0mm 0mm 0mm, clip]{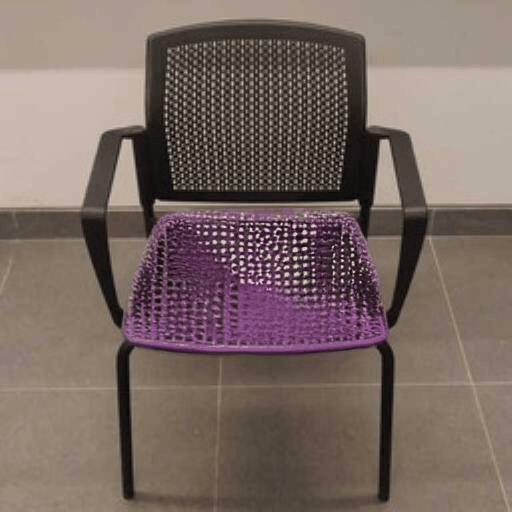}
\caption{A purple chair with a black seat and back.}
\end{subfigure} 
\begin{subfigure}[t]{.49\linewidth}
\centering
	\includegraphics[width=0.22\linewidth, trim=0mm 0mm 0mm 0mm, clip]{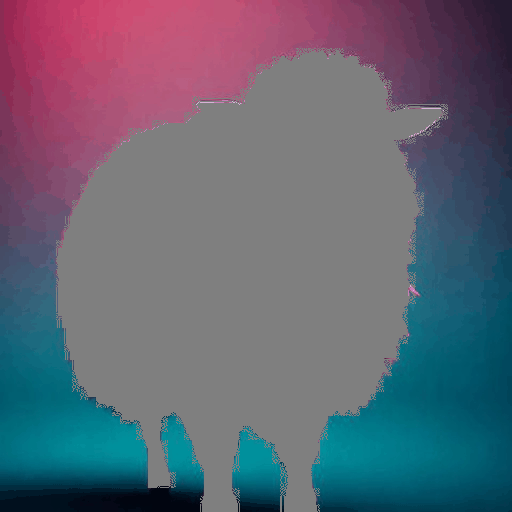}
	\includegraphics[width=0.22\linewidth, trim=0mm 0mm 0mm 0mm, clip]{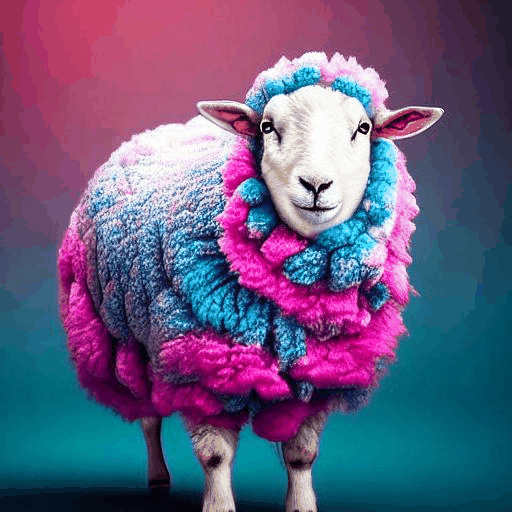}
	\includegraphics[width=0.22\linewidth, trim=0mm 0mm 0mm 0mm, clip]{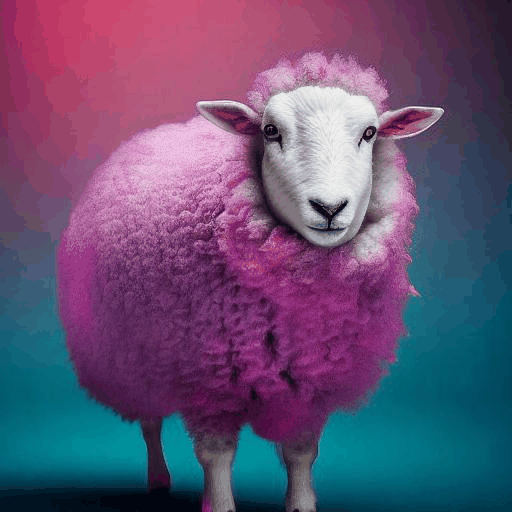}
	\includegraphics[width=0.22\linewidth, trim=0mm 0mm 0mm 0mm, clip]{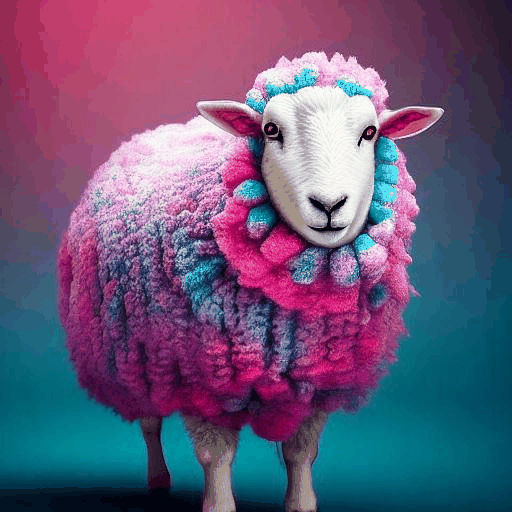}
\caption{A sheep with pink fur is standing.}
\end{subfigure} 
\begin{subfigure}[t]{.49\linewidth}
\centering
	\includegraphics[width=0.22\linewidth, trim=0mm 0mm 0mm 0mm, clip]{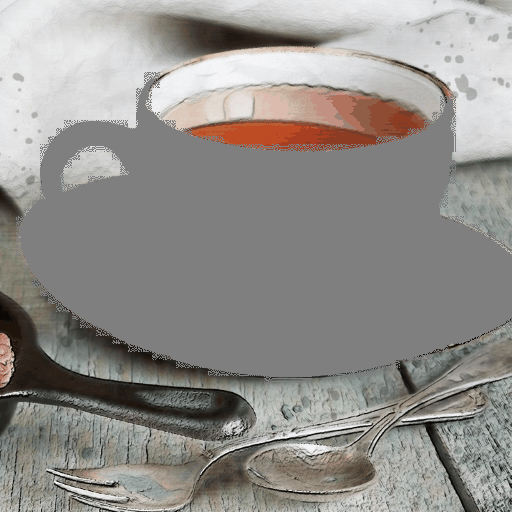}
	\includegraphics[width=0.22\linewidth, trim=0mm 0mm 0mm 0mm, clip]{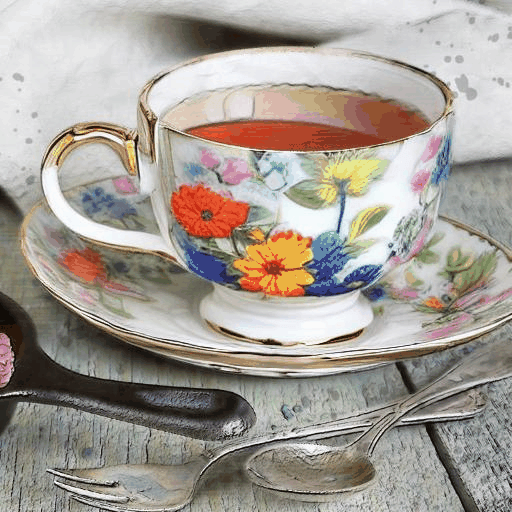}
	\includegraphics[width=0.22\linewidth, trim=0mm 0mm 0mm 0mm, clip]{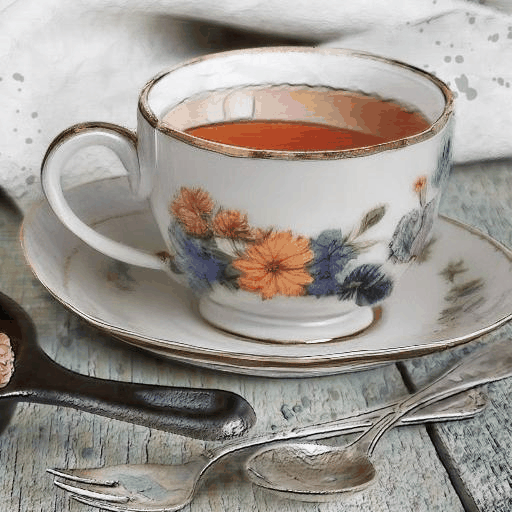}
	\includegraphics[width=0.22\linewidth, trim=0mm 0mm 0mm 0mm, clip]{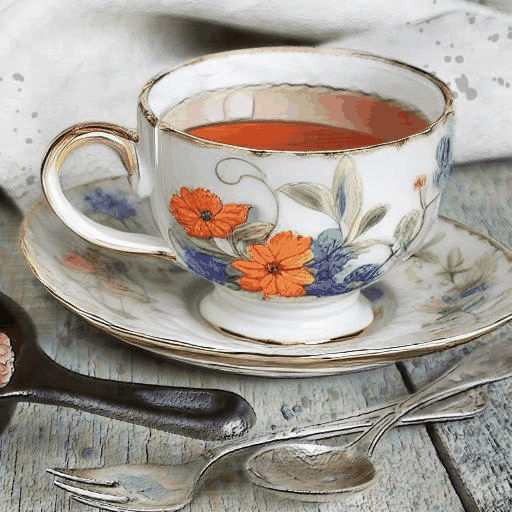}
\caption{A teacup and saucer with spoons.}
\end{subfigure} 
\begin{subfigure}[t]{.49\linewidth}
\centering
	\includegraphics[width=0.22\linewidth, trim=0mm 0mm 0mm 0mm, clip]{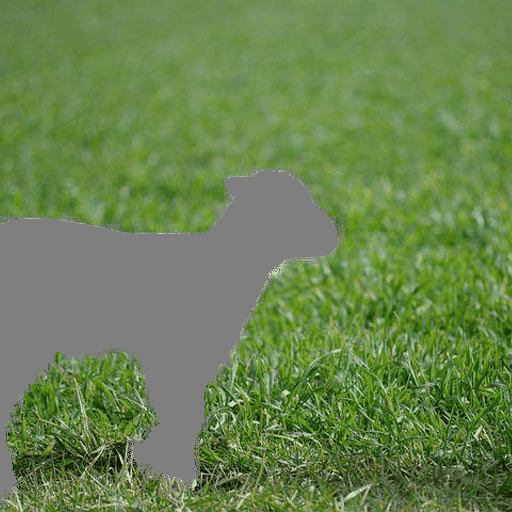}
	\includegraphics[width=0.22\linewidth, trim=0mm 0mm 0mm 0mm, clip]{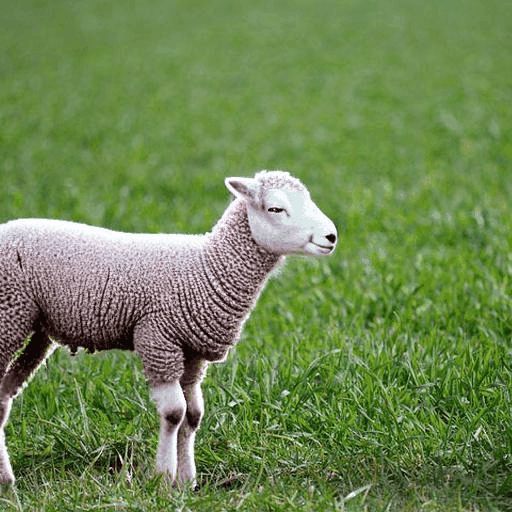}
	\includegraphics[width=0.22\linewidth, trim=0mm 0mm 0mm 0mm, clip]{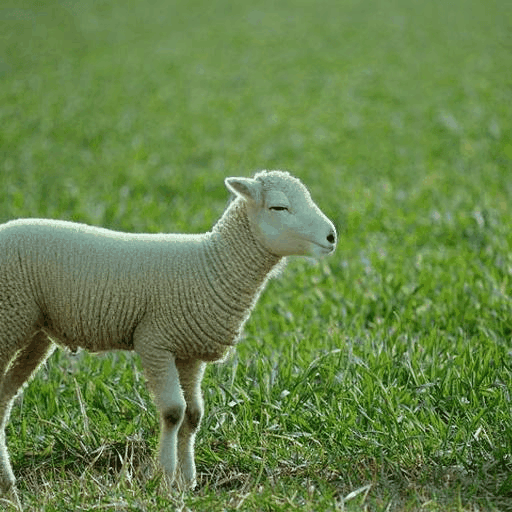}
	\includegraphics[width=0.22\linewidth, trim=0mm 0mm 0mm 0mm, clip]{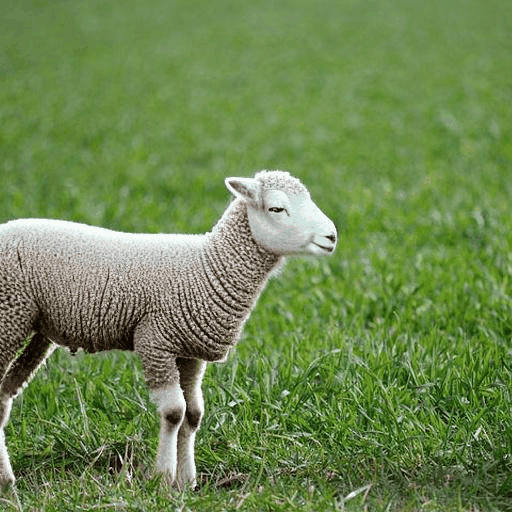}
\caption{A lamb standing in a field of green grass.}
\end{subfigure}
\begin{subfigure}[t]{.49\linewidth}
\centering
	\includegraphics[width=0.22\linewidth, trim=0mm 0mm 0mm 0mm, clip]{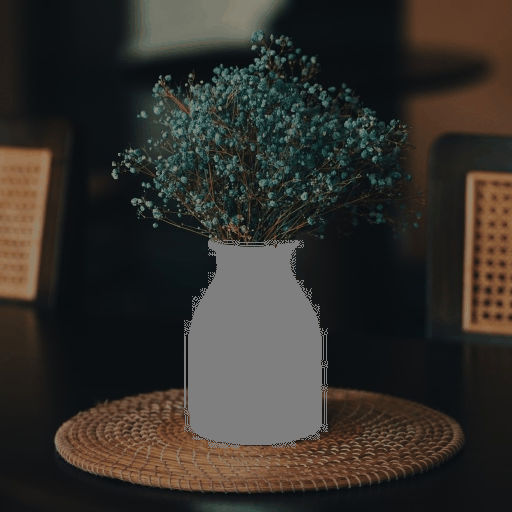}
	\includegraphics[width=0.22\linewidth, trim=0mm 0mm 0mm 0mm, clip]{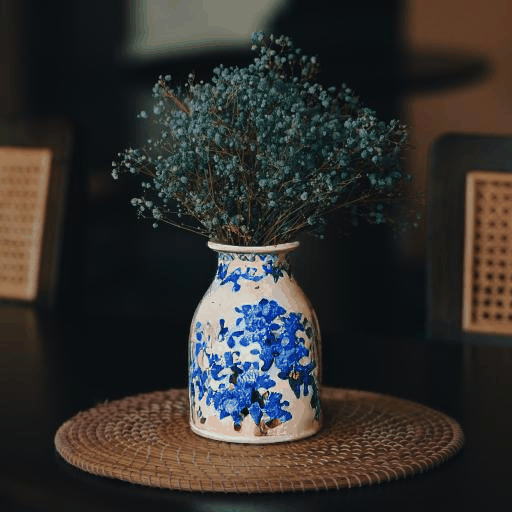}
	\includegraphics[width=0.22\linewidth, trim=0mm 0mm 0mm 0mm, clip]{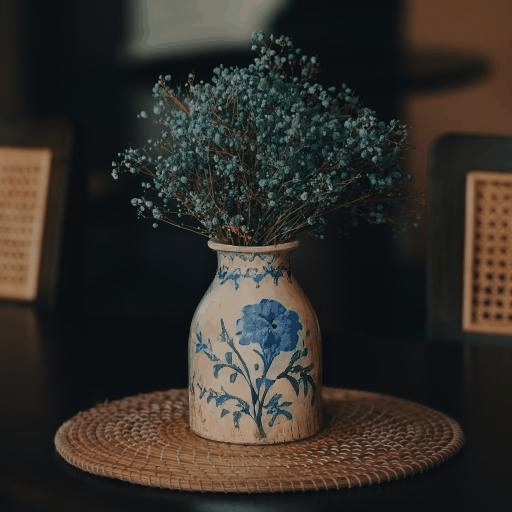}
	\includegraphics[width=0.22\linewidth, trim=0mm 0mm 0mm 0mm, clip]{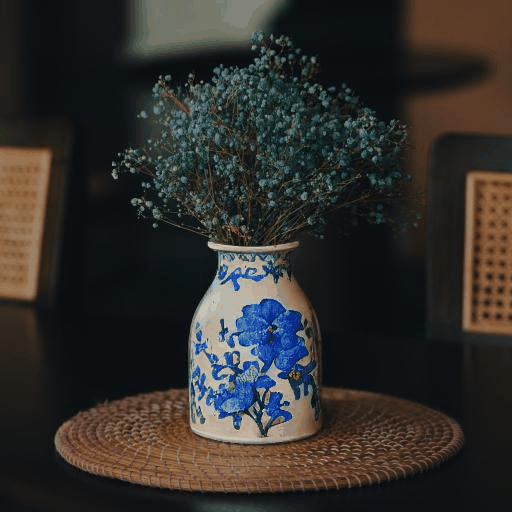}
\caption{A vase with blue flowers sitting on a table.}
\end{subfigure} 
\begin{subfigure}[t]{.49\linewidth}
\centering
	\includegraphics[width=0.22\linewidth, trim=0mm 0mm 0mm 0mm, clip]{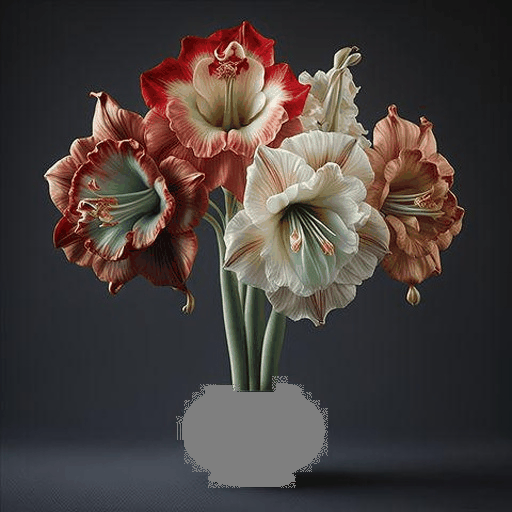}
	\includegraphics[width=0.22\linewidth, trim=0mm 0mm 0mm 0mm, clip]{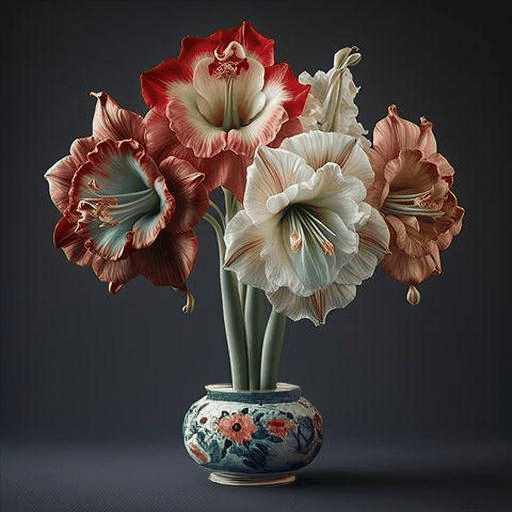}
	\includegraphics[width=0.22\linewidth, trim=0mm 0mm 0mm 0mm, clip]{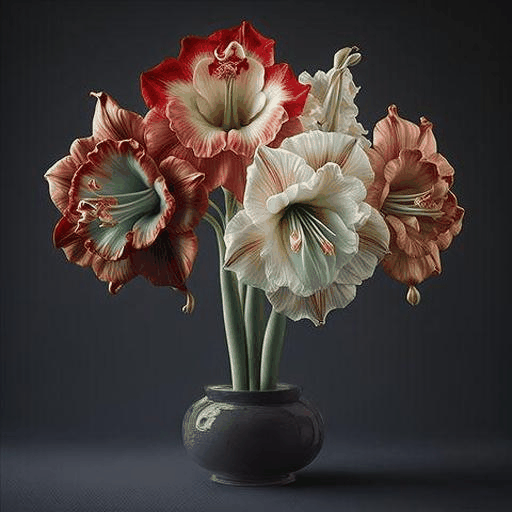}
	\includegraphics[width=0.22\linewidth, trim=0mm 0mm 0mm 0mm, clip]{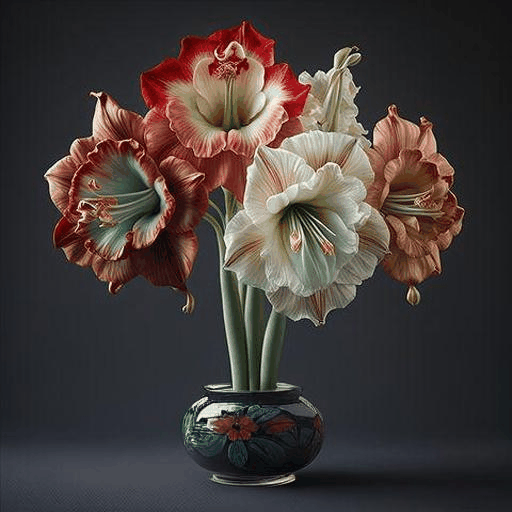}
\caption{A vase with some flowers in it.}
\end{subfigure} 
\hfill
\begin{subfigure}[t]{.49\linewidth}
\centering
	\includegraphics[width=0.22\linewidth, trim=0mm 0mm 0mm 0mm, clip]{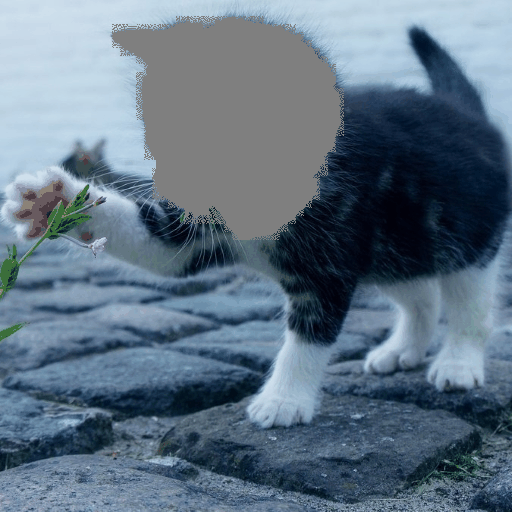}
	\includegraphics[width=0.22\linewidth, trim=0mm 0mm 0mm 0mm, clip]{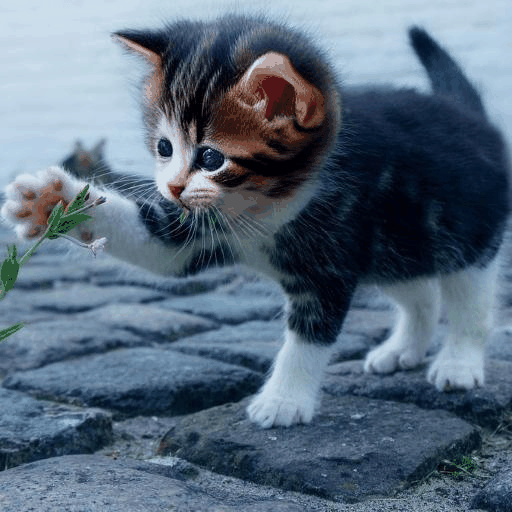}
	\includegraphics[width=0.22\linewidth, trim=0mm 0mm 0mm 0mm, clip]{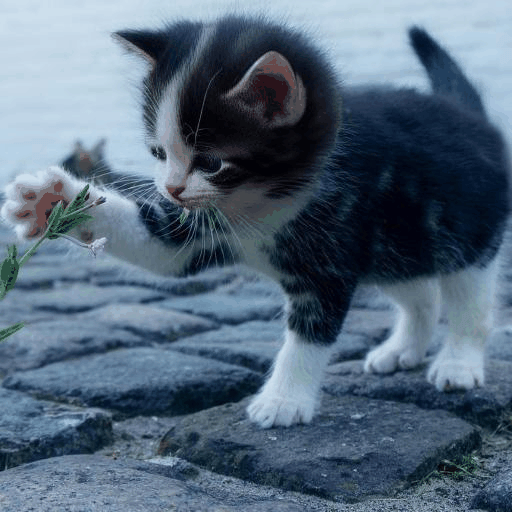}
	\includegraphics[width=0.22\linewidth, trim=0mm 0mm 0mm 0mm, clip]{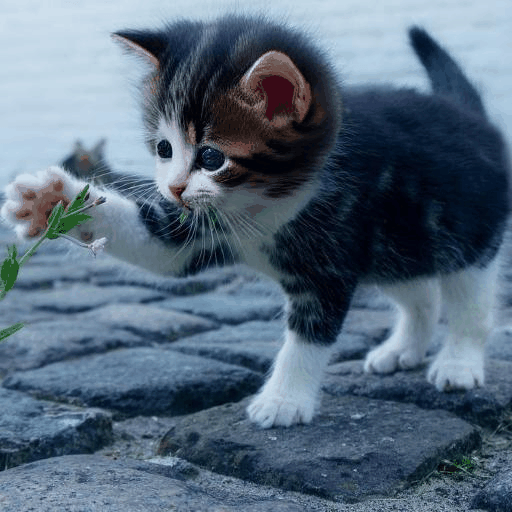}
\caption{A kitten is playing with a flower.}
\end{subfigure}
\caption{\textbf{More results on reward model bias studies using BrushNet.} In each sub-figure, the four images (from left to right) display: the \textit{masked image}, followed by inpainting results from models trained using \textit{HPSv2}, \textit{PickScore}, and \textit{Ensemble}. For optimal detail, view figures zoomed in.}
\label{fig:brushnet_bias_appendix}
\end{figure}

\begin{figure}[!htbp]
\begin{subfigure}[t]{.49\linewidth}
\centering
	\includegraphics[width=0.22\linewidth, trim=0mm 0mm 0mm 0mm, clip]{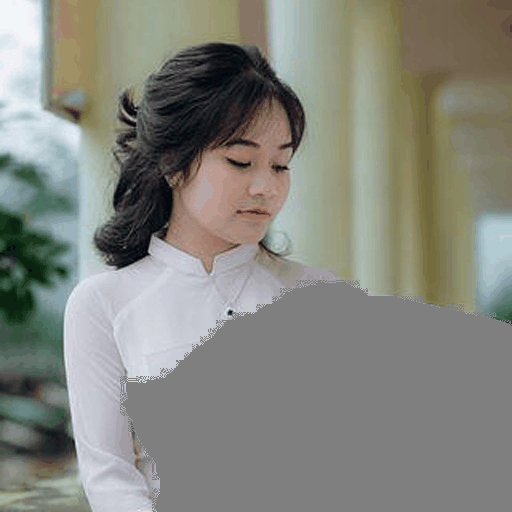}
	\includegraphics[width=0.22\linewidth, trim=0mm 0mm 0mm 0mm, clip]{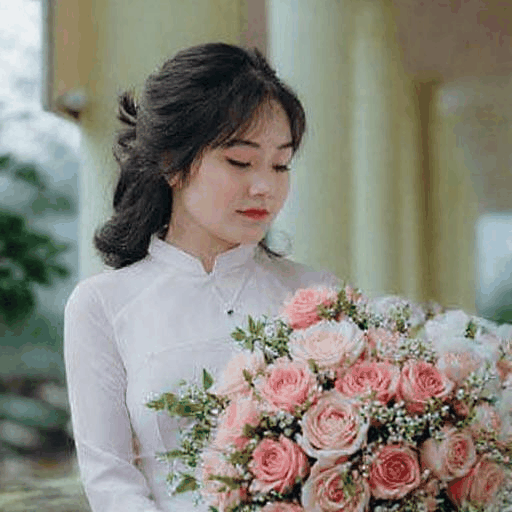}
	\includegraphics[width=0.22\linewidth, trim=0mm 0mm 0mm 0mm, clip]{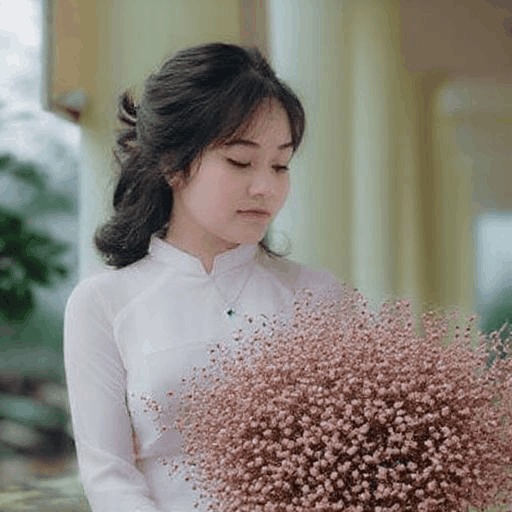}
	\includegraphics[width=0.22\linewidth, trim=0mm 0mm 0mm 0mm, clip]{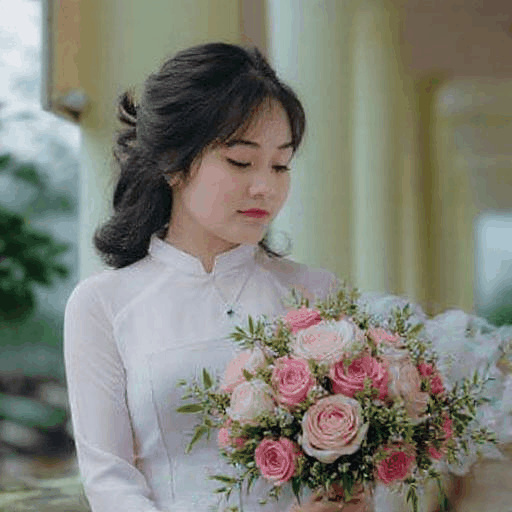}
\caption{A woman in white holding a bouquet of flowers.}
\end{subfigure} 
\begin{subfigure}[t]{.49\linewidth}
\centering
	\includegraphics[width=0.22\linewidth, trim=0mm 0mm 0mm 0mm, clip]{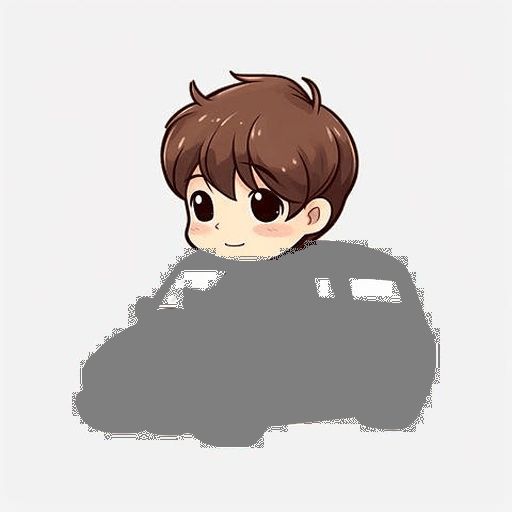}
	\includegraphics[width=0.22\linewidth, trim=0mm 0mm 0mm 0mm, clip]{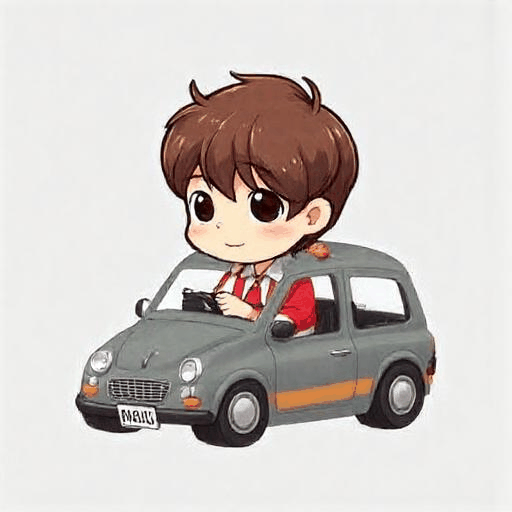}
	\includegraphics[width=0.22\linewidth, trim=0mm 0mm 0mm 0mm, clip]{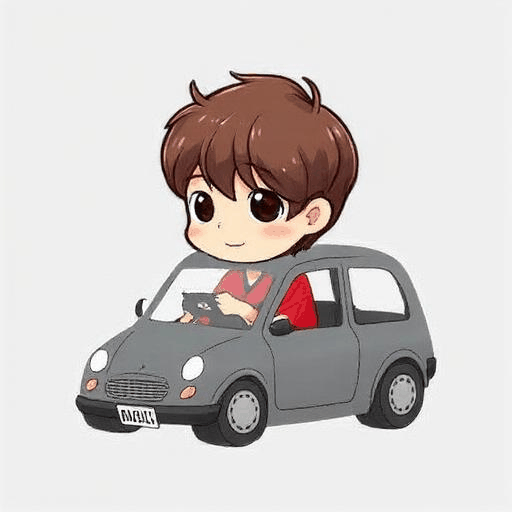}
	\includegraphics[width=0.22\linewidth, trim=0mm 0mm 0mm 0mm, clip]{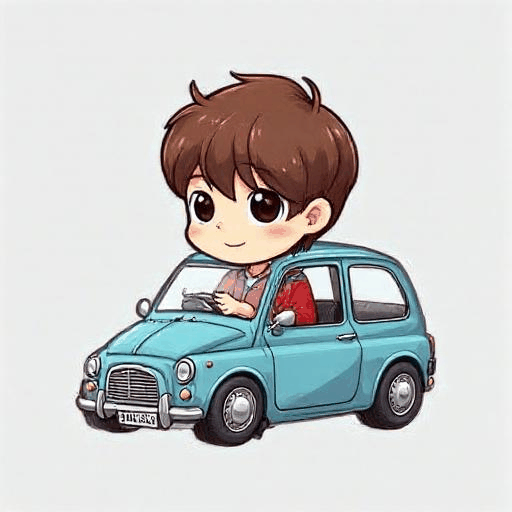}
\caption{A cartoon boy driving a car.}
\end{subfigure} 
\begin{subfigure}[t]{.49\linewidth}
\centering
	\includegraphics[width=0.22\linewidth, trim=0mm 0mm 0mm 0mm, clip]{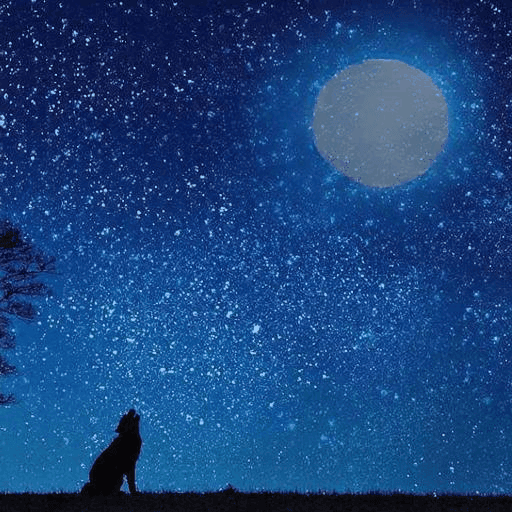}
	\includegraphics[width=0.22\linewidth, trim=0mm 0mm 0mm 0mm, clip]{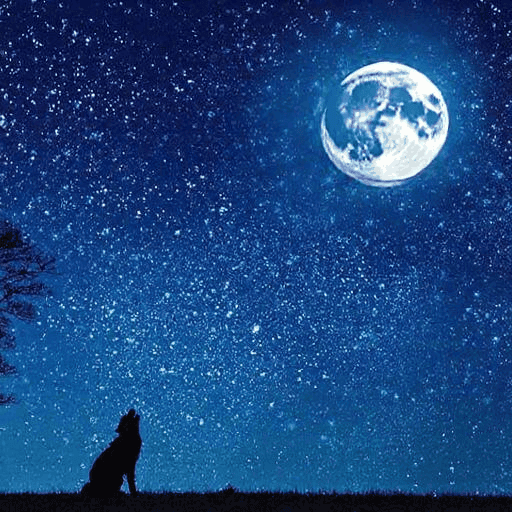}
	\includegraphics[width=0.22\linewidth, trim=0mm 0mm 0mm 0mm, clip]{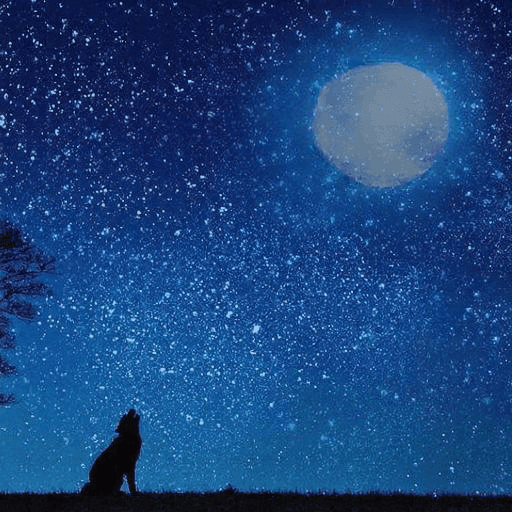}
	\includegraphics[width=0.22\linewidth, trim=0mm 0mm 0mm 0mm, clip]{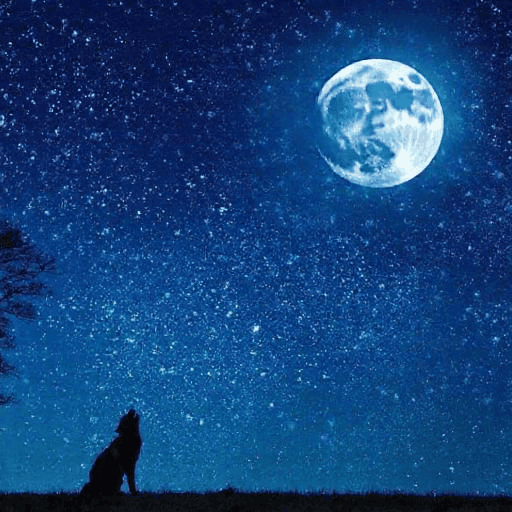}
\caption{Wolf howling at the moon.}
\end{subfigure}
\begin{subfigure}[t]{.49\linewidth}
\centering
	\includegraphics[width=0.22\linewidth, trim=0mm 0mm 0mm 0mm, clip]{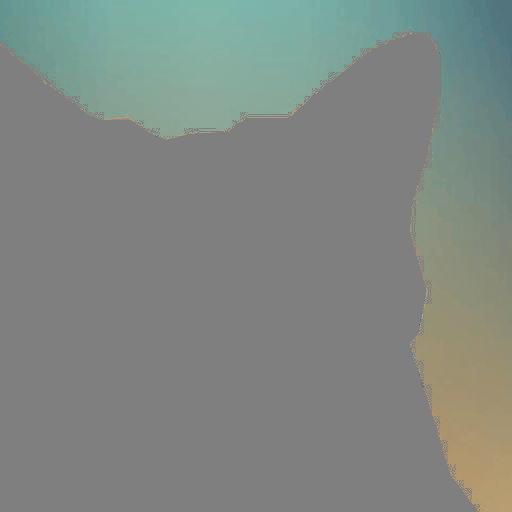}
	\includegraphics[width=0.22\linewidth, trim=0mm 0mm 0mm 0mm, clip]{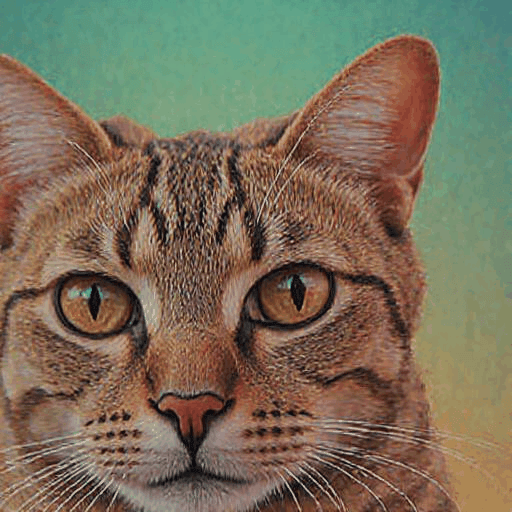}
	\includegraphics[width=0.22\linewidth, trim=0mm 0mm 0mm 0mm, clip]{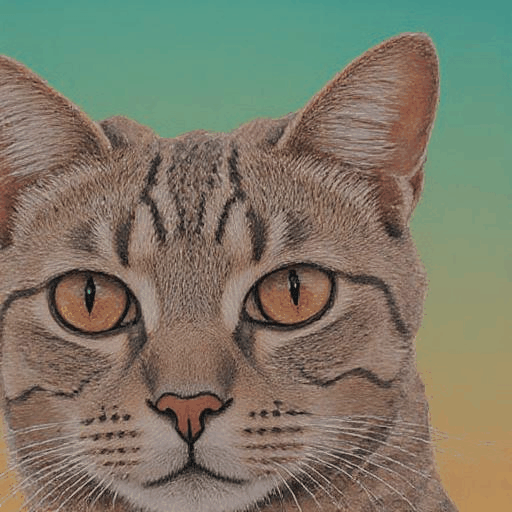}
	\includegraphics[width=0.22\linewidth, trim=0mm 0mm 0mm 0mm, clip]{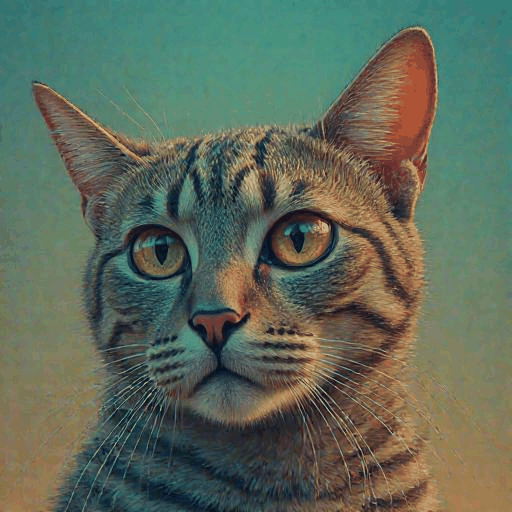}
\caption{A cat is shown in low polygonal style.}
\end{subfigure} 
\begin{subfigure}[t]{.49\linewidth}
\centering
	\includegraphics[width=0.22\linewidth, trim=0mm 0mm 0mm 0mm, clip]{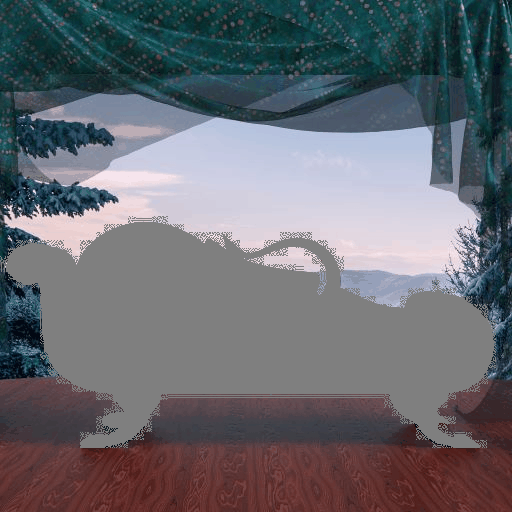}
	\includegraphics[width=0.22\linewidth, trim=0mm 0mm 0mm 0mm, clip]{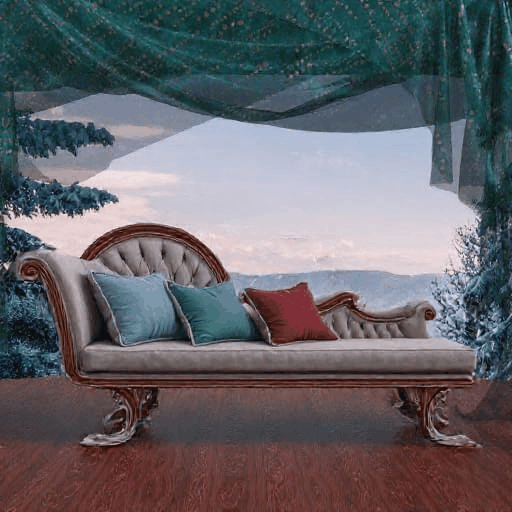}
	\includegraphics[width=0.22\linewidth, trim=0mm 0mm 0mm 0mm, clip]{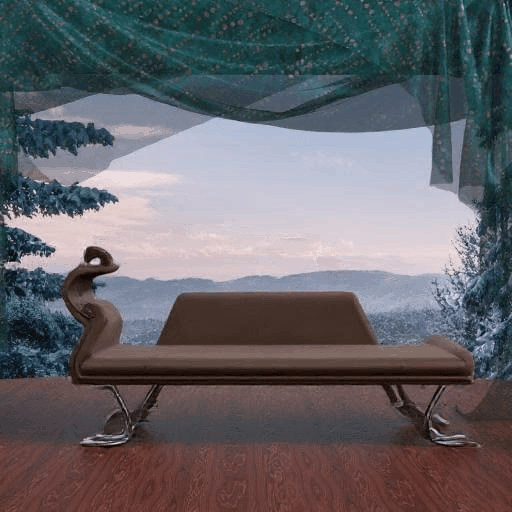}
	\includegraphics[width=0.22\linewidth, trim=0mm 0mm 0mm 0mm, clip]{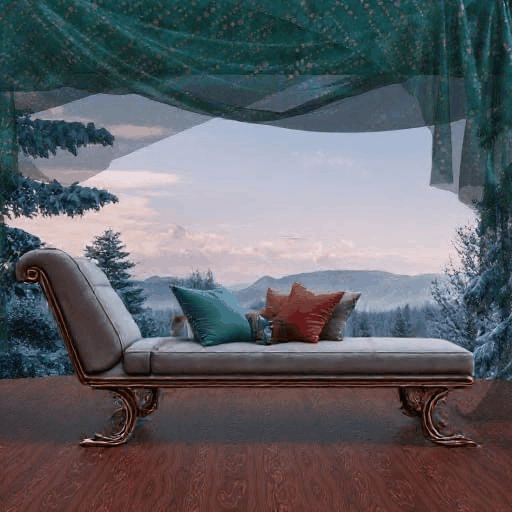}
\caption{A couch with a winged chair at a window.}
\end{subfigure} 
\begin{subfigure}[t]{.49\linewidth}
\centering
	\includegraphics[width=0.22\linewidth, trim=0mm 0mm 0mm 0mm, clip]{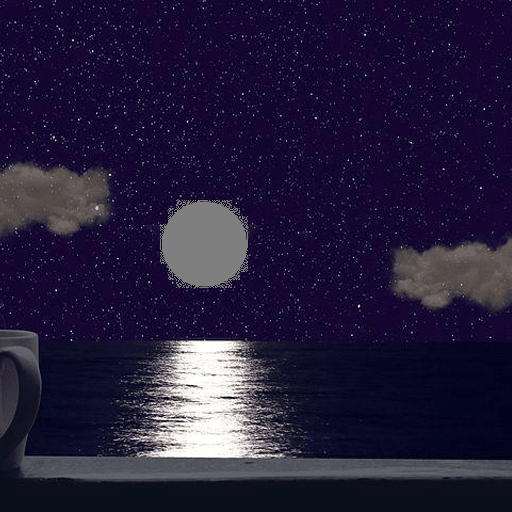}
	\includegraphics[width=0.22\linewidth, trim=0mm 0mm 0mm 0mm, clip]{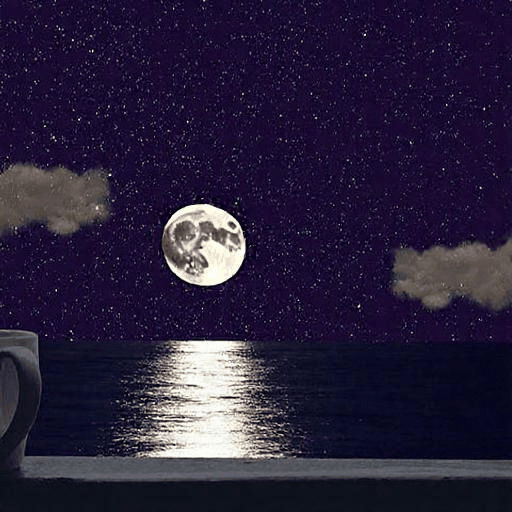}
	\includegraphics[width=0.22\linewidth, trim=0mm 0mm 0mm 0mm, clip]{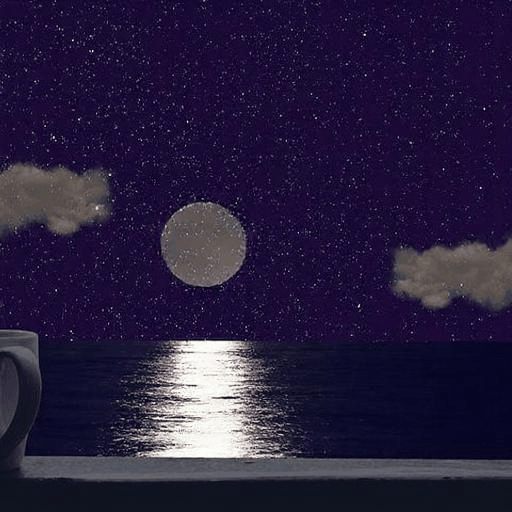}
	\includegraphics[width=0.22\linewidth, trim=0mm 0mm 0mm 0mm, clip]{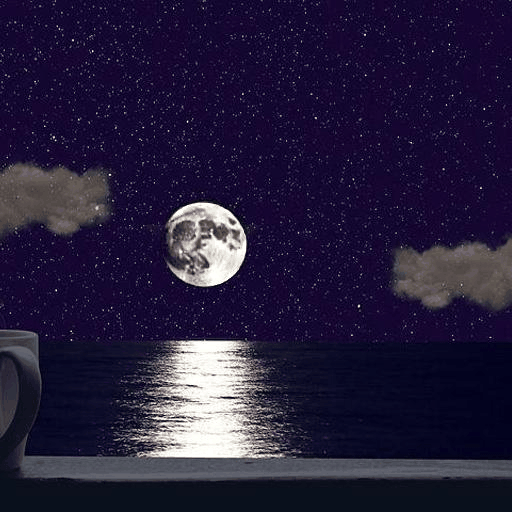}
\caption{A bright moon on the sea.}
\end{subfigure} 
\begin{subfigure}[t]{.49\linewidth}
\centering
	\includegraphics[width=0.22\linewidth, trim=0mm 0mm 0mm 0mm, clip]{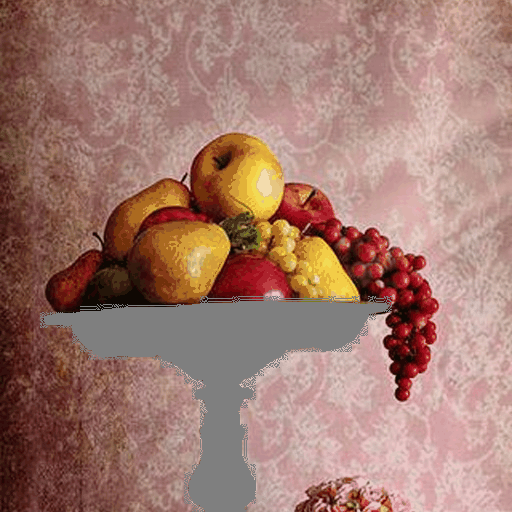}
	\includegraphics[width=0.22\linewidth, trim=0mm 0mm 0mm 0mm, clip]{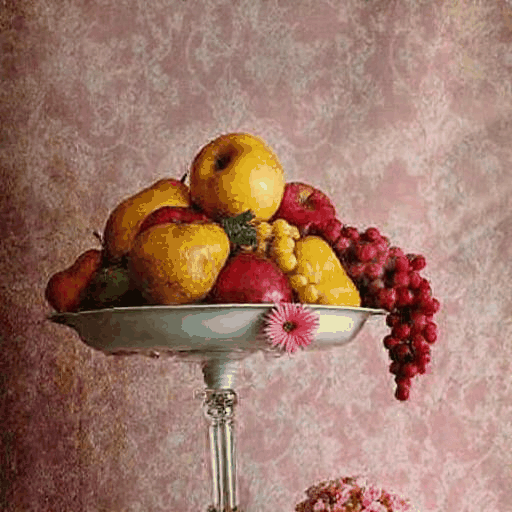}
	\includegraphics[width=0.22\linewidth, trim=0mm 0mm 0mm 0mm, clip]{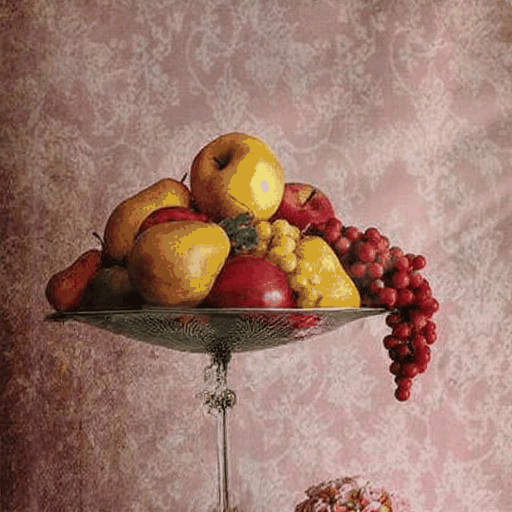}
	\includegraphics[width=0.22\linewidth, trim=0mm 0mm 0mm 0mm, clip]{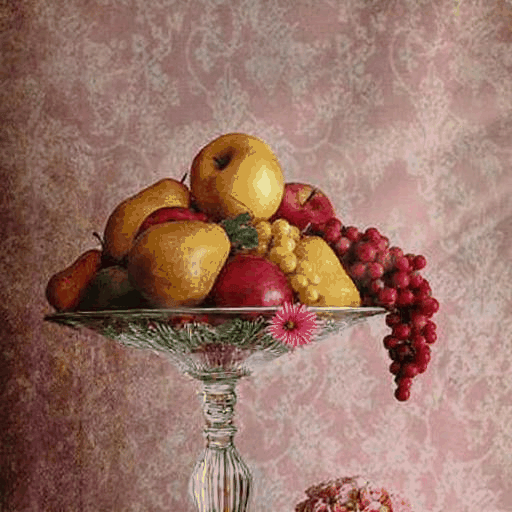}
\caption{A fruit bowl with a pink flower on top.}
\end{subfigure} 
\begin{subfigure}[t]{.49\linewidth}
\centering
	\includegraphics[width=0.22\linewidth, trim=0mm 0mm 0mm 0mm, clip]{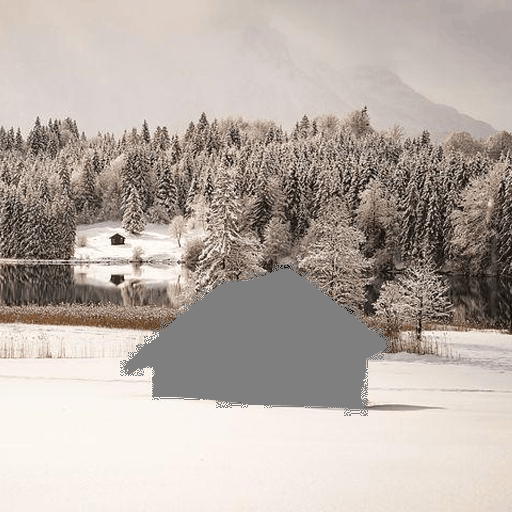}
	\includegraphics[width=0.22\linewidth, trim=0mm 0mm 0mm 0mm, clip]{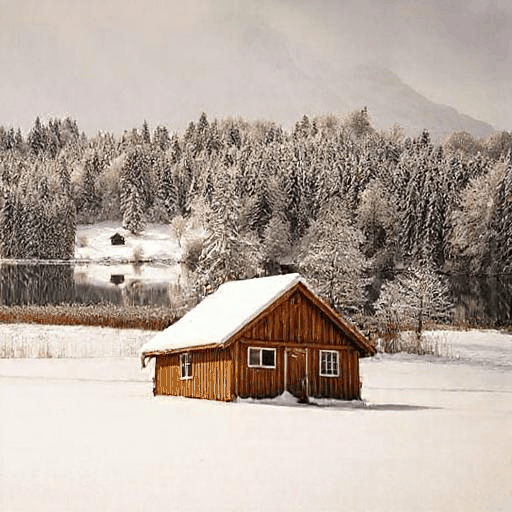}
	\includegraphics[width=0.22\linewidth, trim=0mm 0mm 0mm 0mm, clip]{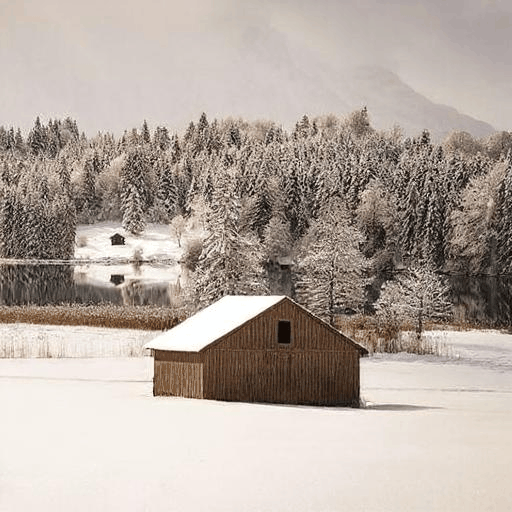}
	\includegraphics[width=0.22\linewidth, trim=0mm 0mm 0mm 0mm, clip]{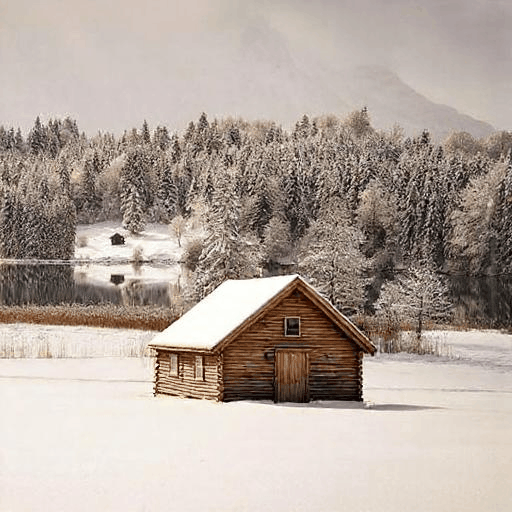}
\caption{A small cabin in the snow near a lake.}
\end{subfigure} 
\begin{subfigure}[t]{.49\linewidth}
\centering
	\includegraphics[width=0.22\linewidth, trim=0mm 0mm 0mm 0mm, clip]{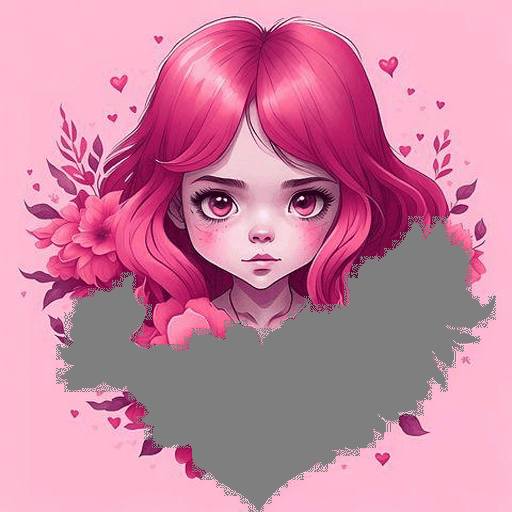}
	\includegraphics[width=0.22\linewidth, trim=0mm 0mm 0mm 0mm, clip]{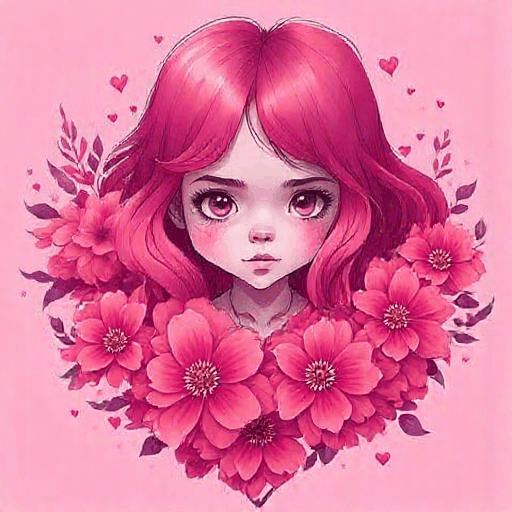}
	\includegraphics[width=0.22\linewidth, trim=0mm 0mm 0mm 0mm, clip]{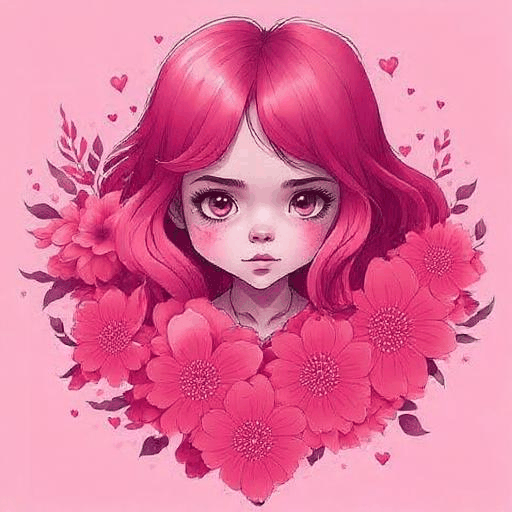}
	\includegraphics[width=0.22\linewidth, trim=0mm 0mm 0mm 0mm, clip]{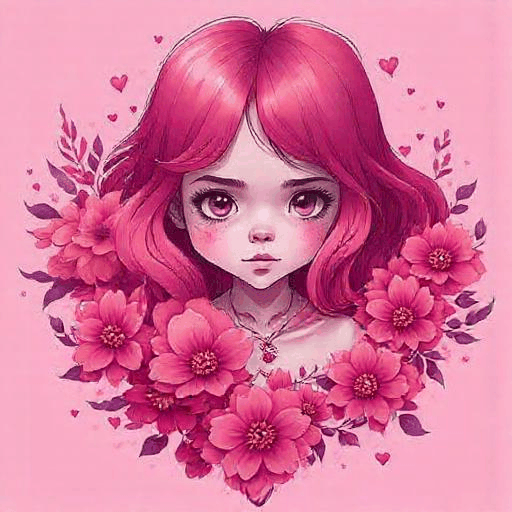}
\caption{A girl with pink hair and flowers on her face.}
\end{subfigure}
\begin{subfigure}[t]{.49\linewidth}
\centering
	\includegraphics[width=0.22\linewidth, trim=0mm 0mm 0mm 0mm, clip]{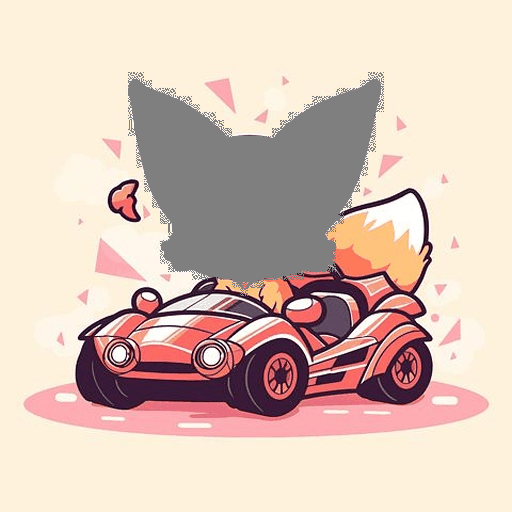}
	\includegraphics[width=0.22\linewidth, trim=0mm 0mm 0mm 0mm, clip]{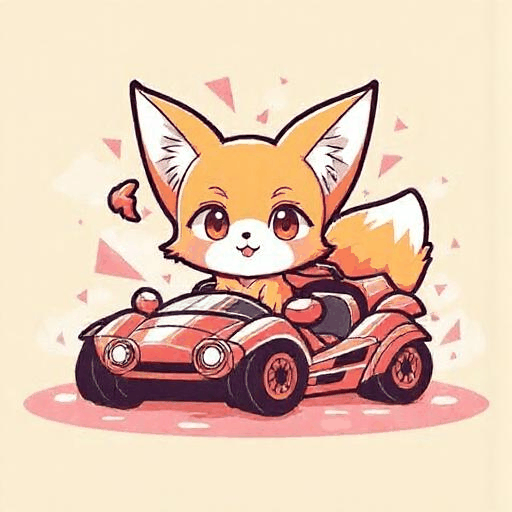}
	\includegraphics[width=0.22\linewidth, trim=0mm 0mm 0mm 0mm, clip]{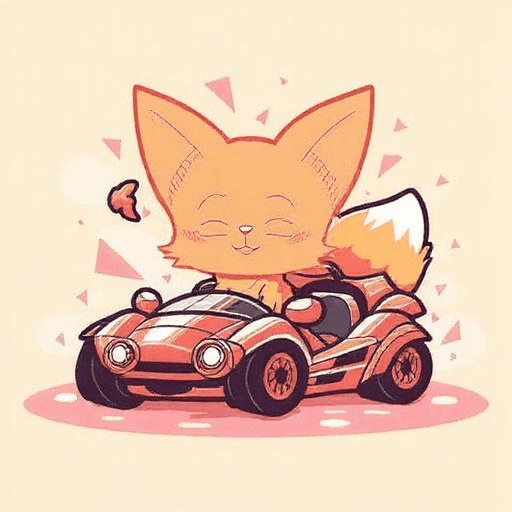}
	\includegraphics[width=0.22\linewidth, trim=0mm 0mm 0mm 0mm, clip]{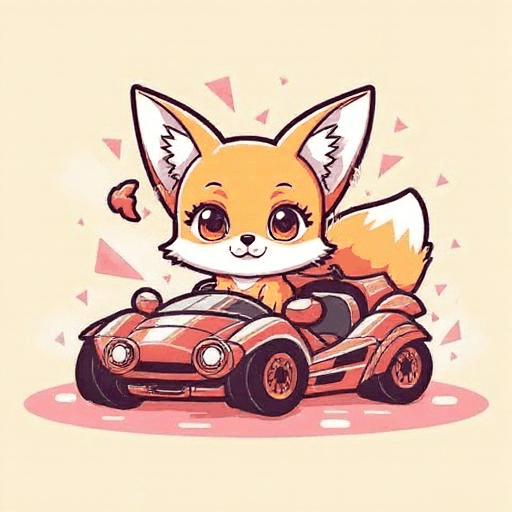}
\caption{Cartoon fox driving a car with a cute face.}
\end{subfigure} 
\begin{subfigure}[t]{.49\linewidth}
\centering
	\includegraphics[width=0.22\linewidth, trim=0mm 0mm 0mm 0mm, clip]{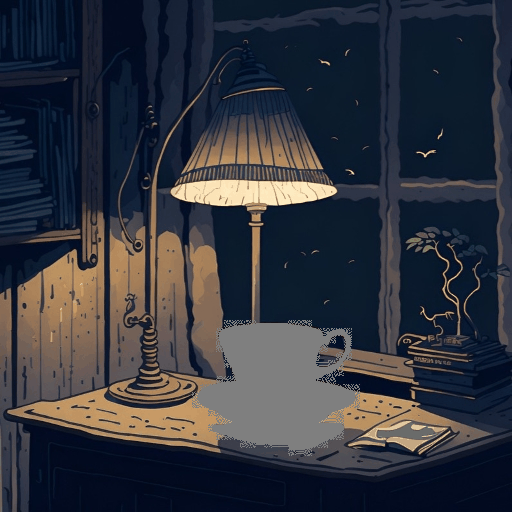}
	\includegraphics[width=0.22\linewidth, trim=0mm 0mm 0mm 0mm, clip]{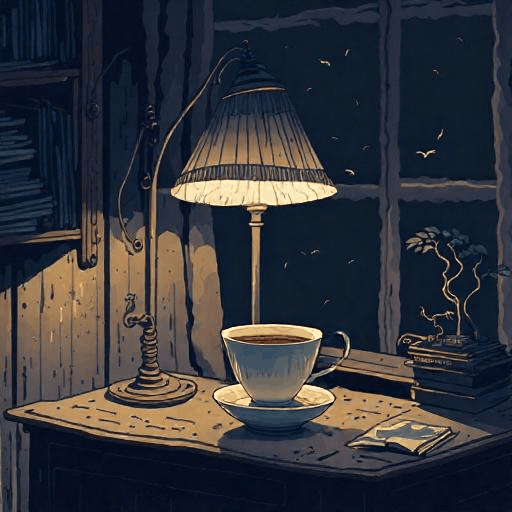}
	\includegraphics[width=0.22\linewidth, trim=0mm 0mm 0mm 0mm, clip]{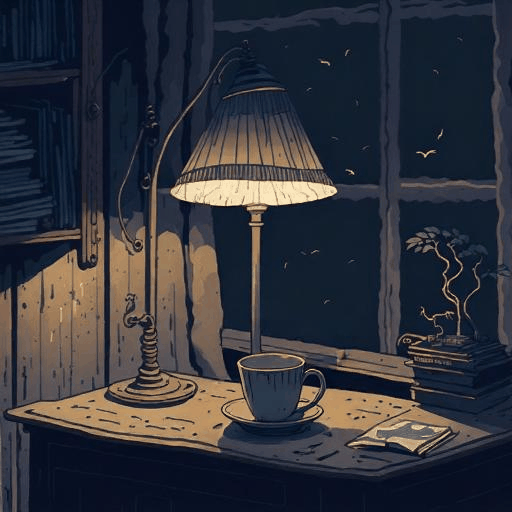}
	\includegraphics[width=0.22\linewidth, trim=0mm 0mm 0mm 0mm, clip]{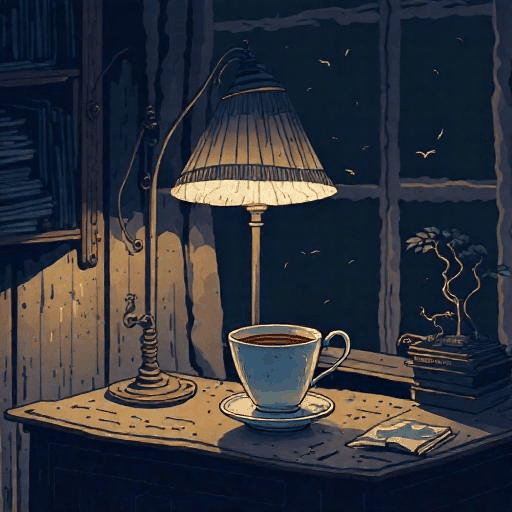}
\caption{A lamp and a cup of tea on a table.}
\end{subfigure} 
\begin{subfigure}[t]{.49\linewidth}
\centering
	\includegraphics[width=0.22\linewidth, trim=0mm 0mm 0mm 0mm, clip]{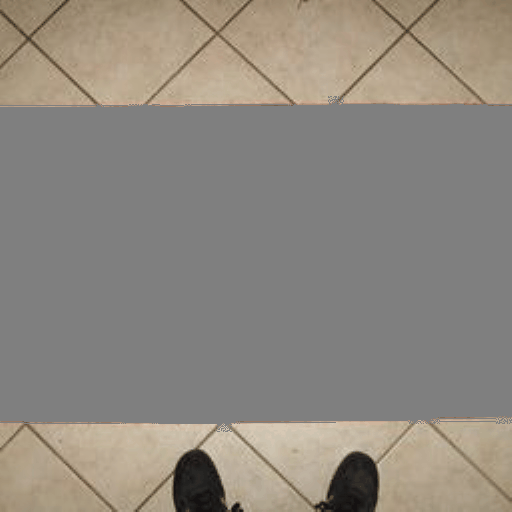}
	\includegraphics[width=0.22\linewidth, trim=0mm 0mm 0mm 0mm, clip]{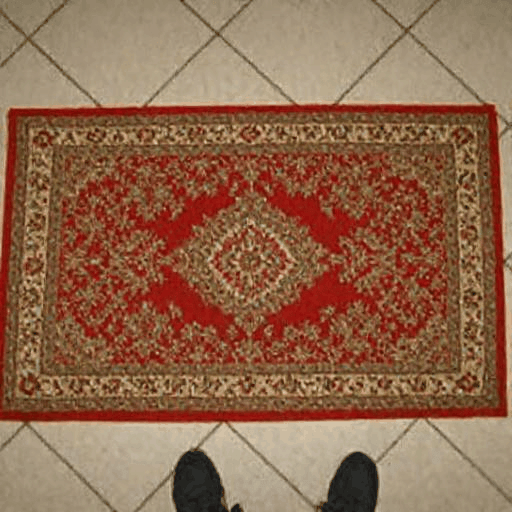}
	\includegraphics[width=0.22\linewidth, trim=0mm 0mm 0mm 0mm, clip]{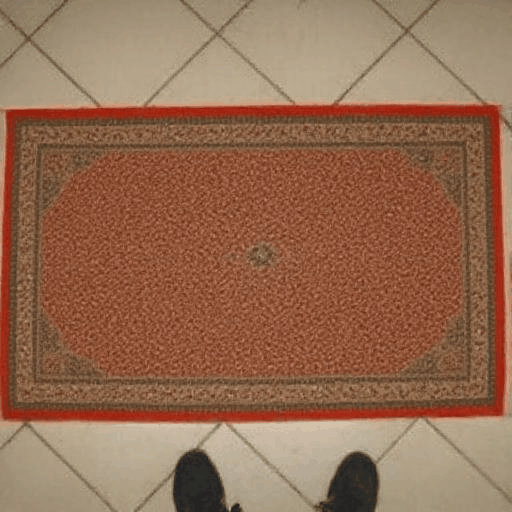}
	\includegraphics[width=0.22\linewidth, trim=0mm 0mm 0mm 0mm, clip]{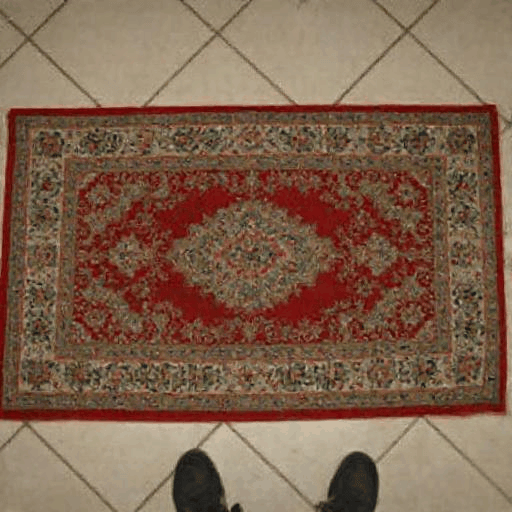}
\caption{A person standing on a tile floor with a rug.}
\end{subfigure} 
\begin{subfigure}[t]{.49\linewidth}
\centering
	\includegraphics[width=0.22\linewidth, trim=0mm 0mm 0mm 0mm, clip]{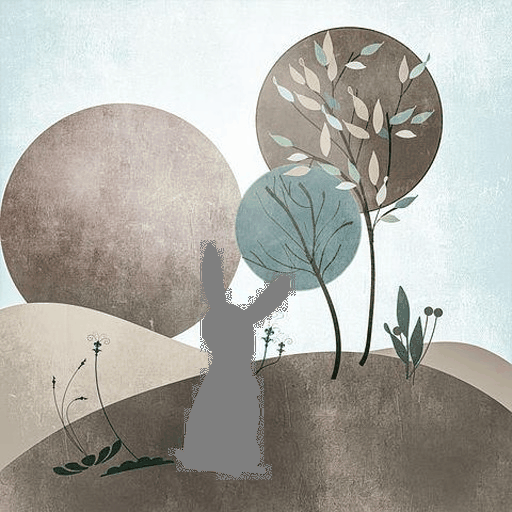}
	\includegraphics[width=0.22\linewidth, trim=0mm 0mm 0mm 0mm, clip]{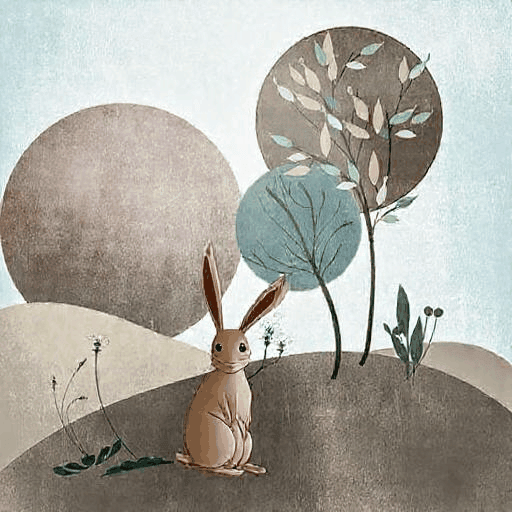}
	\includegraphics[width=0.22\linewidth, trim=0mm 0mm 0mm 0mm, clip]{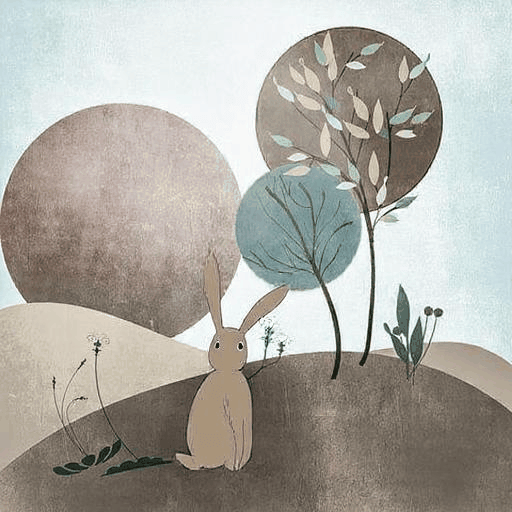}
	\includegraphics[width=0.22\linewidth, trim=0mm 0mm 0mm 0mm, clip]{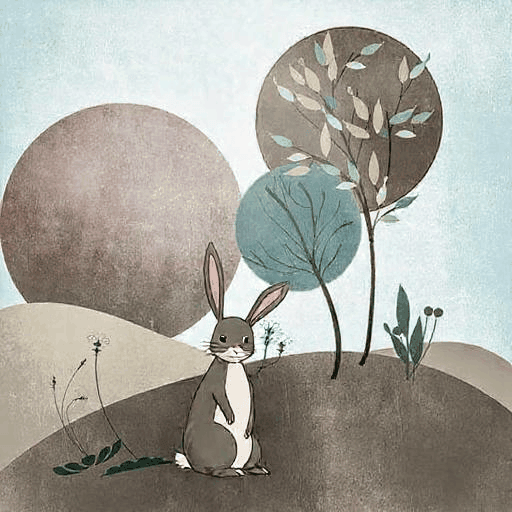}
\caption{A rabbit sitting on a hill with trees.}
\end{subfigure} 
\begin{subfigure}[t]{.49\linewidth}
\centering
	\includegraphics[width=0.22\linewidth, trim=0mm 0mm 0mm 0mm, clip]{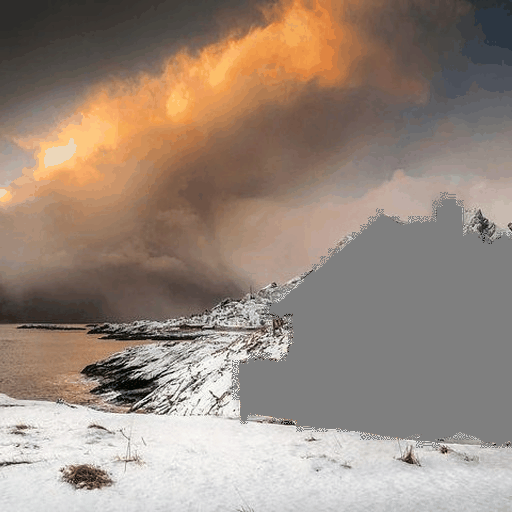}
	\includegraphics[width=0.22\linewidth, trim=0mm 0mm 0mm 0mm, clip]{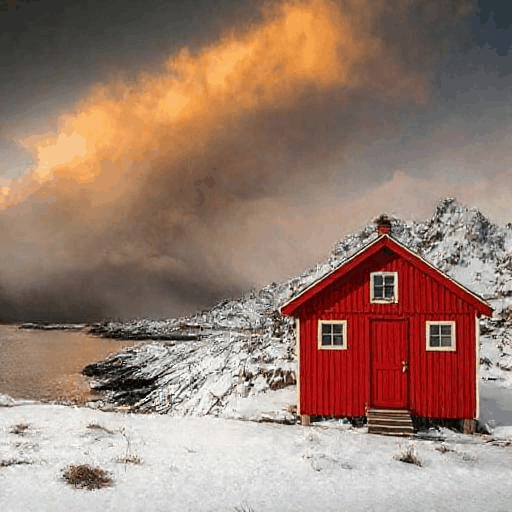}
	\includegraphics[width=0.22\linewidth, trim=0mm 0mm 0mm 0mm, clip]{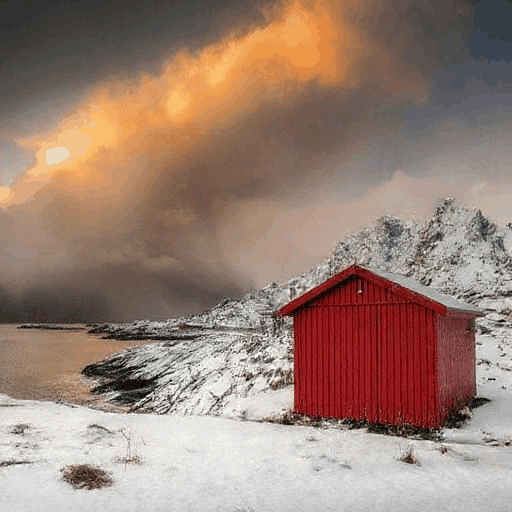}
	\includegraphics[width=0.22\linewidth, trim=0mm 0mm 0mm 0mm, clip]{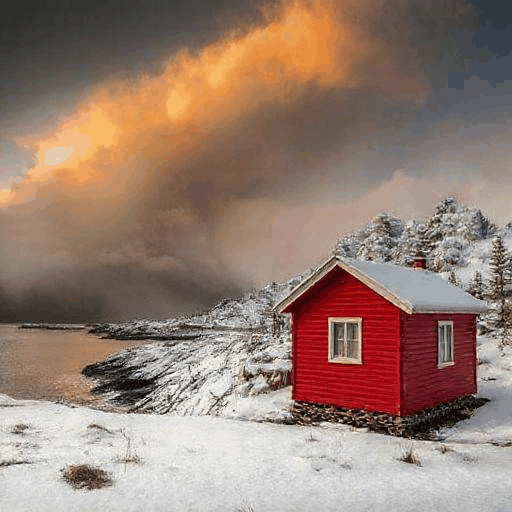}
\caption{A red cabin sits on the shore of a lake.}
\end{subfigure} 
\begin{subfigure}[t]{.49\linewidth}
\centering
	\includegraphics[width=0.22\linewidth, trim=0mm 0mm 0mm 0mm, clip]{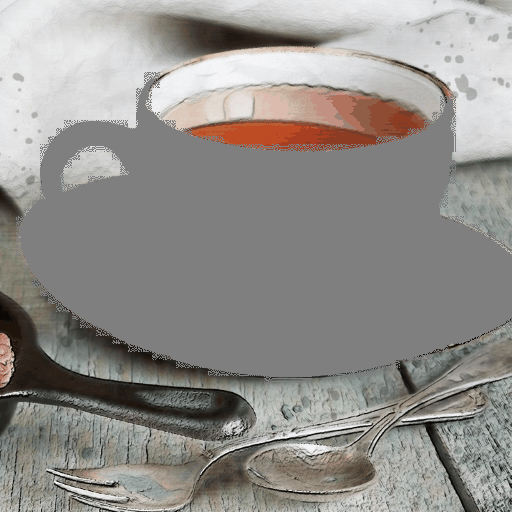}
	\includegraphics[width=0.22\linewidth, trim=0mm 0mm 0mm 0mm, clip]{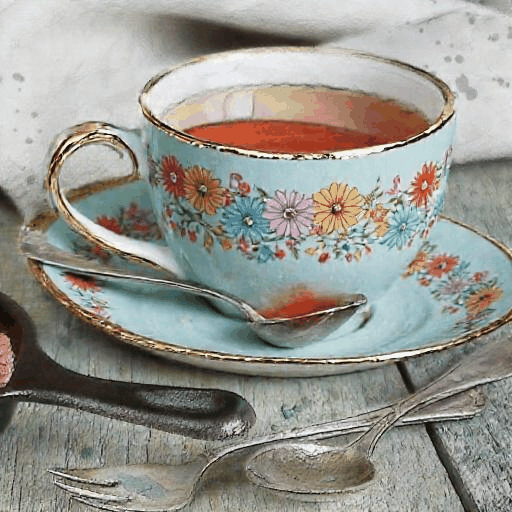}
	\includegraphics[width=0.22\linewidth, trim=0mm 0mm 0mm 0mm, clip]{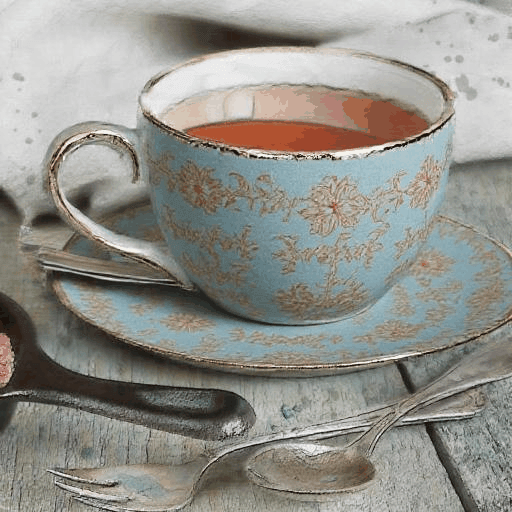}
	\includegraphics[width=0.22\linewidth, trim=0mm 0mm 0mm 0mm, clip]{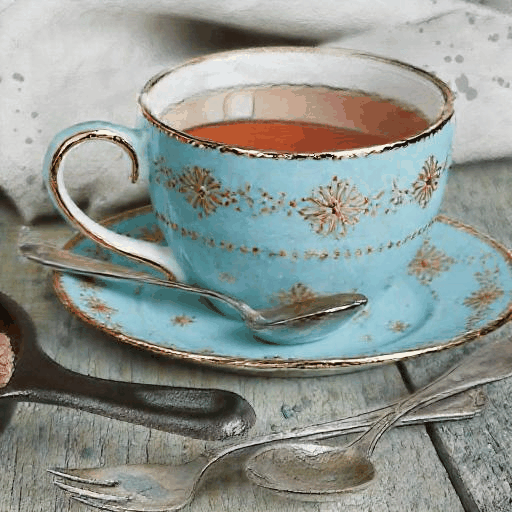}
\caption{A teacup and saucer with spoons.}
\end{subfigure} 
\hfill
\begin{subfigure}[t]{.49\linewidth}
\centering
	\includegraphics[width=0.22\linewidth, trim=0mm 0mm 0mm 0mm, clip]{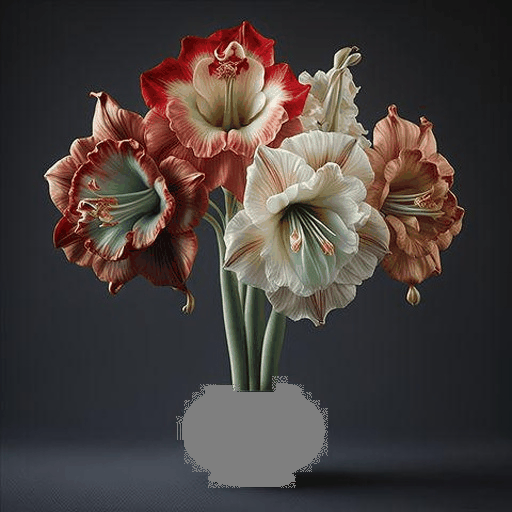}
	\includegraphics[width=0.22\linewidth, trim=0mm 0mm 0mm 0mm, clip]{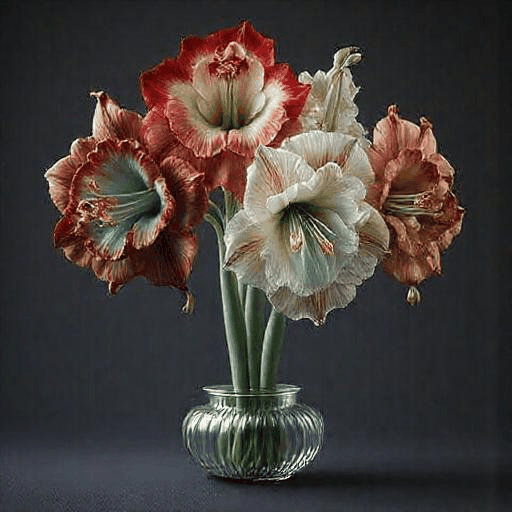}
	\includegraphics[width=0.22\linewidth, trim=0mm 0mm 0mm 0mm, clip]{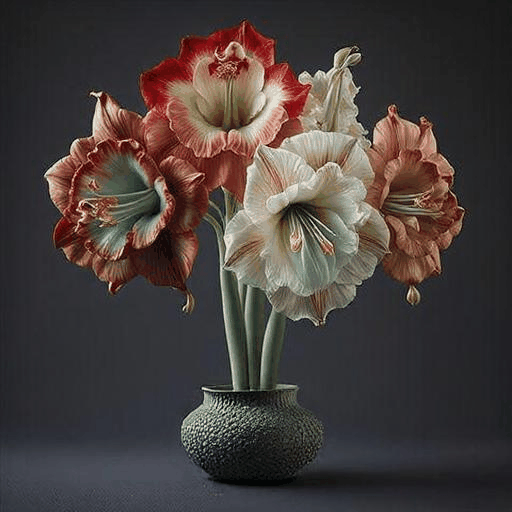}
	\includegraphics[width=0.22\linewidth, trim=0mm 0mm 0mm 0mm, clip]{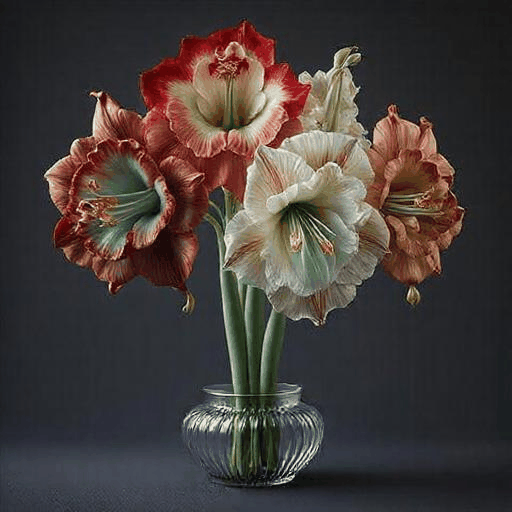}
\caption{A vase with flowers in it on a dark background.}
\end{subfigure}
\caption{\textbf{More results on reward model bias studies using FLUX.1 Fill.} In each sub-figure, the four images (from left to right) display: the \textit{masked image}, followed by inpainting results from models trained using \textit{HPSv2}, \textit{PickScore}, and \textit{Ensemble}. For optimal detail, view figures zoomed in.}
\label{fig:flux_bias_appendix}
\end{figure}

\begin{figure}[!htbp]
\begin{subfigure}[t]{.49\linewidth}
\centering
    \includegraphics[width=0.22\linewidth, trim=0mm 0mm 0mm 0mm, clip]{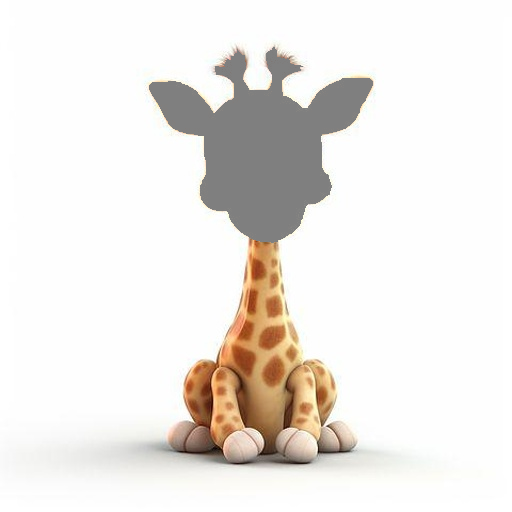}
    \includegraphics[width=0.22\linewidth, trim=0mm 0mm 0mm 0mm, clip]{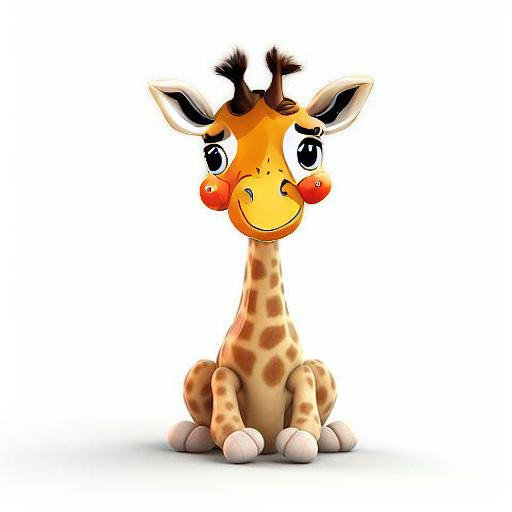}
    \includegraphics[width=0.22\linewidth, trim=0mm 0mm 0mm 0mm, clip]{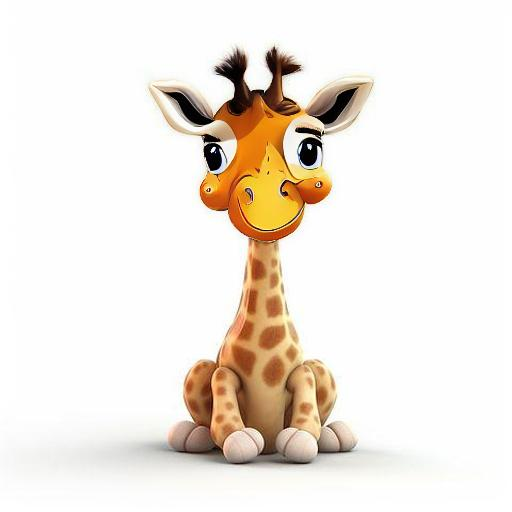}
    \includegraphics[width=0.22\linewidth, trim=0mm 0mm 0mm 0mm, clip]{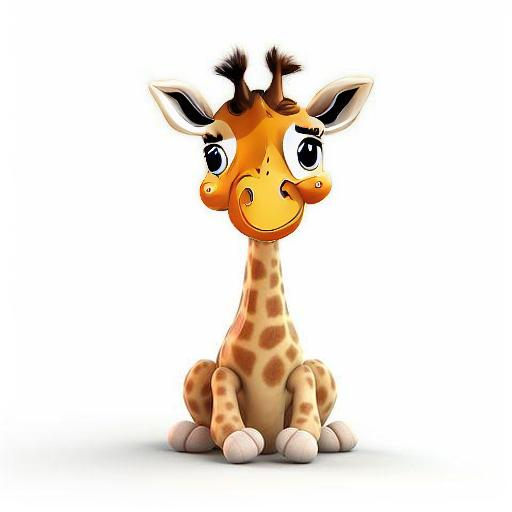}
\caption{a cartoon giraffe sitting down and looking at the camera}
\end{subfigure}
\begin{subfigure}[t]{.49\linewidth}
\centering
    \includegraphics[width=0.22\linewidth, trim=0mm 0mm 0mm 0mm, clip]{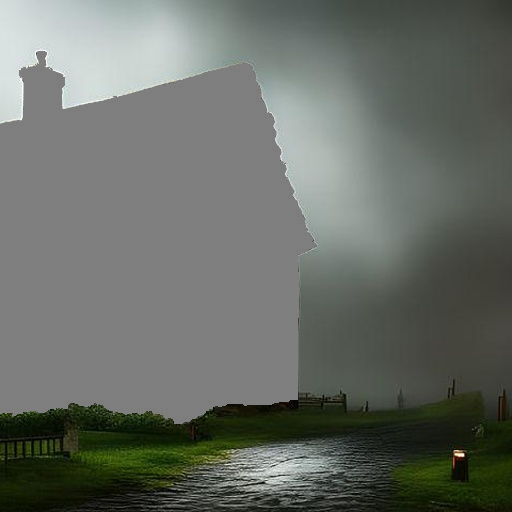}
    \includegraphics[width=0.22\linewidth, trim=0mm 0mm 0mm 0mm, clip]{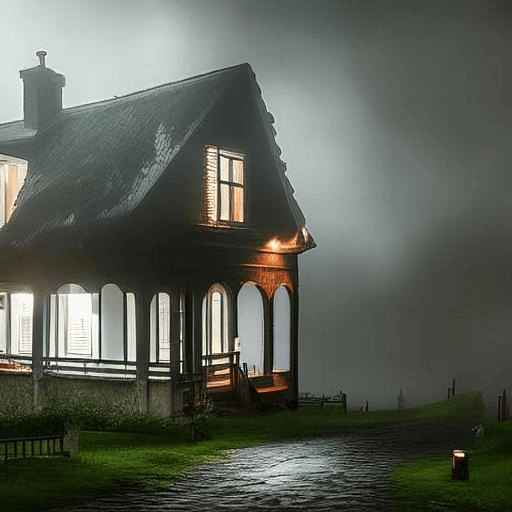}
    \includegraphics[width=0.22\linewidth, trim=0mm 0mm 0mm 0mm, clip]{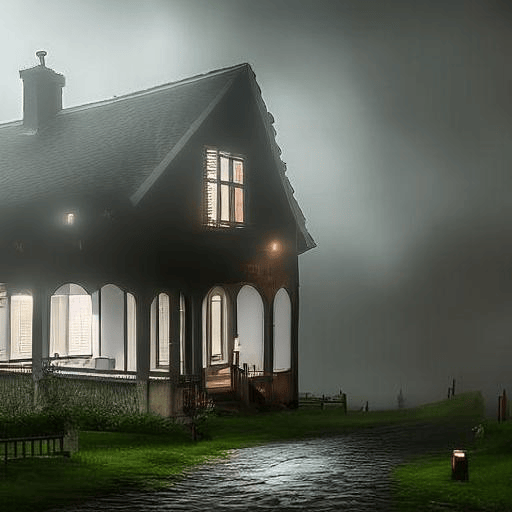}
    \includegraphics[width=0.22\linewidth, trim=0mm 0mm 0mm 0mm, clip]{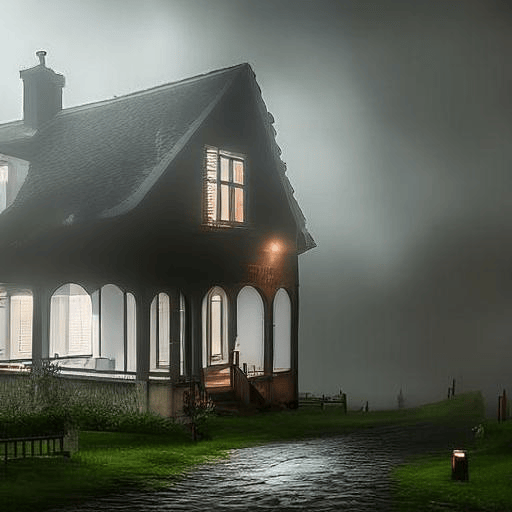}
\caption{a house in the middle of a foggy night}
\end{subfigure}
\begin{subfigure}[t]{.49\linewidth}
\centering
    \includegraphics[width=0.22\linewidth, trim=0mm 0mm 0mm 0mm, clip]{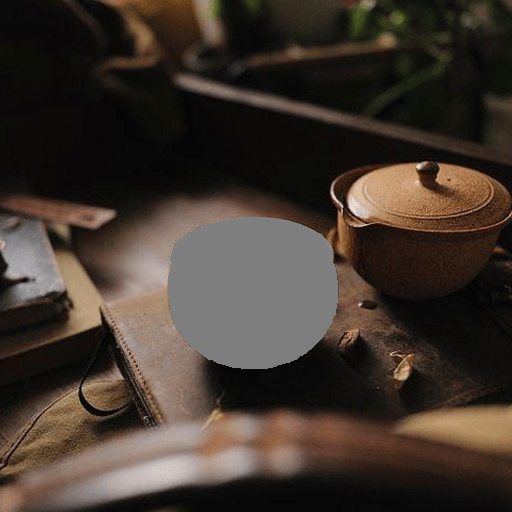}
    \includegraphics[width=0.22\linewidth, trim=0mm 0mm 0mm 0mm, clip]{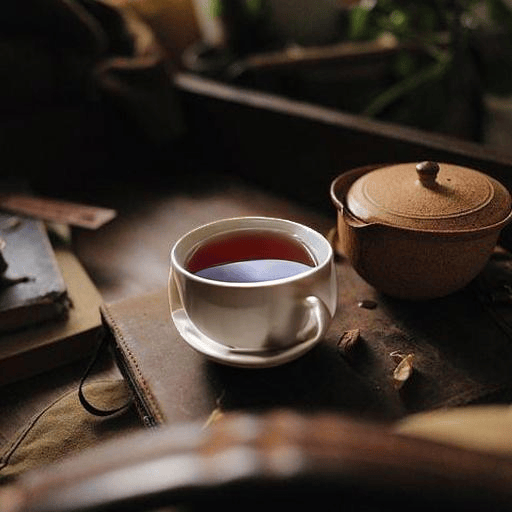}
    \includegraphics[width=0.22\linewidth, trim=0mm 0mm 0mm 0mm, clip]{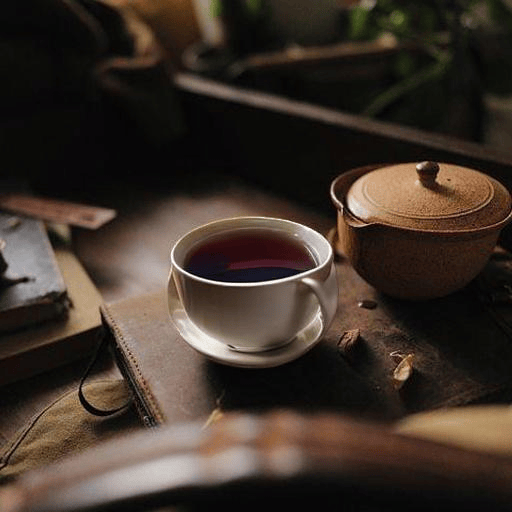}
    \includegraphics[width=0.22\linewidth, trim=0mm 0mm 0mm 0mm, clip]{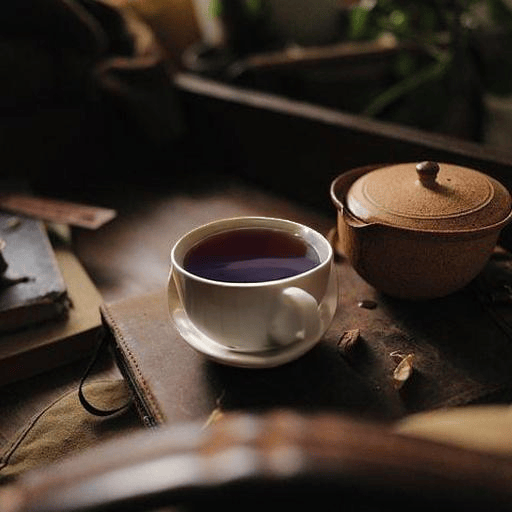}
\caption{a cup of tea sits on top of a wooden table}
\end{subfigure}
\begin{subfigure}[t]{.49\linewidth}
\centering
    \includegraphics[width=0.22\linewidth, trim=0mm 0mm 0mm 0mm, clip]{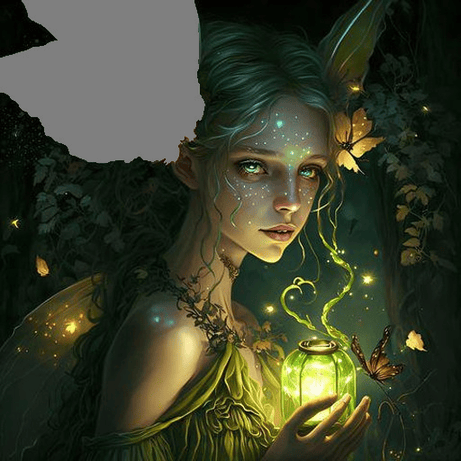}
    \includegraphics[width=0.22\linewidth, trim=0mm 0mm 0mm 0mm, clip]{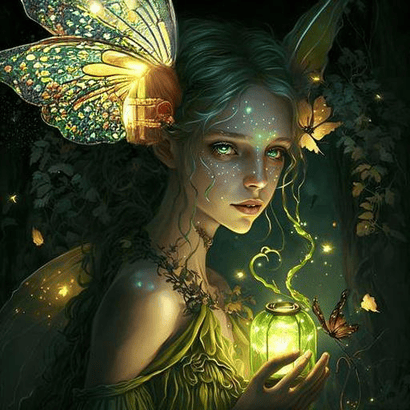}
    \includegraphics[width=0.22\linewidth, trim=0mm 0mm 0mm 0mm, clip]{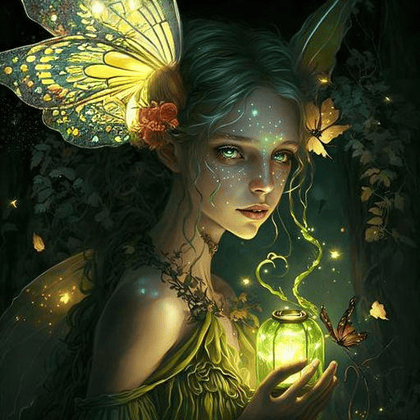}
    \includegraphics[width=0.22\linewidth, trim=0mm 0mm 0mm 0mm, clip]{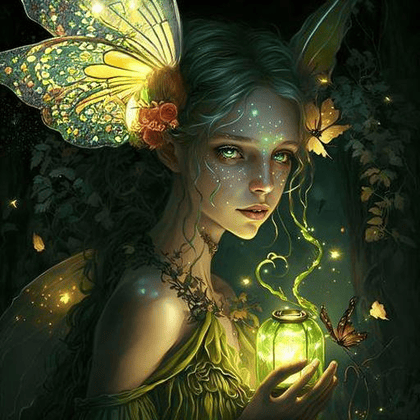}
\caption{a fairy with a lantern in her hand}
\end{subfigure}
\begin{subfigure}[t]{.49\linewidth}
\centering
    \includegraphics[width=0.22\linewidth, trim=0mm 0mm 0mm 0mm, clip]{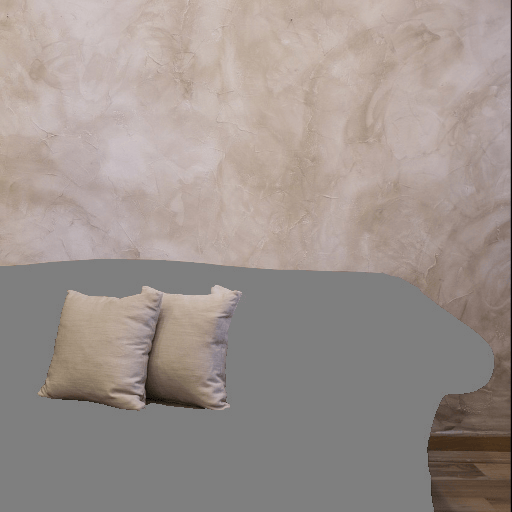}
    \includegraphics[width=0.22\linewidth, trim=0mm 0mm 0mm 0mm, clip]{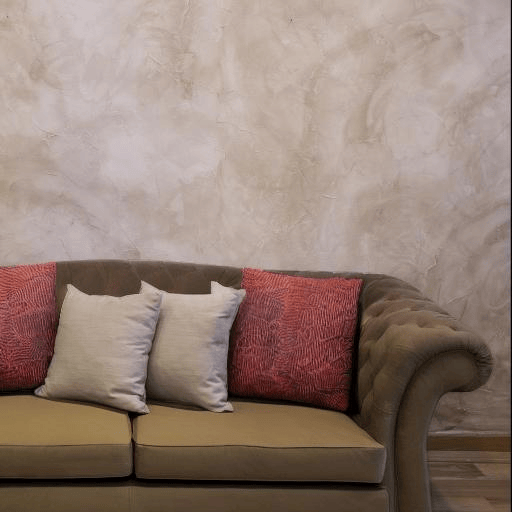}
    \includegraphics[width=0.22\linewidth, trim=0mm 0mm 0mm 0mm, clip]{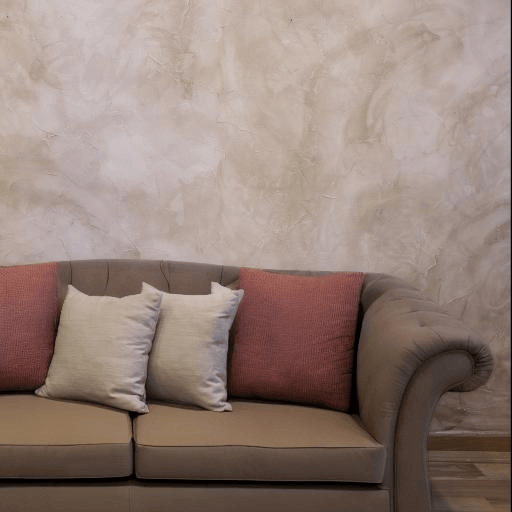}
    \includegraphics[width=0.22\linewidth, trim=0mm 0mm 0mm 0mm, clip]{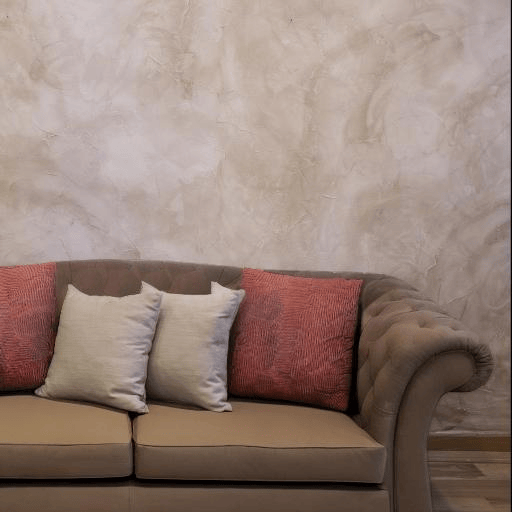}
\caption{a couch with pillows and a wall behind it}
\end{subfigure}
\begin{subfigure}[t]{.49\linewidth}
\centering
    \includegraphics[width=0.22\linewidth, trim=0mm 0mm 0mm 0mm, clip]{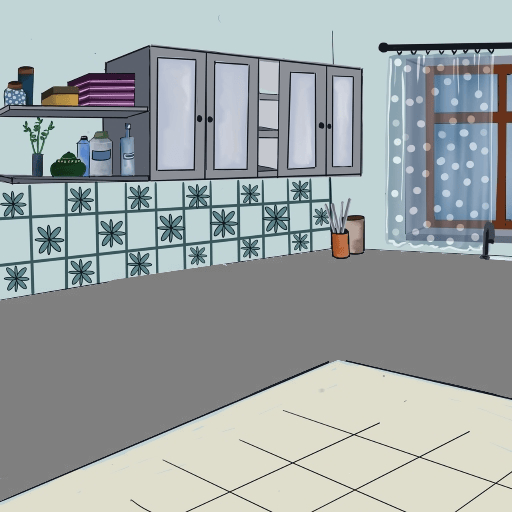}
    \includegraphics[width=0.22\linewidth, trim=0mm 0mm 0mm 0mm, clip]{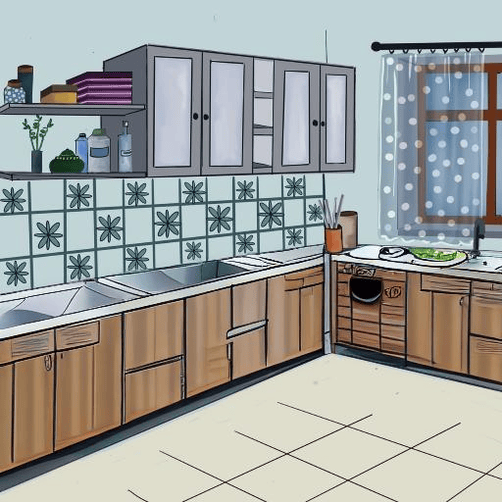}
    \includegraphics[width=0.22\linewidth, trim=0mm 0mm 0mm 0mm, clip]{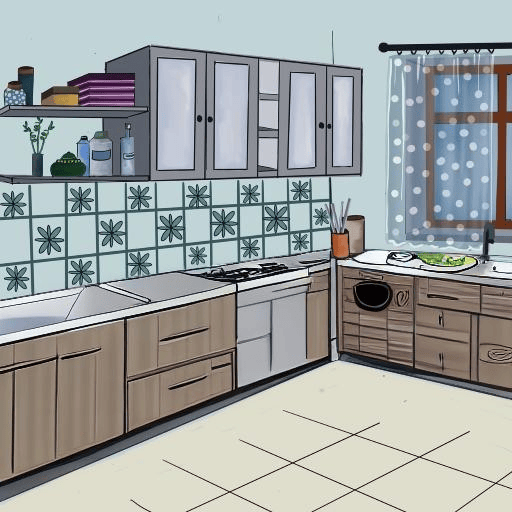}
    \includegraphics[width=0.22\linewidth, trim=0mm 0mm 0mm 0mm, clip]{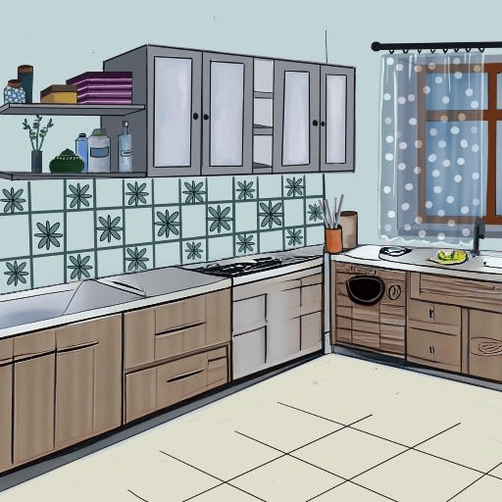}
\caption{a cartoon drawing of a kitchen with a stove and oven}
\end{subfigure}
\begin{subfigure}[t]{.49\linewidth}
\centering
    \includegraphics[width=0.22\linewidth, trim=0mm 0mm 0mm 0mm, clip]{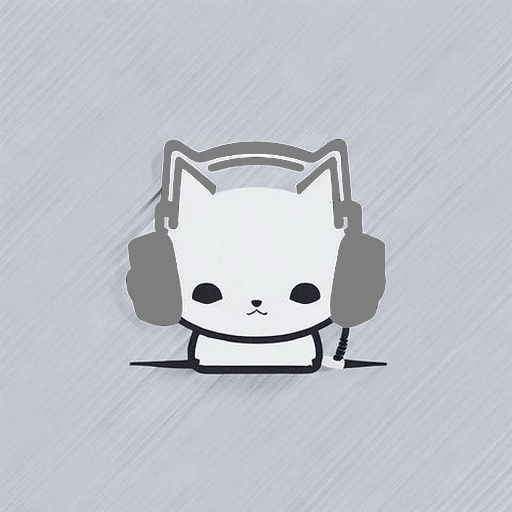}
    \includegraphics[width=0.22\linewidth, trim=0mm 0mm 0mm 0mm, clip]{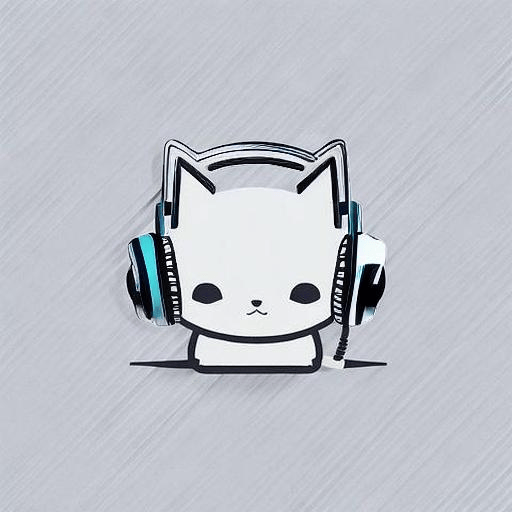}
    \includegraphics[width=0.22\linewidth, trim=0mm 0mm 0mm 0mm, clip]{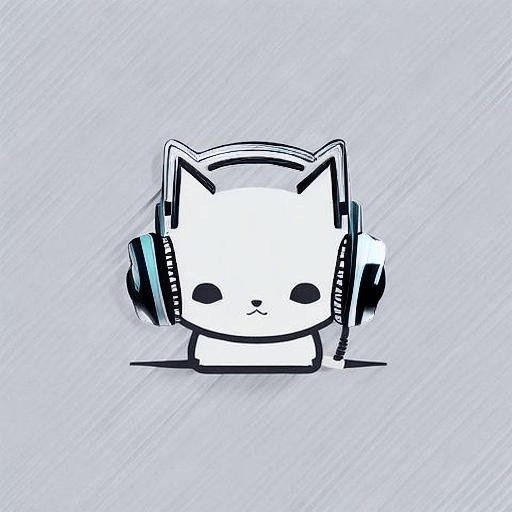}
    \includegraphics[width=0.22\linewidth, trim=0mm 0mm 0mm 0mm, clip]{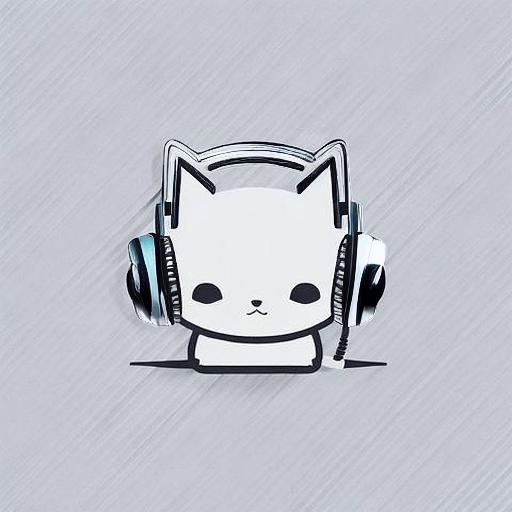}
\caption{a cat with headphones on its head}
\end{subfigure}
\begin{subfigure}[t]{.49\linewidth}
\centering
    \includegraphics[width=0.22\linewidth, trim=0mm 0mm 0mm 0mm, clip]{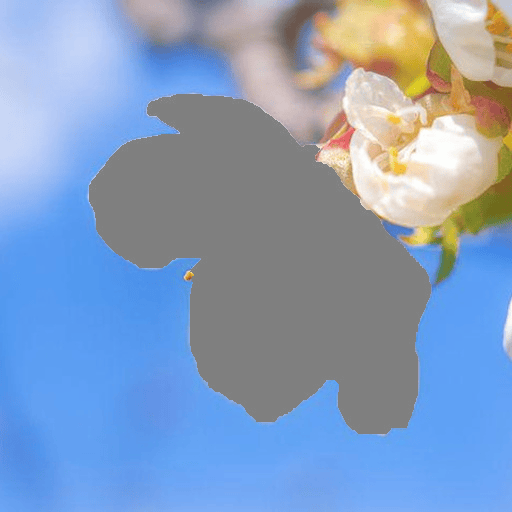}
    \includegraphics[width=0.22\linewidth, trim=0mm 0mm 0mm 0mm, clip]{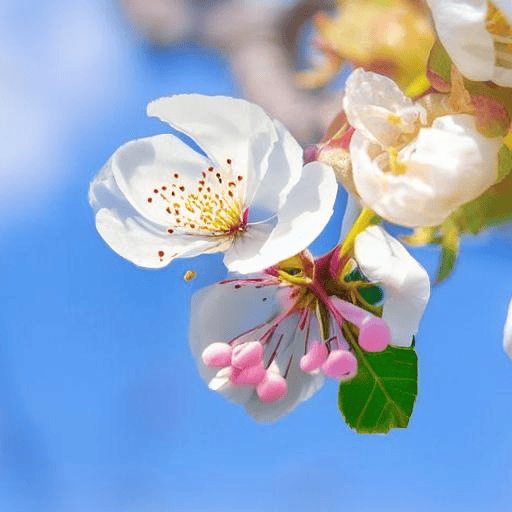}
    \includegraphics[width=0.22\linewidth, trim=0mm 0mm 0mm 0mm, clip]{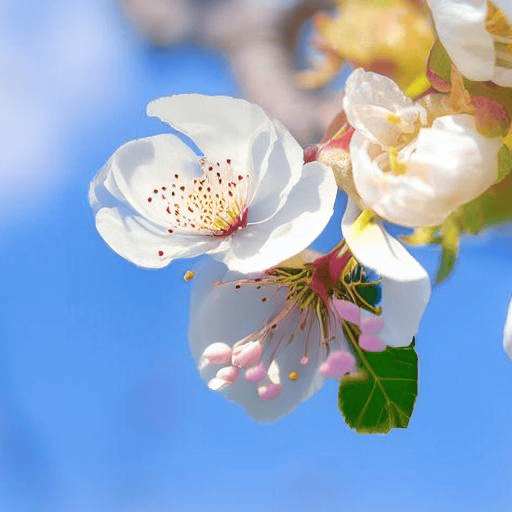}
    \includegraphics[width=0.22\linewidth, trim=0mm 0mm 0mm 0mm, clip]{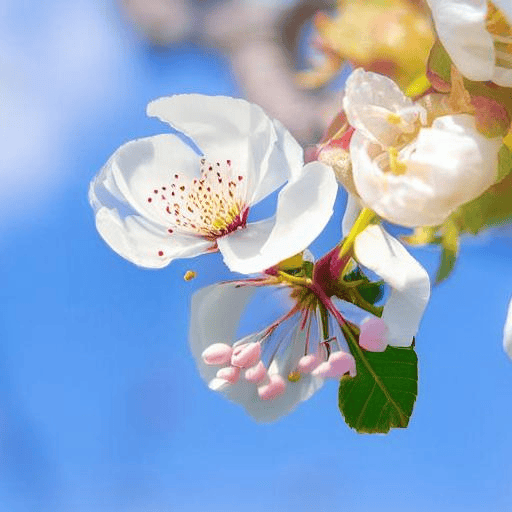}
\caption{a close up of a cherry blossom with white flowers}
\end{subfigure}
\begin{subfigure}[t]{.49\linewidth}
\centering
    \includegraphics[width=0.22\linewidth, trim=0mm 0mm 0mm 0mm, clip]{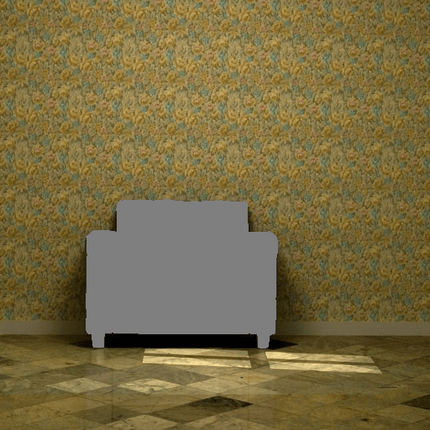}
    \includegraphics[width=0.22\linewidth, trim=0mm 0mm 0mm 0mm, clip]{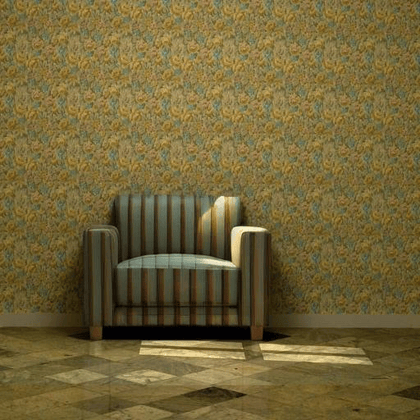}
    \includegraphics[width=0.22\linewidth, trim=0mm 0mm 0mm 0mm, clip]{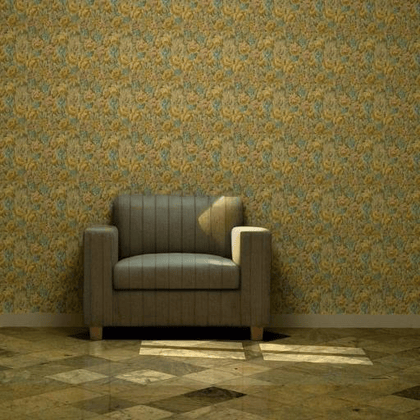}
    \includegraphics[width=0.22\linewidth, trim=0mm 0mm 0mm 0mm, clip]{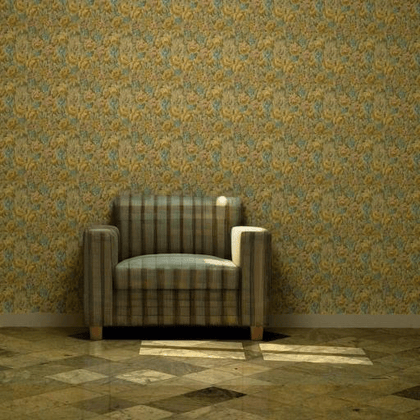}
\caption{a chair in front of a wall with a floral pattern}
\end{subfigure}
\begin{subfigure}[t]{.49\linewidth}
\centering
    \includegraphics[width=0.22\linewidth, trim=0mm 0mm 0mm 0mm, clip]{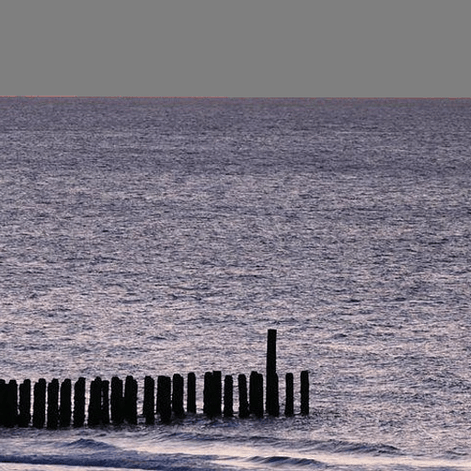}
    \includegraphics[width=0.22\linewidth, trim=0mm 0mm 0mm 0mm, clip]{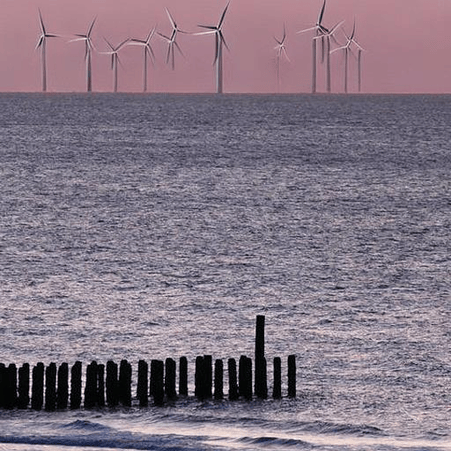}
    \includegraphics[width=0.22\linewidth, trim=0mm 0mm 0mm 0mm, clip]{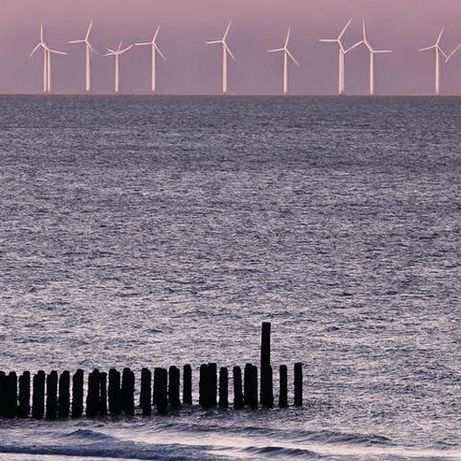}
    \includegraphics[width=0.22\linewidth, trim=0mm 0mm 0mm 0mm, clip]{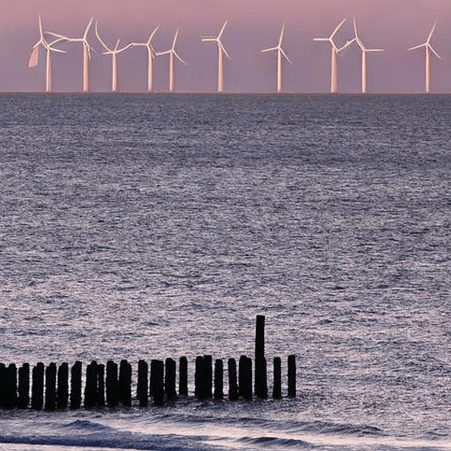}
\caption{wind turbines are seen in the ocean near a pier}
\end{subfigure}
\begin{subfigure}[t]{.49\linewidth}
\centering
    \includegraphics[width=0.22\linewidth, trim=0mm 0mm 0mm 0mm, clip]{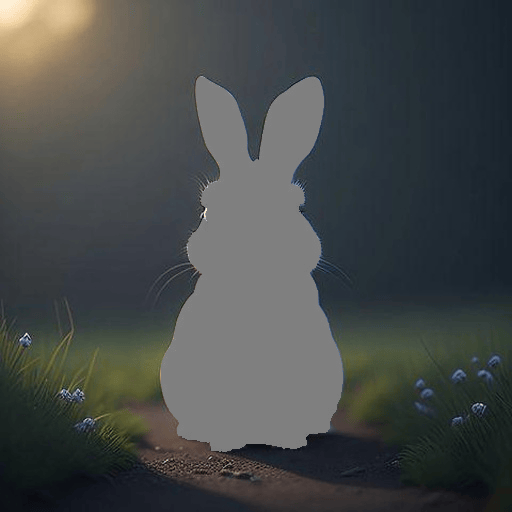}
    \includegraphics[width=0.22\linewidth, trim=0mm 0mm 0mm 0mm, clip]{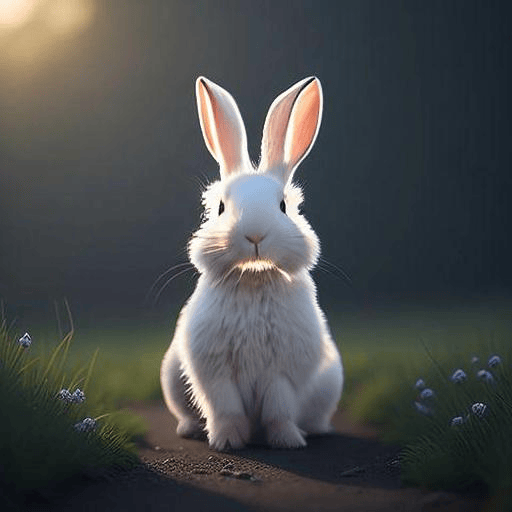}
    \includegraphics[width=0.22\linewidth, trim=0mm 0mm 0mm 0mm, clip]{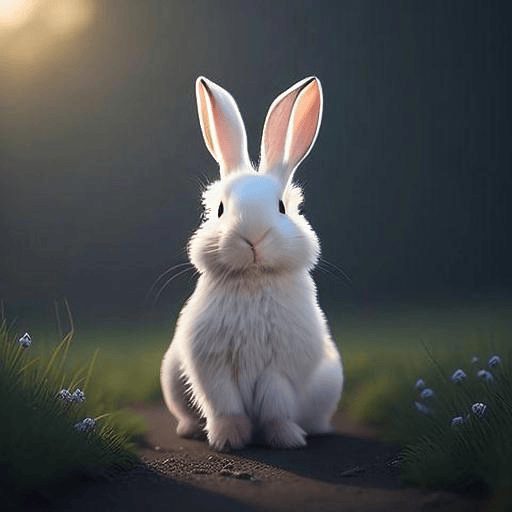}
    \includegraphics[width=0.22\linewidth, trim=0mm 0mm 0mm 0mm, clip]{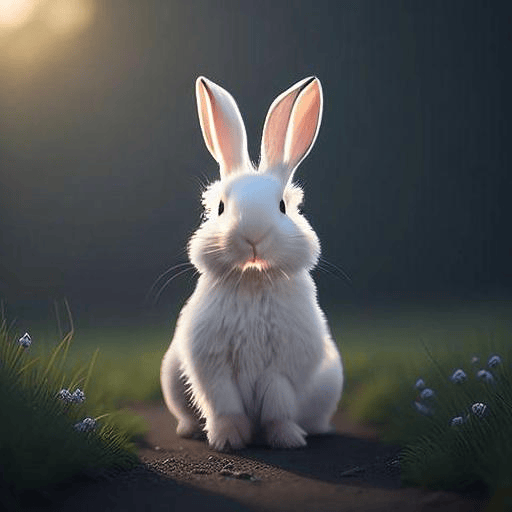}
\caption{a white rabbit sitting on the ground in the grass}
\end{subfigure}
\begin{subfigure}[t]{.49\linewidth}
\centering
    \includegraphics[width=0.22\linewidth, trim=0mm 0mm 0mm 0mm, clip]{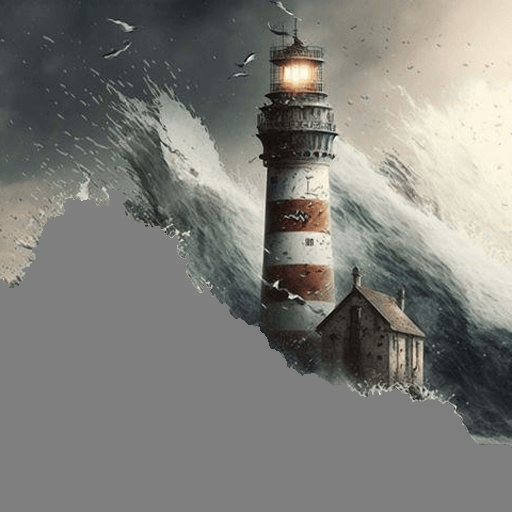}
    \includegraphics[width=0.22\linewidth, trim=0mm 0mm 0mm 0mm, clip]{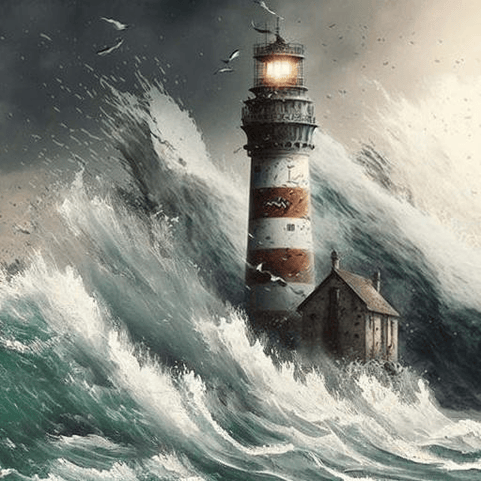}
    \includegraphics[width=0.22\linewidth, trim=0mm 0mm 0mm 0mm, clip]{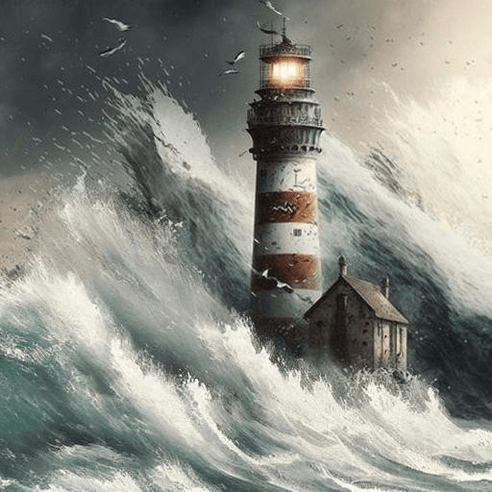}
    \includegraphics[width=0.22\linewidth, trim=0mm 0mm 0mm 0mm, clip]{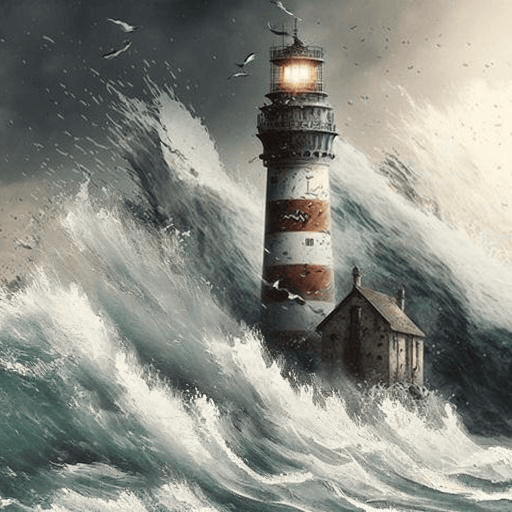}
\caption{a lighthouse in the middle of a stormy sea}
\end{subfigure}
\begin{subfigure}[t]{.49\linewidth}
\centering
    \includegraphics[width=0.22\linewidth, trim=0mm 0mm 0mm 0mm, clip]{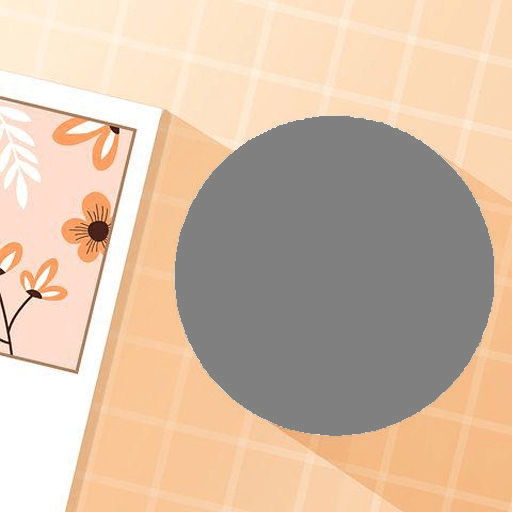}
    \includegraphics[width=0.22\linewidth, trim=0mm 0mm 0mm 0mm, clip]{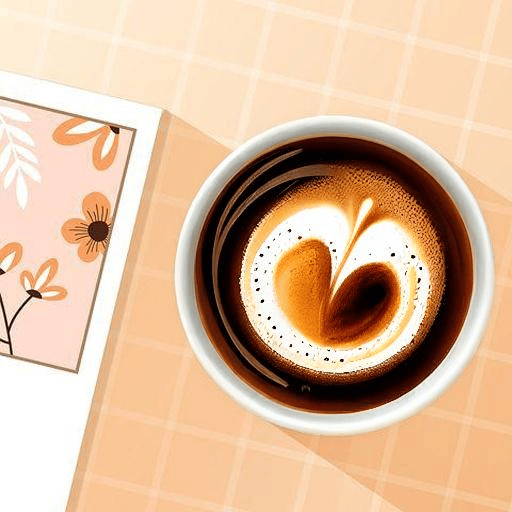}
    \includegraphics[width=0.22\linewidth, trim=0mm 0mm 0mm 0mm, clip]{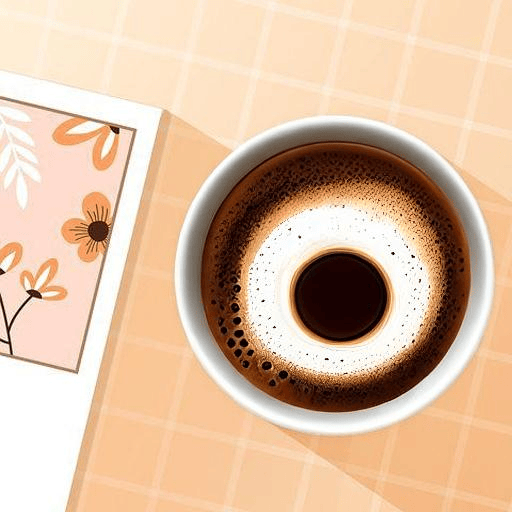}
    \includegraphics[width=0.22\linewidth, trim=0mm 0mm 0mm 0mm, clip]{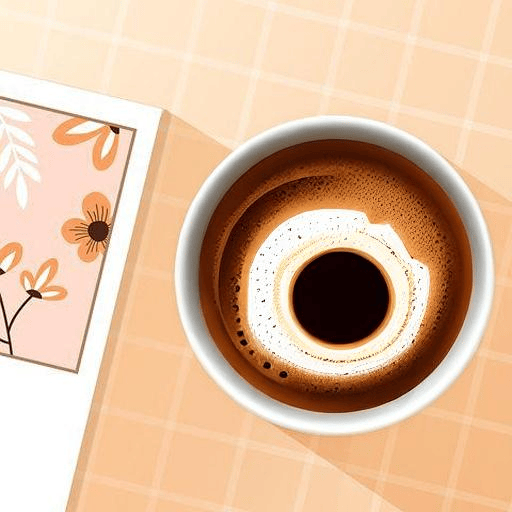}
\caption{a cup of coffee and a notebook on a table}
\end{subfigure}
\begin{subfigure}[t]{.49\linewidth}
\centering
    \includegraphics[width=0.22\linewidth, trim=0mm 0mm 0mm 0mm, clip]{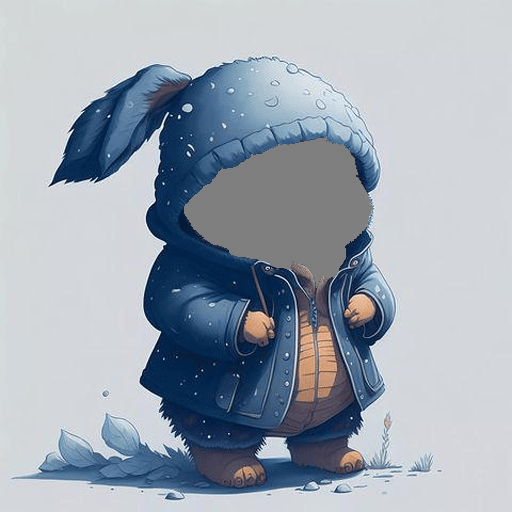}
    \includegraphics[width=0.22\linewidth, trim=0mm 0mm 0mm 0mm, clip]{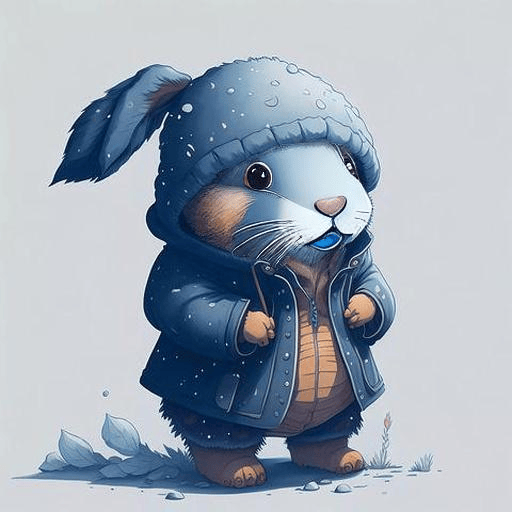}
    \includegraphics[width=0.22\linewidth, trim=0mm 0mm 0mm 0mm, clip]{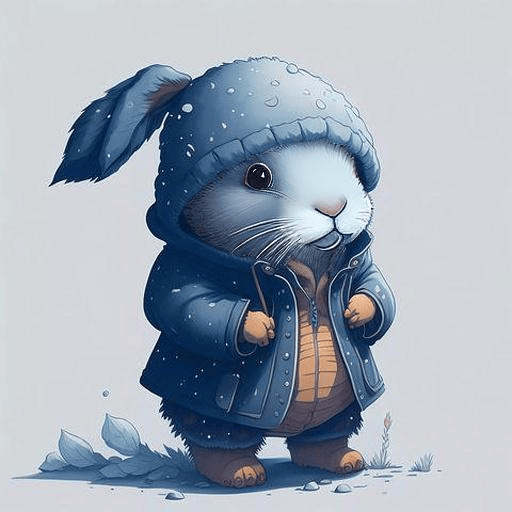}
    \includegraphics[width=0.22\linewidth, trim=0mm 0mm 0mm 0mm, clip]{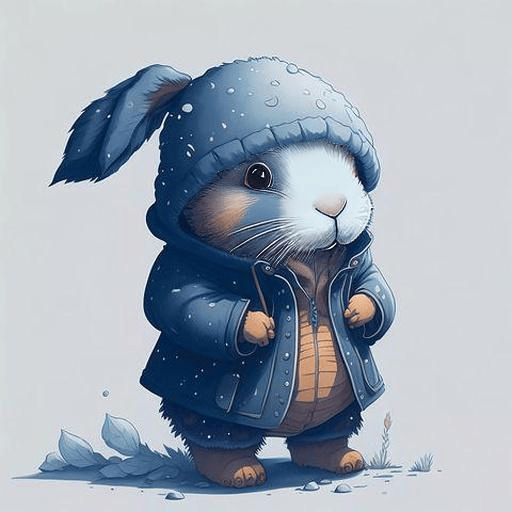}
\caption{a cartoon rabbit wearing a blue coat and hat}
\end{subfigure}
\begin{subfigure}[t]{.49\linewidth}
\centering
    \includegraphics[width=0.22\linewidth, trim=0mm 0mm 0mm 0mm, clip]{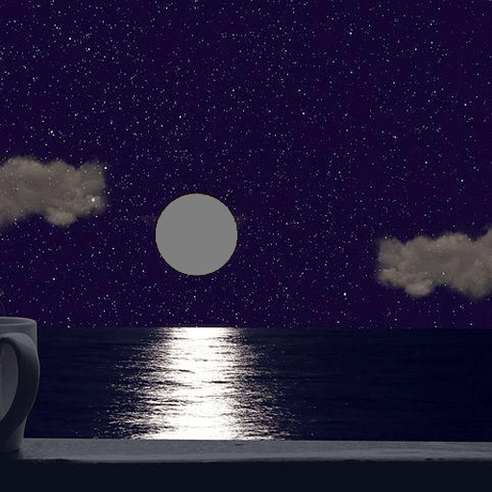}
    \includegraphics[width=0.22\linewidth, trim=0mm 0mm 0mm 0mm, clip]{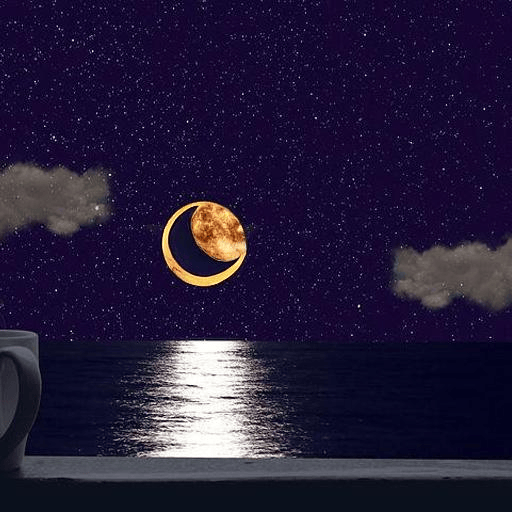}
    \includegraphics[width=0.22\linewidth, trim=0mm 0mm 0mm 0mm, clip]{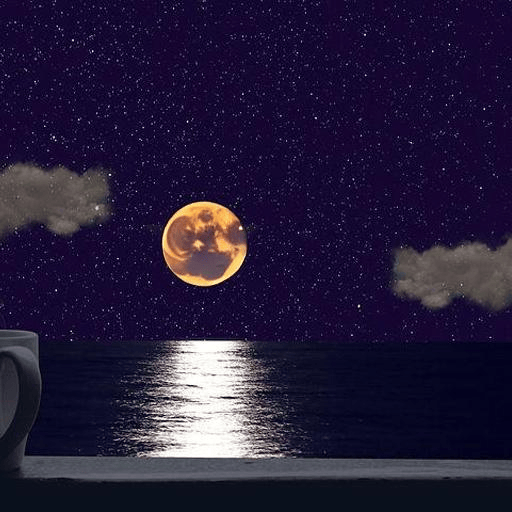}
    \includegraphics[width=0.22\linewidth, trim=0mm 0mm 0mm 0mm, clip]{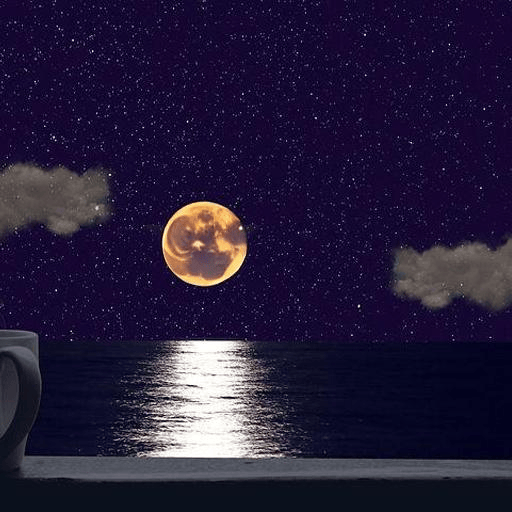}
\caption{a cup of coffee on a table next to a window with the moon and stars in the sky}
\end{subfigure}
\begin{subfigure}[t]{.49\linewidth}
\centering
    \includegraphics[width=0.22\linewidth, trim=0mm 0mm 0mm 0mm, clip]{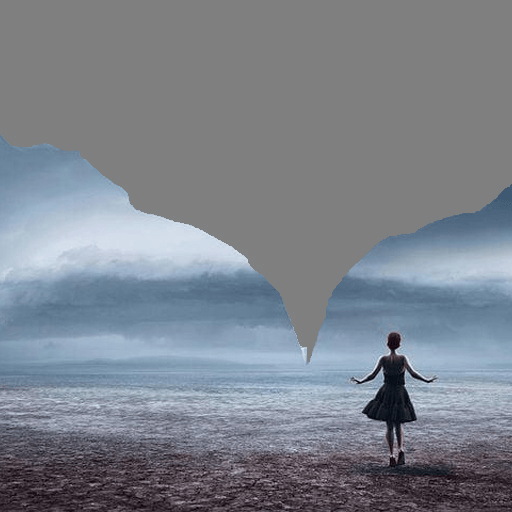}
    \includegraphics[width=0.22\linewidth, trim=0mm 0mm 0mm 0mm, clip]{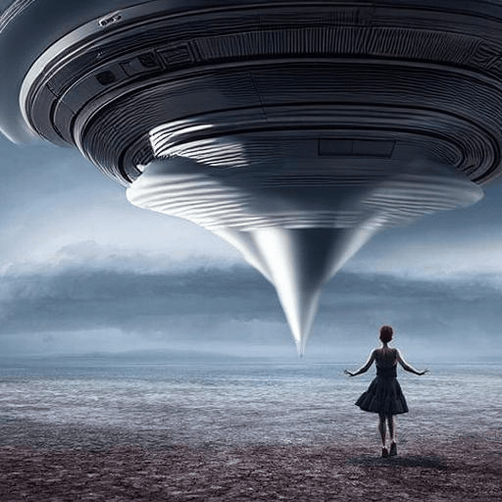}
    \includegraphics[width=0.22\linewidth, trim=0mm 0mm 0mm 0mm, clip]{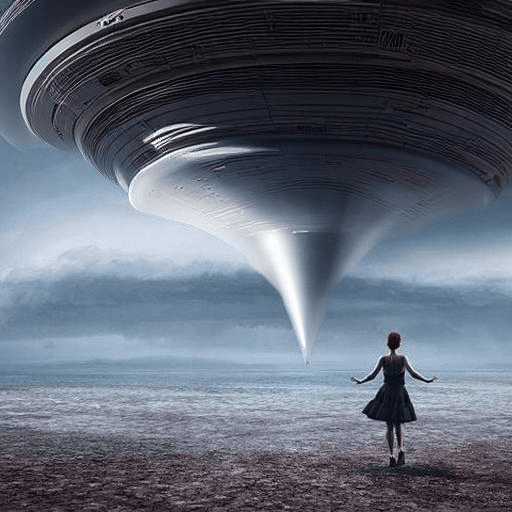}
    \includegraphics[width=0.22\linewidth, trim=0mm 0mm 0mm 0mm, clip]{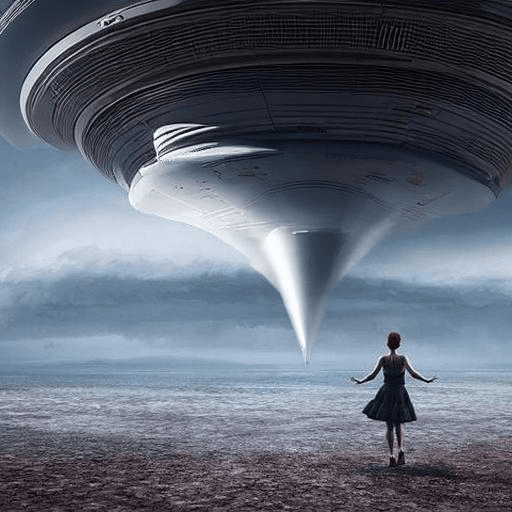}
\caption{a woman standing in front of a large spaceship}
\end{subfigure}
\caption{\textbf{More results on Ensemble bias studies using BrushNet.} In each sub-figure, the four images (from left to right) display: the \textit{masked image}, followed by inpainting results from models trained using \textit{Ensemble}, \textit{CaPO}, and \textit{CaEN}. For optimal detail, view figures zoomed in.}
\label{fig:ensemble_brushnet_bias_appendix}
\end{figure}

\begin{figure}[!htbp]
\begin{subfigure}[t]{.49\linewidth}
\centering
  \includegraphics[width=0.22\linewidth, trim=0mm 0mm 0mm 0mm, clip]{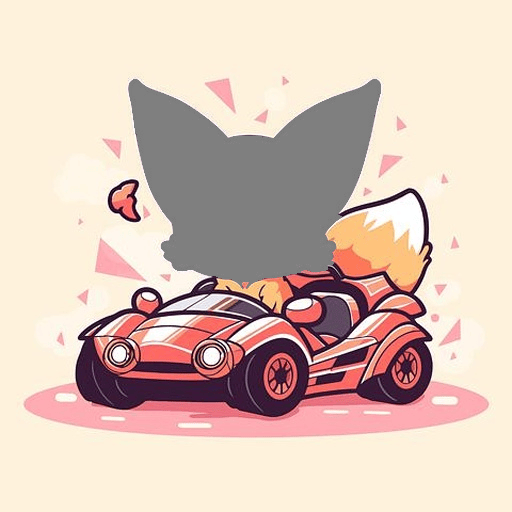}
  \includegraphics[width=0.22\linewidth, trim=0mm 0mm 0mm 0mm, clip]{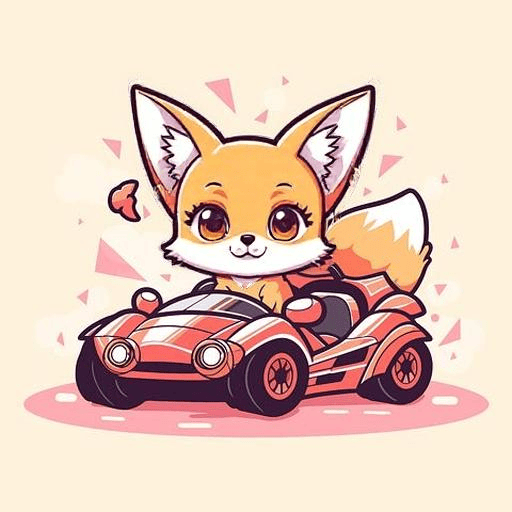}
  \includegraphics[width=0.22\linewidth, trim=0mm 0mm 0mm 0mm, clip]{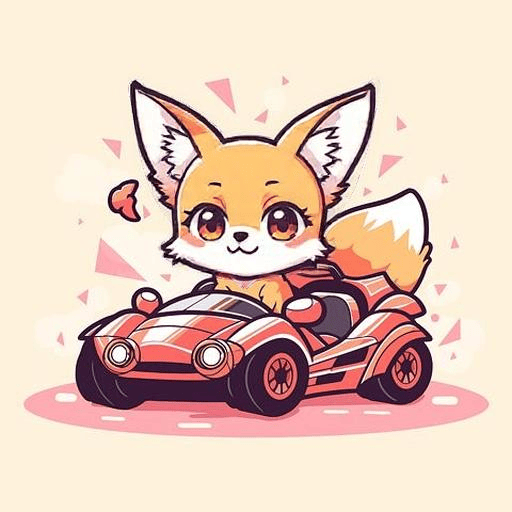}
  \includegraphics[width=0.22\linewidth, trim=0mm 0mm 0mm 0mm, clip]{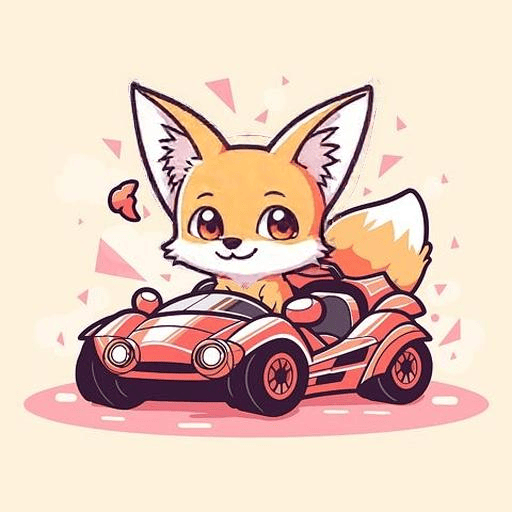}
\caption{cartoon fox driving a car with a cute face}
\end{subfigure}
\begin{subfigure}[t]{.49\linewidth}
\centering
  \includegraphics[width=0.22\linewidth, trim=0mm 0mm 0mm 0mm, clip]{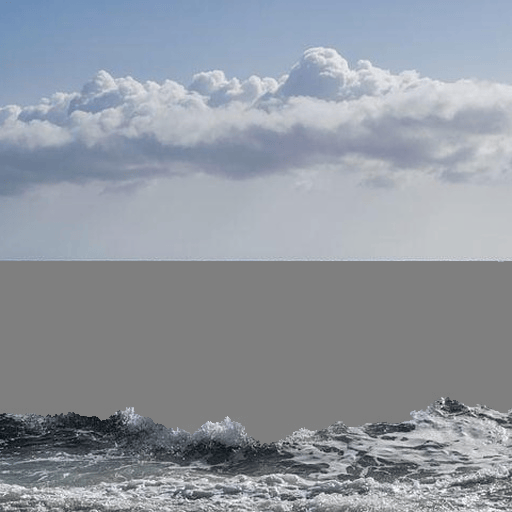}
  \includegraphics[width=0.22\linewidth, trim=0mm 0mm 0mm 0mm, clip]{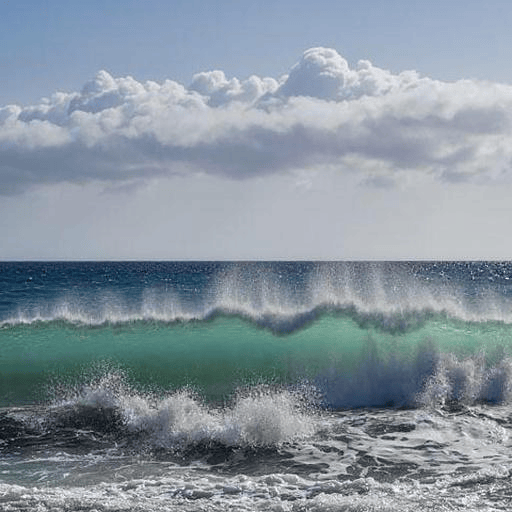}
  \includegraphics[width=0.22\linewidth, trim=0mm 0mm 0mm 0mm, clip]{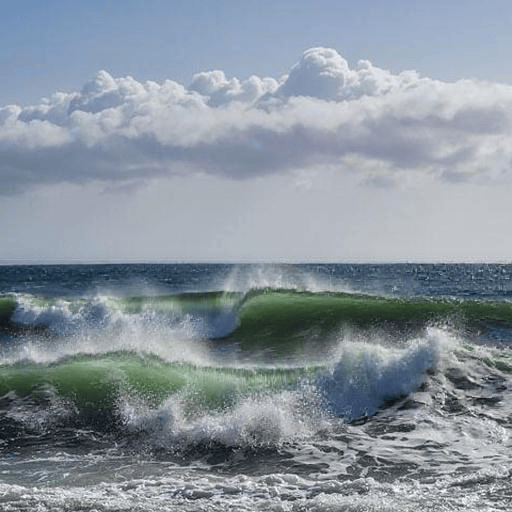}
  \includegraphics[width=0.22\linewidth, trim=0mm 0mm 0mm 0mm, clip]{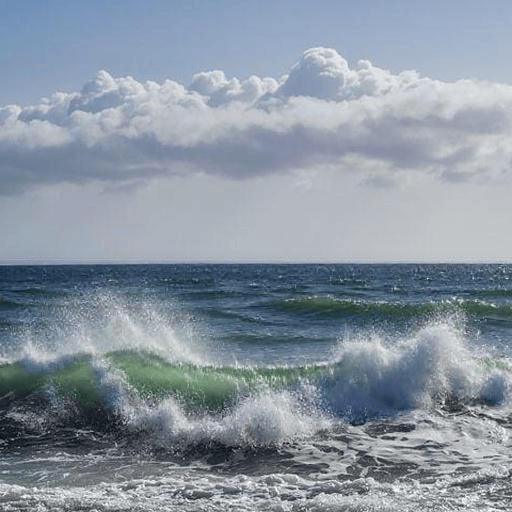}
\caption{a large wave crashing into the ocean with white clouds in the sky}
\end{subfigure}
\begin{subfigure}[t]{.49\linewidth}
\centering
  \includegraphics[width=0.22\linewidth, trim=0mm 0mm 0mm 0mm, clip]{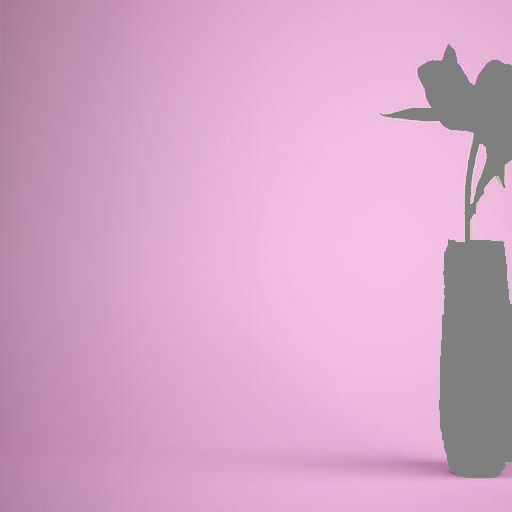}
  \includegraphics[width=0.22\linewidth, trim=0mm 0mm 0mm 0mm, clip]{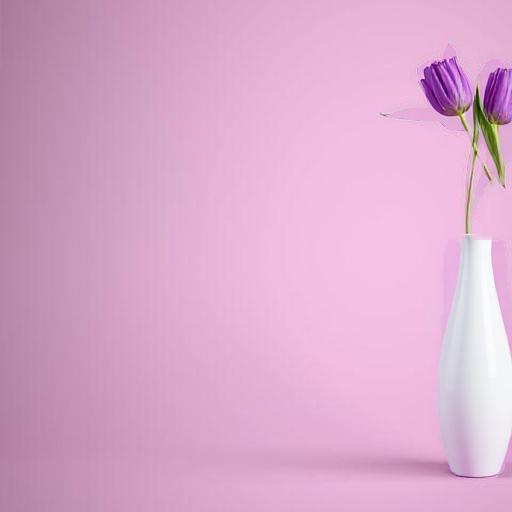}
  \includegraphics[width=0.22\linewidth, trim=0mm 0mm 0mm 0mm, clip]{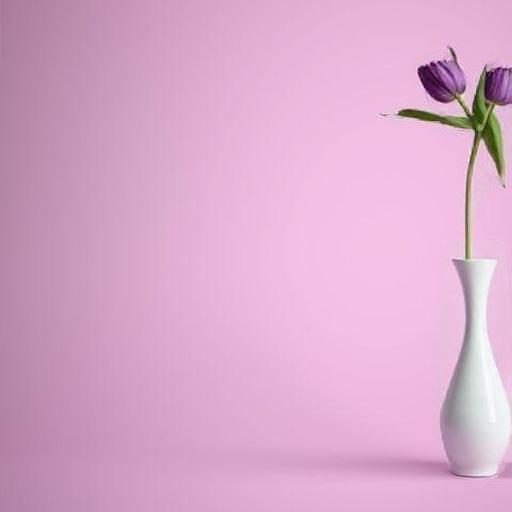}
  \includegraphics[width=0.22\linewidth, trim=0mm 0mm 0mm 0mm, clip]{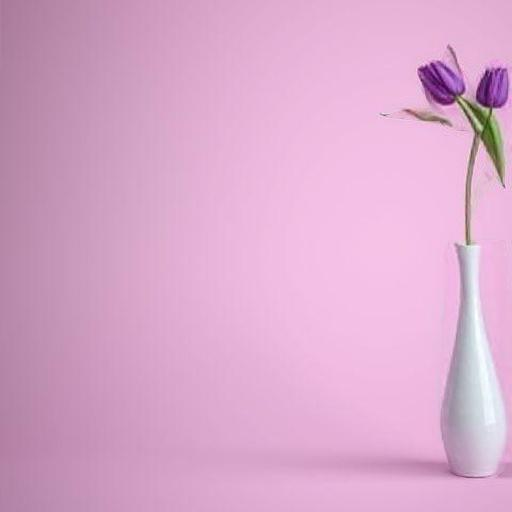}
\caption{a white vase with purple tulips in it}
\end{subfigure}
\begin{subfigure}[t]{.49\linewidth}
\centering
  \includegraphics[width=0.22\linewidth, trim=0mm 0mm 0mm 0mm, clip]{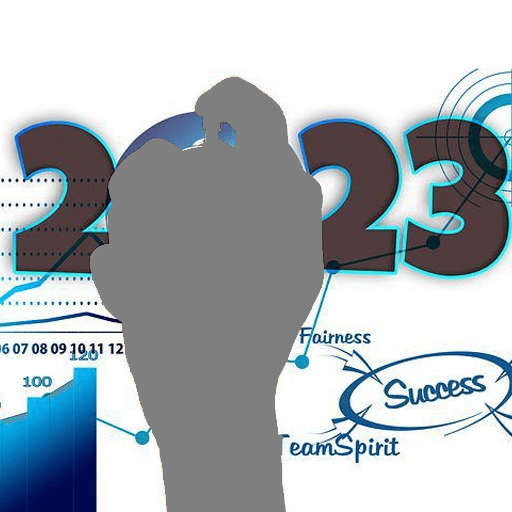}
  \includegraphics[width=0.22\linewidth, trim=0mm 0mm 0mm 0mm, clip]{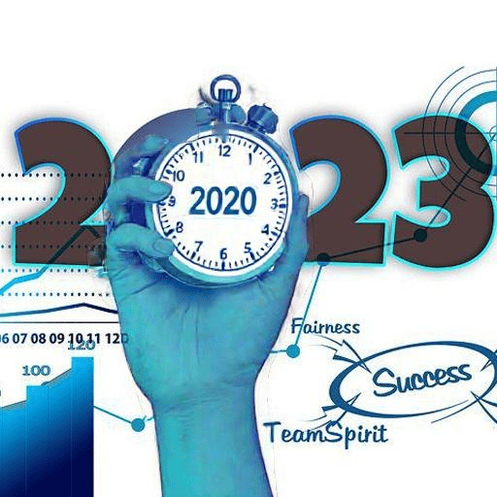}
  \includegraphics[width=0.22\linewidth, trim=0mm 0mm 0mm 0mm, clip]{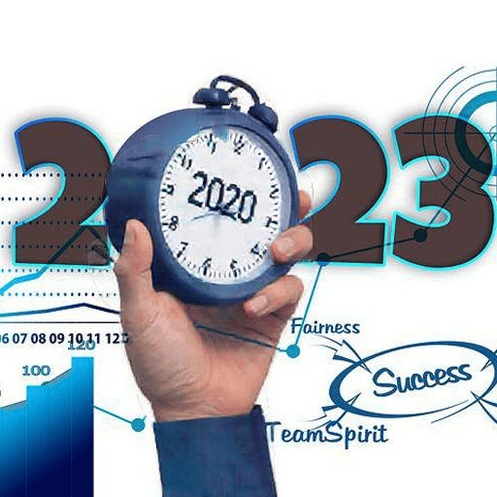}
  \includegraphics[width=0.22\linewidth, trim=0mm 0mm 0mm 0mm, clip]{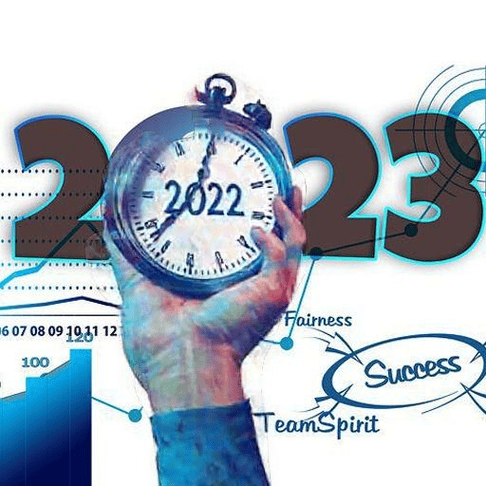}
\caption{a person holding a stopwatch with the words 2020 and 2021}
\end{subfigure}
\begin{subfigure}[t]{.49\linewidth}
\centering
  \includegraphics[width=0.22\linewidth, trim=0mm 0mm 0mm 0mm, clip]{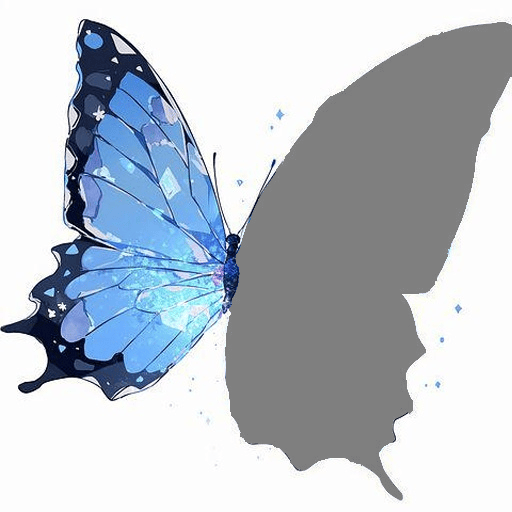}
  \includegraphics[width=0.22\linewidth, trim=0mm 0mm 0mm 0mm, clip]{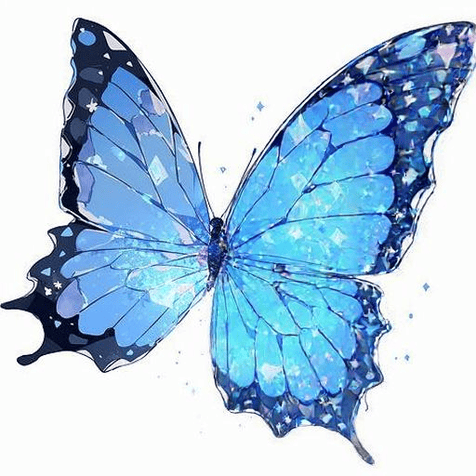}
  \includegraphics[width=0.22\linewidth, trim=0mm 0mm 0mm 0mm, clip]{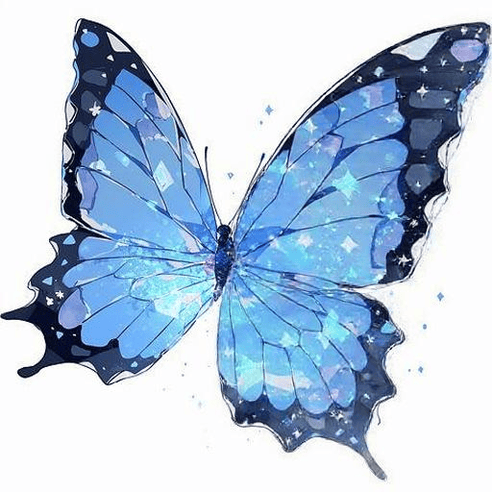}
  \includegraphics[width=0.22\linewidth, trim=0mm 0mm 0mm 0mm, clip]{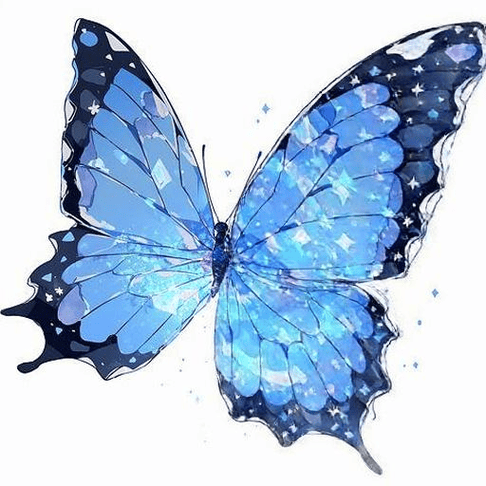}
\caption{a blue butterfly with stars on it}
\end{subfigure}
\begin{subfigure}[t]{.49\linewidth}
\centering
  \includegraphics[width=0.22\linewidth, trim=0mm 0mm 0mm 0mm, clip]{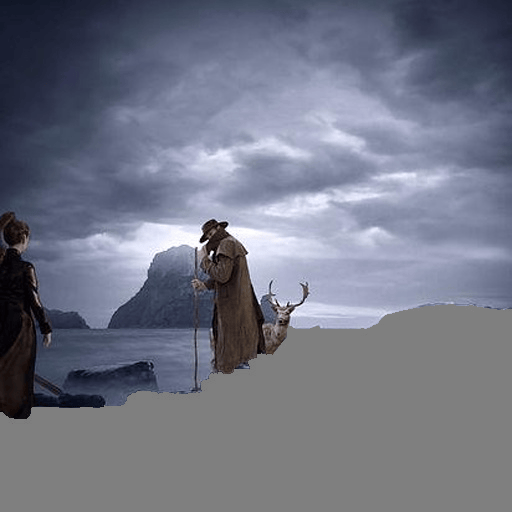}
  \includegraphics[width=0.22\linewidth, trim=0mm 0mm 0mm 0mm, clip]{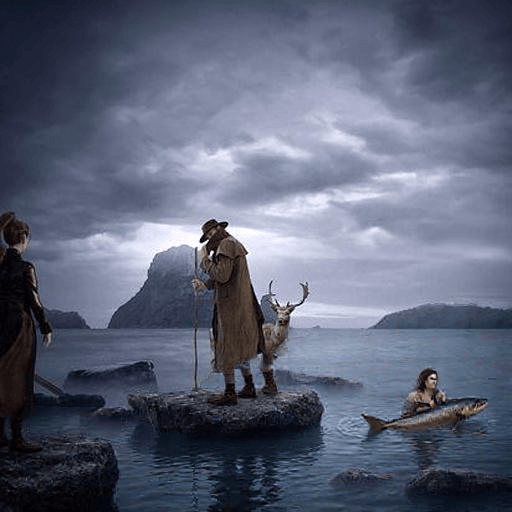}
  \includegraphics[width=0.22\linewidth, trim=0mm 0mm 0mm 0mm, clip]{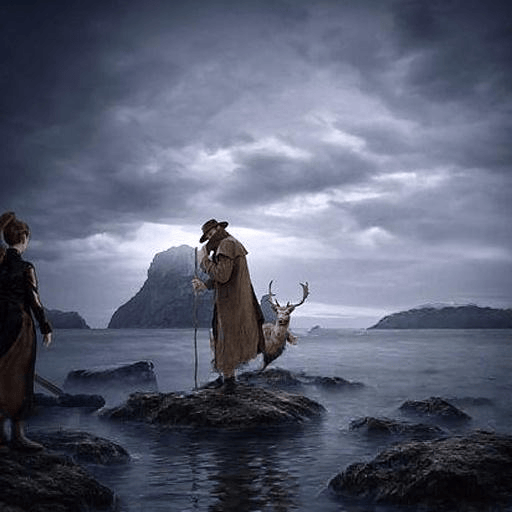}
  \includegraphics[width=0.22\linewidth, trim=0mm 0mm 0mm 0mm, clip]{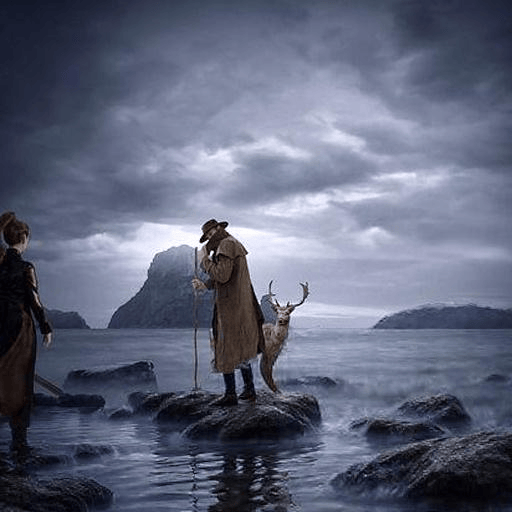}
\caption{a man and woman standing on rocks with a fish in the water}
\end{subfigure}
\begin{subfigure}[t]{.49\linewidth}
\centering
  \includegraphics[width=0.22\linewidth, trim=0mm 0mm 0mm 0mm, clip]{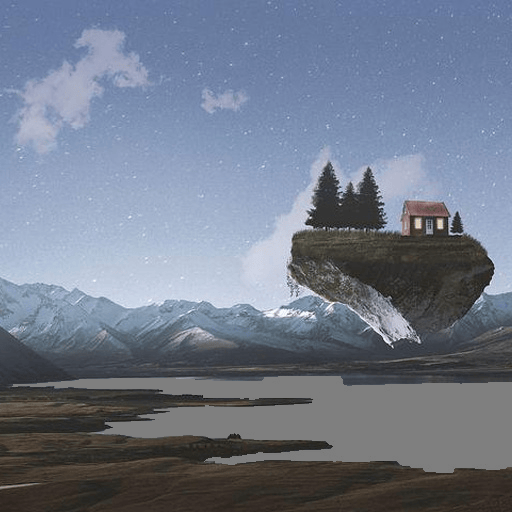}
  \includegraphics[width=0.22\linewidth, trim=0mm 0mm 0mm 0mm, clip]{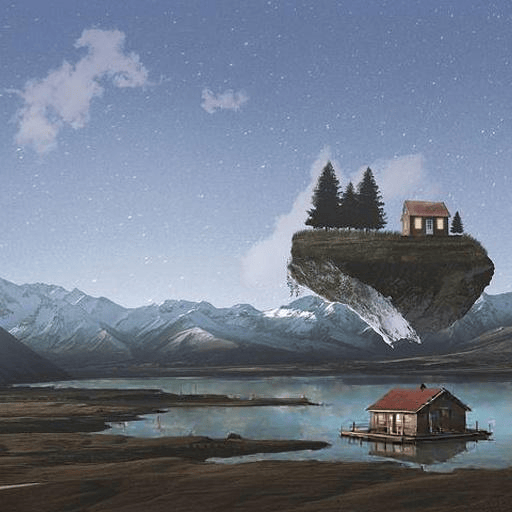}
  \includegraphics[width=0.22\linewidth, trim=0mm 0mm 0mm 0mm, clip]{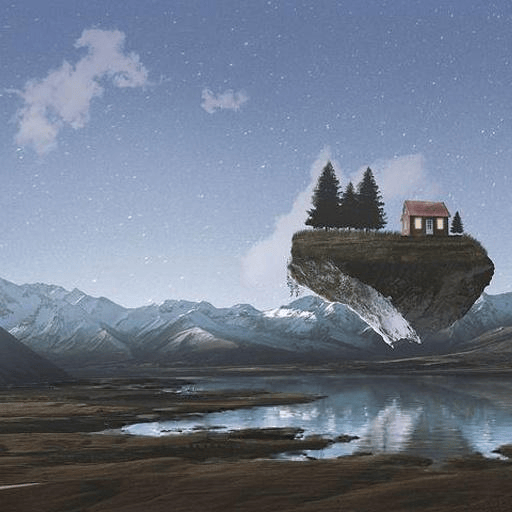}
  \includegraphics[width=0.22\linewidth, trim=0mm 0mm 0mm 0mm, clip]{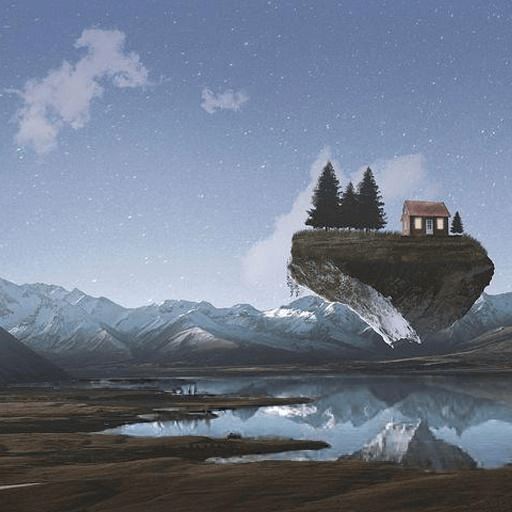}
\caption{a house floating in the air over a lake}
\end{subfigure}
\begin{subfigure}[t]{.49\linewidth}
\centering
  \includegraphics[width=0.22\linewidth, trim=0mm 0mm 0mm 0mm, clip]{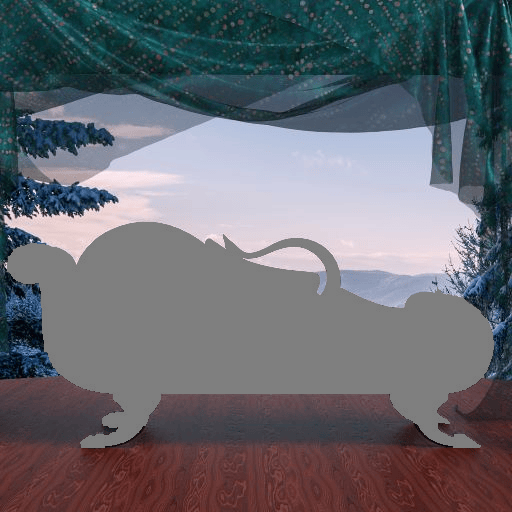}
  \includegraphics[width=0.22\linewidth, trim=0mm 0mm 0mm 0mm, clip]{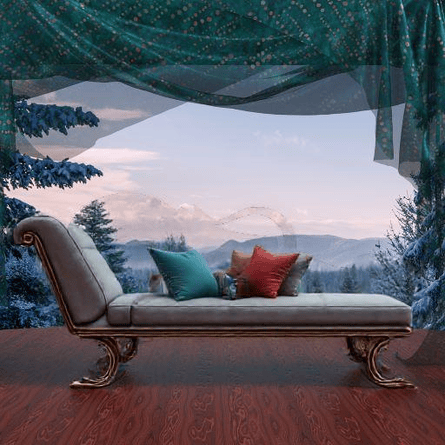}
  \includegraphics[width=0.22\linewidth, trim=0mm 0mm 0mm 0mm, clip]{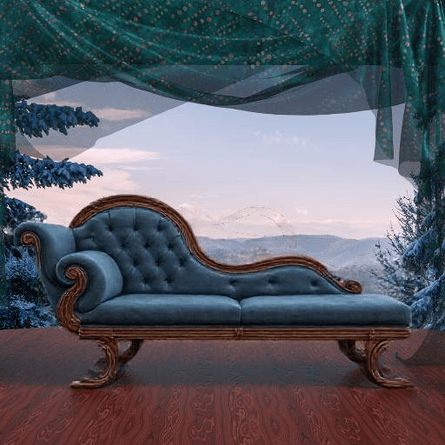}
  \includegraphics[width=0.22\linewidth, trim=0mm 0mm 0mm 0mm, clip]{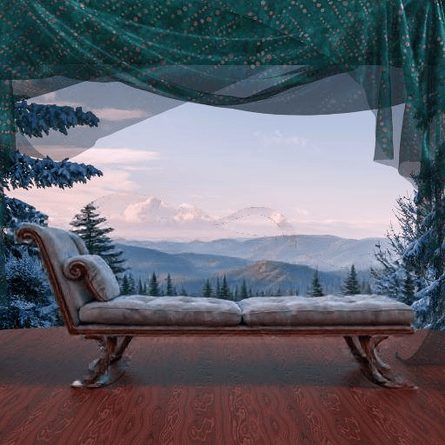}
\caption{a couch with a winged chair in front of a window}
\end{subfigure}
\begin{subfigure}[t]{.49\linewidth}
\centering
  \includegraphics[width=0.22\linewidth, trim=0mm 0mm 0mm 0mm, clip]{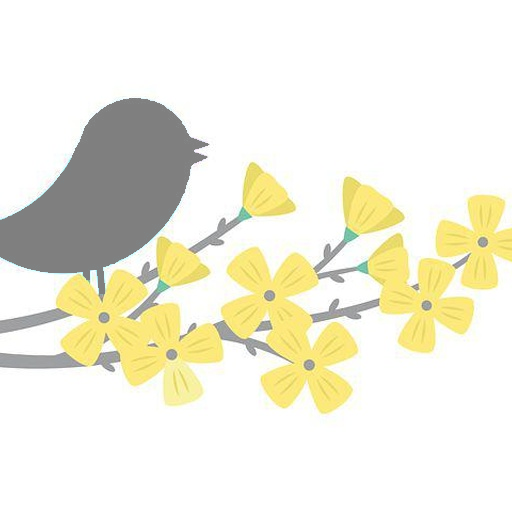}
  \includegraphics[width=0.22\linewidth, trim=0mm 0mm 0mm 0mm, clip]{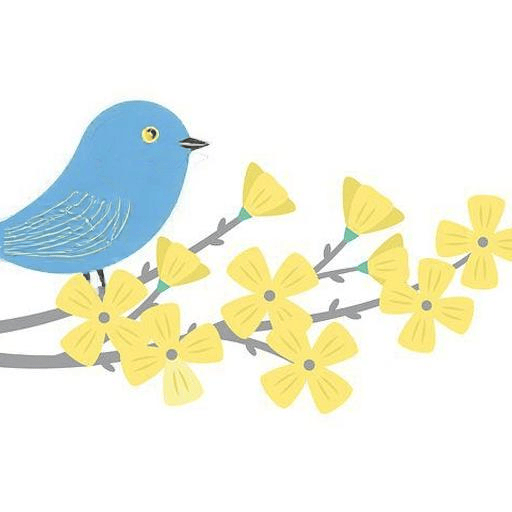}
  \includegraphics[width=0.22\linewidth, trim=0mm 0mm 0mm 0mm, clip]{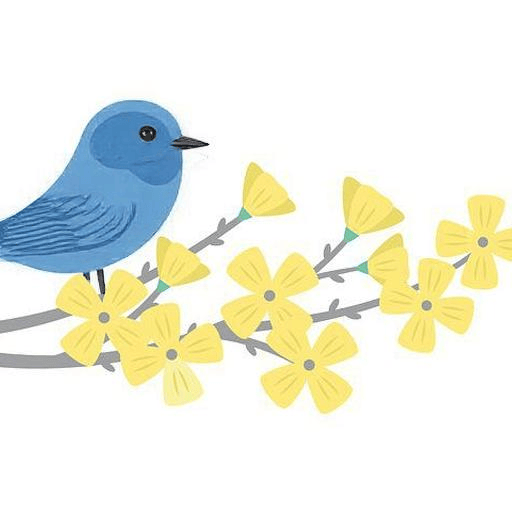}
  \includegraphics[width=0.22\linewidth, trim=0mm 0mm 0mm 0mm, clip]{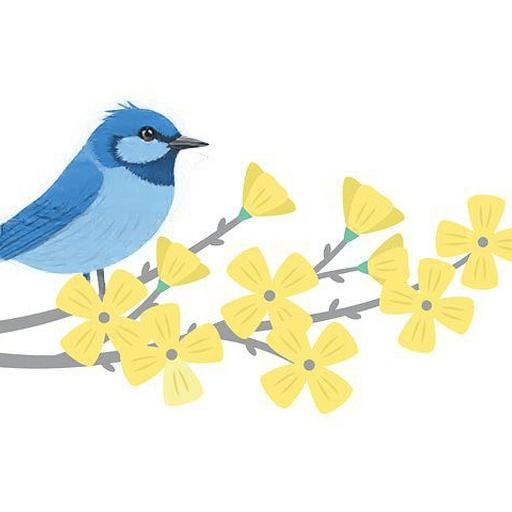}
\caption{a blue bird sitting on a branch with yellow flowers}
\end{subfigure}
\begin{subfigure}[t]{.49\linewidth}
\centering
  \includegraphics[width=0.22\linewidth, trim=0mm 0mm 0mm 0mm, clip]{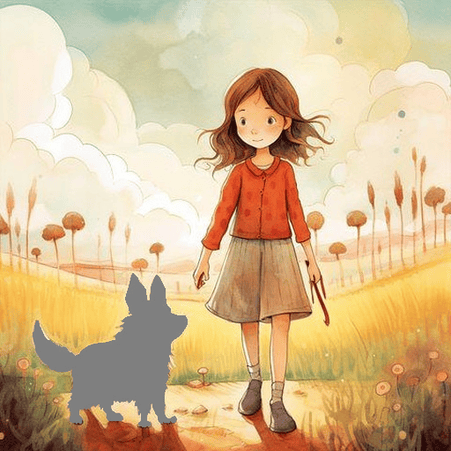}
  \includegraphics[width=0.22\linewidth, trim=0mm 0mm 0mm 0mm, clip]{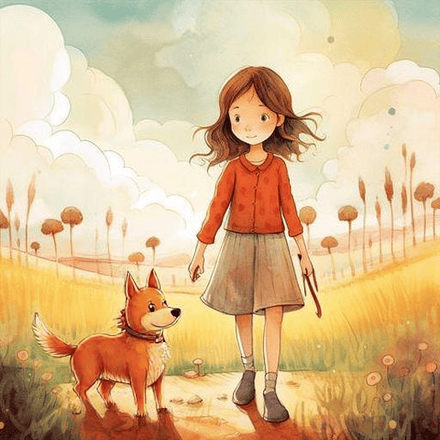}
  \includegraphics[width=0.22\linewidth, trim=0mm 0mm 0mm 0mm, clip]{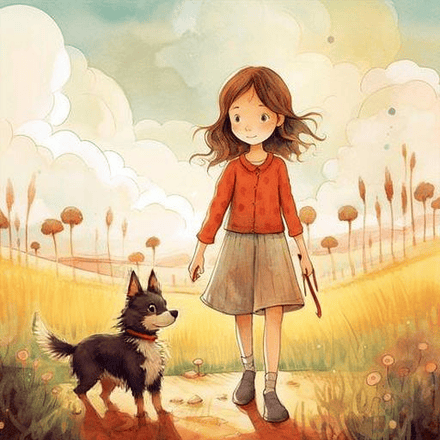}
  \includegraphics[width=0.22\linewidth, trim=0mm 0mm 0mm 0mm, clip]{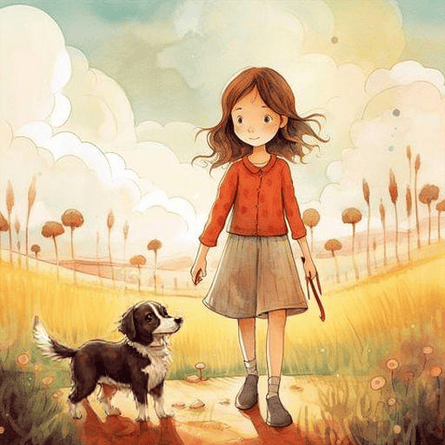}
\caption{the cover of the book, the girl and the dog}
\end{subfigure}
\begin{subfigure}[t]{.49\linewidth}
\centering
  \includegraphics[width=0.22\linewidth, trim=0mm 0mm 0mm 0mm, clip]{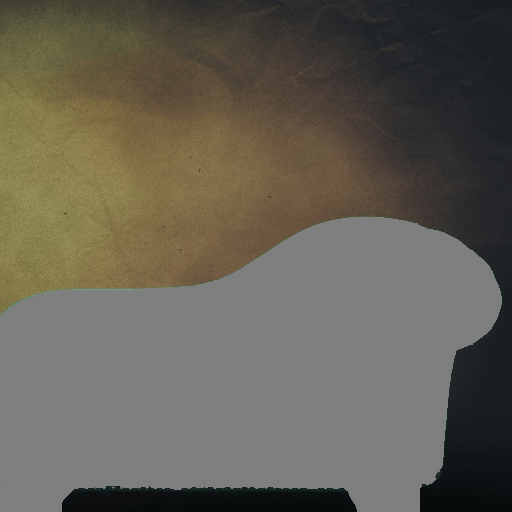}
  \includegraphics[width=0.22\linewidth, trim=0mm 0mm 0mm 0mm, clip]{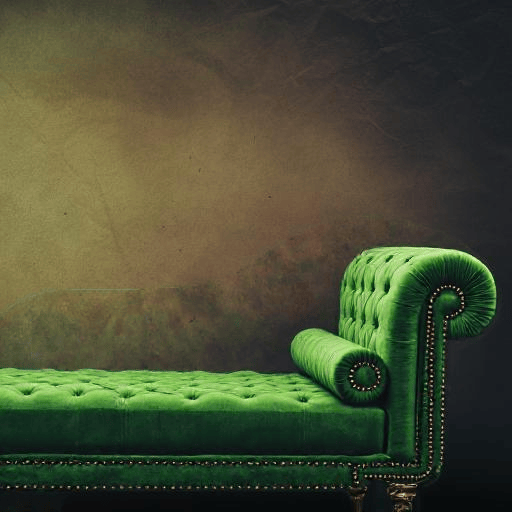}
  \includegraphics[width=0.22\linewidth, trim=0mm 0mm 0mm 0mm, clip]{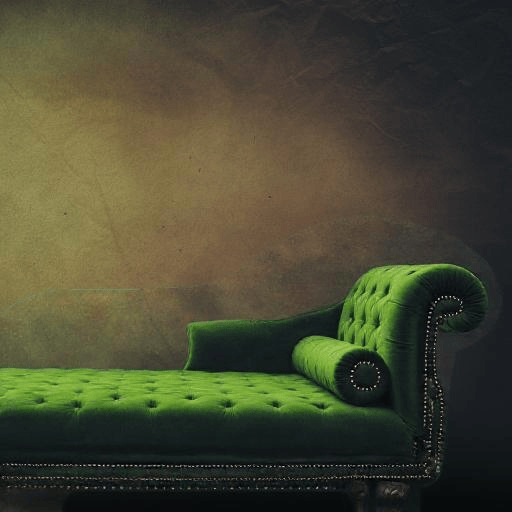}
  \includegraphics[width=0.22\linewidth, trim=0mm 0mm 0mm 0mm, clip]{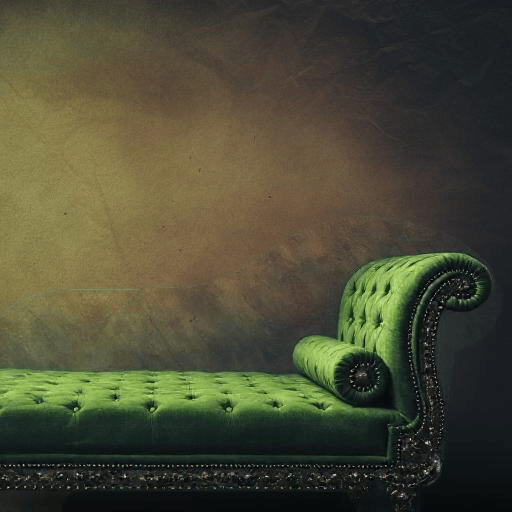}
\caption{a green velvet chaise with studded backrest on a dark background}
\end{subfigure}
\begin{subfigure}[t]{.49\linewidth}
\centering
  \includegraphics[width=0.22\linewidth, trim=0mm 0mm 0mm 0mm, clip]{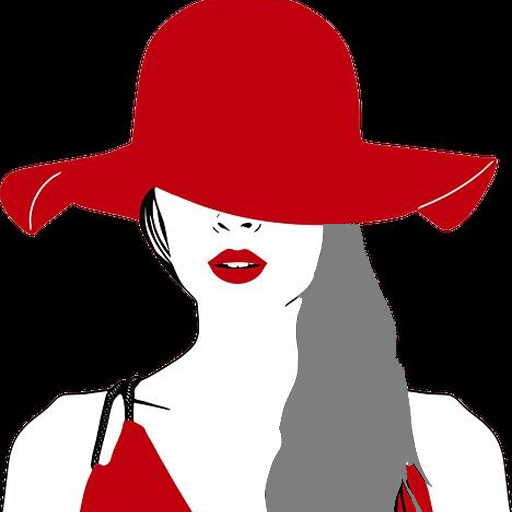}
  \includegraphics[width=0.22\linewidth, trim=0mm 0mm 0mm 0mm, clip]{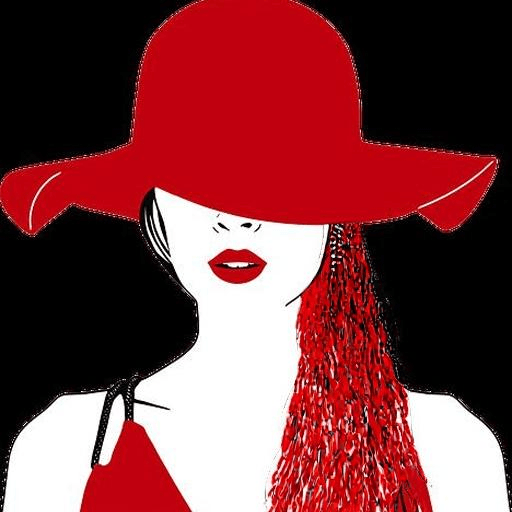}
  \includegraphics[width=0.22\linewidth, trim=0mm 0mm 0mm 0mm, clip]{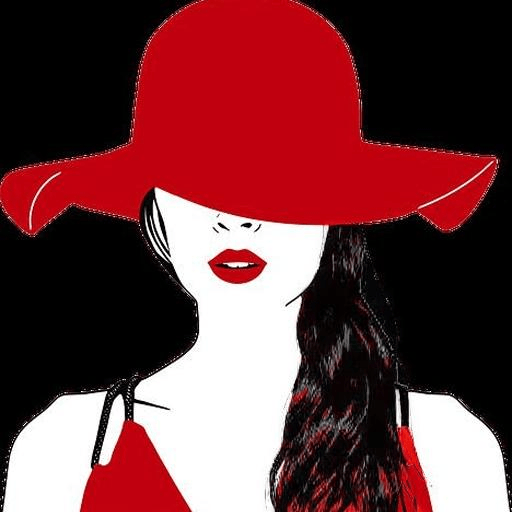}
  \includegraphics[width=0.22\linewidth, trim=0mm 0mm 0mm 0mm, clip]{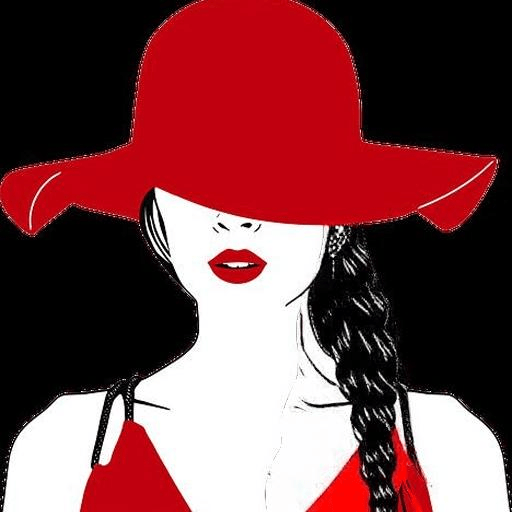}
\caption{a woman in a red hat and red dress}
\end{subfigure}
\begin{subfigure}[t]{.49\linewidth}
\centering
  \includegraphics[width=0.22\linewidth, trim=0mm 0mm 0mm 0mm, clip]{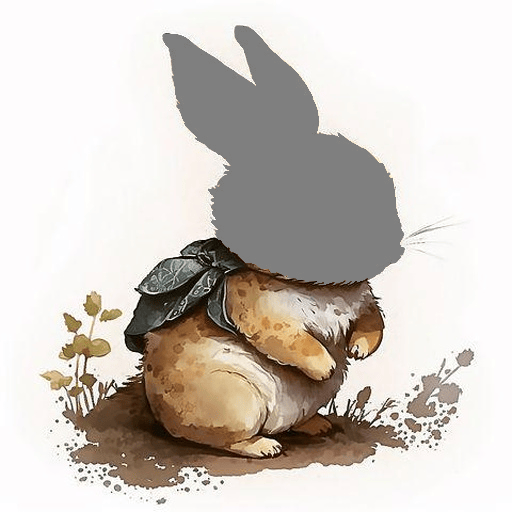}
  \includegraphics[width=0.22\linewidth, trim=0mm 0mm 0mm 0mm, clip]{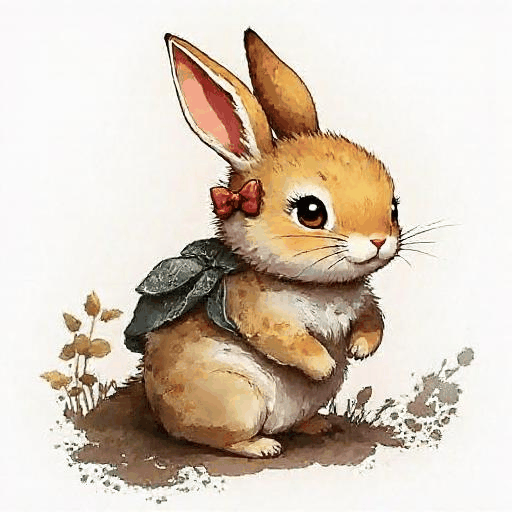}
  \includegraphics[width=0.22\linewidth, trim=0mm 0mm 0mm 0mm, clip]{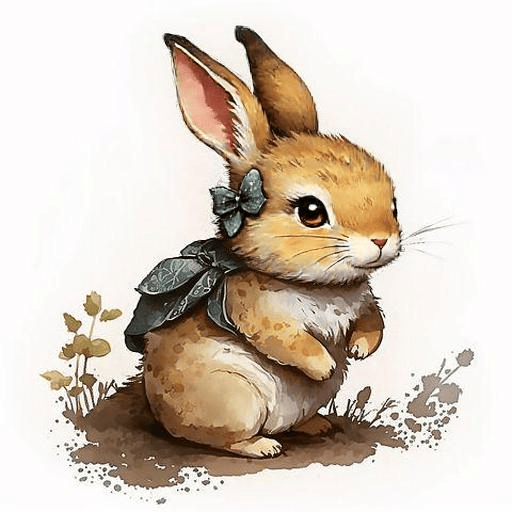}
  \includegraphics[width=0.22\linewidth, trim=0mm 0mm 0mm 0mm, clip]{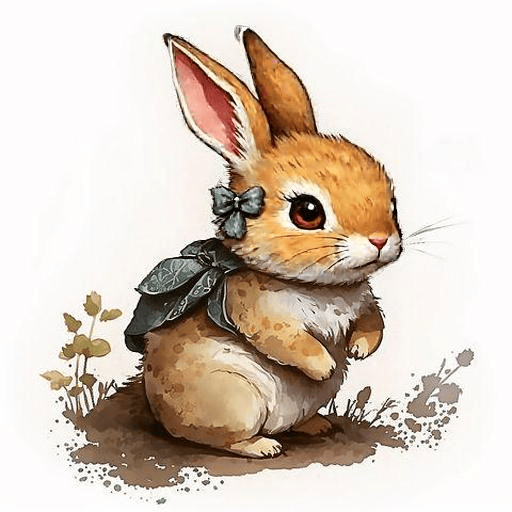}
\caption{a cute little rabbit with a bow on its back}
\end{subfigure}
\begin{subfigure}[t]{.49\linewidth}
\centering
  \includegraphics[width=0.22\linewidth, trim=0mm 0mm 0mm 0mm, clip]{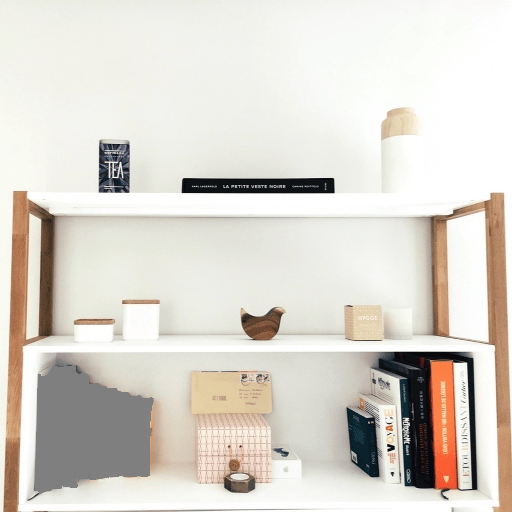}
  \includegraphics[width=0.22\linewidth, trim=0mm 0mm 0mm 0mm, clip]{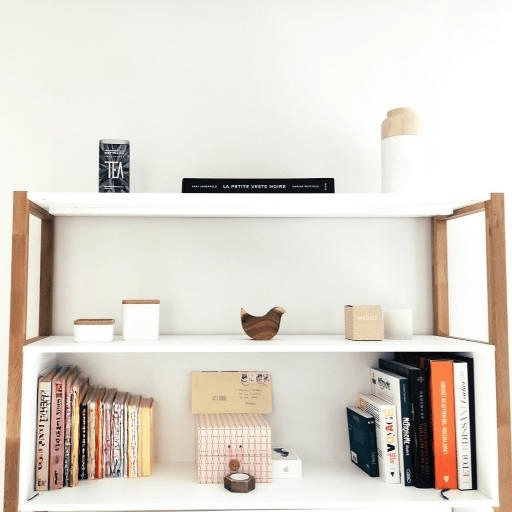}
  \includegraphics[width=0.22\linewidth, trim=0mm 0mm 0mm 0mm, clip]{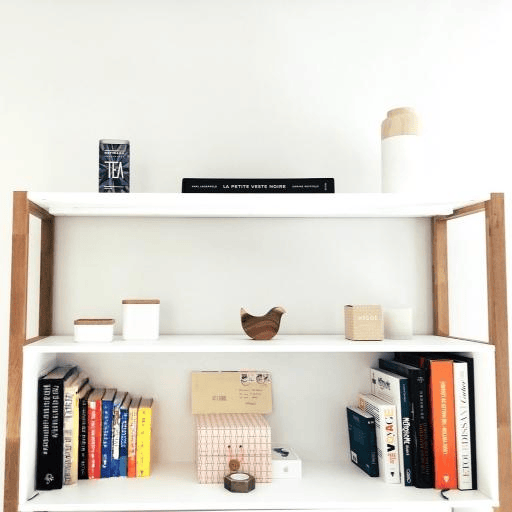}
  \includegraphics[width=0.22\linewidth, trim=0mm 0mm 0mm 0mm, clip]{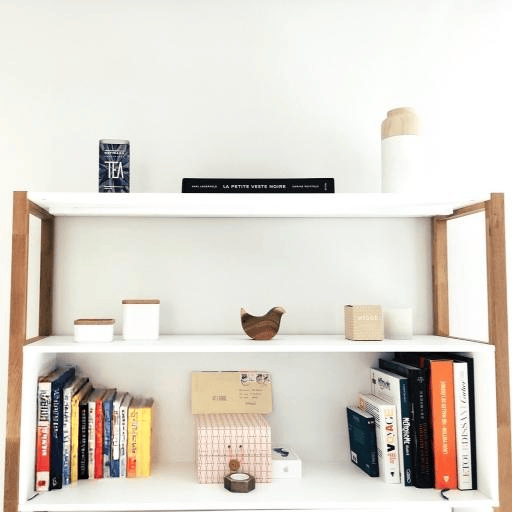}
\caption{a white shelf with books and other items on it}
\end{subfigure}
\begin{subfigure}[t]{.49\linewidth}
\centering
  \includegraphics[width=0.22\linewidth, trim=0mm 0mm 0mm 0mm, clip]{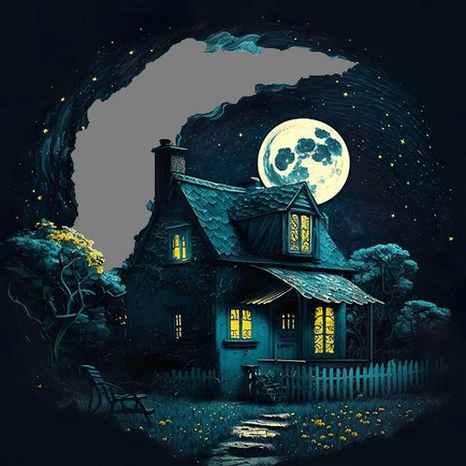}
  \includegraphics[width=0.22\linewidth, trim=0mm 0mm 0mm 0mm, clip]{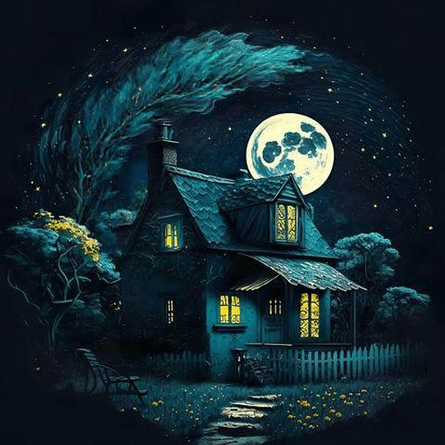}
  \includegraphics[width=0.22\linewidth, trim=0mm 0mm 0mm 0mm, clip]{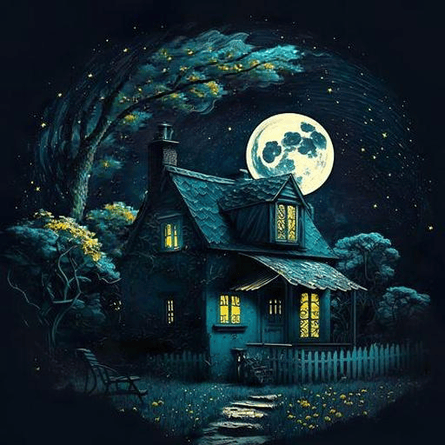}
  \includegraphics[width=0.22\linewidth, trim=0mm 0mm 0mm 0mm, clip]{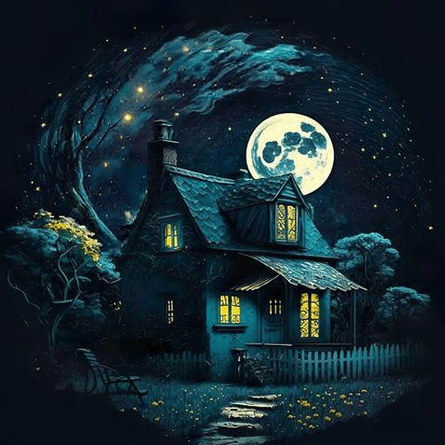}
\caption{a painting of a house with a full moon in the sky}
\end{subfigure}
\begin{subfigure}[t]{.49\linewidth}
\centering
  \includegraphics[width=0.22\linewidth, trim=0mm 0mm 0mm 0mm, clip]{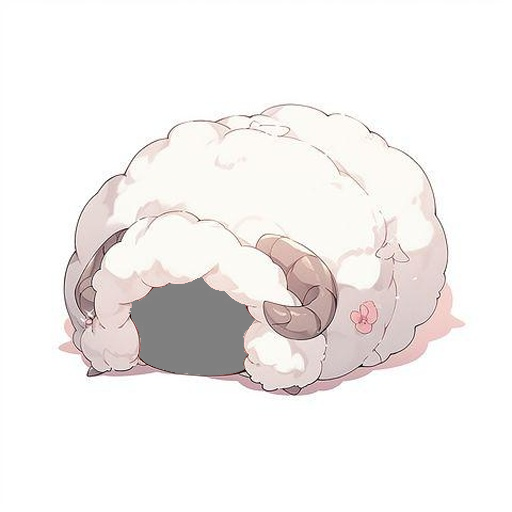}
  \includegraphics[width=0.22\linewidth, trim=0mm 0mm 0mm 0mm, clip]{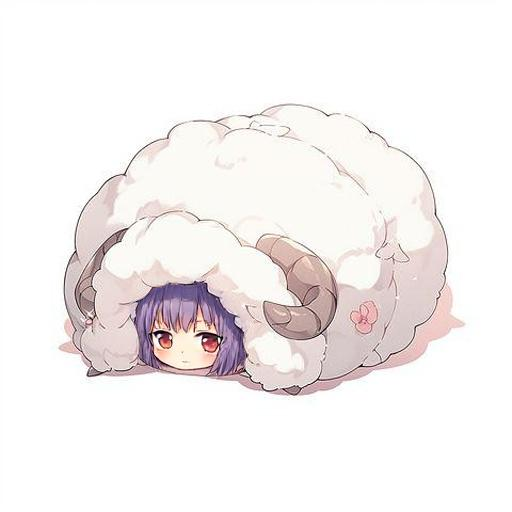}
  \includegraphics[width=0.22\linewidth, trim=0mm 0mm 0mm 0mm, clip]{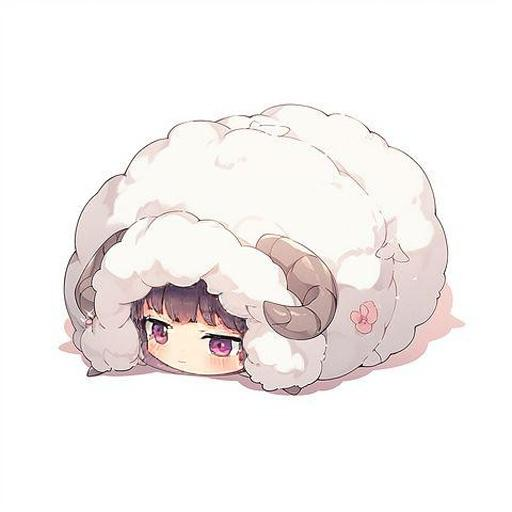}
  \includegraphics[width=0.22\linewidth, trim=0mm 0mm 0mm 0mm, clip]{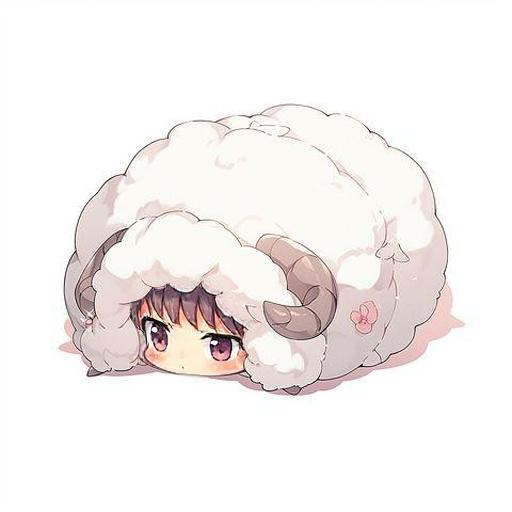}
\caption{anime girl}
\end{subfigure}
\caption{\textbf{More results on Ensemble bias studies using FLUX.1 Fill.} In each sub-figure, the four images (from left to right) display: the \textit{masked image}, followed by inpainting results from models trained using \textit{Ensemble}, \textit{CaPO}, and \textit{CaEN}. For optimal detail, view figures zoomed in.}
\label{fig:ensemble_flux_bias_appendix}
\end{figure}

\clearpage

\end{document}